\documentclass[tocnosub,noragright,centerchapter,fullpagesingle,12pt]{uiuc_csthesis21}

\makeatletter

\usepackage{setspace}  %
\usepackage[numbers, sort]{natbib}  %
\usepackage{url}  %

\usepackage{lscape}  %
\def\fillandplacepagenumber{
	\par
	\pagestyle{empty}
	\vbox to 0pt{\vss}
	\vfill
	\vbox to 0pt{
		\baselineskip 0pt
		\hbox to \linewidth{\hss}
		\baselineskip\footskip
		\hbox to \linewidth{\hfil\thepage\hfil}\vss
	}
}
\usepackage{siunitx}
\usepackage{graphicx}  %
\usepackage{caption}
\usepackage{subfigure}  %

\usepackage[labelfont=bf, textfont={rm,it}, margin=0.25in]{caption}

\usepackage{booktabs}  %
\usepackage{multicol}
\usepackage{multirow}
\usepackage{tabularx}
\usepackage{amsfonts}
\usepackage{amsmath}
\usepackage{amssymb}
\usepackage{amstext}
\usepackage{amsthm}
\usepackage{enumitem}

\theoremstyle{definition}
\newtheorem{definition}{Definition}[chapter]

\newtheorem{corollary}{Corollary}[chapter]

\newtheorem{prop}{Proposition}[chapter]
\newtheorem{problem}{Problem}[chapter]
\newtheorem{hypothesis}{Observation}[chapter]

\usepackage{listings}  %

\usepackage{algorithm}
\usepackage[noend]{algpseudocode}

\usepackage{chngcntr}
\counterwithin{algorithm}{chapter}
\usepackage{newtxmath}
\usepackage[labelfont=bf, textfont={rm,it}, margin=0.25in]{caption}
\usepackage{xspace}
\usepackage[colorlinks=true, linkcolor=black, citecolor=blue, bookmarks=false]{hyperref} %
\usepackage{cleveref}

\usepackage{amsmath,amsfonts,bm}

\def\eqref#1{equation~\ref{#1}}

\def\1{\bm{1}}

\def\rv{{\textnormal{v}}}

\def\rva{{\mathbf{a}}}
\def\rvb{{\mathbf{b}}}

\def\rve{{\mathbf{e}}}

\def\rvg{{\mathbf{g}}}
\def\rvh{{\mathbf{h}}}

\def\rvk{{\mathbf{k}}}

\def\rvm{{\mathbf{m}}}

\def\rvp{{\mathbf{p}}}

\def\rvs{{\mathbf{s}}}

\def\rvv{{\mathbf{v}}}

\def\rvx{{\mathbf{x}}}

\def\rvz{{\mathbf{z}}}

\def\va{{\bm{a}}}
\def\vb{{\bm{b}}}

\def\ve{{\bm{e}}}

\def\vg{{\bm{g}}}
\def\vh{{\bm{h}}}

\def\vp{{\bm{p}}}

\def\vx{{\bm{x}}}

\def\mA{{\bm{A}}}

\def\mD{{\bm{D}}}
\def\mE{{\bm{E}}}
\def\mF{{\bm{F}}}

\def\mI{{\bm{I}}}

\def\mL{{\bm{L}}}

\def\mP{{\bm{P}}}

\def\mR{{\bm{R}}}

\def\mV{{\bm{V}}}
\def\mW{{\bm{W}}}
\def\mX{{\bm{X}}}
\def\mY{{\bm{Y}}}
\def\mZ{{\bm{Z}}}

\DeclareMathAlphabet{\mathsfit}{\encodingdefault}{\sfdefault}{m}{sl}
\SetMathAlphabet{\mathsfit}{bold}{\encodingdefault}{\sfdefault}{bx}{n}

\def\gC{{\mathcal{C}}}

\def\gE{{\mathcal{E}}}

\def\gG{{\mathcal{G}}}

\def\gI{{\mathcal{I}}}

\def\gK{{\mathcal{K}}}
\def\gL{{\mathcal{L}}}
\def\gM{{\mathcal{M}}}
\def\gN{{\mathcal{N}}}

\def\gQ{{\mathcal{Q}}}

\def\gS{{\mathcal{S}}}
\def\gT{{\mathcal{T}}}
\def\gU{{\mathcal{U}}}
\def\gV{{\mathcal{V}}}

\def\gZ{{\mathcal{Z}}}

\def\sD{{\mathbb{D}}}

\def\sR{{\mathbb{R}}}

\def\sT{{\mathbb{T}}}

\def\sZ{{\mathbb{Z}}}

\newcommand{\E}{\mathbb{E}}

\newcommand{\sigmoid}{\sigma}

\newcommand{\KL}{D_{\mathrm{KL}}}

\phdthesis

\title{Sparsity-Aware Neural User Behavior Modeling in Online Interaction Platforms}
\author{Aravind Sankar}
\department{Computer Science}
\degreeyear{2021}

\advisor{Hari Sundaram}

\committee{Associate Professor Hari Sundaram, Chair\\
		Professor Jiawei Han \\
		Associate Professor Hanghang Tong\\
        Dr. Neil Shah, Senior Research Scientist at Snap Inc.} %
                
\begin{document}

\newcommand{\subscript}[2]{$#1 _ #2$}

\newcommand{\infomotif}{{\textsc{InfoMotif}}}
\newcommand{\protocf}{{\textsc{ProtoCF}}}
\newcommand{\groupim}{{\textsc{GroupIM}}}
\newcommand{\infvae}{{\textsc{InfVAE}}}
\newcommand{\grafrank}{{\textsc{GraFRank}}\xspace}

\makeatletter
\newcommand{\algmargin}{\the\ALG@thistlm}
\makeatother

\newcommand\independent{\protect\mathpalette{\protect\independenT}{\perp}}
\def\independenT#1#2{\mathrel{\rlap{$#1#2$}\mkern2mu{#1#2}}}

\algnewcommand{\LeftComment}[1]{\Statex \(\triangleright\) #1}
\algnewcommand{\parState}[1]{\State%
  \parbox[t]{\dimexpr\linewidth-\algmargin}{\strut #1\strut}}
  
\newcolumntype{K}[1]{>{\centering\arraybackslash}p{#1}}
\newcolumntype{R}[1]{>{\raggedright\arraybackslash}p{#1}}

\maketitle

\parindent 1em%

\frontmatter

\begin{abstract}
Modern online platforms offer users an opportunity to participate in a variety of content-creation, social networking, and shopping activities.
With the rapid proliferation of such online services, learning data-driven user behavior models is indispensable to enable personalized user experiences.
Recently, representation learning has emerged as an effective strategy for user modeling, powered by neural networks trained over large volumes of interaction data.
Despite their enormous potential, we encounter the unique challenge of data sparsity for a vast majority of entities, \textit{e.g.}, sparsity in ground-truth labels for entities and in entity-level interactions (cold-start users, items in the long-tail, and ephemeral groups).

In this dissertation, we develop generalizable neural representation learning frameworks for user behavior modeling designed to address different sparsity challenges across applications.
Our problem settings span transductive and inductive learning scenarios, where transductive learning models entities seen during training and inductive learning targets entities that are only observed during inference.
We leverage different facets of information reflecting user behavior (\textit{e.g.}, interconnectivity in social networks, temporal and attributed interaction information) to enable personalized inference at scale.
Our proposed models are complementary to concurrent advances in neural architectural choices and are adaptive to the rapid addition of new applications in online platforms.

First, we examine two transductive learning settings: inference and recommendation in \textit{graph-structured} and \textit{bipartite user-item} interactions.
In chapter~\ref{chap:infomotif}, we formulate user profiling in social platforms as semi-supervised learning over graphs given sparse ground-truth labels for node attributes.
We present a graph neural network framework that exploits higher-order connectivity structures (network motifs) to learn attributed structural roles of nodes that identify structurally similar nodes with co-varying local attributes.
In chapter~\ref{chap:protocf}, we design neural collaborative filtering models for few-shot recommendations over user-item interactions.
To address item interaction sparsity due to heavy-tailed distributions, our proposed meta-learning framework learns-to-recommend few-shot items by knowledge transfer from arbitrary base recommenders.
We show that our framework consistently outperforms state-of-art approaches on overall recommendation (by 5\% Recall) while achieving significant gains (of 60-80\% Recall) for tail items with fewer than 20 interactions.

Next, we explored three inductive learning settings: modeling spread of \textit{user-generated content} in social networks; item recommendations for \textit{ephemeral groups}; and \textit{friend ranking} in large-scale social platforms.
In chapter~\ref{chap:infvae}, we focus on diffusion prediction in social networks where a vast population of users rarely post content.
We introduce a deep generative modeling framework that models users as probability distributions in the latent space with variational priors parameterized by graph neural networks.
Our approach enables massive performance gains (over 150\% recall) for users with sparse activities, while being faster than state-of-the-art neural models by an order of magnitude.
In chapter~\ref{chap:groupim}, we examine item recommendations for ephemeral groups with limited or no historical interactions together.
To overcome group interaction sparsity, we present self-supervised learning strategies that exploit the preference co-variance in observed group memberships for group recommender training. Our framework achieves significant performance gains (over 30\% NDCG) over prior state-of-the-art group recommendation models. 
In chapter~\ref{chap:grafrank}, we introduce multi-modal inference with graph neural networks that captures knowledge from multiple feature modalities and user interactions for multi-faceted friend ranking.
Our approach achieves notable higher performance gains for critical
populations of less-active and low-degree users.

\end{abstract}

\begin{acknowledgments}
There are so many people who contributed to my Ph.D. journey at the University of Illinois, Urbana-Champaign (UIUC).

First and foremost, I would like to thank my advisor, Professor Hari Sundaram, whose guidance and encouragement has been instrumental in shaping this dissertation.
I am ever-grateful to Hari for taking me as a student, his kindness and genuine interest in my mental and physical well-being meant a lot to me during my initial Ph.D. years.
Hari taught me invaluable art of technical writing --  the ability to express sophisticated concepts in a simplified and accessible language without missing out on the essential details.
His guidance has greatly influenced how I envision the big picture and impact of a research problem, and enhanced my ability to relate and connect ideas across different research areas.
He has also been incredibly supportive of my independent research pursuits with other professors and external collaborators.
Over the years, he has been integral in shaping me into the researcher that I have become, and I would be fortunate to work with him again in the future. 

I also want to thank my other thesis committee members, Professor Jiawei Han, Professor Hanghang Tong, and Dr. Neil Shah. 
I would especially like to thank Jiawei for his kind words of encouragement during my uncertain moments.
His courses laid the foundation for my foray into data mining, and my numerous collaborations with him and other students from his research group have contributed immensely to different threads of work in this thesis.
I am thankful to Hanghang for his insightful questions and feedback during my thesis proposal, which were greatly instructive and helpful in the direction of this work.
I am also immensely grateful to Neil Shah for his guidance during my internship at Snap, which motivated me to explore practically impactful implications of my research beyond publishing objectives.
Despite the fully remote internship experience, I had an awesome time working with Neil, and got to know how much fun research in the industry can be.

I would also like to thank Yanhong Wu for his guidance and encouragement during my two summers that I spent at Visa Research. 
Yanhong went above and beyond his mandated responsibilities to mentor me, and spent countless hours outside of his regular work schedule to assist me with additional experiments and paper writing after the end of my internship.

I'm extremely fortunate to have been advised by my undergraduate thesis advisors Professors Sayan Ranu and Karthik Raman at Indian Institute of Technology Madras (IIT-Madras).
Even after I moved to the U.S. for my Ph.D. studies, they actively continued mentoring me to ensure that our work reached its best possible conclusion with a publication in a top bioinformatics journal. 
Without their efforts in instructing and mentoring me at different levels, I could not have possibly had the opportunities that led me to UIUC.

I would also like to acknowledge my colleagues at Crowd Dynamics Lab for their numerous research discussions, constructive feedback, and collaborations that have shaped my work. 
A big shout out to my research collaborators Junting Wang, Adit Krishnan, and Xinyang Zhang, who have closely worked with me on several projects.

Finally, I am forever indebted to my parents, sister, and brother-in-law for their unwavering love and support through the ups and downs of my Ph.D. journey.

\end{acknowledgments}

\tableofcontents

\mainmatter

\chapter{Introduction}
\label{chap:intro}

\section{Introduction}
In recent times, users participate in a multitude of content-creation, social networking and e-commerce platforms, with diverse interactional settings.
User behavior is characterized by their activities and interactions with other users and functionalities within the platform.
For instance, users in social networking platforms (such as Facebook and Snapchat) form friendships where they communicate with friends, post or re-share user-generated content, and interact in diverse groups. In e-commerce platforms (such as Amazon and Walmart), users browse, view, and purchase products and often create product reviews.
With the rapid growth and pervasiveness of such online platforms, learning data-driven models of user behavior is indispensable to enable highly personalized user experiences.

Modeling user behavior in online interactional platforms poses exciting new challenges in handling behavioral data at the scale of over millions of entities (\textit{e.g.}, users, content, groups, etc.).
First, \textit{data sparsity}, despite the enormous potential to learn nuanced behavioral patterns from massive interaction logs, a fundamental challenge is data \textit{sparsity} for a vast majority of entities towards model learning.
Second, \textit{interaction heterogeneity}, users participate in diverse interactional scenarios, including structural \textit{graph} connectivity in social networks, \textit{bipartite} user-item interactions in e-commerce platforms, and \textit{multipartite} user-group-item interactions in group activities.
Sparsity challenges manifest in a variety of different ways, \textit{e.g.}, a substantial fraction of users in social networks are inactive with limited \textit{structural} and \textit{engagement} information~\cite{barabasi};  further, ground-truth \textit{labels} are typically very limited and expensive to obtain in large-scale platforms.
Similarly, we encounter \textit{interaction} sparsity for a large proportion of the item inventory in rapidly expanding e-commerce platforms~\cite{popularity_bias}.
Modeling user behavior in the face of \textit{sparsity} challenges across different interactional scenarios, is critical to personalize experiences for diverse user populations.

Recently, neural representation learning, powered by deep neural network architectures, has emerged as one of the most promising approaches for user modeling, with state-of-the-art performance in several established benchmarks and widespread adoption in various industrial settings, \textit{e.g.}, video recommendations on YouTube~\cite{youtube}, personalized search ranking and listing recommendations at AirBnb~\cite{airbnb}, related pin recommendations on Pinterest~\cite{pinsage, pinnersage} etc. 
Neural networks have the representational capacity to learn sophisticated behavioral patterns from extensive volumes of data without the need for manual feature engineering.
However, developing sparsity-aware neural user modeling frameworks to enable personalized inference at scale is important, challenging, and has remained largely beyond reach.

In this thesis, we develop neural representation learning frameworks to characterize user behavior in online interactional platforms, broadly divided into \textit{transductive} and \textit{inductive} learning scenarios. Transductive learning involves behavior modeling and prediction for entities seen during model training, while inductive learning concerns unseen entities that are only observed during inference.
We design neural user modeling frameworks with two desirable properties, \textit{sparsity-awareness}: we leverage different facets of information reflecting user behavior (\textit{e.g.}, interconnectivity in social networks, temporal and attributed interaction information) to enable personalized behavior inference at scale in the face of interaction sparsity; \textit{generalizability}: we present architecture-agnostic frameworks with generalizable learning strategies; our models are complementary to concurrent advances in neural architectural choices and are adaptive to the rapid addition of new applications in online platforms.

\section{Technical Challenges}
Designing neural user modeling frameworks poses several technical challenges in handling the massive scale of behavioral data and the diversity of interaction types involving over millions of entities.
Despite the potential to learn neural models that harness massive interaction logs, a central theme of this dissertation is addressing \textit{data sparsity} challenges that manifest in different ways across applications, which are briefly summarized below:
\begin{description}
\item \textbf{Ground-truth Label Sparsity}: In large-scale online platforms, we typically have access to very limited ground-truth labels for entity attributes, \textit{e.g.}, age, gender of users in social networks or aspect ratings of products in e-commerce platforms.

\item \textbf{Entity-level Interaction Sparsity}:
Heavy-tailed distributions in user interests and interaction patterns are commonly observed in user behavioral data~\cite{barabasi_tail} across several online platforms.
Despite the availability of behavioral data at massive scales, we have very limited historical records at the granularity of individual entities due to the highly skewed interaction distribution. 
Thus, learning meaningful trends or insights for a vast majority of entities is challenging due to \textit{entity-level interaction sparsity}.

\item \textbf{Feature Diversity and Skew}: In inductive learning applications, entities are often represented using a combination of diverse features, \textit{e.g.}, user features in social networking platforms may include static profile attributes, dynamic communication and engagement activities, while item features in e-commerce platforms may include textual descriptions, product reviews, and other attributes (such as brand or price).
Here, the key modeling challenges are feature heterogeneity and skew: the features often belong to different modalities and exhibit non-trivial correlations; we also observe skewed occurrences of features across different entities.

\end{description}

In this dissertation, we primarily develop architecture-agnostic frameworks for user behavior modeling, with generalizable learning strategies that are targeted towards addressing the aforementioned sparsity challenges.
Our learning frameworks enable personalized inference at scale for broad application scenarios across diverse platforms.
Furthermore, we also present an example of a specific deep learning architecture design that effectively handles sparsity concerns in a large-scale industrial application.

Our problem settings span transductive and inductive learning scenarios, where transductive learning models behavior of entities seen during training and inductive learning targets unseen entities that are only observed during inference.
Thus, our key research contributions are also organized below into \textit{transductive} and \textit{inductive} user behavior modeling applications.

\section{Transductive User Behavior Modeling}
In this part, we explore two fundamental \textit{transductive} interactional scenarios with access to interactions/connectivity information of all participating entities during model training.

First, we consider a \textit{graph-structured} data representation to model interactions between various entities, such as how users connect and communicate with each other in social networks, or how they purchase and rate products in e-commerce platforms.
We examine transductive or semi-supervised learning given structural graph connectivity between the interacting entities and a few manually labeled examples.
This is a natural way to formulate user profiling in online platforms given \textit{sparse} ground-truth labels for a certain attribute (\textit{e.g.}, age, gender), and the goal is to exploit the graph-structured interactions to infer the attribute for the remaining users.
In Chapter~\ref{chap:infomotif}, we leverage \textit{higher-order network structures} to develop a semi-supervised learning framework (\textsc{InfoMotif}) over graphs.

We then investigate the \textit{bipartite} user-item interactional setting, which is the foundation for a wide variety of Collaborative Filtering (CF) methods~\cite{conventional_CF} that personalize item recommendations based on historical user interactions.
A close examination of prior neural recommendation models reveals poor accuracy levels for the vast majority of items (with sparse interactions) in the item inventory~\cite{cikm18adv, popularity_bias}.
In Chapter~\ref{chap:protocf}, we introduce a \textit{few-shot learning} framework (\textsc{ProtoCF}) that extracts and transfers meta-knowledge from data-rich head entities (such as popular items), to enable accurate personalized recommendations for data-poor tail entities (such as long-tail items) with severe interaction sparsity.

\subsection{{Higher-Order Structures in Graph-based Interactions}}
Graph Neural Networks (GNNs) have emerged as a popular paradigm for semi-supervised learning on graphs given a few labeled examples, and have recently enabled substantial advances due to their ability to learn node representations combining graph structure and node/link attributes.
Localized \textit{message passing} between neighboring nodes is the basis of node representation learning in graphs, including the popular GNN models. %
We identify two key imitations due to localized message passing: \textit{over-smoothing} (poor resolution in distinguishing structural node neighborhoods) and \textit{localization} ($k$-layer message-passing is restricted to the $k$-hop neighborhood of the labeled training nodes), which degrade performance of prior GNNs due to limited availability of labeled information.

To address the limitations of localized message passing, we propose the novel concept of \textit{attributed structural roles}, grounded on \textit{network motifs}, to regularize arbitrary GNN models for semi-supervised learning.
Network motifs are a general class of \textit{higher-order structures} (such as dense subgraphs and cliques) that indicate connectivity patterns between nodes, are crucial for understanding the organization and properties of complex networks~\cite{network_motif}.
Our key hypothesis is that leveraging the higher-order connectivity structures between nodes is critical to accurately label the nodes,  effectively compensating the lack of sufficient labeled information in the local neighborhoods.
Attributed structural roles identify structurally similar nodes with co-varying local attributes, independent of network proximity.

In this work, we propose an \textit{architecture-agnostic} framework (\textsc{InfoMotif}) to capture attributed structural roles by dynamically prioritizing multiple motifs in the learning process without relying on distributional assumptions in the underlying graph or the learning task.
Our experiments across a wide variety of assortative and disassortative networks, indicate significant gains (3-10\% accuracy) for  \textsc{InfoMotif} over prior approaches, with stronger gains for nodes with sparse training labels and diverse attributes in local neighborhood structures.

\subsection{Few-shot Collaborative Filtering via Meta-Transfer}
Neural Collaborative Filtering (NCF) methods have recently revolutionized modern recommender systems with impressive gains over conventional collaborative filtering approaches.
However, a closer analysis of prior neural recommendation models reveals skewed performance gains towards popular items with abundant historical interactions, with poor accuracy levels for a significant chunk of \textit{long-tail} (niche) items in the inventory~\cite{cikm18adv, popularity_bias}. Moreover, we empirically show that prior neural recommenders lack the \textit{resolution power} to accurately rank relevant items within the long-tail owing to severe interaction sparsity. This restricts personalization and impedes suppliers of long-tail items in under-represented categories (genres or styles).
Targeting long-tail items can enhance recommendation diversity and bring relatively larger revenues compared to popular items with competitive markets.

In this work, we focus on learning to accurately recommend few-shot items with interaction sparsity. We formulate few-shot item recommendation as a metric-based meta-learning problem of \textit{learning-to-embed} items with few interactions.
Our key insight is \textit{episodic training} to eliminate the interaction distribution inconsistency between head items (with abundant interactions) and tail items (with sparse interactions); we enable few-shot generalization by sub-sampling interactions from head items to mimic tail items during model training.

We propose a meta-learning framework (\textsc{ProtoCF}) that \textit{extracts}, \textit{relates}, and \textit{transfers} the meta-knowledge learned by a neural base recommender over head items to learn robust and discriminative latent representations for tail items.
Our experimental results indicate significant gains (Recall@50 gains of 4-20\%) for \textsc{ProtoCF} on few-shot personalized item recommendations (with over 70\% of the item inventory in the tail) over state-of-art models.

\section{Inductive User Behavior Modeling}
In this part, we consider a few diverse \textit{inductive} interactional scenarios where modeling user behavior entails new entities that are only observed during model inference.

First, we examine \textit{content re-sharing} behavior of users in social micro-blogging platforms (such as Twitter and Tumblr) where user-generated content is often noisy and transient with a short lifespan.
The spread or diffusion of user-generated content in social platforms hinges primarily on the contextual influence of recent re-shares and the extent of social connectivity.
In Chapter~\ref{chap:infvae}, we present a social regularization framework (\textsc{InfVAE}) to carefully model the co-variance of temporal context (recent posting activities) with structural graph connectivity information, for predicting spread of user-generated content in social networks.

Next, we study \textit{multipartite} user interactions (users, groups, items) in \textit{ephemeral group} activities where groups have dynamic memberships and \textit{sparse} item interactions. Group interactions are becoming increasingly popular in a variety of contemporary social platforms such as Meetup, Facebook and Snapchat groups, etc.
In Chapter~\ref{chap:groupim}, we introduce a self-supervised learning framework (\textsc{GroupIM}) to generate personalized item recommendations tailored to ephemeral groups with no previous interactions together.

Finally, we explore \textit{multi-faceted} user interactions in an industrial social networking setting with rich multi-modal individual (such as interests, demographics, etc.) and pairwise attributes (such as communications).
This is common in modern online platforms that routinely track many aspects of user behavior across different functionalities to gain multi-faceted insights.
In Chapter~\ref{chap:grafrank}, we present our inductive learning approach (\textsc{GraFRank}) that incorporates \textit{multi-modal} features to successfully overcome sparsity challenges for \textit{less-active} and \textit{low-degree} users, in the industrially ubiquitous application of friend suggestion in social platforms. We briefly summarize our key technical contributions below.

\subsection{Social Regularization to Predict Information Diffusion}
Modeling the spread or \textit{diffusion} of information through content posting or re-sharing on social media platforms (such as Twitter, Facebook, etc.) has widespread applications, including news feed ranking, limiting misinformation spread, viral marketing, etc.
To predict the set of influenced users who re-share a piece of user-generated content, prior work mainly consider the impact of \textit{temporal} user-user influence learned from historical posting activities; this results in poor predictions for the vast population of passive users who seldom post content.
Incorporating knowledge of \textit{social graph connectivity} structures is crucial to overcome sparsity challenges for users with sparse diffusion interactions.

Social regularization builds on correlation theories such as \textit{homophily} and \textit{influence}.
Social homophily~\cite{homophily} suggests stronger ties between users with shared latent interests, inducing similar behaviors without direct causal influence; in contrast, temporal influence captures direct peer-to-peer effects, resulting in temporally correlated diffusion behaviors~\cite{aral_inf}.
Homophily and Influence are fundamentally \textit{confounded} in observational studies of diffusion processes~\cite{shalizi}, which makes it challenging to contextually model the impact of both factors.

In this work, we propose a \textit{social regularization} framework (\textsc{InfVAE}) to model the co-variance of social homophily (indicated by graph connectivity) and temporal influence (recent posting activities), for predicting diffusion behaviors.
We differentiate diffusion \textit{roles} of users (influential versus susceptible) and adopt \textit{variational} priors to regularize the latent space of user representations via graph autoencoders designed to preserve structural connectivity information.
Our experiments show significant gains (22\% MAP@10) for \textsc{InfVAE} over state-of-the-art models, with massive gains for users with sparse diffusion interactions.

\subsection{Self-supervised Ephemeral Group Interaction Modeling}
With the emergence of social networking platforms such as Meetup and Facebook Event, social group activities have gained popularity; it is essential to provide groups with relevant item recommendations (\textit{e.g.}, restaurant or a concert).
Existing studies target \textit{persistent} groups which are fixed, stable groups (\textit{e.g.}, families watching movies) where the members have interacted with numerous items together as a group.
We examine interactions of \textit{ephemeral} groups casually formed by ad-hoc users with \textit{limited or no} historical interactions together.
Ephemeral group interactions are pervasive in several real-world scenarios, including dining with strangers, traveling in group tours, and attending social events.

To address group interaction \textit{sparsity}, our key technical insight is to exploit the preference covariance among group members and contextually prioritize their individual preferences, to make group recommendations. 
In contrast to prior work that design customized group preference aggregators, we achieve \textit{inductive generalization} to ephemeral groups by designing \textit{self-supervised} learning objectives to \textit{regularize} base neural group recommenders with arbitrary individual preference encoders and group preference aggregators.

In this work, we present a \textit{self-supervised} learning framework (\textsc{GroupIM}) that relies on \textit{mutual information} estimation and maximization over group memberships, to overcome group interaction sparsity for ephemeral group recommendation.
Empirically, \textsc{GroupIM} achieves significant performance gains (31-62\% NDCG@20) over state-of-the-art group recommendation models, and is particularly effective for \textit{large} and \textit{diverse} groups.

\subsection{Multi-Faceted Friend Ranking in Social Platforms}
Graph Representation Learning methods (including GNNs), have recently advanced graph learning in multiple prominent academic applications, such as link prediction, community discovery, and node classification.
Most prior transductive learning approaches directly learn latent representations per node, resulting in \textit{prohibitive model sizes} for large-scale social networks with over multiple millions of users.
It is vital to develop inductive learning techniques for \textit{large-scale social modeling applications} where users interact with different functionalities, communicate with diverse groups, and have multifaceted interaction patterns.

In this work, we consider the industrially ubiquitous  application of \textit{friend suggestions} (recommending new candidate users to befriend) in social platforms. %
Compared to academic datasets, sparsity challenges are also typically exacerbated in industrial settings owing to the significant fraction of inactive users with limited structural and engagement information.
To overcome structural and interactional sparsity, we exploit the rich knowledge of in-platform user actions to formulate friend suggestion as \textit{multi-faceted friend ranking} on an evolving friendship graph, with multi-modal user features and link communication features.

We design a graph neural network architecture (\textsc{GraFRank}) that captures knowledge from multiple correlated feature modalities and user-user interactions to learn multi-faceted user representations for friend ranking.
\textsc{GraFRank} outperforms state-of-the-art baselines on friend candidate retrieval (by 30\% MRR) and ranking (by 20\% MRR) tasks.
Notably, \textsc{GraFRank} achieves higher gains for the critical population of \textit{less-active} and \textit{low-degree} users, and is currently being implemented by Snapchat to enhance their quick-add feature.

\section{Thesis Organization}
This thesis is organized in two parts:  \textit{transductive} and \textit{inductive} user behavior modeling.

In the first part, we model user behavior in \textit{transductive} interactional scenarios where we have access to historical interactions and connectivities among all participating entities during model training, and the goal is to infer their missing properties (\textit{e.g.}, profile attributes) or predict future behaviors (\textit{e.g.}, item purchases) during model inference.

In the second part, we examine \textit{inductive} interactional scenarios with real-time behavior prediction for new entities that are only observed during model inference.
This scenario is common when dealing with user-generated content and new users in social media platforms and ephemeral group interactions with dynamic memberships (whose members may be together for the first time).
Our primary focus here is the design of efficient sparsity-aware models that inductively generalize to new transient entities with limited interaction data. Table~\ref{tab:thesis_overview} provides a brief overview of different chapters in this thesis.

Table~\ref{tab:thesis_organization} summarizes the interaction types of user interactions in each chapter and their associated technical challenges.
In chapter~\ref{chap:infomotif}, we examine semi-supervised learning over graphs with sparse ground-truth training labels.
We handle entity-level interaction sparsity challenges in Chapter~\ref{chap:protocf} (long-tailed items), Chapter~\ref{chap:infvae} (long-tailed users), and Chapter~\ref{chap:groupim} (ephemeral groups). 
In chapter~\ref{chap:grafrank}, we further incorporate multi-faceted features and interactions across different modalities, with skewed distributions across different users.

\begin{table}[t]
\centering
\begin{tabular}{@{}p{0.203\linewidth}|p{0.11\linewidth}|p{0.6\linewidth}}
\toprule
\textbf{Part} & \textbf{Chapter}  & \textbf{Chapter Title} \\
\midrule
\multirow{2}{\linewidth}{Transductive User Modeling (Part I)}  
&  Chapter~\ref{chap:infomotif} & Higher-Order Structures in Graph-based Interactions \\
& Chapter~\ref{chap:protocf} & Few-shot Collaborative Filtering via Meta-Transfer\\
\midrule
\multirow{3}{\linewidth}{Inductive User Modeling (Part II)}
& Chapter~\ref{chap:infvae} & Social Regularization to Predict Information Diffusion\\
& Chapter~\ref{chap:groupim} & Self-supervised Ephemeral Group Interaction Modeling \\
& Chapter~\ref{chap:grafrank} & Multi-Faceted Friend Ranking in Social Platforms \\
\bottomrule
\end{tabular}
\caption{Thesis Organization into transductive and inductive user modeling.}
\label{tab:thesis_overview}
\end{table}

In this thesis, our user modeling applications operate on a variety of interaction types, including bipartite user-item interactions (Chapter~\ref{chap:protocf}), multipartite user-group-item interactions (Chapter~\ref{chap:groupim}), and graph-structured interactions (Chapters~\ref{chap:infomotif},~\ref{chap:infvae}, and~\ref{chap:grafrank}).

In particular, our graph-based user modeling frameworks, although primarily targeted at social networks, generalize to a broad class of graphs that satisfy two key structural properties: scale-free graphs and homophily.
These assumptions directly stem from the inherent nature of human interactions; 
we typically expect skewed distributions to be present in interaction data produced by human activity, resulting in graphs that exhibit a power law distribution (scale-free); in most user-user interactions networks, similar users are expected to form social connections (homophily).

However, we note that general graphs that arise in non-social contexts, (\textit{e.g.,} transportation networks or brain networks) may exhibit different structural properties. In such a scenario, it is critical to revisit the inductive biases in designing graph learning models (\textit{e.g.}, graph embedding methods and graph neural networks).

\begin{table}[h]
\centering
\begin{tabular}{@{}p{0.203\linewidth}|p{0.11\linewidth}|p{0.27\linewidth}|p{0.33\linewidth}}
\toprule
\textbf{Part} & \textbf{Chapter}  & \textbf{Data Type} & \textbf{Technical Challenge} \\
\midrule
\multirow{2}{\linewidth}{Transductive User Modeling (Part I)}
& Chapter~\ref{chap:infomotif}  & Graph Connectivity & Ground-truth Label Sparsity \\
& Chapter~\ref{chap:protocf}  & Bipartite Interactions & Entity Interaction Sparsity\\
\midrule
\multirow{3}{\linewidth}{Inductive User Modeling (Part II)} 
& Chapter~\ref{chap:infvae}  & Graph Connectivity &  Entity Interaction Sparsity \\
& Chapter~\ref{chap:groupim}  & Multipartite Interactions & Entity set Interaction Sparsity \\
& Chapter~\ref{chap:grafrank} & Graph Connectivity &  Entity Interaction Sparsity and Feature Diversity \\
\bottomrule
\end{tabular}
\caption{Comparison of various technical challenges and data characteristics across the different thesis chapters.}
\label{tab:thesis_organization}
\end{table}

\chapter{Literature Review}
\label{chap:literature_review}
In this chapter, we provide a broad overview of prior work in user inference tasks and recommendation methodologies and applications in different interaction contexts.

First, we discuss related work on user modeling approaches centered around addressing sparsity challenges in social network analysis and mining applications.
Then, we discuss prior sparsity-aware approaches research directions in a variety of different recommender systems applications.
Finally, we discuss related work on various learning paradigms, including semi-supervised learning, deep generative modeling, self-supervised learning, few-shot learning, and meta-learning, that tackle sparsity challenges in diverse machine learning domains.

\section{Social Network Mining}
There has been a lot of interest in the past on analyzing, characterizing, and predicting user behavior in online social networking platforms.
In this dissertation, we primarily focus on machine learning techniques for modeling (understanding and predicting) user behavior in social networks, which are based on recent advances in representation learning over graphs. 

We first briefly review prior work on unsupervised and semi-supervised graph representation learning techniques, which are targeted at node-level property inference and link prediction (or completion) applications in static graphs.
Then, we review related work on modeling different kinds of temporally evolving user behaviors in social networks, including friendship creation, inter-user interactions, and content-sharing activities.

\subsection{Static Graph Representation Learning}
Graph embedding methods learn latent node representations in graphs to capture the structural properties of a node and its local neighborhoods.
Broadly, they fall into two categories: the first is unsupervised graph embedding models that learn universal node representations that capture different structural graph properties (\textit{e.g.}, community, roles, etc.); the second is the class of semi-supervised learning models that further exploit supervision (\textit{e.g.}, ground-truth labels for nodes or links) to learn task-driven graph embeddings.

\textit{Unsupervised Graph Representation Learning:}
There are a number of techniques that have been proposed to learn unsupervised latent node embeddings.
The earliest techniques such as Principal component analysis~\cite{graph_pca} and Laplacian Eigenmaps~\cite{laplacian_eigenmap} were based on dimensionality reduction, optimize an objective over an affinity matrix of the entire graph to maximizes the variance of the latent representations.
Inspired by advances in neural language models such as word2vec~\cite{word2vec}, stochastic graph embedding techniques such as Deepwalk~\cite{deepwalk} and node2vec~\cite{node2vec} learn unsupervised embeddings to maximize the likelihood of co-occurrence of nodes in fixed-length random walks. 
A related approach is the design of learning objectives to learn node embeddings that capture different orders of proximity between nodes in a graph~\cite{line, hope, arope}.

As an alternative to proximity-preserving objectives to learn graph embeddings, some methods learn role-aware embeddings that embed structurally similar nodes close in the latent space, independent of network position~\cite{rossi2019community, rolx}; such methods typically utilize structural node features (\textit{e.g.}, node degrees, motif count statistics) to extend classical proximity-preserving embedding methods based on random walks~\cite{struc2vec} and matrix factorization~\cite{hone}.

Neural graph embedding techniques were first introduced in SDNE~\cite{sdne} which utilized graph autoencoders to learn node embeddings that preserve both local and graph measures of graph proximity.
More generally, the class of graph autoencoders~\cite{sdne, vgae, graph_enc_dec} employ various encoding and decoding architectures to embed graph structure and learn node embeddings through a variety of unsupervised training objectives.
A few methods extended the above techniques to also incorporate knowledge of node features (or attributes) in the embedding learning process~\cite{dane, asne, anrl}.

\textit{Semi-supervised Graph Representation Learning:}
Graph-based semi-supervised learning is a well-studied problem, where the goal is to classify nodes in a graph given a small set of labeled examples~\cite{ssl-survey}.
Traditional methods propagate labels through linked nodes in the graph based on different smoothness assumptions~\cite{transductive}.
Collective classification techniques~\cite{citation} generalize label propagation by further utilizing node features and allowing more flexible local updates.
Semi-supervised variants of graph embedding techniques include label information in the learning process through supervised training objectives~\cite{planetoid, lane}.

\textit{Graph Neural Networks:}
Graph Neural Networks are neural networks over graphs designed to learn node-level or graph-level representations towards a variety of learning tasks.
Initial GNNs generalized Convolutional Neural Networks (CNNs) to graphs in the spectral domains; spectral GNNs~\cite{spectral, cheby} defined graph convolutions through the Graph Fourier Transform described by eigenvectors of the Graph Laplacian.
However, spectral methods (being a function of the graph Laplacian) cannot generalize to graphs with different structural properties.

Message-passing GNNs (spatial) generalize label propagation through localized message passing over node neighborhoods; they learn node representations by recursively propagating features (\textit{i.e.}, message passing) from local neighborhoods through the use of aggregation and activation functions~\cite{gcn}.
Graph Convolutional Networks (GCNs)~\cite{gcn} learn degree-weighted neighborhood aggregators, which can be interpreted as a form of Laplacian smoothing.
Many models generalize GCN with a variety of learnable neighborhood aggregators, \textit{e.g.}, self-attentions~\cite{gat}, mean and max pooling functions~\cite{graphsage}; these approaches have consistently outperformed proximity-preserving graph embedding techniques on several benchmarks.

In this dissertation, we identify two key limitations in prior message-passing GNN architectures: k-hop localization and over-smoothing. A few recent methods have attempted to address either of these issues.
To handle localization, non-local GNN variants incorporate information from nodes in different localities via varying influence radii~\cite{jknet}, shortest paths~\cite{pgnn}, and global summaries~\cite{dgi}.
To address over-smoothing, a few methods design structural GCNs via degree-aware~\cite{demonet} and motif-based aggregators~\cite{motifcnn, motif_attention_cikm19}; however, such methods are highly localized.
In chapter~\ref{chap:infomotif}, we design a learning framework over GNNs to learn attributed structural roles based on the co-variance of features in motifs (higher-order connectivity structures), thus simultaneously enhancing the distinguishability of node representations and identifying structurally similar nodes independent of graph proximity.

\textit{Heterogeneous Graphs:} Recent work generalize graph embedding techniques and graph neural networks to heterogeneous graphs containing nodes and edges of different types.
The key idea is to incorporate structural and semantic information indicated by meta-path or meta-graph structures.  A few popular heterogeneous graph embedding approaches capture proximity between nodes connected via meta-paths~\cite{hine}, meta-graphs~\cite{metagraph2vec, aspem}, and meta-path guided random walks~\cite{hin2vec, metapath2vec}.
Heterogeneous graph neural networks conduct message-passing aggregation over local neighborhoods induced by 
specific node types~\cite{hetgnn, hgt}, meta-paths~\cite{graphinception, han, magnn} and meta-graphs~\cite{metagnn}.
Despite advances in modeling rich heterogeneous semantics into message-passing,  the key limitation of localization remains. In our work in chapter~\ref{chap:infomotif}, we also demonstrate the benefits of structure role learning (via typed motif structures) in our framework for heterogeneous graph mining applications.

In addition to reviewing prior graph representation learning models designed for node property inference or link prediction applications in static graphs, we also discuss prior work related to modeling temporally evolving user behaviors in social networks.

\subsection{Temporal User Behavior Modeling}
In modern social platforms (such as Facebook, Twitter, and Snapchat), users participate in a wide variety of temporally evolving interactions, \textit{e.g.}, form friendships and following connections, post or re-share user-generated content, communicate with other users, and interact with a variety of different functionalities within the platform.

In this dissertation, we examine two central applications: modeling temporal friendship formation (or broadly, temporal link evolution) and the spread (or diffusion) of user-generated content in social networks. We now review prior work on these research directions.

\textit{Temporal Link Evolution:}
The earliest methods designed for modeling friendship formation in social networks were motivated by the principles of homophily~\cite{homophily} and triadic closure~\cite{triadic}.
Initial methods include carefully designed heuristics to model user-proximity, \textit{e.g.,} path-based Katz centrality~\cite{katz} or common neighbor-based Adamic/Adar~\cite{adamic_adar} and learning techniques that exploit such pairwise features for link ranking~\cite{social_bpr}.
However, heuristic feature extraction for each potential link is infeasible in large-scale time-evolving social networks.

Recent methods introduced graph representation learning models over dynamic graphs, which are commonly represented as a sequence of evolving graph snapshots at multiple discrete time steps. Such techniques broadly fall into two categories: the first category includes temporal smoothness methods that ensure node embedding stability across consecutive time-steps, guided by triadic closure~\cite{smoothness, dynamictriad} and incremental models that update embeddings from the previous time step~\cite{dyngem}. However, these methods cannot capture long-range variations in graph structure, and are inadequate when nodes exhibit vastly differing evolutionary behaviors.
The second category includes dynamic graph neural networks that model temporal graph evolution; popular approaches include recurrent methods~\cite{dyngraph2vec, evolvegcn} that capture temporal dynamics via hidden states to summarize historical snapshots, and self-attentional models~\cite{dysat} that flexibly weight historical representations.

Despite the recent advances in dynamic GNN architectures, they remain unexplored for large-scale user-user social modeling applications (such as friend suggestion) where users exhibit multifaceted behaviors.
In chapter~\ref{chap:grafrank}, we design dynamic GNNs for the important application of friend suggestion, through a novel multi-faceted friend ranking formulation with multi-modal user features and link communication features.

\textit{Information Diffusion Modeling:}
Prior diffusion models in social networks were designed towards one or more objectives: \textit{microscopic} diffusion prediction, which learns a model from social graph connectivity and diffusion cascade sequences, to predict the next (or final set of) influenced user(s) given a sequence of initial activated seed users; and \textit{macroscopic} cascade prediction, which predicts global properties of information spreading in the social network, including size, growth, and shape of cascades over time~\cite{deepcas}.
In this dissertation, our main focus is modeling individual user behavior; thus, we focus on microscopic diffusion modeling.
  
Historically, modeling spread of user-generated content in social networks has been studied through two seminal diffusion models: Independent Cascade (IC)~\cite{ic} and Linear Threshold (LT)~\cite{lt}; IC models information spreading as cascades of activations over the network, while LT determines user activations according to thresholds on incoming neighbor influence.
The earliest data-driven diffusion models extend IC and LT via probabilistic generative models to incorporate knowledge of topics~\cite{topic-ic}, continuous timestamps~\cite{ctic}, user profiles~\cite{node_attribute}, and community structures~\cite{cic-icdm, cold}. A key limitation of such methods is the reliance on extensive feature engineering and the limited modeling capacity of the chosen probability distributions.

Recently, representation learning models have incorporated user influencing capability and susceptibility in diffusion to learn latent user embeddings from ordered pairwise co-occurrences in diffusion cascades and social graph connectivity structures~\cite{cdk, aaai15, wsdm16, inf2vec}.
To handle time-stamped diffusion cascade sequences, neural diffusion models have generalized sequential models (such as RNNs and LSTMs) to consider DAG propagation structures by  projecting cascades onto local social neighborhoods~\cite{topolstm, cikm18_attention, deepdiffuse}.
As such, prior diffusion models capture the temporal correlation in diffusion behaviors based on activations observed in the cascades, which may be inadequate for the vast majority of users that seldom post content and appear in very few cascades (sparse diffusion interactions).

In our work in chapter~\ref{chap:infvae}, we address the key challenge of diffusion interaction sparsity by utilizing variational priors to regularize latent user representations in diffusion models via variational graph autoencoders designed to preserve structural connectivity information. 
By parameterizing latent space priors through flexible deep generative models, our framework also generalizes classical probabilistic generative modeling approaches.

\section{Recommender Systems}
Recommender systems appear in a variety of e-commerce, advertising, and social networking applications. This has led to several recent advances in developing neural recommendation models designed to handle the massive scale and diversity in user interactions. Our key focus is motivated by the fundamental data sparsity challenge for the vast majority of entities towards model learning (owing to heavy-tailed interaction distributions).

In this dissertation, we examine few-shot recommendation settings targeted at entities with very few interactions. Specifically, we explore sparsity challenges in personalized item recommendations to users (collaborative filtering), and groups of users (group recommendation). Below, we review prior work on collaborative filtering and group recommendation.

\subsection{Collaborative Filtering}
Collaborative Filtering (CF) is one of the most popular techniques
for user modeling in recommender systems. First, we briefly discuss classical CF models, followed by recent advances in neural CF models and techniques that address interaction sparsity challenges.

\textit{Classical Collaborative Filtering:}
Classical CF models operated on the explicit feedback setting by formulating a matrix completion problem to infer the unknown user-item ratings; the key idea towards rating inference is to derive knowledge from those of ``similar” items or ``similar" users.
There are two broad classes of techniques: neighborhood based collaborative filtering~\cite{neighborhood_CF} and latent-factor matrix factorization~\cite{mf} models.
Neighborhood based CF models determine similar items (or users) based on their rating similarity or co-occurrence, and estimate unknown ratings by a weighted average of ratings over the similar items.
Matrix factorization models decompose the user-item rating matrix into user and item specific latent factors, which are later used to infer missing ratings via inner products between the user and item latent factors.
The earliest methods designed for implicit feedback data (commonly observed in massive online interactions), extended matrix factorization models with different pointwise and pairwise ranking objectives that were designed to handle missing data using sample re-weighting~\cite{implicit_mf} or negative sampling~\cite{bpr}.

\textit{Neural Collaborative Filtering:}
Neural CF models generalize matrix factorization to parameterize users and items with vectorized representations and interaction functions that are powered by deep neural networks.
To enhance the embedding layer that generates user and item representations, much effort has been devoted to examine several modeling choices including latent embeddings~\cite{ncf}, denoising~\cite{cdae} and variational autoencoders~\cite{vae-cf} and graph neural networks~\cite{pinsage, ngcf}.
Other recent efforts to improve the interaction function exploit deep learning techniques such as nonlinear neural networks~\cite{ncf}, memory networks~\cite{cmn}, factorization machines~\cite{nfm}, and euclidean distance metrics~\cite{lrml}, to capture non-linear feature interactions between users and items.
While these neural models learn expressive models to significantly outperform conventional CF approaches, sparsity
concerns owing to long-tail items and cold-start users remain a critical challenge.

\textit{Few-shot and Cold-start Recommendation:}
Few-shot and cold-start recommendation are two important and related sub-problems that have received attention in prior research. 
Most neural recommendation models do not learn meaningful latent item (or user) representations in the cold-start and few-shot settings, where new items (or users) provide either no interactions or a handful of interactions, respectively.
We briefly review prior work in both the few-shot (commonly known as long-tail) and cold-start recommendation scenarios.

Long-tail challenges in recommender systems arise due to interaction sparsity for the vast majority of items (and users).  
Clustering is one popular way to address interaction sparsity by exploiting modeling group-level information to account for the lack of entity-level interactions; early methods generate recommendations over tail items at the granularity of item/user clusters, \textit{e.g.}, cluster-based smoothing~\cite{clustering}, user-item co-clustering~\cite{cccf} and joint clustering and collaborative filtering~\cite{dbrec}.
Other efforts leverage item-item co-occurrence statistics to regularize recommenders, \textit{e.g.}, joint factorization of an item co-occurrence matrix that shares latent factors with a CF model.
In a few neural recommenders, variational auto-encoders (VAEs)~\cite{vae-cf, enhanced_vae} employ distributional regularization via Gaussian priors on the latent space, and have effectively alleviated sparsity challenges.
Another recent strategy to alleviate sparsity is data augmentation for items (or users) in the tail via adversarial regularization~\cite{cikm18adv} or rating generation~\cite{rating_gan} techniques. However, adversarial learning is often computationally expensive and does not scale to massive online platforms.
While the few-shot learning problem also appears in other prediction problems with few-shot classes~\cite{matching_nets, protonet} (given a handful of samples), the massive scale of interaction data (note that each user/item is a few-shot instance, unlike few-shot classes) renders these solutions inapplicable.

In chapter~\ref{chap:protocf}, we formulate long-tail item recommendations in the most general user-item interaction setting, as a few-shot learning problem of learning-to-recommend few-shot items with very few interactions~\cite{protocf}. Our proposed meta-learning framework is complementary to advances in neural recommender architectures and enables flexible adaptation to the tail.

Cold-start recommendation is a related problem that specifically targets new users or new items with \textit{no} historical interactions.
In such a scenario, the key idea is to rely on auxiliary side information, \textit{e.g.}, item content~\cite{cvae}, textual reviews~\cite{deepconn}, contextual factors~\cite{cross-domain}, social connections~\cite{social_rec}, and external knowledge graphs~\cite{kgat}.
A wide variety of techniques have been proposed for each of the sub-problems, \textit{i.e.,} content-based collaborative filtering, review-based recommendation, context-aware recommendation, social recommendation, and knowledge-aware recommendation respectively.
Generalizable techniques include randomized feature dropouts to enable generalization to missing inputs~\cite{dropoutnet, randomized}, and meta-learning frameworks for zero-shot (cold-start) recommendation.
Meta-learning methods for cold-start recommendation utilize gradient-based parameter adaptation~\cite{melu, ml2e, MetaHIN, mamo_kdd, dual_tales}, hyper-parameter initialization~\cite{s2meta}, and shared layers with user-specific (or item) parameter adaptation~\cite{meta_rec, wei2020fast, adaptive}.
In this dissertation, we focus on the few-shot setting without requiring access to any side information to ensure generalizability; a deeper exploration of different cold-start recommendation scenarios is an important future direction.

\subsection{Group Recommendation}
Group interactions have become increasingly prevalent in a variety of contemporary social networking platforms such as Meetup, Facebook, and Snapchat. 
Prior literature on developing recommender systems to suggest items relevant to groups of users, can be broadly divided into two categories based on group types: persistent and ephemeral.
Persistent groups have stable members with rich activity history together, while ephemeral groups comprise users who interact with very few items together and may appear only during inference time.

\textit{Persistent Group Recommendation:} Since persistent groups have stable groups with rich historical interactions together, previous studies treat groups as virtual users, and adopt conventional personalized recommendation techniques to make group recommendations.
A few key approaches include genetic algorithm-based group interaction modeling~\cite{genetic_group}, power balance between group members towards content selection~\cite{group_balance}, and deep neural architectures modeling individual choices and group decisions~\cite{dlgr}.
applied. However, such methods cannot handle new groups that are only observed during inference time.

\textit{Ephemeral Group Recommendation:}  Prior efforts focus on investigating different strategies for aggregating individual preferences and modeling group interactions.
To generate ephemeral group recommendations, they either aggregate recommendation results (or item scores) for each member, or aggregate individual member preferences, thus falling into two classes: score (or late) aggregation~\cite{least_misery} and preference (or early) aggregation~\cite{com}.

The score (or late) aggregation strategies first generate recommendation results for each group member, and then generate group recommendations by aggregating these individual results based on static predefined aggregation strategies.
Popular \textit{score aggregation} strategies include least misery~\cite{least_misery}, average~\cite{average}, maximum satisfaction~\cite{max_satisfaction}, and relevance and disagreement~\cite{rd}, inspired by social choice theories~\cite{social_choice}.
An empirical comparison of different heuristic strategies~\cite{least_misery} has demonstrated the absence of a clear winner, especially with the variance in group sizes and coherence levels.
However, the hand-crafted heuristics are static modeling hypotheses that overlook real-world group interactions.

The preference (or early) aggregation strategies first aggregate the preferences of group members, which are then utilized to produce group recommendations.
The first approaches in this category fuse the profiles (raw item histories) of members into a group profile and utilize conventional recommenders to generate group recommendations~\cite{merging}.
Recent approaches introduced latent-factor models based on probabilistic generative models and deep neural networks.
Probabilistic methods~\cite{com, crowdrec} model the group generative process by considering both the personal preferences and relative influence of group members, to differentiate their contributions towards group decisions.
Neural methods explore attention mechanisms~\cite{attention} to learn data-driven preference aggregators~\cite{agree, agr, wang2019group} jointly from individual and group interactions; yet, these models severely suffer from the interaction sparsity challenges for ephemeral groups, which often results in degenerate solutions.

In chapter~\ref{chap:groupim}, we address interaction sparsity for ephemeral group recommendation by designing self-supervised learning strategies based on mutual information estimation and maximization, to regularize arbitrary neural base group recommenders. In the following section, we review machine learning paradigms that are relevant to the proposed user behavior modeling approaches in this dissertation.

\section{Relevant Machine Learning Paradigms}
In this dissertation, we address different sparsity challenges that manifest in online user interactions. Our proposed frameworks and models have deeper connections to broader machine learning paradigms and their applications in other domains, including computer vision and natural language. 
We briefly review relevant work in semi-supervised learning, deep generative modeling, self-supervised learning, and few-shot and meta-learning paradigms.

\subsection{Semi-supervised Learning} 
To account for lack of labeled data in supervised learning tasks, semi-supervised learning leverages large amounts of unlabeled data to learn additional structure about the input distribution, with the goal of enhancing model performance on the learning task.
Different modeling assumptions have been examined in prior work, including label smoothness, low-density (or clustering), and manifold regularization~\cite{ssl-survey, ssl-survey-2}.
One popular approach is transductive learning that constructs a large graph connecting neighboring objects (similar data points are connected by an edge), and propagates labels from the initial labeled set to the unlabeled objects (following label smoothness assumptions).
More recently, graph neural networks~\cite{gcn} generalize classical label propagation techniques through trainable neighbor aggregation over feature-rich local node neighborhoods in attributed graphs.

Graph-based semi-supervised learning has enormous potential for social interaction modeling tasks where users are already connected and typically satisfy homophily assumptions along several attributes.
In this dissertation, we have designed novel graph neural network frameworks for semi-supervised learning and ranking over graphs in chapters~\ref{chap:infomotif} and~\ref{chap:grafrank}.

\subsection{Deep Generative Modeling}
Deep Generative Models are a powerful class of neural networks that aim to approximate any kind of data distribution.
Variational Autoencoders (VAEs)~\cite{vae} and Generative Adversarial Networks (GANs)~\cite{gan} are the most popular frameworks for deep generative modeling, and have achieved tremendous success in several computer vision and natural language processing applications, including image and video generation~\cite{vqvae2, MoCoGAN}, editing~\cite{inpainting}, and enhancement~\cite{gan_enhance}, and text synthesis~\cite{RelGAN} and summarization~\cite{gan_summarize}.
Deep generative models have recently been examined in graph mining and recommender systems applications
to address sparsity challenges via data augmentation~\cite{rating_gan}, enhanced hard negative mining~\cite{irgan}, and distributional priors on the latent space~\cite{vae-cf}.

Flexible distribution-aware and data-driven prior regularization enables better characterization of long-tail and cold-start entities. In chapter~\ref{chap:infvae}, we explored deep generative models for robust behavior modeling in social networks with sparse user interactions.

\subsection{Self-supervised Learning}
To further alleviate sparsity challenges, the recently emerging paradigm of self-supervised learning defines auxiliary (pretext) tasks which are formulated using only unlabeled data; solving the pretext task learns effective latent representations that can benefit a variety of different learning tasks.
The most effective framework is contrastive learning, which aims to maximizes the agreement between different views of the data, by contrasting against negative samples.
Contrastive learning techniques have benefited a broad range of computer vision tasks by formulating pretext tasks directly from local-global contrast~\cite{dim} or by rotating, cropping, or colorizing images~\cite{cpc, simclr}.
Self-supervised learning has also been effectively utilized in recent natural language representation learning models by carefully designing pretext texts, \textit{e.g.}, masked word prediction and next sentence prediction in BERT~\cite{bert}, sentence order prediction in ALBERT~\cite{albert}, sentence permutation in BART~\cite{bart} etc.

Despite promising recent advances in self-supervised learning techniques across computer vision and natural language processing applications, this paradigm is relatively unexplored in the graph mining and recommendation domains.
In chapter~\ref{chap:infomotif}, we explore self-supervised learning strategies to train graph neural networks by exploiting higher-order structures (via network motifs) to formulate pretext tasks.
In chapter~\ref{chap:groupim}, we presented a self-supervised learning framework that utilized group membership structure learning to overcome group interaction sparsity for ephemeral group recommendation.

\subsection{Few-shot Learning and Meta-Learning}
Few-shot learning is the paradigm of designing models capable of learning new tasks (\textit{e.g.}, classification tasks) rapidly given a limited number of training examples; such techniques are useful when training examples are hard to find, or where the cost of data labeling is high~\cite{fsl_survey}.
Recently, meta-learning (learning to learn) framework has emerged as an effective approach for few-shot learning. 
In the meta-learning framework, we learn how to learn (\textit{e.g.}, to classify) given a set of training tasks with the expectation of generalization to new tasks that have never been encountered during training time.
The adaptation process is a mini learning session that occurs during inference time with limited exposure to the new task configurations.
There are three common prior meta-learning modeling approaches: (a) learning an efficient distance metric (metric-based)~\cite{protonet}; (b) deep neural network (\textit{e.g.}, recurrent) with external or internal memory (model-based)~\cite{fsl_opt}; (c) optimize the model parameters explicitly for fast learning (optimization-based)~\cite{maml}.

Designing meta-learning frameworks to handle interaction sparsity in recommendation settings poses unique scaling challenges. In chapter~\ref{chap:protocf}, we introduce metric-based meta-learning frameworks for few-shot item recommendations.

\chapter{\textsc{InfoMotif}: Higher-Order Structures in Graph-based Interactions}
\label{chap:infomotif}

\section{Introduction}
In this chapter, we propose a new class of motif-regularized graph neural networks (GNNs); GNNs have emerged as a popular paradigm for semi-supervised learning on graphs due to their ability to learn representations combining topology and attributes, without relying on expensive feature engineering.
GNNs are typically formulated as a message passing framework~\cite{gnn_review},
where the representation of a node is computed by a GNN layer aggregating features from its graph neighbors via learnable aggregators.
Long-range dependencies are captured by using $k$ layers to incorporate features from $k$-hop neighborhoods.
GNNs have demonstrated promising results in several application domains spanning \textit{homogeneous} graphs (\textit{e.g.}, user-user friendship networks) comprising nodes and edges of a single type, and \textit{heterogeneous} graphs (\textit{e.g.}, academic citation networks) containing nodes and edges of different types. 
In particular, GNNs have achieved state-of-the-art results in several applications including node classification~\cite{gcn}, link prediction~\cite{gnn_linkpred}, and personalized recommendation~\cite{pinsage}.

\begin{figure}[htbp]
    \centering
    \includegraphics[width=0.9\linewidth, trim={0in 6.1in 3.5in 0in}, clip]{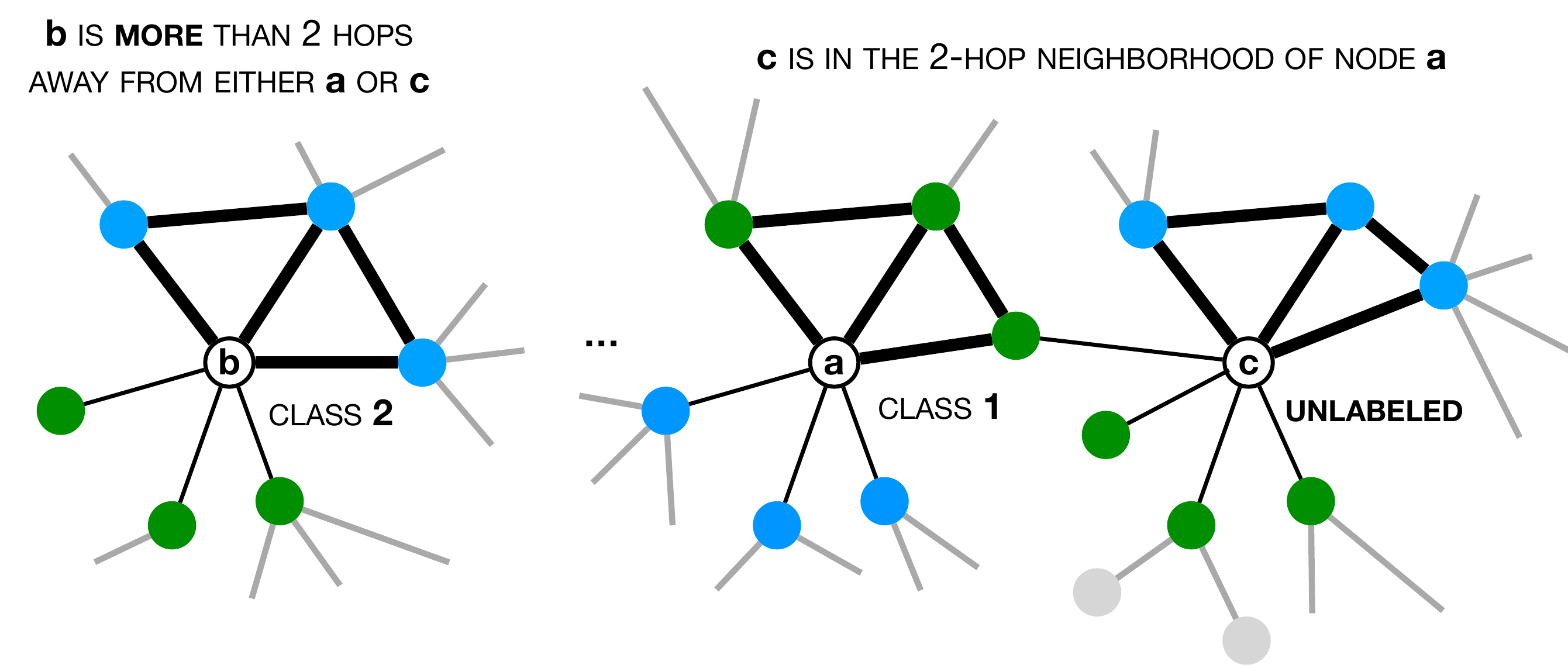}
    \caption{\textbf{Localized message passing limitations: } A stylized example in a homogeneous graph with a 2-layer GNN model (node colors indicate node attributes). Node $a$ is in the 2-hop range of node $c$. Node $c$ \textbf{does not receive} gradient updates from node $b$ (class 2) since node $b$ is more than 2 hops away. A 2-layer GNN will likely label node $c$ as class 1. Notice that $c$ is in class 2 since $c$ and $b$ have identical local structure and attribute co-variation.
    }
    \label{fig:example}
\end{figure}

\textbf{Localized message passing limitations: } 
We illustrate two key limitations of prior $k$-layer GNN architectures: \textit{$k$-hop localized} and \textit{over-smoothed} representations (\Cref{fig:example}).

\begin{enumerate}
    \item GNNs, while highly expressive, are inherently \textit{localized}: a $k$-layer GNN cannot utilize features of nodes that lie outside the $k$-hop neighborhood of the labeled training nodes.  In~\Cref{fig:example}, nodes $a$ and $b$ belong to different classes.
          A 2-layer GNN
          sees unlabeled node $c$ within the aggregation range of $a$ (class 1) and outside the influence of $b$ (class 2 and more than 2 hops away). Thus,
          a GNN will more likely label $c$ with class 1 (than class 2).
          However, in reality, $c$ and $b$ display identical attributes (node color) in the local structure; a localized GNN fails to incorporate this factor.
    \item
          GNNs with multiple layers learn \textit{over-smoothed} node representations by iteratively aggregating neighbor features
          ~\cite{gcn_oversmoothing}.
          In~\Cref{fig:example}, nodes $c$ and $a$ share the same number of neighbors with blue and green attributes; however, green neighbors of node $a$ form triangles, while blue neighbors of node $b$ (and $c$) form triangles.
          Considering \textit{local nodal attribute arrangements}, node $c$ is more similar to $b$ than to $a$.
          The over-smoothing effect in GNNs obscures this attribute co-variation difference when classifying node $c$.
\end{enumerate}

Thus, we require a new learning framework over graphs, to overcome the above key limitations due to localized message passing in popular GNN models.

One way to overcome these limitations is the paradigm of \textit{role discovery}~\cite{role_discovery} that identifies nodes with structurally similar neighborhoods.
In contrast to the notion of communities defined by network proximity, structural roles characterize nodes by their local connectivity and subgraph patterns independent of their location in the network~\cite{rossi2019community}; thus, two nodes with similar roles may lie in different parts of the graph.
Prior role-aware models learn similar representations for structurally similar nodes while ignoring nodal attributes~\cite{struc2vec}, \textit{i.e.}, they will assign the same role to nodes $a$ and $b$ in~\Cref{fig:example} with topologically identical local structures; however, nodes $a$ and $b$ differ in their local attribute arrangements (blue vs. green attributes in triangles), and thus belong to different classes.
Furthermore, structural role learning is relatively unexplored in heterogeneous graphs with typed nodes and edges.

\textbf{Present Work: } To enable the expressivity to distinguish attributed structures, we propose the concept of \textit{attributed structural roles} that identify structurally similar nodes with co-varying attributes, independent of network proximity.
We ground structural roles on \textit{network motifs}\footnote{The terms network motif, graphlet, and induced subgraph are used interchangeably in literature}, which are induced subgraph structures over a few nodes (\textit{e.g.}, triangles).
Networks motifs are a broad class of \textit{higher-order structures}  that indicate connectivity patterns between nodes, and are crucial for understanding the organization and properties of complex networks~\cite{network_motif}.
In addition, network motifs can be easily generalized to capture type semantics in rich heterogeneous graphs through heterogeneous (typed) higher-order structures~\cite{heterogeneous_motif} (also known as metagraphs).
Leveraging higher-order connectivity structures between nodes is extremely valuable to overcome the lack of sufficient training labels in local neighborhoods during semi-supervised learning.
We define two nodes as sharing attributed structural roles if they participate in topologically similar motif instances over co-varying sets of attributes. We note that attribute co-variance permits for multiple discrete and continuous attributes, rather than stricter notions such as regular equivalence~\cite{rossi2019community}.

We propose~\infomotif, a GNN architecture-agnostic regularization framework that exploits the co-variance of attributes and motif structures.
~\infomotif~learns regularizers based on a set of network motifs, which vary in their task-specific significance.
Specifically, across instances of the same motif (\textit{e.g.}, a triangle structure), we learn discriminative attribute correlations to regularize the underlying GNN node representations;  
this encourages the GNN to learn statistical correspondences between distant nodes that participate in similarly attributed instances of that motif.
We propose a novel training curriculum to integrate multiple motif regularizers while attending to motif types and  skewed motif distributions. We summarize our key contributions below:

\begin{description}
    \item \textbf{Attributed Structural Role Learning}:
          We propose the novel concept of \textit{attributed structural roles} to regularize GNN models for semi-supervised learning over graphs. In contrast to prior role discovery work that identify structurally similar nodes agnostic to attributes~\cite{struc2vec}, we adopt the paradigm of \textit{self-supervised learning} to regularize node representations to capture \textit{attribute correlations} in motif structures. Our framework~\infomotif~unifies the expressive local neighborhood aggregation power of message-passing GNNs with the paradigm of structural role discovery.
    \item \textbf{Architecture-agnostic Regularization Framework:}
              To the best of our knowledge,~\infomotif~is the first to address the limitations of localized message passing in GNNs through an architecture-agnostic framework.
          Unlike prior attempts that design new aggregators~\cite{demonet,jknet},  we achieve architecture independence by modulating the node representations learned by the base GNN through motif-based \textit{mutual information maximization}, to capture attributed structural roles. We regularize three state-of-the-art GNNs within our framework to demonstrate significant performance gains.
    \item \textbf{Distribution-agnostic Multi-Motif Curriculum}:
          We propose two learning progress indicators, \textit{task-driven utility} and \textit{distributional novelty}, to integrate multiple motif regularizers within our~\infomotif framework.
          Unlike prior strategies~\cite{transductive, dualgcn} that incorporate regularizers via tunable hyper-parameters, our training curriculum dynamically prioritizes different motifs in the learning process without relying on any distributional assumptions on the underlying graph or on the learning task.
\end{description}

We regularize three state-of-the-art GNN models in our~\infomotif~framework for semi-supervised node classification.
Our experiments are conducted on a wide variety of real-world datasets spanning homogeneous and heterogeneous networks.
In \textit{homogeneous} graphs, our experimental results indicate significant gains (3-10\% accuracy) for~\infomotif~over prior approaches on two diverse classes of datasets: assortative \textit{citation} networks that exhibit strong homophily and dis-assortative \textit{air-traffic} networks that depend on structural roles.
We also demonstrate the utility of our framework in three \textit{heterogeneous} graph datasets where~\infomotif~outperforms a number of state-of-the-art methods with significant performance gains (5\% accuracy) on average.
Our qualitative analysis %
indicates stronger gains for nodes with \textit{sparse training labels} and \textit{diverse attributes} in local neighborhood structures.

We organize the rest of the chapter as follows. In Section~\ref{sec:infomotif_formulation}, we
present the problem formulation, and introduce preliminaries on graph neural networks and network motifs.
We describe our proposed framework~\infomotif~in Sections~\ref{sec:infomotif_methods} and~\ref{sec:infomotif_model_details}, present
experimental results in Section~\ref{sec:infomotif_experiments}, discuss limitations in Section~\ref{sec:infomotif_discussion}, finally concluding in Section~\ref{sec:infomotif_conclusion}.

\section{Preliminaries}
\label{sec:infomotif_formulation}
In this section, we formalize semi-supervised node classification on graphs via graph neural networks and introduce network motifs in homogeneous and heterogeneous graphs.

\subsection{{Problem Definition}}
\label{sec:infomotif_prob_defn}
Let $\mathcal{G} = (\mathcal{V}, \mathcal{E})$ be an attributed graph, with nodes $\mathcal{V}$ and edges $\mathcal{E} \in \mathcal{V} \times \mathcal{V}$. Note, $\mathcal{V} = \mathcal{V}_L \cup \mathcal{V}_U$, the sets of labeled ($\mathcal{V}_L$) and unlabeled ($\mathcal{V}_U$) nodes in the graph.
Let $\mathcal{N}(v)$ denote the neighbor set of node $v \in \mathcal{V}$ in $\mathcal{G}$, and $\mathbf{X} \in \mathbb{R}^{|\mathcal{V}| \times F}$ denotes the attribute matrix with rows $\mathbf{x}_v \in \mathbb{R}^F$ for node $v \in \mathcal{V}$.
In our work, the graph may be \textit{heterogeneous} with multiple types of nodes and edges; in such a scenario, 
we have a node type mapping $\psi: \gV \mapsto \gT_V$ where $\gT_V$ is the set of $T_V$ node types that identifies each node in $\gV$ with a type in $\gT_V$, and a corresponding edge type mapping $\xi: \gE \mapsto \gT_E$ where $\gT_E$ is the set of $T_E$ edge types.
Each labeled node $v \in \mathcal{V}_L$ belongs to one of $C$ classes, encoded by a one-hot vector $\mathbf{y}_v \in \mathbb{B}^C$ ($\mathbb{B} = \{0,1\}$). Our goal is to predict the labels %
of the unlabeled nodes $v \in \mathcal{V}_U$. 
This is the familiar transductive or semi-supervised learning setup for node classification in large-scale graphs~\cite{transductive}.

\subsection{{Graph Neural Networks}}
\label{sec:infomotif_base_gnn}
Graph Neural Networks (GNNs) use multiple \textit{message-passing} layers to learn node representations. At each layer $l>0$, where $0$ is the input layer, GNNs compute a representation for node $v$ by aggregating features from its local neighborhood, through a learnable aggregator function $f_{\theta, l}$ per layer. Using $k$ layers allows for the $k$-hop neighborhood of a node to influence its representation.  
Let $\mathbf{h}_{v,l-1} \in \mathbb{R}^{D}$ denote the representation of node $v$ in layer $l-1$. The $l$-th layer of the GNN follows a message passing rule given by:

\begin{equation}
\mathbf{h}_{v,l} = f_{\theta,l} \Big( \mathbf{h}_{v, l-1}, \{  \mathbf{h}_{u, l-1} \} \Big), \quad u \in \mathcal{N}_v
\label{eq:basic_agg}
\end{equation}	

~\Cref{eq:basic_agg} says that the node embedding $\mathbf{h}_{v,l} \in \mathbb{R}^{D}$ for node $v$ at the $l$-th layer is a non-linear aggregation $f_{\theta, l}$ of the embeddings from layer $l-1$ of node $v$ and the embeddings of immediate network neighbors $u \in \mathcal{N}(v)$ of node $v$. The function $f_{\theta, l}$ defines the message passing mechanism at layer $l$
and we can use a variety of aggregator architectures, including graph convolution~\cite{gcn}, graph attention~\cite{gat}, and pooling~\cite{graphsage} to instantiate $f_{\theta, l}$. The node representation for $v$ at the input layer is denoted by $\mathbf{h}_{v, 0}$  (\textit{i.e.}, $l=0$), where $\mathbf{h}_{v, 0} = \rvx_v$ and $\rvx_v \in \mathbb{R}^{F}$.
We designate the representation of node $v$ at the final GNN layer $\rvh_v \in \sR^{D}$, as its \textbf{base GNN representation}. %
In this work, we use GNNs as a collective term for neural networks that operate over graphs using localized message passing~\cite{gilmer2017neural}, as opposed to spectral methods~\cite{spectral} that learn convolutional filters from the entire graph.

\subsection{{Network Motifs}}
\label{sec:infomotif_network_motif}
\textit{Network motifs} are a general class of higher-order connectivity patterns, with a history of use in network science~\cite{network_motif,temporal_motif}. A motif has several topologically equivalent appearances in the network called \textit{motif instances}. Prior work~\cite{graphlet_counts, subgraph_counting_survey} shows how to efficiently compute motif instances for large graphs.

\begin{figure}[t]
    \centering
    \includegraphics[width=0.85\linewidth]{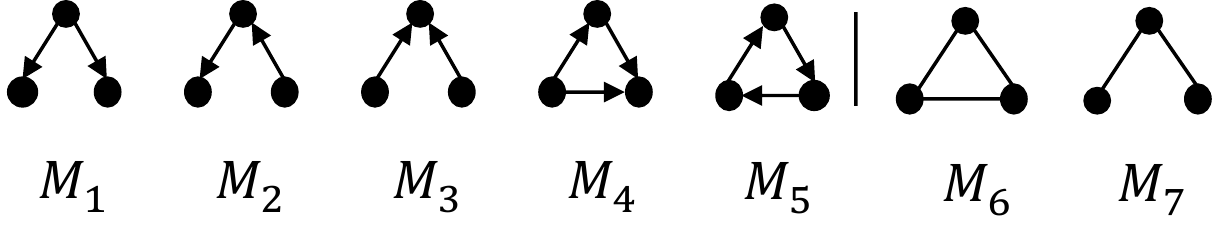}
    \caption{Topologically distinct, directed ($M_1$ to $M_5$) and undirected ($M_6$ to $M_7$) connected, network motifs over three nodes.}
    \label{fig:motifs}
\end{figure}

\begin{definition}[Network Motif]
A network motif $M_t = (\gV_t, \gE_t)$ is a connected, induced subgraph consisting of a subset $\mathcal{V}_t \subset \mathcal{V}$ and $\mathcal{E}_t = \{ e \in \mathcal{E} \mid e = (u, v),  u, v \in \mathcal{V}_t \}$. Let $k_t$ denote the number of nodes in network motif $M_t$; that is, $k_t=|\mathcal{V}_t|$. 
\end{definition}

Here, we consider 3-node connected network motifs, \textit{e.g.},~\Cref{fig:motifs} shows all 3-node, topologically distinct, directed (\textit{e.g.}, citations) and undirected, connected network motifs.

In a heterogeneous graph, nodes/edges are of many different types which makes it essential to explicitly (and jointly) model the connectivity patterns 
and the participating types.
We define \textit{typed} network motifs that generalize network motifs through additional constraints on the types of participating nodes, which are formally described below as:

\begin{definition}[Typed Network Motif]
A typed network motif $M_t = (\gV_t, \gE_t, \psi_t, \xi_t)$ is a connected, induced subgraph consisting of a subset $\mathcal{V}_t \subset \mathcal{V}$ and $\mathcal{E}_t = \{ e \in \mathcal{E} \mid e = (u, v),  u, v \in \mathcal{V}_t \}$ such that 
node and edge type mappings $\psi_t=\psi |_{\gE_t}$ and $\xi_t |_{\gV_t}$ are restrictions of $\psi$ and $\xi$ to $\gV_t$ and $\gE_t$ respectively, and $k_t = |\mathcal{V}_t|$ is the number of nodes in $M_t$.
\end{definition}

We assume that the given graph $\gG$ has a set of unique network motifs $\mathcal{M} = \{ M_1, \dots, M_T\}$.

\begin{figure}[t]
\subfigure[Schema of DBLP]{
        \centering
        \includegraphics[width=0.23\linewidth]{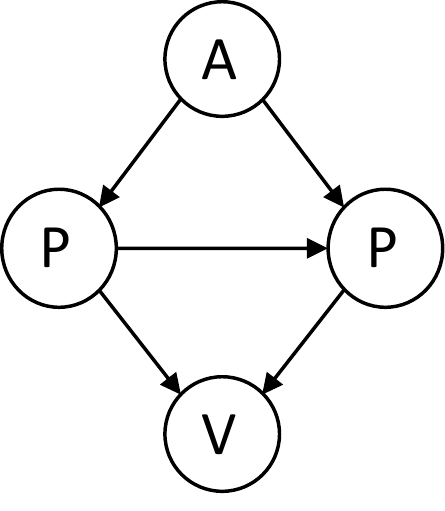}
        \label{fig:dblp_schema}}
        \hspace{0.02\linewidth}
\subfigure [Examples of typed 3-node motifs in DBLP]{
        \centering
        \includegraphics[width=0.63\linewidth]{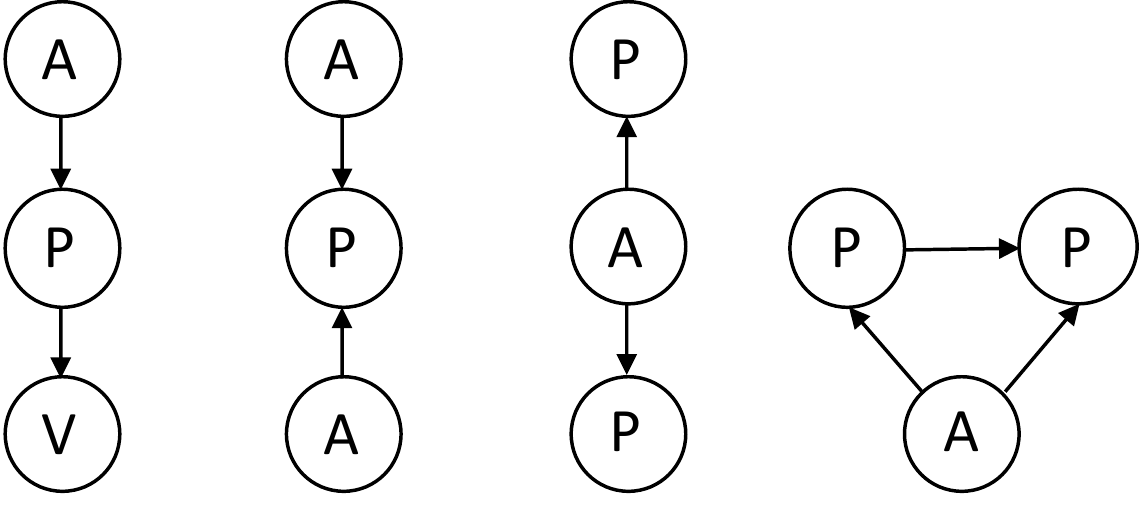}
        \label{fig:dblp_motifs}}
        \caption{ (a) Heterogeneous network schema of academic citation network DBLP with three node types: Author (A), Paper (P) and Venue (V) and three edge types $A-P, P-V$ and $P-A$.  (b) Examples of 3-node connected typed network motifs.}         
\end{figure}

\begin{definition}[Motif Instance] Let $I_t$ be an induced subgraph of $\gG$. We define $I_t$ to be a motif instance of $M_t$ if $I_t$ is isomorphic to $M_t$. A motif $M_t$ can have several motif instances in $\gG$. While each such motif instance has a unique node set, two motif instances can share nodes. We denote the set of unique instances of $M_t$ in $\mathcal{G}$ that contain node $v$ as $\mathcal{I}_v (M_t)$.
\end{definition}

\subsection{{Model Regularization}}
\label{sub:Regularization}
We plan to use these local structural properties (\textit{i.e.}, network motifs) to \textit{regularize} the graph neural network model during training. Typically, we train GNNs by minimizing the cross-entropy loss $L_B$, between model predictions $\mathbf{\hat{y}}_v \in \mathbb{R}^{C}$ and ground-truth labels $\mathbf{y}_v \in \mathbb{B}^{C}$ of labeled training nodes in $v \in \mathcal{V}_L$, which is defined below:
\begin{equation}
    L_B = - \sum\limits_{v \in \mathcal{V}_L} \sum\limits_{c=1}^C y_{v, c} \log \hat{y}_{v, c}
    \label{eqn:infomotif_base_supervised_loss}
\end{equation}
where the $c$-th index of the one-hot vector $\hat{y}_{v,c}$ is the probability that $v$ belongs to the true class $c$. 
Notice that the loss $L_B$ is agnostic to any local structural properties (\textit{e.g.}, mixing patterns in social networks~\cite{multiscale}) that may be indicative of the true node class. Thus, we develop a modified loss $L'_B = L_B + \lambda L_R$, where $L_R$ is the regularization loss that incorporates attributed motif structure and $\lambda$ is a constant. %
Our goal is to design $L_R$ to overcome the two limitations of message-passing models: localized and over-smoothed node representations.

\begin{table}[t]
    \centering
    \begin{tabular}{@{}rl@{}}
        \toprule
        Symbol                & Description                                                               \\
        \midrule
        $\mathcal{M}$         & Set $\{M_1, \dots, M_T\}$ of $T$ network motifs                           \\
        $\mathcal{I}_v (M_t)$ & Set of instances of motif $M_t$ in $\mathcal{G}$ that contain node $v$    \\
        $\mathbf{h}_{v, l}$   & Representation of node $v$ at layer $l$ of GNN                            \\
        $\mathbf{h}_v$        & Base GNN representation of node $v$ (final layer)                         \\
        $\mathbf{h}^{t}_v$    & Motif-gated representation of node $v$ for motif $M_t$                    \\
        $\mathbf{e}_{v, I_t}$ & Instance-specific representation of $v$ in $I_t \in \mathcal{I}_v (M_t) $ \\
        $\mathbf{s}_{v, t}$   & Motif-level representation of node $v$ for motif $M_t$                    \\
        $\mathbf{z}_v$        & Final Representation of node $v$                                          \\
        $\alpha_{vt}$         & Task-specific importance of motif $M_t$ to node $v$                       \\
        $\beta_v$             & Novelty score for training node $v \in \gV_L$                             \\
        \bottomrule
    \end{tabular}
    \caption{Notation}
    \label{tab:infomotif_notations}
\end{table}
\section{InfoMotif Framework}
\label{sec:infomotif_methods}
In this section, we first discuss the structural properties of GNNs to motivate the notion of attributed structural roles. In section~\ref{sec:infomotif_single_motif}, we present our motif-based self-supervised learning framework~\infomotif~to regularize GNNs based on a single motif. Finally, in section~\ref{sec:infomotif_multi_motif}, we introduce our overall framework with a novel multi-motif training curriculum.

\subsection{{Motivating Insights: Attributed Structural Roles}}
\label{sec:infomotif_role_learning}
A $k$-layer GNN computes a localized representation $\mathbf{h}_{v, k}$ for each node $v$ that incorporates information from its $k$-hop neighborhood, denoted by $\mathcal{N}_k (v)$. %
For a node set $S \subseteq \gV$, let $\mathcal{N}_k(S) = \bigcup_{v \in S} \gN_k (v) $ define its $k$-hop neighborhood, and $\mX (S)$ denote its set of input node features.
Let $\mY (\gV_L)$ comprise the training labels of nodes in the labeled set $\gV_L$.
For a $k$-layer GNN trained on 
$\gV_L$ using loss $L_B$ (\Cref{eqn:infomotif_base_supervised_loss}), let 
$\Theta^* = \{ \Theta_{1}, \dots, \Theta_{k} \}$ be the optimal parameters computed by its training algorithm.
Now, we have the following proposition.
\begin{prop}
The optimal parameter set $\Theta^{*}$ is a function of $\mX (\mathcal{N}_k(\gV_L))$ and $\mY (\gV_L)$; further, any changes in inputs $\mX (\gV \setminus \mathcal{N}_k(\gV_L))$ do not affect $\Theta^{*}$.
\label{lem:lemma}
\end{prop}

\textit{Proof Sketch.}
By an induction argument, the supervised loss $L_B$ can be written as $g(\Theta_{1}, \dots, \Theta_{k}, \mY (\gV_L), \mX (\mathcal{N}_k(\gV_L))$ for some function $g(\cdot)$.
Thus, when the GNN is trained on $L_B$ using gradient updates, the optimal $\Theta^{*}$ must be independent of $\mX (\gV \setminus \mathcal{N}_k(\gV_L))$.

Notice that the addition of a standard regularization term (\textit{e.g.}, $L_1$ or $L_2$ regularization) only impacts $\{\Theta_{1}, \dots, \Theta_{k} \}$; thus, the overall loss function still remains independent of $\gV \setminus \mathcal{N}_k(\gV_L)$, satisfying proposition~\ref{lem:lemma}.

Hence, the optimal parameters $\Theta^{*}$ of a $k$-layer GNN are only affected by node features in the $k$-hop neighborhood $\gN_k(\gV_L)$ of the labeled set $\gV_L$, \textit{i.e.}, the features and structural connectivities of nodes in $\gV \setminus \mathcal{N}_k(\gV_L)$ are ignored during the training process.

Let the $k$-hop neighborhood of class $c$ be $\gN_k ( \gV_L (c) )$ where $\gV_L (c) = \{ v \in \gV_L : y_{vc} = 1\}$ is the set of nodes labeled with class $c$.
Let $L_B(c)$
denote the supervised loss term specific to class $c$. 
Now, the corollary directly follows from proposition~\ref{lem:lemma}:
\begin{corollary}
If node $v \not \in \gN_k ( \gV_L (c) ) $, the $k$-hop neighborhood of class $c$, then the supervised loss term $L_B (c)$ for class $c$ is independent of $v$.
\label{lem:corr}
\end{corollary}

The above corollary states that gradients from the supervised loss $L_B (c)$ for class $c$ cannot reach nodes that lie outside the $k$-hop neighborhood of class $c$, \textit{i.e.}, $\gN_k ( \gV_L (c) )$.
To illustrate its implications, we revisit~\Cref{fig:example}. Since node $c$ lies beyond the $2$-hop neighborhood of node $b$, node $c$ does not affect the training loss at node $b$ (which belongs to class 2). 
Thus, despite nodes $c$ and $b$ having identical co-variation of attributes and structure (blue
neighbors form triangles), node $c$ does not influence the training loss for nodes with class 2.

\begin{figure*}[t]
    \centering
    \includegraphics[width=0.9\linewidth]{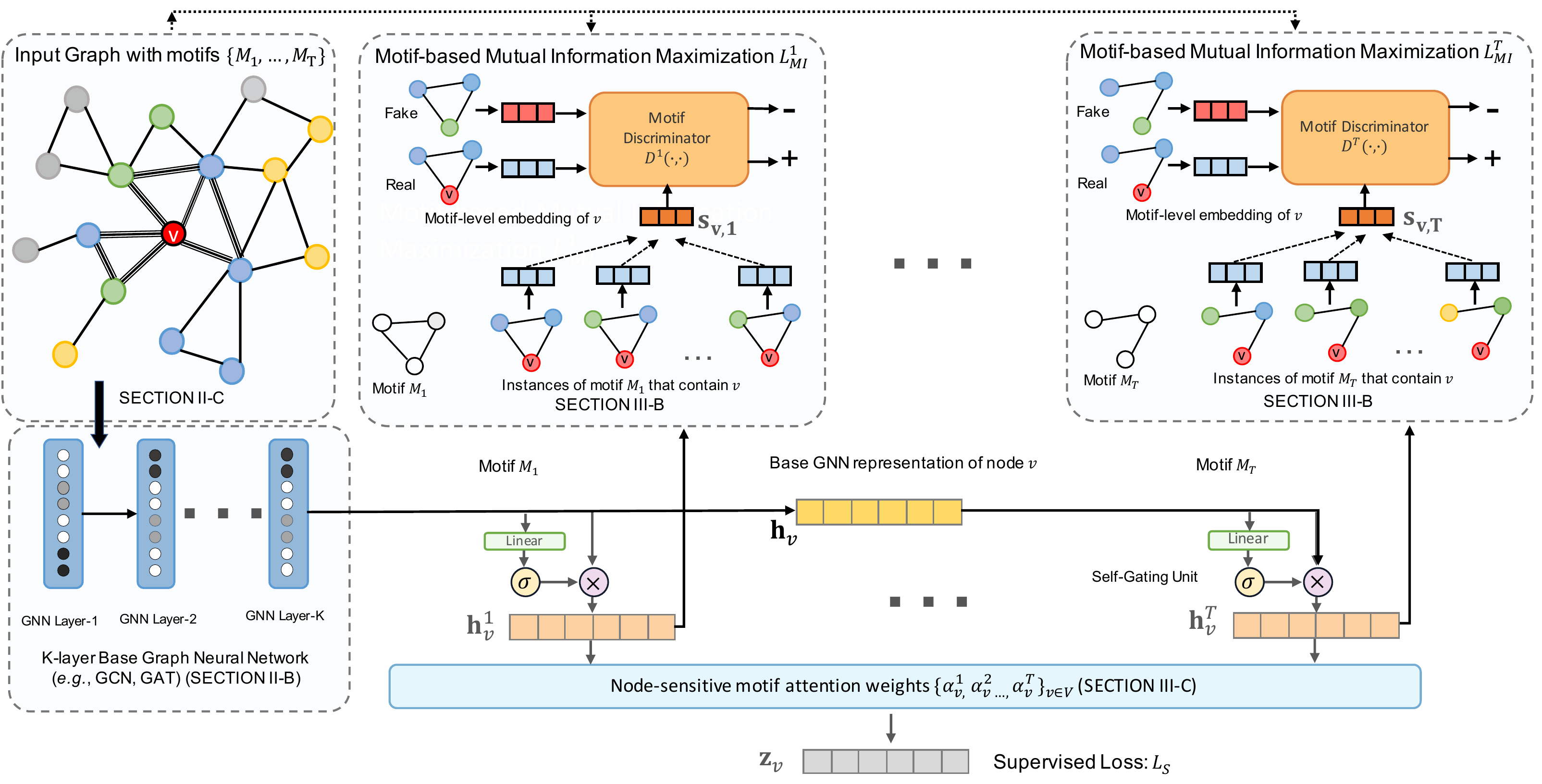}
    \caption{Neural architecture diagram of~\infomotif~depicting the different model components: base GNN $f_{\theta, l}$ with $k$ layers (bottom left), motif-based mutual information maximizing regularizers $L^{t}_{MI}$ (top right), and motif-attention module to compute final node representations $\rvz_v$ (bottom right).
    Instances of motif $M_1$ are shown in the graph (top left) with textured lines and colors indicate node attributes.}
    \label{fig:framework}
\end{figure*}

\subsection{{Self-supervised Single Motif Regularization}}
\label{sec:infomotif_single_motif}
In this section, we introduce~\infomotif, a framework to regularize node representations of the base GNN by exploiting the co-variance of node attributes and motif structures. We define \textit{attributed structural roles} by assigning the same role to nodes that participate in motif instances over \textit{co-varying} sets of attributes.
Compared to prior role-aware models~\cite{struc2vec} that discover structurally similar nodes \textit{agnostic} to attributes, 
we define roles based on attribute occurrence in higher-order connectivity structures.
In heterogeneous graphs, attributed structural roles further incorporate the semantics of node and edge types described by the connectivity structures of typed network motifs.

Now, we describe our self-supervised learning strategy to learn attribute co-variance for a single motif. In the next section, we extend these formulations to handle multiple motifs.

\subsubsection*{\textbf{Motif-based Mutual Information}}
We first consider a single network motif type $M_t \in \mathcal{M}$ and a specific node $v \in \gV$ to learn attribute co-variance across instances $\mathcal{I}_v (M_t)$ that contain $v$ in the graph.
To learn attributed structural roles, it is necessary to \textit{contrast} the attributed instances of motif $M_t$ against attributed node combinations that are not present in any instances of $M_t$.

We maximize the motif-based \textit{mutual information} (MI) between a \textit{motif-level} representation of $v$ and corresponding \textit{instance-specific} representations centered at $v$.
By introducing motif-based MI maximization as a regularizer, the GNN is encouraged to learn \textit{discriminative} statistical correspondences between nodes that participate in instances of the same motif.
Motif-based MI maximization is an example of the broader paradigm of self-supervised learning that derives auxiliary supervision signals from the intrinsic structure (\textit{e.g.,} connectivity patterns in a network motif) of the underlying data.

We first adapt the base GNN representation $\mathbf{h}_v$ (see~\Cref{sec:infomotif_base_gnn}), specific to motif $M_t$ through a \textit{motif gating} function $f^{t}_{\textsc{gate}} :  \mathbb{R}^D \mapsto  \mathbb{R}^D$ resulting in a gated embedding $\mathbf{h}_v^{t}$.
Then, we introduce a \textit{motif instance encoder} $f^{t}_{\textsc{enc}}:  \mathbb{R}^D \times \mathbb{R}^{(k_t \times D)} : \mapsto \mathbb{R}^D $ to compute the instance-specific representation $\mathbf{e}_{v, I_t} \in \mathbb{R}^D$ of node $v$ conditioned on other co-occurring nodes in instance $I_t \in \mathcal{I}_v (M_t)$.
Finally, the motif-level representation $\mathbf{s}_{v, t} \in \mathbb{R}^D$ of node $v$ summarizes the set of instance-specific representations $\{ \mathbf{e}_{v, I_t} \}_{I_t \in \mathcal{I}_v (M_t)}$ through a permutation-invariant   \textit{motif readout} function $f^{t}_{\textsc{read}} (\cdot)$, \textit{e.g.}, averaging or pooling functions.

For each node $v \in \mathcal{V}$, we maximize motif-based mutual information $L^{t}_{MI}$ between its instance-specific representations $\{ \rve_{v, I_t} \}_{I_t \in \mathcal{I}_v (M_t)}$ and motif-level representation $\rvs_{v, t}$, by defining $I_{\psi^t}$ as a mutual information estimator for motif $M_t$ that is \textit{shared} across all nodes.
The resulting learning objective is given by:

\begin{equation}
L^{t}_{MI} (\theta, \phi^t, \psi_t) = \frac{1}{|\gV|} \sum\limits_{v \in \gV} \sum\limits_{I_t \in \gI_v(M_t)} I_{ \psi_t} ( \rve_{v, I_t} ; \rvs_{v, t} )
\end{equation}
\normalsize

where 
$\theta$ and $\phi^t$ denote the parameters of the layers $\{ f_{\theta, l}\}_{l=1}^k$, and motif-specific transforms $\{f^{t}_{\textsc{gate}}, f^{t}_{\textsc{enc}}, f^{t}_{\textsc{read}} \}$ respectively.
By maximizing MI across all instances of motif $M_t$ in the graph through a shared MI estimator $I^{t}_{\psi}$, we enable the GNN to learn correspondences between a pair of potentially distant nodes that participate in instances of motif $M_t$.

\subsubsection*{\textbf{Mutual Information Maximization}}
Following prior neural MI estimation methods~\cite{mine, dim},
we model the MI estimator $I_{\psi^t}$ as a \textit{discriminator} network that learns a decision boundary to accurately distinguish between \textit{positive} samples drawn from the joint distribution and \textit{negative} samples drawn from the product of marginal distributions.
We train a \textit{constrastive} discriminator network $\mD^t_{\psi} : \sR^D \times \sR^D \mapsto \sR^{+} $, where $\mD_{\psi}^{t}(\rve_{v, I_t} , \rvs_{v, t})$ denotes the probability score assigned to this instance-motif pair (higher scores for observed instances of motif $M_t$).
The positive samples $(\rve_{v, I_t}, \rvs_{v, t})$ for $\mD_{\psi}^{t}$ are the representations $\rve_{v, I_t}$ of observed instances $I_t \in \gI_v(M_t)$ of motif $M_t$ paired with the motif-level representation $\rvs_{v,t}$. 
The negative samples  $(\rve_{v, \widetilde{I}_t}, \rvs_{v,t})$ are derived by pairing $\rvs_{v, t}$ with the representations $\rve_{v, \widetilde{I}_t} $ of negative instances $\widetilde{I}_t$ sampled from a negative sampling distribution $P_{\gN} (\widetilde{I}_t | M_t)$.
The discriminator network $\mD^t_{\psi}$ is trained on a noise-contrastive objective $L^{t}_{MI}$ with a binary cross-entropy loss  between samples drawn from the joint distribution (positive pairs), and the product of marginals (negative pairs), which is defined as in the following objective below:

\begin{align}
    L^{t}_{MI} = \frac{1}{|\gV|} \sum\limits_{v \in \gV} L^{t}_{MI} (v)  = - \frac{1}{2Q |\gV|}
    \sum\limits_{v \in \gV}  %
       \sum\limits_{i=1}^Q \Big[ & \E_{I_t}  \log  \mD_{\psi}^{t} (\rve_{v, I_t} , \rvs_{v, t})  \nonumber \\   & +
     \E_{\widetilde{I}_t} \log (1 - \mD_{\psi}^{t} (\rve_{v, \widetilde{I}_t} , \rvs_{v, t}) ) \Big]
    \label{eqn:infomotif_mi_loss}
\end{align}

where $Q$ is the number of observed motif instances sampled per node. 
The above objective~\Cref{eqn:infomotif_mi_loss} maximizes MI between $\rvs_{v ,t}$ and  $\{ \mathbf{e}_{v, I_t} \}_{I_t \in \mathcal{I}_v (M_t)}$ based on the Jensen-Shannon Divergence between their joint distribution and  product of marginals~\cite{dgi}.

We design the negative sampling distribution $P_{\gN} (\widetilde{I}_t | M_t)$ to learn attribute co-variance in instances of motif $M_t$.
For each positive instance $I_t$, the generated negative instance $\widetilde{I}_t$ is topologically equivalent but contains attributes that do not occur in instances of $M_t$ in $\gG$.
By contrasting the observed instances of $M_t$ against fake instances with perturbed attributes, $\mD_{\psi}^{t}$ learns attributed structural roles with respect to network motif $M_t$.

\subsection{{Multi-Motif Regularization Framework}}
\label{sec:infomotif_multi_motif}
Now, we extend our framework for any graph that includes a set of motifs $\gM = \{ M_1, \dots, M_T\}$. 
A typical way to include regularizers (\Cref{eqn:infomotif_mi_loss}) from multiple motifs is given by:
\begin{equation}
L^{'} = L_B + \lambda L^{'}_{MI} = L_B + \lambda \cdot \frac{1}{T} \sum\limits_{t=1}^T L^{t}_{MI}
\label{eqn:infomotif_base_reg}
\end{equation}

where $\lambda$ is a tunable hyper-parameter to balance the supervised classification loss $L_B$ and motif regularizers.
Intuitively, each motif $M_t \in \mathcal{M}$ is a connectivity pattern that can be viewed as defining one kind of structural role, \textit{e.g.}, bridge nodes.
Each motif has a different significance towards the learning task.
Thus, a multi-motif framework should automatically identify the significance of different motifs without manual hand tuning.

In addition, real-world networks exhibit heavy-tailed degree and community distributions~\cite{tail}, which manifest as \textit{skewed} (imbalanced) motif occurrences among nodes as well as across motif types. 
This further complicates the learning process of incorporating multiple motifs as regularizers.
We identify three key aspects \textit{task-oriented}, \textit{node-sensitive}, and \textit{skew-aware} that are critical to the design of a multi-motif learning framework:
\begin{itemize}
\item \textbf{Task}: Distinguish the significance of different motifs to compute node representations conditioned on the underlying semi-supervised learning task.
\item \textbf{Node}: Expressive power to control the extent of regularization exerted by each motif at a node-level granularity.
\item \textbf{Skew}: Adapt to varying levels of motif occurrence skew without any distributional assumptions on the input graph.
\end{itemize}

To address these objectives, we first describe our approach to compute final node representations conditioned on multiple motifs, followed by two novel online reweighting strategies. %

\subsubsection*{\textbf{Task-driven Representations}}
The base GNN is trained by a supervised task loss $L_B$ (\Cref{eqn:infomotif_base_supervised_loss}) over the labeled node set $\gV_L$.
We instead aggregate the set of motif-gated representations ($\rvh_v^t$ for motif $M_t \in \gM$), to compute the final representation $\rvz_v \in \sR^{D}$ for node $v$.
We learn attention weights $\alpha_{vt}$ to characterize the task-driven importance of motif $M_t$ to node $v$ and compute $\rvz_v$ through a weighted average, given by:
\begin{equation}
\rvz_v = \sum\limits_{t=1}^{T} \alpha_{vt} \rvh^{t}_v  \hspace{10 pt} 
\alpha_{vt} = \frac{\exp \big(\vp \cdot  \rvh^{t}_v  \big) }{\sum\limits_{t'=1}^T \exp \big(\vp  \cdot \rvh^{t'}_{v} \big)} 
\label{eqn:infomotif_motif_attention}
\end{equation}
\normalsize

where $\vp \in \sR^D$ defines the attention function and is learned by optimizing the final representations $\{ \rvz_v \}_{v \in \gV_L}$ of labeled nodes $\gV_L$ using the supervised loss $L_B$ (\Cref{eqn:infomotif_base_supervised_loss}). The final representation $\rvz_v$ of each node $v \in \gV$ is used for node classification.

\subsubsection*{\textbf{Node-sensitive Motif Regularization}}

Instead of using static uniform weights to incorporate motif regularizers (\Cref{eqn:infomotif_base_reg}), we contextually weight the contributions of different motif regularization terms $ L^t_{MI} (v)  $ from \Cref{eqn:infomotif_mi_loss} at a node-level granularity through the attention weights $\alpha_{vt}$ of motif $M_t$ for node $v$. This results in the following motif-regularization loss: 

\begin{equation}
 L_{MI} = \frac{1}{nT}\sum\limits_{t=1}^T \sum\limits_{v \in \gV} \alpha_{vt} L^t_{MI} (v) 
 \label{eqn:infomotif_full_mi_loss}
\end{equation}

\normalsize
The loss $L_{MI}$ varies the extent of regularization per node in proportion to the task-specific importance $\alpha_{vt}$ of motif $M_t$ to node $v$.
Notice that the attention function is learned by training the final representations $\rvz_v$ of labeled nodes $v \in \gV_L$ on the supervised loss $L_B$; in contrast, the motif-regularization loss $L_{MI}$ (which operates on all nodes) re-weights each motif loss term $L^t_{MI} (v)$  per node $v$ with the estimated attention weight $\alpha_{vt}$ from~\Cref{eqn:infomotif_motif_attention}.

\begin{algorithm}[t]
\caption{The framework of~\infomotif-GNN.}
\begin{algorithmic}[1]
\Require Graph $\gG$, Labeled node set $\gV_L$, Base GNN $\{f_{\theta, l} \}_{l=1}^k$
\Ensure Motif-regularized embedding $\rvz_v$ for each node $v \in \gV$
\State Initialize sample novelty weights $\beta_v = 1 \; \forall \; v \in \gV_L$
\While{\textit{not converged}}
\LeftComment {\textbf{\textit{Supervised loss over labeled node set $\gV_L$}}}
\For{each batch of nodes $\gV_B \subseteq \gV_L$}
\State Fix sample weights $\{ \beta_v\}_{v \in \gV_B}$ and optimize $L_S$ on $\gV_B$ using mini-batch gradient descent (Equation~\ref{eqn:infomotif_supervised_loss}).
\EndFor
\State Compute motif attention weights $\{ \bm{\alpha_v}\}_{v \in \gV}$ (Equation~\ref{eqn:infomotif_motif_attention}).
\LeftComment {\textbf{\textit{Motif-based InfoMax loss over entire node set $\gV$}}}
\For{each batch of nodes $\gV_B \subseteq \gV$}
\State Fix motif weights $\{ \bm{\alpha}_v\}_{v \in \gV}$ and optimize $L_{MI}$ on $\gV_B$ using mini-batch gradient descent (Equation~\ref{eqn:infomotif_full_mi_loss})
\EndFor
\State Compute sample weights $\{\beta_v\}_{v \in \gV_L}$ (Equation~\ref{eqn:infomotif_novelty}).
\EndWhile
\State Compute $\rvz_v \in \sR^D \; \forall \; v \in \gV$ (Equation~\ref{eqn:infomotif_motif_attention})
\end{algorithmic}
\label{alg:opt}
\end{algorithm}

\subsubsection*{\textbf{Skew-aware Sample Weighting}}
Prior work in curriculum and meta learning has shown the importance of re-weighting training examples to overcome training set biases~\cite{reweight}.
In particular, re-weighting strategies that emphasize harder examples are effective at handling imbalanced data distributions~\cite{imbalance}.
We propose a \textit{novelty-driven} re-weighting strategy to handle skew in motif occurrences across nodes and motif types.

The novelty $\beta_v$ of node $v$ is a function of its motif distribution, \textit{i.e.}, novel nodes contain uncommon motif types in their neighborhood, which in turn reflects in their attention weight distribution over motifs.
Let $\bm{\alpha}_v \in \sR^{T}$ denote the vector of attention weights for a labeled node $v$ over the motif set $\gM$.
Now, the novelty $\beta_v$ of node $v$ is high
if its motif distribution $\bm{\alpha}_v$ significantly diverges from those of other nodes.
We quantify $\beta_v$ by the deviation (measured by euclidean distance) of $\bm{\alpha}_v$ from the mean motif distribution of labeled nodes  $v \in \gV_L$. %
\begin{equation}
 \beta_v = \frac{exp ( \left\lVert \bm{\alpha}_v - \bm{\mu} \right\rVert^2 ) }{ \sum\limits_{u \in \gV_L} \exp (\left\lVert \bm{\alpha}_u - \bm{\mu} \right\rVert^2) }  \hspace{10pt} \bm{\mu} = \frac{1}{|\gV_L|} \sum\limits_{v \in \gV_L} \bm{\alpha}_v
 \label{eqn:infomotif_novelty}
\end{equation}
\normalsize
The novelty scores are normalized over $\gV_L$ using a softmax function, to give non-negative sample weights $0 < \beta_v \leq 1$. 
We now define the novelty-weighted supervised loss $L_S$ as:
\begin{equation}
    L_S = - \sum\limits_{v \in \mathcal{V}_L} \beta_v \sum\limits_{c=1}^C y_{vc} \log \hat{y}_{vc}
    \label{eqn:infomotif_supervised_loss}
\end{equation}
\normalsize
In contrast to the original supervised loss $L_B$ (\Cref{eqn:infomotif_base_supervised_loss}), the re-weighted objective $L_S$ induces a novelty-weighted training curriculum that  progressively focuses on harder samples.

\subsubsection*{\textbf{Model Training}}
The overall objective of~\infomotif~is composed of two terms, the re-weighted supervised loss $L_S$ (\Cref{eqn:infomotif_supervised_loss}), and motif regularizers (\Cref{eqn:infomotif_full_mi_loss}), given by:
\begin{equation}
L = L_S + \lambda L_{MI}
\label{eqn:infomotif_combined_loss}
\end{equation}
\normalsize

We optimize $L_S$ and $L_{MI}$ alternatively at each training epoch, which removes the need to tune balance hyper-parameter $\lambda$.~\Cref{alg:opt} summarizes the training procedure.

\subsubsection*{\textbf{Complexity Analysis}}
On the whole, the complexity of our model is $O(\mF)+ O(nTQD + nTD^2) $ where $O(\mF)$ is the base GNN complexity, $T$ is the number of motifs, $Q$ is sampled instance count per motif, and $D$ the latent space dimensionality. 
Since $T \ll n$ and $Q \ll n$, the added complexity of our \infomotif~framework scales linearly with respect to the number of nodes.

\section{Model Details}
\label{sec:infomotif_model_details}
We now discuss the architectural details of our framework: motif instance encoder, gating, readout, and discriminator.

\subsection{{Motif Gating}}
We design a pre-filter with \textit{self-gating units} (SGUs) to regulate the flow of information from the base GNN embedding $\mathbf{h}_v$ to the motif-based regularizer.
The SGU  for motif $M_t$, denoted by $f^t_{\textsc{gate}} (\cdot)$, learns a non-linear gating function to modulate the input $\mathbf{h}_v$  at a feature-wise granularity through dimension re-weighting. %
In particular, we define the self-gating unit for motif $M_t$ as:
\begin{equation}
    \mathbf{h}^{t}_v = f^t_{\textsc{gate}} ( \mathbf{h}_v) = \mathbf{h}_v \odot \sigma (\mathbf{W}_g^t \mathbf{h}_v  + \mathbf{b}_g^t)
    \label{eqn:gating}
\end{equation}
\normalsize
where $\mathbf{W}^t \in \mathbb{R}^{D \times D}, \mathbf{b}^t \in \mathbb{R}^{D}$ are learned parameters, $\odot$ denotes the element-wise product operation, and $\sigma$ is the sigmoid non-linearity. The self-gating mechanism effectively serves as a multiplicative skip-connection~\cite{glu} that facilitates gradient flow from the motif-based regularizer to the layers of the base GNN.

\subsection{{Motif Instance Encoder}} The encoder $f_{\textsc{enc}} (\cdot)$ computes the instance-specific representation $\rve_{v, I_t}$ for node $v$ conditioned on the gated representations $\{ \rvh^t_u\}_{u \in I_t}$ of the nodes in instance $I_t$.
We apply self-attentions~\cite{self_attention} to compute a weighted average of the gated node representations $\{ \rvh^t_u\}_{u \in I_t}$ in $I_t$.
Specifically, $f_{\textsc{enc}}$ attends over each node $u \in I_t$ to compute attention weight $\alpha_{u}$ by comparing its gated representation $\rvh^t_u$ with that of node $v$, $\rvh^t_v$ using an attention network that is parameterized by a single MLP layer, defined by:

\begin{equation}
\rve_{v, I_t} = \sum\limits_{u \in I_t} \alpha_u \rvh^{t}_u  \hspace{10 pt} \alpha_u = \frac{\exp \big(\va^{t} \cdot [ \rvh^{t}_u || \rvh^{t}_v ] \big) }{\sum\limits_{u^{'} \in I_t} \exp \big(\va^t  \cdot [ \rvh^{t}_{u^{'}} || \rvh^{t}_v] \big)}
\label{eqn:infomotif_encoder}
\end{equation}
\normalsize
where $\va^{t} \in \sR^{2D}$ is a weight vector parameterizing the attention function and $||$ denotes concatenation. We empirically find self-attentional encoders to outperform pooling alternatives.

\subsection{{Motif Readout}}
The motif readout function $f^t_{\textsc{read}} (\cdot)$ summarizes the set of instance-specific representations $\{\rve_{v, I_t}\}_{I_t \in \gI_v (M_t)}$ to compute the motif-level representation $\rvs_{v ,t}$ for motif $M_t$.
Here, we use a simple averaging of the instance-specific representations to formulate the motif readout function $f^t_{\textsc{read}} (\cdot)$ as defined below:

\begin{equation}
   \rvs_{v ,t} = f^t_{\textsc{read}} \Big(\{\rve_{v, I_t}\}_{I_t \in \gI_v (M_t)} \Big) = \sigma \Big( \sum\limits_{I_t \in \gI_v (M_t) } \frac{\rve_{v, I_t}}{|\gI_v (M_t)|} \Big)
   \label{eqn:infomotif_readout}
\end{equation}
 \normalsize
 
where $\sigma$ is the sigmoid non-linearity.
We adopt a batch-wise model training strategy with motif instance sampling ($\sim$ 20 per node) to compute the motif-level representation $\rvs_{v ,t}$. 
While we find this simple readout function to be empirically effective, sophisticated architectures~\cite{set2vec} are  likely necessary to handle larger sample sizes.

\subsection{{Motif Discriminator}}
The discriminator $D^{t}_{\psi}$ learns a motif-specific scoring function to assign higher likelihoods to observed instance-motif pairs relative to negative examples.
Similar to prior work~\cite{dgi, groupim}, we use a bilinear scoring function defined by:
\begin{equation}
    \mD^{t}_{\psi} (\rve_{v, I_t}, \rvs^{t}_v) = \sigma (\rve_{v, I_t} \cdot \mW_d^{t} \rvs^t_v)
\end{equation}
\normalsize
where $\mW_d^{t} \in \sR^{D \times D} $ is a trainable scoring matrix and $\sigma$ is the sigmoid non-linearity to convert raw scores into probabilities of $(\rve_{v, I_t}, \rvs^{t}_v)$ being a positive example for motif $M_t$.

\section{Experiments}
\label{sec:infomotif_experiments}

We present extensive quantitative and qualitative analyses on multiple diverse datasets spanning homogeneous and heterogeneous graphs.
We first introduce datasets, baselines, and experimental setup (Section~\ref{sec:infomotif_datasets}, \ref{sec:infomotif_baselines}, and~\ref{sec:infomotif_setup}), followed by node classification results in Sections~\ref{sec:infomotif_homogeneous_results} and~\ref{sec:infomotif_heterogeneous_results}~by integrating three GNN models in our framework.
We propose four research questions to guide our experiments:

\begin{enumerate}[label=(\subscript{\textbf{RQ}}{{\textbf{\arabic*}}}),leftmargin=*]
\item How does~\infomotif~compare with state-of-the-art graph neural networks and embedding learning methods on \textit{node classification} over \textit{homogeneous} graphs?

\item Is~\infomotif~effective~for \textit{node classification} on \textit{heterogeneous} information networks (multiple types of nodes and edges) compared to state-of-the-art approaches?

\item How do the different \textit{architectural} design choices and \textit{training} strategies in~\infomotif~impact performance?

\item  What is the impact of \textit{node degree}, local training \textit{label sparsity}, and local \textit{attribute diversity}, on the classification performance of~\infomotif?

\item How do the motif-based \textit{regularization} strategies and associated \textit{hyper-parameters} in~\infomotif~affect model training time and performance?

\end{enumerate}

 \begin{table}[t]
 \centering
 \begin{tabular}{@{}p{0.17\linewidth}R{0.09\linewidth}R{0.09\linewidth}R{0.09\linewidth}R{0.09\linewidth}R{0.09\linewidth}R{0.09\linewidth}@{}}
 \toprule
 \multirow{2}{*}
 & \multicolumn{3}{c}{ \textbf{Citation Networks}}  & \multicolumn{3}{c}{ \textbf{Air-Traffic Networks}} \\
 \cmidrule(lr){2-4} \cmidrule(lr){5-7}

 \textbf{Dataset} &  \textbf{Cora} & \textbf{Citeseer} & \textbf{Pubmed}  & \textbf{Brazil} & \textbf{Europe} & \textbf{USA} \\ 
 \midrule
 \textbf{\# Nodes} &  2,485 & 2,110 & 19,717 &  131 & 399 & 1,190\\ 
 \textbf{\# Edges} & 5,069 & 3,668 & 44,324  & 1,038 & 5,995 &13,599\\
 \textbf{\# Attributes} & 1,433  & 3,703 & 500 & - & - & -\\
 \textbf{\# Classes} & 7 & 6 & 3 & 4 & 4 & 4 \\
 \bottomrule
 \end{tabular} 
\caption{Dataset statistics of homogeneous network benchmarks, including three assortative citation~\cite{citation} networks and three dis-assortative air-traffic~\cite{struc2vec} networks. Ground-truth classes in citation networks exhibit attribute homophily; ground-truth classes in flight networks indicate node structural roles.}
\label{tab:infomotif_stats}
\end{table}

\begin{table}[t]
\centering
\begin{tabular}{@{}p{0.25\linewidth}R{0.2\linewidth}R{0.2\linewidth}R{0.2\linewidth}@{}}
\toprule
\textbf{Dataset} & \textbf{DBLP-A } &  \textbf{DBLP-P} & \textbf{Movie} \\
\midrule
\textbf{\# Nodes} & \num[group-separator={,}]{11170} & \num[group-separator={,}]{35770} &  \num[group-separator={,}]{10441} \\
\textbf{\# Edges} &  \num[group-separator={,}]{24846}  & \num[group-separator={,}]{131636} & \num[group-separator={,}]{99509} \\
\textbf{\# Attributes} &  4,479  & 11,680 &  4,577\\
\textbf{\# Node Types} & 3 & 3 & 4 \\
\textbf{\# Classes} & 4 & 10 & 6 \\
\bottomrule
\end{tabular}
\caption{Dataset statistics of three heterogeneous graphs across bibliographic and movie information networks, with multiple node and edge types. Networks schemas are shown in Figures~\ref{fig:dblp_schema} and~\ref{fig:movie_schema}.}
\label{tab:hin_dataset_stats}
\end{table}

\subsection{{Datasets}}
\label{sec:infomotif_datasets}
Our experiments are designed towards semi-supervised node classification on nine real-world benchmark datasets, divided across homogeneous and heterogeneous networks.
In homogeneous graphs, we experiment on two diverse types of datasets: \textit{citation networks} that exhibit strong homophily and \textit{air-traffic networks} that depend on structural roles. Table~\ref{tab:infomotif_stats} presents detailed statistics for each dataset.

\begin{itemize}
\item \textbf{Citation Networks:} We consider three benchmark datasets: Cora, Citeseer, and PubMed~\cite{citation}, where nodes correspond to documents and edges represent citation links. Each document is associated with a bag-of-words feature vector and the task is to classify documents into different research topics.
\item \textbf{Air-Traffic Networks:} We use three undirected networks: Brazil, Europe, and USA~\cite{struc2vec} where nodes correspond to airports and edges indicate the existence of commercial flights.
Class labels are assigned based on activity level, measured by the cardinality of flights or people that passed the airports. We use one-hot indicator vectors as node attributes.
Here, class labels are related to the role played by airports.
\end{itemize}

We conduct experiments on three real-world datasets over heterogeneous graphs whose statistics are shown in Table~\ref{tab:hin_dataset_stats}:
\begin{itemize}[leftmargin=*]
\item  \textbf{DBLP-A}: This is a bibliographic network composed of 3 node types: author ($A$), paper ($P$) and venue ($V$), connected by three link types: $P \rightarrow P$, $A\mbox{-}P$ and $P\mbox{-}V$ (Figure~\ref{fig:dblp_schema}). We use a subset of DBLP~\cite{pathsim} with text attributes of papers to classify authors based on their research areas.
\item \textbf{DBLP-P}: This dataset has the same schema as DBLP-A, but the learning task is to classify research papers into ten categories, which are obtained from Cora~\cite{cora}.
\item \textbf{Movie}: We use MovieLens~\cite{movielens} with four node types: movie ($M$), user ($U$), actor ($A$) and tag ($T$) linked by four types: $U\mbox{-}M$, $A\mbox{-}M$, $U\mbox{-}T$ and $M\mbox{-}T$, with attributes for actors and movies. The classification task is movie genre prediction.
\end{itemize}

\begin{figure}[t]
\centering
\subfigure[Schema of Movie]{
        \centering
        \includegraphics[width=0.22\linewidth]{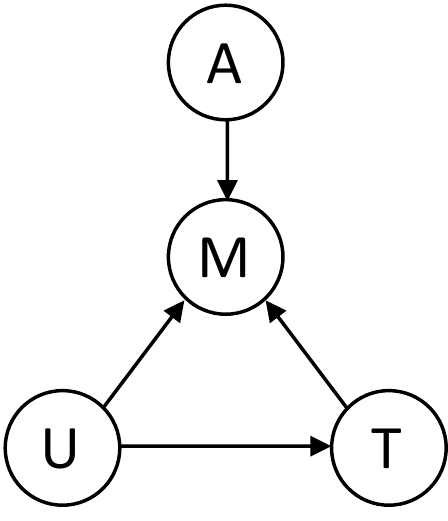}
        \label{fig:movie_schema}}
        \hspace{0.02\linewidth}
\subfigure [Examples of typed 3-node motifs in MovieLens]{
        \centering
        \includegraphics[width=0.62\linewidth]{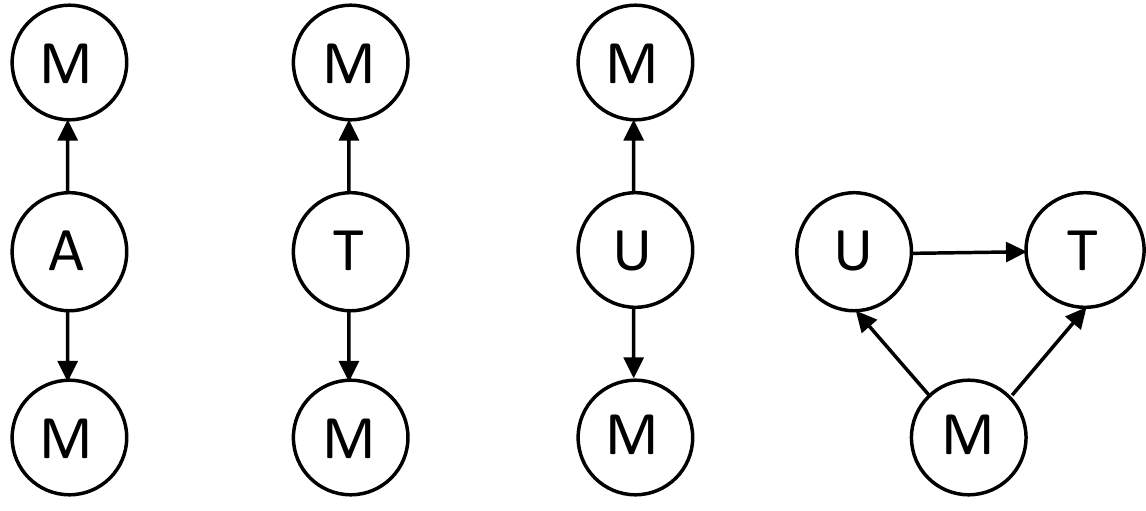}
        \label{fig:movie_motifs}}
        \caption{ (a) Heterogeneous network schema of movie network with four node types: Actor (A), Movie (M), User (U) and Term (T) and four edge types $A-M, U-M, T-M$, and $U-T$.  (b) Examples of 3-node connected typed network motifs.}    
\end{figure}

\subsection{{Baselines}}
\label{sec:infomotif_baselines}
We first introduce baseline methods designed for learning over homogeneous graphs, organized into four categories based on whether they are \textit{proximity-based vs. structural}; and the paradigm of \textit{embedding learning vs. graph neural networks}:

\begin{itemize}
    \item \textbf{Proximity-based embedding methods}:  Conventional methods node2vec~\cite{node2vec} that employs second-order proximity, and motif2vec~\cite{motif2vec} based on higher-order proximity.

\begin{itemize}   
\item    node2vec~\cite{node2vec}: Skip-gram model to learn unsupervised node representations that capture node co-occurrences in second-order random walk samples.
\item motif2vec~\cite{motif2vec}: Higher-order extension of skip-gram embedding models to model higher-order proximity induced by motif structures.
\end{itemize}    
    
    \item \textbf{Structural embedding methods}: Structural role-aware embedding learning models struc2vec~\cite{struc2vec}, GraphWAVE~\cite{graphwave}, and DRNE~\cite{drne}.
    
    \begin{itemize}  
    \item struc2vec~\cite{struc2vec}: Learns node representations capturing structural identity by utilizing degree sequences at different local neighborhoods.
    
    \item GraphWAVE~\cite{graphwave}: Learns node embeddings that capture structural similarity by leveraging spectral graph wavelet diffusion patterns.
    
    \item DRNE~\cite{drne}: Deep recursive neural network to learn node embeddings that preserve the property of regular equivalence.
\end{itemize}
    
    \item \textbf{Standard Graph Neural Networks}: We compare against several state-of-the-art GNNs based on localized message passing:

\begin{itemize}
  
\item GCN~\cite{gcn}: Graph convolutional networks operating on the graph Laplacian matrix using degree-weighted neighborhood aggregation.

\item GraphSAGE~\cite{graphsage}: Scalable graph neural networks with neighborhood sampling and aggregation using a variety of pooling functions (max, mean, and LSTM)

\item GAT~\cite{gat}: Graph attention networks with self-attentional neighbor aggregation.

\item JK-Net~\cite{jknet}: Jumping knowledge networks that use skip-connections to vary the influence radius (per node) for neighborhood aggregation.

\item DGI~\cite{dgi}: Deep Graph Infomax that learns node embeddings by maximizing mutual information between local and global representations of the same graph.
\end{itemize}
     
    \item \textbf{Structural Graph Neural Networks}: Motif-based GNN models Motif-CNN~\cite{motifcnn}, MCN~\cite{motif_attention_cikm19} and degree-specific DEMO-Net~\cite{demonet}.

\begin{itemize}
\item Motif-CNN~\cite{motifcnn}: Motif-based graph convolutional networks with attention-based cross-motif aggregation.

\item MCN~\cite{motif_attention_cikm19}: Motif-based graph attentional model that uses weighted multi-hop motif adjacency matrices to capture higher-order neighborhood information.

\item DEMO-Net~\cite{demonet}: Degree-specific graph neural network that uses multi-task graph convolutions to learn degree-aware node embeddings.

\end{itemize}

\end{itemize}

Next, we introduce graph neural networks designed specifically for semi-supervised learning over heterogeneous graphs.

\begin{itemize}
 \item \textbf{Heterogeneous Graph Neural Networks}: State-of-the-art \textit{metapath}-based GNN models MAGNN~\cite{magnn}, HAN~\cite{han}, and \textit{metagraph}-aware GNN model Meta-GNN~\cite{metagnn}.
 
\begin{itemize}
\item  MAGNN~\cite{magnn}: Learns heterogeneous node embeddings via intra-metapath aggregation to incorporate intermediate semantic nodes, and the inter-metapath aggregation to combine messages from multiple metapaths.

\item HAN~\cite{han}: Learns metapath-specific node embeddings from different metapath-based homogeneous graphs using an attention mechanism.

\item Meta-GNN~\cite{metagnn}: Learns node embeddings to extract and aggregate features from local metagraph-structured neighborhoods in heterogeneous graphs.

\end{itemize}

\end{itemize}

\newcommand*{\factortraining}{0.02}
\newcommand*{\factor}{0.106}
\begin{table}[hbt]
\centering
\footnotesize
\noindent\setlength\tabcolsep{2.5pt}
\begin{tabular}{@{}p{0.2\linewidth}@{\hspace{0pt}}
K{\factortraining\linewidth}K{\factortraining\linewidth}@{\hspace{0pt}}
K{\factor\linewidth}K{\factor\linewidth}@{\hspace{10pt}}
K{\factor\linewidth}K{\factor\linewidth}@{\hspace{10pt}}
K{\factor\linewidth}K{\factor\linewidth}@{}} \\
\toprule
\multirow{1}{*}{\textbf{}} &  \multicolumn{2}{c}{\textbf{Data}} & \multicolumn{2}{c}{\textbf{Cora}}  & \multicolumn{2}{c}{\textbf{Citeseer}} & \multicolumn{2}{c}{\textbf{PubMed}} \\
\cmidrule{2-3} \cmidrule(lr){4-5} \cmidrule(lr){6-7} \cmidrule(lr){8-9}
\multirow{1}{*}{\textbf{Training Ratio}} & \textbf{X} & \textbf{Y} & \textbf{20\%} & \textbf{40\%}  & \textbf{20\%} & \textbf{40\%} &  \textbf{20\%} & \textbf{40\%}  \\
\midrule
\multicolumn{9}{c}{\textsc{Proximity-based Graph Embedding Methods}} \\
\midrule[0pt]
\textbf{{Node2Vec~\cite{node2vec}} } &  &    
& 75.7 $\pm$ 0.5& 	76.1 $\pm$ 0.5& 	
68.1 $\pm$ 0.5	& 69.1 $\pm$ 0.6& 	
80.1 $\pm$ 0.6& 	80.2 $\pm$ 0.6 	\\
\textbf{Motif2Vec~\cite{motif2vec}}  & &   
& 79.0 $\pm$ 0.4& 	79.2 $\pm$ 0.4& 		
66.6 $\pm$ 0.4& 	67.1 $\pm$ 0.3& 	
79.8 $\pm$ 0.2& 	79.8 $\pm$ 0.4   \\
\midrule[0pt]
\multicolumn{9}{c}{\textsc{Structural Graph Embedding Methods}} \\
\midrule[0pt]
\textbf{Struct2Vec~\cite{struc2vec}} & &    &
35.4 $\pm$ 1.0 & 37.6 $\pm$ 1.3 & 
31.2 $\pm$ 0.8 & 35.1 $\pm$ 0.9 & 
48.5 $\pm$ 0.3 & 49.2 $\pm$ 0.4 \\
\textbf{GraphWave~\cite{graphwave}} &   &    &
39.5 $\pm$ 2.1 & 41.1 $\pm$ 1.5 & 
38.5 $\pm$ 1.2 & 40.6 $\pm$ 0.9 & 
43.0 $\pm$ 2.0 & 43.3 $\pm$ 1.3  \\
\textbf{DRNE~\cite{drne}} &   &    &
34.9 $\pm$ 1.5 & 36.5 $\pm$ 1.5 & 
30.8 $\pm$ 1.2 & 32.2 $\pm$ 1.2 & 
40.4 $\pm$ 0.7 & 41.6 $\pm$ 0.4  \\
\midrule[0pt]
\multicolumn{9}{c}{\textsc{Standard Graph Neural Networks}} \\
\midrule[0pt]
\textbf{GCN~\cite{gcn}}  & \checkmark & \checkmark 
 & 81.6 $\pm$ 0.5 &	82.0 $\pm$ 0.4&
75.8 $\pm$ 0.5 & 	76.6 $\pm$ 0.3 &	
85.7 $\pm$ 0.7 &	86.1 $\pm$ 0.5 \\
\textbf{GAT~\cite{gat}}  & \checkmark   & \checkmark   
& 80.9 $\pm$ 0.7 &	81.4 $\pm$ 0.2	 & 
74.5 $\pm$ 0.7 &	75.5 $\pm$ 0.7 &	
 83.3 $\pm$ 0.3 &	84.2 $\pm$ 0.3  \\
\textbf{GraphSAGE~\cite{graphsage}} & \checkmark   & \checkmark   
& 81.3 $\pm$ 0.3 &	83.5 $\pm$ 0.3 &
72.9 $\pm$ 0.3 &	73.8 $\pm$ 0.2 &	
86.6 $\pm$ 0.2 &	87.2 $\pm$ 0.3  \\
\textbf{JKNet~\cite{jknet}} &\checkmark  & \checkmark  
 &81.3 $\pm$ 0.8&83.6 $\pm$ 0.8	&
	71.5 $\pm$ 0.8&	72.5 $\pm$ 0.7	&
		82.2 $\pm$ 0.4&	83.8 $\pm$ 0.5   \\
\textbf{DGI~\cite{dgi}}  & \checkmark  & 
&  76.2 $\pm$ 0.8& 	77.3 $\pm$ 0.9& 
74.5 $\pm$ 0.7& 	74.7 $\pm$ 0.7& 
78.2 $\pm$ 0.9& 	78.5 $\pm$ 0.9 \\	
\midrule[0pt]
\multicolumn{9}{c}{\textsc{Structural Graph Neural Networks}} \\
\midrule[0pt]
\textbf{DemoNet~\cite{demonet}} & \checkmark   & \checkmark   
& 81.0 $\pm$ 0.6 &	82.4 $\pm$ 0.5 &	
67.9 $\pm$ 0.7 &	68.5 $\pm$ 0.6 &	
79.5 $\pm$ 0.4	 &80.5 $\pm$ 0.4	   \\
\textbf{Motif-CNN~\cite{motifcnn}} & \checkmark  & \checkmark  
 &81.6 $\pm$ 0.5 &	82.8 $\pm$ 0.5 &	
73.4 $\pm$ 0.3 &	76.8 $\pm$ 0.3 &	
87.3 $\pm$ 0.1 &	87.5 $\pm$ 0.1     \\
\textbf{MCN~\cite{motif_attention_cikm19}} & \checkmark  & \checkmark  
 &81.1 $\pm$ 0.9 &	82.4 $\pm$ 0.8 &	
73.2 $\pm$ 0.4 &	75.9 $\pm$ 0.7 &	
85.2 $\pm$ 0.6 &	85.9 $\pm$ 0.5   \\
\midrule[0pt]
\multicolumn{9}{c}{\textsc{Motif-regularized Graph Neural Networks~(\textbf{\infomotif}) }} \\
\midrule[0pt]
\textbf{{InfoMotif-GCN}}  & \checkmark & \checkmark & 
 \textbf{85.7 $\pm$ 0.4} &	\textbf{87.4 $\pm$ 0.4}&
\textbf{77.7 $\pm$ 0.5}&	\textbf{78.5 $\pm$ 0.5}&
\textbf{87.5 $\pm$ 0.2}&	\textbf{88.3 $\pm$ 0.2}  \\
\textbf{{InfoMotif-JKNet}}  & \checkmark & \checkmark & 
85.5 $\pm$ 0.3 &	86.5 $\pm$ 0.5&		
74.5 $\pm$ 0.8&	 76.7 $\pm$ 0.9&	
87.0 $\pm$ 0.2&	 87.9 $\pm$ 0.3   \\
\textbf{{InfoMotif-GAT}} & \checkmark & \checkmark & 
85.5 $\pm$ 0.3 & 87.2 $\pm$ 0.7 & 
76.5 $\pm$ 0.5 & 77.0 $\pm$ 0.4 & 
85.9 $\pm$ 0.4 & 86.2 $\pm$ 0.5 \\

\bottomrule
\end{tabular}
\caption{Node classification results (\% test accuracy) on assortative citation networks. %
$\mathbf{X}$ and $\mathbf{Y}$ denote the use of attributes and labels respectively. %
We report mean classification accuracy and standard deviation over 5 trials. We show GraphSAGE results with the best performing aggregator (among Max, Mean and LSTM).~\infomotif~improves results of all base GNNs by 3.5\% on average across datasets.}
\label{tab:citation_results}
\end{table}

\subsection{{Experimental Setup}}
We tested~\infomotif~by integrating GCN, JK-Net and GAT as base graph neural network models within our framework.
We only consider the largest connected component in each dataset and evaluate different train/validation/test splits to fairly compare different models~\cite{pitfalls}.
We create 10 random data splits per training ratio and evaluate the mean test classification accuracy along with standard deviation.

All experiments were conducted on a Tesla K-80 GPU using PyTorch. Our implementation of~\infomotif~is publicly available\footnote{\url{
https://github.com/CrowdDynamicsLab/InfoMotif}}.
For citation networks, we use two-layer base GNNs with layer sizes of 256 each, while using 64 for the smaller air-traffic networks.
We train the base JK-Net model using 4 GCN layers and maxpool layer aggregation, while the base GAT model learns 8 attention heads per layer.
The model is trained for a maximum of 100 epochs with a batch size of 256 nodes with Adam optimizer.
We also apply dropout with a rate of 0.5, and tune the learning rate in the range $\{ 10^{-4}, 10^{-3}, 10^{-2}\}$.

\label{sec:infomotif_setup}

\subsection{{Homogeneous Graphs ($\text{RQ}_1$)}}
\label{sec:infomotif_homogeneous_results}

In homogeneous graphs, we train~\infomotif~using the set of all directed 3-node motifs in citation networks and undirected 3-node motifs in air-traffic networks (\Cref{fig:motifs}).
We evaluate different train/validation/test splits (training ratios of 20\%, and 40\%) and report experimental results comparing~\infomotif~with three base GNNs, against competing baselines on citation and air-traffic networks, in Tables~\ref{tab:citation_results} and~\ref{tab:airport_results} respectively.

\renewcommand*{\factortraining}{0.02}
\renewcommand*{\factor}{0.106}
\begin{table}[bt]
\centering
\footnotesize
\noindent\setlength\tabcolsep{2.5pt}
\begin{tabular}{@{}p{0.2\linewidth}@{\hspace{0pt}}
K{\factortraining\linewidth}K{\factortraining\linewidth}@{\hspace{0pt}}
K{\factor\linewidth}K{\factor\linewidth}@{\hspace{10pt}}
K{\factor\linewidth}K{\factor\linewidth}@{\hspace{10pt}}
K{\factor\linewidth}K{\factor\linewidth}@{}} \\
\toprule
\multirow{1}{*}{\textbf{}} &  \multicolumn{2}{c}{\textbf{Data}} & \multicolumn{2}{c}{\textbf{USA}}  & \multicolumn{2}{c}{\textbf{Europe}} & \multicolumn{2}{c}{\textbf{Brazil}} \\
\cmidrule{2-3} \cmidrule(lr){4-5} \cmidrule(lr){6-7} \cmidrule(lr){8-9}
\multirow{1}{*}{\textbf{Training Ratio}} & \textbf{X} & \textbf{Y} & \textbf{20\%} & \textbf{40\%} & \textbf{20\%} & \textbf{40\%} & \textbf{20\%} & \textbf{40\%}  \\
\midrule
\multicolumn{9}{c}{\textsc{Proximity-based Graph Embedding Methods}} \\
\midrule[0pt]
\textbf{{Node2Vec}~\cite{node2vec}} &   &
 & 24.6 $\pm$ 0.9	&24.8 $\pm$ 0.9	&
36.5 $\pm$ 1.0&	37.4 $\pm$ 1.1	&
26.3 $\pm$ 1.4	&30.4 $\pm$ 1.3    \\
\textbf{{Motif2Vec}~\cite{motif2vec}}  &   &
 & 51.3 $\pm$ 1.1 	&54.8 $\pm$ 1.1	&
37.1 $\pm$ 1.2&	38.1 $\pm$ 1.2	&
27.2 $\pm$ 1.5	&33.9 $\pm$ 1.5  \\
\midrule[0pt]
\multicolumn{9}{c}{\textsc{Structural Graph Embedding Methods}} \\
\midrule[0pt]
\textbf{{Struct2Vec}~\cite{struc2vec}}  &  &
 & 50.4 $\pm$ 0.8&51.3 $\pm$ 0.8		&
42.5 $\pm$ 0.7&45.6 $\pm$ 0.8	&
45.8 $\pm$ 1.1&51.8 $\pm$ 1.1   \\
\textbf{GraphWave~\cite{graphwave}} &   &    &
45.2 $\pm$ 1.4 & 48.0 $\pm$ 1.4 & 
38.1 $\pm$ 1.9 & 41.1 $\pm$ 1.6 & 
40.2 $\pm$ 2.0 & 43.1 $\pm$ 1.8 \\
\textbf{DRNE~\cite{drne}} & & & 
51.3 $\pm$ 1.1 & 52.4 $\pm$ 1.1 &
43.1 $\pm$ 1.7 & 47.6 $\pm$ 1.3 &
46.5 $\pm$ 2.7 & 50.2 $\pm$ 2.3  \\
\midrule[0pt]
\multicolumn{9}{c}{\textsc{Standard Graph Neural Networks}} \\
\midrule[0pt]
\textbf{GCN~\cite{gcn}}  & \checkmark  & \checkmark  
& 51.9 $\pm$ 0.9	&56.0 $\pm$ 0.9&
37.4 $\pm$ 0.9&40.1 $\pm$ 0.8	&
	36.5 $\pm$ 1.5	&38.9 $\pm$ 1.6  \\
\textbf{GAT~\cite{gat}}  & \checkmark   & \checkmark   
& 52.7 $\pm$ 1.0	&53.5 $\pm$ 0.9 	&
31.5 $\pm$ 1.0	&34.3 $\pm$ 1.0		&
37.3 $\pm$ 1.6	&37.9 $\pm$ 1.6  \\
\textbf{GraphSAGE~\cite{graphsage}} & \checkmark  & \checkmark   
& 45.3 $\pm$ 1.2	&49.4 $\pm$ 1.2&	
28.8 $\pm$ 1.0&	32.5 $\pm$ 1.0	&
	36.1 $\pm$ 1.6	&37.5 $\pm$ 1.6	\\
\textbf{JKNet~\cite{jknet}} &\checkmark  & \checkmark  
 &53.8 $\pm$ 1.2&56.1 $\pm$ 1.0	&
	49.7 $\pm$ 1.1&	53.8 $\pm$ 1.1	&
		55.9 $\pm$ 1.5&	58.4 $\pm$ 1.8  \\
\textbf{DGI~\cite{dgi}}  & \checkmark   &
&  46.4 $\pm$ 1.3&	47.3 $\pm$ 1.2&
37.5 $\pm$ 1.5	&39.9 $\pm$ 1.5	&
41.4 $\pm$ 1.6	&45.2 $\pm$ 1.7\\

\midrule[0pt]
\multicolumn{9}{c}{\textsc{Structural Graph Neural Networks}} \\
\midrule[0pt]

\textbf{DemoNet~\cite{demonet}} & \checkmark   & \checkmark  
 & 58.6 $\pm$ 1.2	&58.8 $\pm$ 1.1&	
40.4 $\pm$ 1.3& 46.2 $\pm$ 1.2	& 
46.1 $\pm$ 1.4	&48.9 $\pm$ 1.5  \\
\textbf{Motif-CNN~\cite{motifcnn}} & \checkmark  & \checkmark  
 &53.6 $\pm$ 1.0&54.2 $\pm$ 1.0&
	37.9 $\pm$ 1.0&	41.1 $\pm$ 1.1	&
	28.9 $\pm$ 1.6&	35.7 $\pm$ 1.7	   \\
\textbf{MCN~\cite{motif_attention_cikm19}} & \checkmark  & \checkmark  
 &54.8 $\pm$ 1.4&54.9 $\pm$ 1.3	&
	36.8 $\pm$ 1.2&	39.6 $\pm$ 1.5	&
	42.9 $\pm$ 1.6&	43.6 $\pm$ 1.4  \\

\midrule[0pt]
\multicolumn{9}{c}{\textsc{Motif-regularized Graph Neural Networks~(\textbf{\infomotif}) }} \\
\midrule[0pt]

\textbf{{InfoMotif-GCN}}  & \checkmark  & \checkmark &
59.5 $\pm$ 0.9	&62.9 $\pm$ 0.7	&
\textbf{53.5 $\pm$ 0.6}	&56.9 $\pm$ 0.6 &
	56.6 $\pm$ 1.2	&60.7 $\pm$ 1.2  \\
\textbf{{InfoMotif-JKNet}}  & \checkmark & \checkmark & 
 \textbf{61.8 $\pm$ 1.6} &	\textbf{64.3 $\pm$ 1.2}&
53.1 $\pm$ 1.2&	\textbf{56.9 $\pm$ 0.6}& 
\textbf{62.7 $\pm$ 1.8}&	\textbf{67.9 $\pm$ 1.5}\\
\textbf{{InfoMotif-GAT}} & \checkmark & \checkmark & 
58.0 $\pm$ 0.4 & 60.4 $\pm$ 0.3 & 
 46.0 $\pm$ 1.5 & 50.0 $\pm$ 2.0 &
50.6 $\pm$ 1.3 & 56.3 $\pm$ 1.1\\
\bottomrule
\end{tabular}
\caption{Node classification results (\% test accuracy) on dis-assortative air-traffic networks. $\mathbf{X}$ and $\mathbf{Y}$ denote the use of attributes and labels respectively.  Structural embedding methods and structural GNNs typically outperform proximity-based models.~\infomotif~JK-Net achieves significant performance gains of 4\% to 14\% across different datasets.
}
\label{tab:airport_results}
\end{table}

In citation network datasets, GNN models generally outperform conventional embedding methods.
Moreover, attribute-agnostic structural embedding methods perform poorly and structural GNNs perform comparably to standard message-passing GNNs.
Citation networks exhibit strong attribute homophily in local neighborhoods;
thus, structural GNNs do not provide much performance benefits over state-of-the-art message-passing GNNs.
In contrast, our framework~\infomotif~regularizes GNNs to discover distant nodes with similar attributed structures across the entire graph.
~\infomotif~achieves consistent average classification accuracy gains of 3\%  for all three base GNN variants, with the GCN model variant achieving best empirical results overall.

In air-traffic networks, structural embedding methods outperform their proximity-based counterparts. We also observe a similar trend for structural GNNs over standard message-passing GNNs. Class labels rely more on node structural roles than the labels of neighbors in air-traffic networks.
JK-net outperforms other competing GNNs, signifying the importance of long-range dependencies in air-traffic networks.
Our framework~\infomotif~enables GNNs to learn structural roles agnostic to network proximity, and achieves significant gains of 10\% on average across all datasets.

\renewcommand*{\factortraining}{0.02}
\renewcommand*{\factor}{0.107}

\begin{table}[H]
\centering
\footnotesize
\noindent\setlength\tabcolsep{2.5pt}
\begin{tabular}{@{}p{0.2\linewidth}@{\hspace{0pt}}
K{\factortraining\linewidth}K{\factortraining\linewidth}@{\hspace{0pt}}
K{\factor\linewidth}K{\factor\linewidth}@{\hspace{10pt}}
K{\factor\linewidth}K{\factor\linewidth}@{\hspace{10pt}}
K{\factor\linewidth}K{\factor\linewidth}@{}} \\
\toprule
\multirow{1}{*}{\textbf{}} &  \multicolumn{2}{c}{\textbf{Data}} & \multicolumn{2}{c}{\textbf{DBLP-A}}  & \multicolumn{2}{c}{\textbf{DBLP-P}} & \multicolumn{2}{c}{\textbf{Movie}} \\
\cmidrule{2-3} \cmidrule(lr){4-5} \cmidrule(lr){6-7} \cmidrule(lr){8-9}
\multirow{1}{*}{\textbf{Training Ratio}} & \textbf{X} & \textbf{Y} & \textbf{10\%} & \textbf{20\%} & \textbf{10\%} & \textbf{20\%} & \textbf{10\%} & \textbf{20\%} \\
\midrule
\multicolumn{9}{c}{\textsc{Proximity-based Graph Embedding Methods}} \\
\midrule[0pt]
\textbf{{Node2Vec}~\cite{node2vec}} &   &
 &63.9 $\pm$ 0.4& 65.3 $\pm$ 0.6&
68.5 $\pm$ 0.4& 70.1 $\pm$ 0.5
 &54.1 $\pm$ 0.3  & 56.7 $\pm$ 0.3     \\
\textbf{{Motif2Vec}~\cite{motif2vec}}  &   &
 & 62.7 $\pm$ 0.7 & 64.5 $\pm$ 0.6&
68.9 $\pm$ 0.9& 71.3 $\pm$ 1.1
 &55.0 $\pm$ 0.3 & 57.4 $\pm$ 0.5    \\
\midrule[0pt]
\multicolumn{9}{c}{\textsc{Structural Graph Embedding Methods}} \\
\midrule[0pt]
\textbf{{Struct2Vec}~\cite{struc2vec}}  &  &
 & 34.2 $\pm$ 0.4 & 36.1 $\pm$ 0.4&
 34.9 $\pm$ 0.2 & 35.7 $\pm$ 0.3
 &32.7 $\pm$  0.4 &  34.9 $\pm$  0.1   \\
\textbf{GraphWave~\cite{graphwave}} &   &    
 & 34.8 $\pm$ 0.5 &37.0 $\pm$ 0.6&
 35.6 $\pm$ 0.2 & 36.3 $\pm$ 0.3
 & 33.5 $\pm$  0.4 & 36.0 $\pm$   0.4  \\

\textbf{DRNE~\cite{drne}} &  &    
 & 33.9 $\pm$ 0.3& 36.5 $\pm$ 0.3&
  35.1 $\pm$ 0.5 & 35.5 $\pm$ 0.4 
 & 31.0 $\pm$ 0.2 &  35.1 $\pm$  0.2   \\

\midrule[0pt]
\multicolumn{9}{c}{\textsc{Standard Graph Neural Networks}} \\
\midrule[0pt]
\textbf{GCN~\cite{gcn}}  & \checkmark  & \checkmark  
 & 65.3 $\pm$ 1.1& 69.6 $\pm$ 0.9&
 71.3 $\pm$ 0.7& 73.4 $\pm$ 0.8 
 & 55.7 $\pm$ 1.0  &  57.3 $\pm$ 0.7  \\
\textbf{GAT~\cite{gat}}  & \checkmark   & \checkmark   
 & 67.5 $\pm$ 0.8 & 71.7 $\pm$ 0.8&
 71.9 $\pm$ 0.5& 73.0 $\pm$ 0.7
 &58.6 $\pm$ 0.9 & 59.9 $\pm$  1.0   \\
\textbf{GraphSAGE~\cite{graphsage}} & \checkmark  & \checkmark   
 & 65.3 $\pm$ 0.7&69.0 $\pm$ 0.6&
 70.9 $\pm$ 0.8&72.7 $\pm$ 0.8
 &55.6 $\pm$  0.4&  56.4 $\pm$ 0.8    \\
\textbf{JKNet~\cite{jknet}} &\checkmark  & \checkmark  
 & 69.6 $\pm$ 1.0&73.2 $\pm$ 1.2&
69.8 $\pm$ 1.1&72.0 $\pm$ 1.3
 &58.3 $\pm$ 0.7 &  60.5 $\pm$  0.9   \\
\textbf{DGI~\cite{dgi}}  & \checkmark   &
 &64.7 $\pm$ 0.5 &68.5 $\pm$ 0.7&
 41.9 $\pm$ 0.8& 61.1 $\pm$ 0.6
 &38.6 $\pm$ 0.9 &  40.4 $\pm$  0.9   \\

\midrule[0pt]
\multicolumn{9}{c}{\textsc{Structural Graph Neural Networks}} \\
\midrule[0pt]

\textbf{DemoNet~\cite{demonet}} & \checkmark   & \checkmark  
 & 70.7 $\pm$ 1.3&72.3 $\pm$ 1.1&
72.6 $\pm$ 0.9 & 73.5 $\pm$ 0.6
 & 59.5 $\pm$ 0.8 &  61.2 $\pm$  0.8  \\
\textbf{Motif-CNN~\cite{motifcnn}} & \checkmark  & \checkmark  
 & 66.4 $\pm$ 1.1& 70.1 $\pm$ 1.3&
 71.5 $\pm$ 0.8 & 72.2 $\pm$ 0.6 
 & 54.3 $\pm$ 0.3 & 56.9 $\pm$ 0.5    \\
\textbf{MCN~\cite{motif_attention_cikm19}} &  \checkmark & \checkmark  
 & 67.1 $\pm$ 1.2 & 71.2 $\pm$ 1.1 &
 71.9 $\pm$ 0.9 & 72.5 $\pm$ 0.6 
 & 54.7 $\pm$ 0.4 &  57.2 $\pm$ 0.6    \\

\midrule[0pt]
\multicolumn{9}{c}{\textsc{Heterogeneous graph Neural Networks}}\\
\midrule[0pt]
\textbf{HAN~\cite{han}} & \checkmark  & \checkmark  
 & 68.2 $\pm$ 1.0 &72.0 $\pm$ 1.3 &
 73.1 $\pm$ 0.9& 74.0 $\pm$ 0.7
 &60.7 $\pm$  1.1& 62.1 $\pm$ 0.8    \\
\textbf{MAGNN~\cite{magnn}}& \checkmark  & \checkmark  
 & 68.9 $\pm$ 0.7&72.5 $\pm$ 0.7&
74.7 $\pm$ 0.6 & 75.8 $\pm$ 0.7
 & 62.1 $\pm$ 0.9 &  63.0 $\pm$   0.5  \\
\textbf{Meta-GNN~\cite{metagnn}}& \checkmark  & \checkmark  
 & 71.3 $\pm$ 1.2 &73.9 $\pm$ 1.4&
 74.6 $\pm$ 0.6 &75.8 $\pm$ 0.6
 &61.7 $\pm$ 0.5 &  63.7 $\pm$  0.7   \\

\midrule[0pt]
\multicolumn{9}{c}{\textsc{Motif-regularized Graph Neural Networks~(\textbf{\infomotif}) }} \\
\midrule[0pt]

\textbf{{InfoMotif-GCN}}  & \checkmark  & \checkmark
 &73.7 $\pm$ 1.2 &77.4 $\pm$ 1.1&
 \textbf{78.8 $\pm$ 0.4} &\textbf{79.0 $\pm$ 0.7}
 &\textbf{64.7 $\pm$  0.8} &\textbf{65.0  $\pm$ 1.1}    \\
\textbf{{InfoMotif-JKNet}}  & \checkmark & \checkmark 
 &\textbf{75.6 $\pm$ 0.9} &\textbf{79.9 $\pm$ 0.8} &
 75.5 $\pm$ 1.0 & 76.3 $\pm$ 1.2
 & 60.7 $\pm$  0.6 &  62.0 $\pm$   0.6  \\
\textbf{{InfoMotif-GAT}} & \checkmark & \checkmark 
  & 72.4 $\pm$ 1.0 &75.3 $\pm$ 1.4&
77.1 $\pm$ 0.7&77.9 $\pm$ 0.6
 & 62.8 $\pm$  0.5& 64.2  $\pm$    0.5 \\
\bottomrule
\end{tabular}
\caption{Node classification results (\% test accuracy) on heterogeneous graphs from bibliographic and movie networks, with schemas shown in figures~\ref{fig:dblp_schema} and~\ref{fig:movie_schema}. $\mathbf{X}$ and $\mathbf{Y}$ denote the use of node attributes and training labels respectively.  Heterogeneous GNN models (MAGNN, Meta-GNN) typically outperform structural GNNs (DemoNet, Motif-CNN, MCN) and type-agnostic message-passing GNNs (GCN, GraphSAGE, GAT). Our framework~\infomotif~(with typed network motif regularization) achieves consistent and significant performance gains of 5\% on average across the three heterogeneous network datasets.}
\label{tab:hin_results}
\end{table}

\subsection{{Heterogeneous Graphs ($\text{RQ}_2$)}}
\label{sec:infomotif_heterogeneous_results}
In heterogeneous graphs, we train~\infomotif~and other baselines that use motifs/metagraphs using typed 3-node network motifs (shown in Figures~\ref{fig:dblp_motifs} and~\ref{fig:movie_motifs}) that are defined based on the heterogeneous type schema (Figure~\ref{fig:dblp_schema} and~\ref{fig:movie_schema}). Our experimental results comparing our framework~\infomotif~with baselines on DBLP and Movie networks, are shown in Table~\ref{tab:hin_results}.

We find that message-passing GNNs (such as GCN, GAT) generally outperform conventional embedding methods (such as node2vec).
Structural embedding methods (\textit{e.g.}, struc2vec) perform poorly; this reveals their inability to capture structural aspects relevant to heterogeneous graphs with rich type semantics. 
Heterogeneous GNNs such as MAGNN and Meta-GNN outperform homogeneous GNNs owing to their type-aware semantic neighbor aggregation via metapaths and metagraphs respectively.
Our framework~\infomotif~further learns type-aware attributed structural roles which results in significant performance gains of 5\% on average over prior approaches.

\renewcommand*{\factor}{0.142}
\begin{table}[hbtp]
    \centering
    \noindent\setlength\tabcolsep{2.9pt}
\begin{tabular}{@{}p{0.4\linewidth}@{\hspace{9pt}}K{\factor\linewidth}K{\factor\linewidth}K{\factor\linewidth}@{}}
        \toprule
        \textbf{Dataset} & \textbf{Cora} & \textbf{Citeseer} & \textbf{Pubmed} \\ 
        \midrule
        InfoMotif-GCN ($L_S + \lambda L_{MI}$) &  \textbf{87.4 $\pm$ 0.4} & \textbf{78.5 $\pm$ 0.5} & \textbf{88.3 $\pm$ 0.2}  \\
        w/o novelty weights ($\beta_v=1$ in~\cref{eqn:infomotif_supervised_loss}) & 86.4 $\pm$ 0.5  & 77.6 $\pm$ 0.5 & 87.8 $\pm$ 0.3\\
        w/o task weights ($\alpha_{vt} = 1$ in~\cref{eqn:infomotif_full_mi_loss}) & 84.6 $\pm$ 0.4 & 77.3 $\pm$ 0.4 & 87.3 $\pm$ 0.2\\ 
        w/o novelty and task weights & 84.0 $\pm$ 0.5 & 76.4 $\pm$ 0.6 & 87.3 $\pm$ 0.2\\
        Base model GCN  ($L_B$) & 82.0 $\pm$ 0.4 &  76.6 $\pm$ 0.3 & 86.1 $\pm$ 0.5 \\
        \bottomrule
    \end{tabular}
    \caption{Ablation study results with 40\% training ratio on citation networks.
    The novelty and task weighting strategies improve classification accuracies by 2\% on average.}
    \label{tab:ablation}
\end{table}

\subsection{{Model Ablation Study ($\text{RQ}_3$)}}
\label{sec:infomotif_ablation}
We present an ablation study on two citation networks Cora and Citeseer, to analyze the importance of major components in~\infomotif~(Table~\ref{tab:ablation}). In our experiments, we choose GCN as the base GNN model due to its consistently high performance.

\begin{itemize}[leftmargin=*]
    \item \textbf{Remove novelty-driven sample weighting}. 
    We set the novelty weight $\beta_v=1$ (\Cref{eqn:infomotif_supervised_loss}) for each labeled node $v \in \gV_L$ to test the importance of addressing motif occurrence skew. We observe consistent 1\% gains due to our novelty-driven sample weighting.
    \item \textbf{Remove task-driven motif weighting}. 
    We remove the node-sensitive motif weights from the motif regularization loss (\Cref{eqn:infomotif_full_mi_loss}) by setting $\alpha_{vt}=1$ for every node-motif pair.
    Contextually weighting the different motif regularizers at a node-level granularity results in 2\% average accuracy gains across both datasets.
    \item \textbf{Remove both novelty and task driven weighting}.
This variant applies a uniform motif regularization over all nodes without  distinguishing the node-sensitive relevance of each motif; 
this significantly degrades classification accuracy.
\end{itemize}

\subsection{{Qualitative Analysis ($\text{RQ}_4$)}}
\label{sec:infomotif_analysis}
We qualitatively examine the source of~\infomotif's gains over the base GNN (GCN due to its consistent performance).
by analyzing \textit{node degree}, \textit{label sparsity} and \textit{attribute diversity} in local node neighborhoods, on the Cora and Citeseer networks.

\subsubsection*{\textbf{Node Degree}}
\begin{figure}[hbtp]
    \centering
    \includegraphics[width=0.9\linewidth]{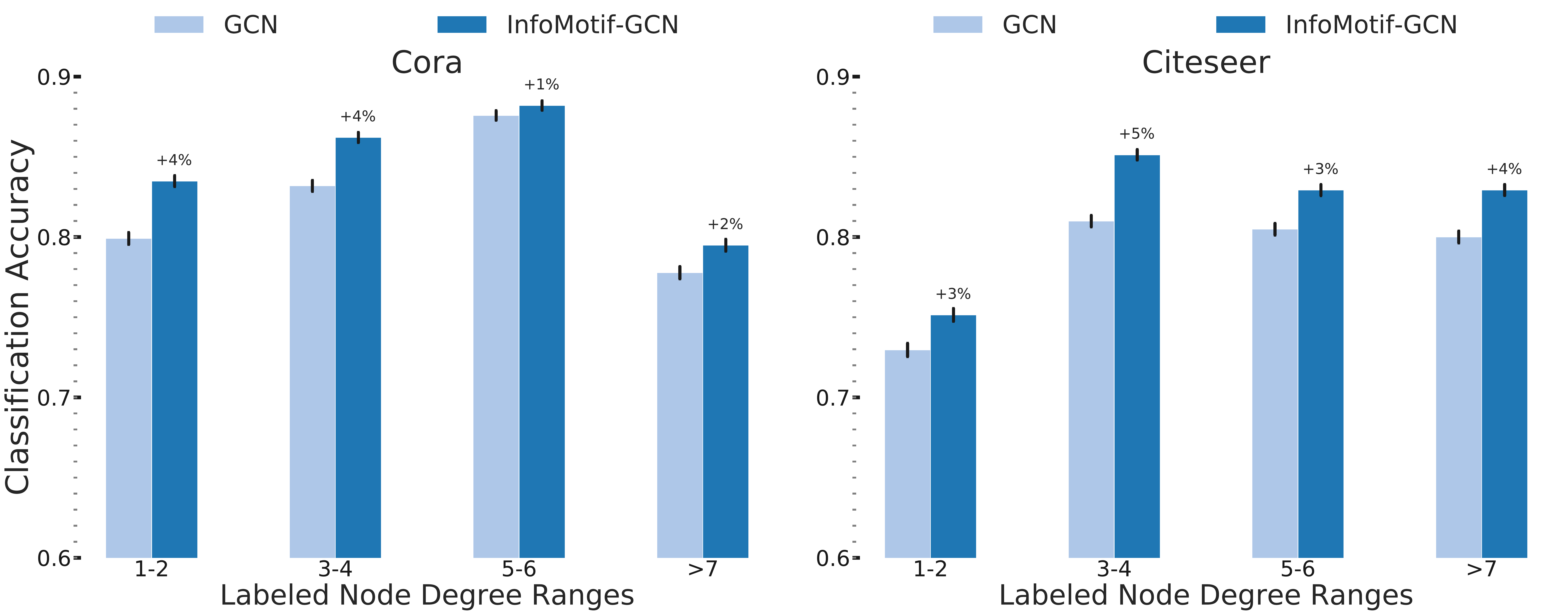}
    \caption{Classification accuracy with respect to node degree.~\infomotif~has consistent gains across all segments with higher gains for low-to-medium degree nodes (quartiles Q1 \& Q2).}
    \label{fig:node_deg}
\end{figure}

To evaluate performance variance with node degree, we divide the set of test nodes into bins based on four degree ranges. Figure~\ref{fig:node_deg} depicts the 
variation in classification accuracy for GCN and~\infomotif-GCN~across degree segments, on Cora and Citeseer datasets.

\infomotif~has consistent performance improvements over GCN across all degree segments, with notably higher gains for low-to-medium degree nodes (quartiles Q1 and Q2). Learning structural roles through self-supervised motif regularization is beneficial for nodes with limited local structural information.

\subsubsection*{\textbf{Label Sparsity}}
\begin{figure}[hbtp]
    \centering
    \includegraphics[width=0.9\linewidth]{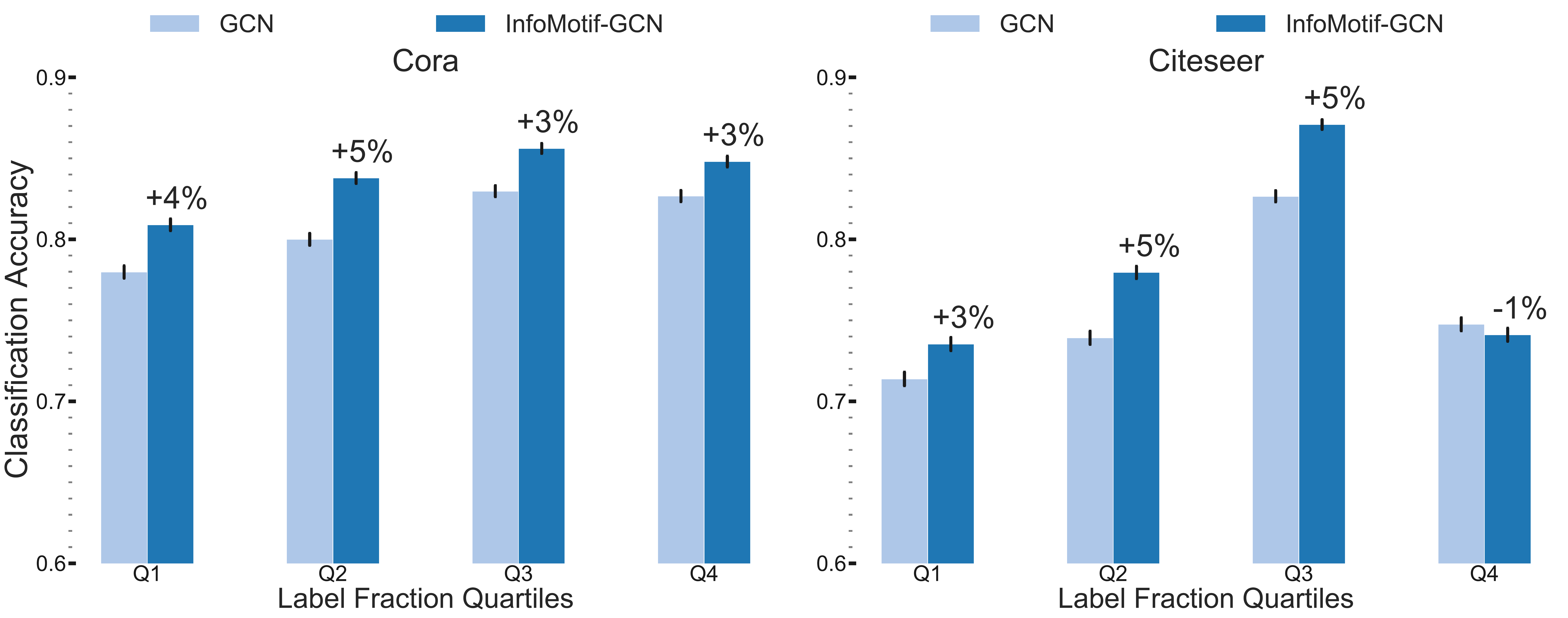}
    \caption{Classification accuracy over \textit{label fraction} quartiles. (Q1: smaller label fraction).~\infomotif~has larger gains over GCN in Q1 \& Q2 (nodes that exhibit label sparsity)}
    \label{fig:label_sparsity}
\end{figure}

We define the \textit{label fraction} for a node as the fraction of labeled training nodes in its 2-hop neighborhood, \textit{i.e.}, a node exhibits label sparsity if it has very few or no labeled training nodes within its 2-hop aggregation range.
We separate the set of test nodes into four quartiles by their label fraction.
Figure~\ref{fig:label_sparsity} depicts classification results for GCN and~\infomotif-GCN under each quartile (Q1 has nodes with small label fractions).

~\infomotif~has stronger performance gains over GCN for nodes with smaller label fractions (quartiles Q1 and Q2); this empirically validates the efficacy of our motif-based regularization framework in addressing the key limitation of $k$-hop localization in message-passing GNNs~(\Cref{sec:infomotif_role_learning}), \textit{i.e.},~\infomotif~benefits nodes with very few or no labeled nodes within their $k$-hop aggregation ranges, thus effectively addressing \textit{label sparsity} challenges.

\subsubsection*{\textbf{Attribute Diversity}}
\begin{figure}[hbtp]
    \centering
    \includegraphics[width=0.9\linewidth]{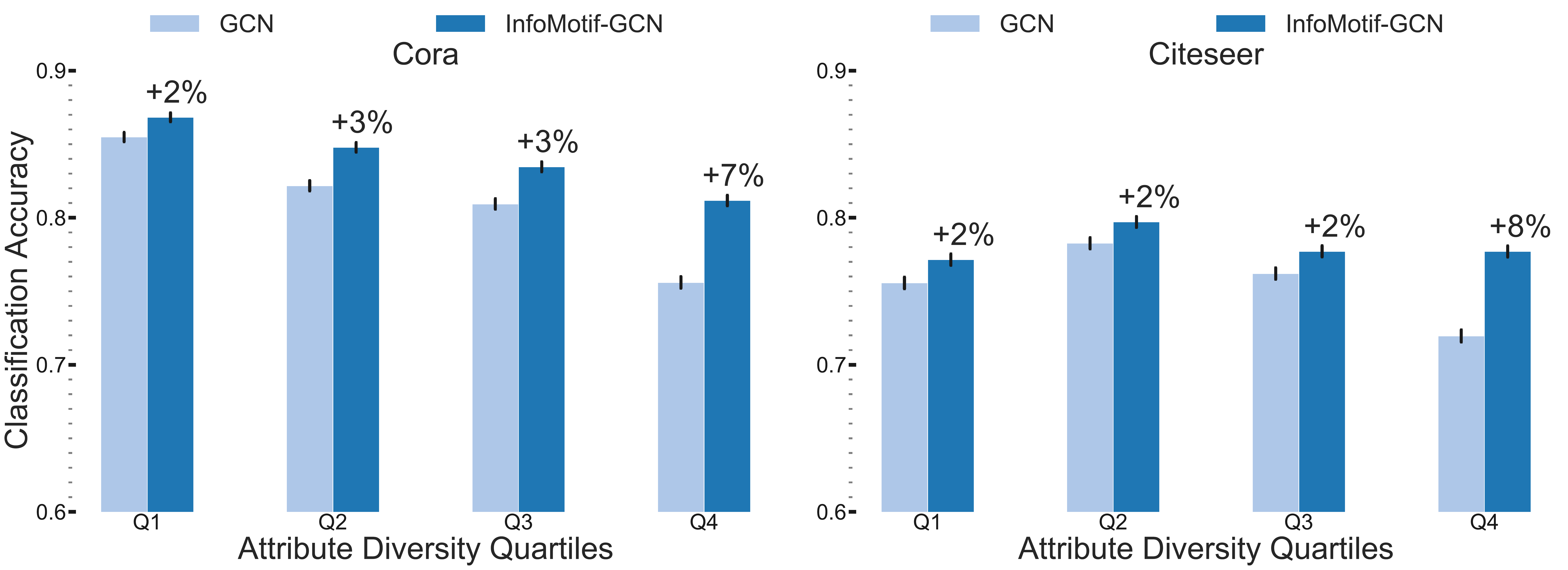}
    \caption{Classification accuracy across attribute diversity quartiles. (Q4: high attribute diversity).~\infomotif~has stronger gains in Q3 \& Q4 (nodes with diverse attributed neighborhoods).}
    \label{fig:cosine_sim}
\end{figure}

We measure the local \textit{attribute diversity} of a node by the mean pair-wise attribute dissimilarity (computed by cosine distance) of itself with other nodes in its 2-hop neighborhood, \textit{i.e.}, a node that exhibits strong homophily with its neighbors has low attribute diversity. 
We report classification results across attribute diversity quartiles in Figure~\ref{fig:cosine_sim}.

Nodes with diverse attributed neighborhoods are typically harder examples for classification.
Regularizing GNNs to learn attributed structures via motif occurrences can accurately classify diverse nodes, as evidenced by the higher relative gains of~\infomotif~for diverse nodes (quartiles Q3 and Q4).

\subsection{{Parameter Sensitivity}}
\label{sec:infomotif_sensitivity}
\begin{figure}[hbtp]
    \centering
    \includegraphics[width=0.9\linewidth]{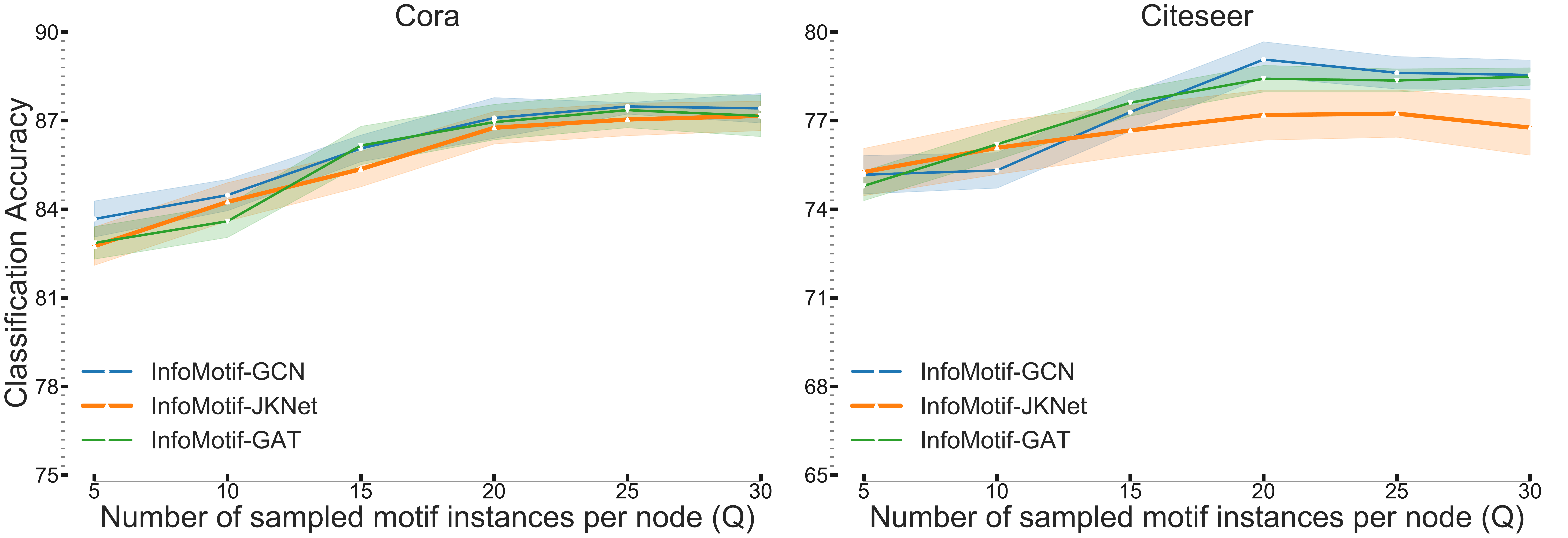}
    \caption{Classification accuracy increases slowly with the number of sampled motif instances and stabilizes around 15 to 20. Variance bands indicate 95\% confidence intervals over 10 runs.}
    \label{fig:sensitivity}
\end{figure}

We examine the effect of hyper-parameter $Q$ that controls the number of motif instances sampled per node to train our motif-based discriminators (\Cref{eqn:infomotif_mi_loss}).~\Cref{fig:sensitivity} shows variation in accuracies of our three GNN variants with the number of sampled instances (5 to 30), on Cora and Citeseer networks.

Performance of all GNN variants stabilize with 20 instances across both datasets.
Since the complexity of our framework scales linearly with $Q$, we fix $Q=20$ across datasets to provide an effective trade-off between compute-cost and performance

\subsection{{Efficiency Analysis}}
\label{sec:infomotif_efficiency}
\begin{figure}[ht]
    \centering
    \includegraphics[width=0.9\linewidth]{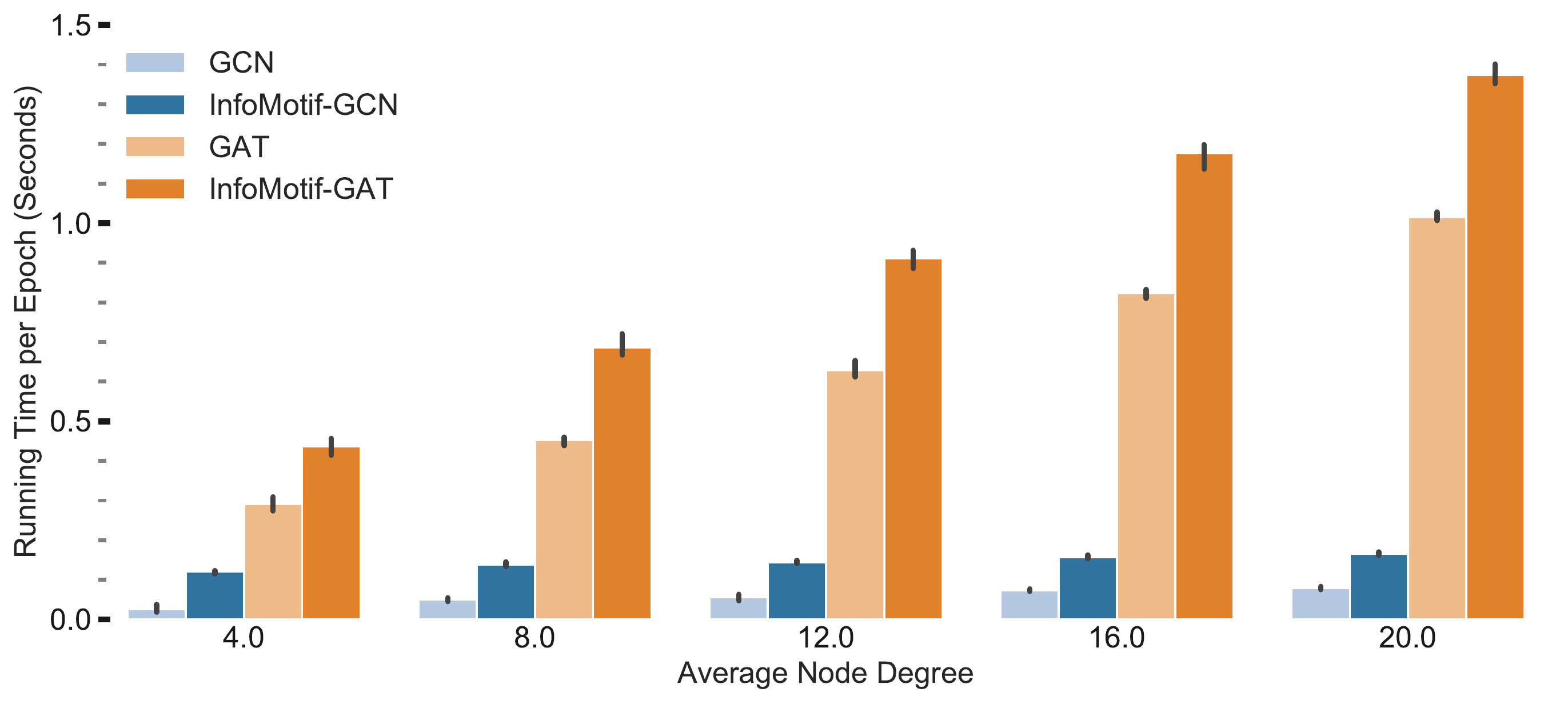}
    \caption{Runtime comparison of~\infomotif~variants with its base GNNs.~\infomotif~has minimal computational overheads; notice the nearly constant runtime gap with increasing node degree.}
    \label{fig:runtime}
\end{figure}

We empirically evaluate the added complexity of~\infomotif~on two GNN models, GCN and GAT.
We report the time per epoch %
on synthetically generated Barabasi-Albert networks~\cite{barabasi} with 5000 nodes and increasing link density (\Cref{fig:runtime}).

\infomotif~adds a small fraction of the base GNN runtime, and the added complexity scales linearly with the number of nodes, as evidenced by its nearly constant runtime gap over increasing link density (\Cref{fig:runtime}).
Furthermore, our GCN variant~\infomotif-GCN is significantly more efficient than GAT.

\section{{Discussion}}
\label{sec:infomotif_discussion}
Our framework is designed to address the limitations of \textit{oversmoothing} and \textit{localization} in prior message-passing GNNs.

~\infomotif~is orthogonal to advances in GNN architectures that improve the \textit{structural distinguishability} of node representations through carefully designed neighborhood aggregators.
In contrast, we enhance the structural resolution of node representations by regularizing base GNNs through self-supervised learning objectives designed to capture connectivity in higher-order motif structures.
By training contrastive discriminators to discover attribute correlations among motif instances across the entire graph, our approach learns generalizable \textit{global} roles in addition to modeling local connectivity patterns.
The enhanced quality of our learned node representations is further evidenced by our superior empirical performance for nodes with diverse attributed neighborhoods (Section~\ref{sec:infomotif_analysis}).

~\infomotif~addresses the challenge of \textit{localization} by regularizing base GNNs to learn attributed structural roles through self-supervised training objectives. 
Specifically, our approach statistically relates distant nodes in the graph with co-varying attributed structures, to effectively overcome label sparsity in local neighborhoods (Section~\ref{sec:infomotif_analysis}).
Instead of adopting deeper GNNs that directly expand neighborhood aggregation ranges, we demonstrate the effectiveness of regularizing shallow base GNNs to learn attributed structural roles.
Compared to deeper GNNs that scale poorly with neighborhood sizes, our regularization strategy enables efficient model inference.

In our work, we choose network motif structures as the central basis to formulate structural roles.
In contrast to alternative approaches to quantify structural similarity based on coarse properties like degree sequences~\cite{struc2vec} or rigid notions of structural equivalence~\cite{rossi2019community}, network motifs are fundamental higher-order connectivity structures that enable flexible generalization to complex heterogeneous graphs with rich semantics.
We empirically demonstrate the utilities of untyped motifs in homogeneous graphs and typed motifs in heterogeneous graphs.

Our key modeling hypothesis is the importance of attribute co-variance in local structures towards the learning application (\textit{e.g.}, classification in social networks).
Our substantial gains on two diverse classes of datasets indicates broad applicability for~\infomotif~across graphs with varied structural characteristics.
However, the performance gains may diminish in application scenarios (\textit{e.g.}, learning in regular mesh graphs) where modeling such co-variance is not beneficial or even necessary.

\section{Related Work}
Our work is related to semi-supervised learning approaches over \textit{homogeneous} and \textit{heterogeneous} graphs, and recent advances in the paradigm of \textit{self-supervised} learning.

\textbf{Homogeneous Graphs:}
Semi-supervised learning over graphs is a well-studied problem, where the goal is to classify nodes in a graph given a small set of labeled examples.
The most popular label spreading~\cite{ssl-survey} techniques propagate labels through linked nodes in the graph based on smoothness assumptions.
Graph Neural Networks (GNNs) generalize label spreading through localized message passing over feature-rich node neighborhoods and have achieved state-of-the-art results in several benchmarks~\cite{gcn}.
GNNs learn node representations by recursively aggregating features from local neighborhoods in an end-to-end manner, with diverse applications, including information diffusion prediction~\cite{infvae}, social recommendation~\cite{social_adv}, and community question answering~\cite{irgcn}.
Graph Convolutional Networks (GCNs)~\cite{gcn} learn degree-weighted aggregators, which can be interpreted as a special form of Laplacian smoothing~\cite{gcn_oversmoothing}.
Many models generalize GCN with a wide range of neighborhood aggregators, \textit{e.g.}, self-attentions~\cite{gat, dysat}, mean and max pooling functions~\cite{graphsage}, etc.
However, all these models learn node representations that inherently overfit to the $k$-hop neighborhood around each node.

There are two broad categories of prior graph representation learning approaches that aim to overcome the key limitations of \textit{oversmoothing} and \textit{localization} in GNNs:  \textit{non-local GNNs} that capture contributions from distant nodes in the graph; and \textit{structural role learning} techniques that enhance the structural distinguishability of the learned node representations.

Non-local methods expand the propagation range of GNNs to aggregate node representations of differing localities, \textit{e.g.}, JKNet~\cite{jknet} uses skip-connections to vary the influence radius per node, PGNN~\cite{pgnn} captures global network positions via shortest-paths, and DGI~\cite{dgi} maximizes MI between node representations and a summary representation of the entire graph.
However, they either operate on a local scale~\cite{gmi}, or learn coarse structural properties,  which limits their ability to capture features from distant yet structurally similar nodes.

Role-aware models embed structurally similar nodes close in the latent space, independent of network position~\cite{rolx, rossi2019community}.
A few approaches~\cite{drne} employ strict definitions of structural equivalence to embed nodes with identical local structures to the same point in the latent space, while others utilize structural node features (\textit{e.g.}, node degrees, motif count statistics) to extend classical proximity-preserving embedding methods, \textit{e.g.}, feature-based matrix fa  ctorization~\cite{hone} and random walk methods~\cite{struc2vec}.
Notably, a few methods design structural GCNs via motif adjacency matrices~\cite{motif_attention_cikm19, motifcnn, metagnn}.
However, all these methods model structural roles without considering node attributes.
A related direction is higher-order network representation learning that models proximity via network motifs~\cite{motif2vec}.
However, such representations are highly localized and cannot identify structurally similar nodes independent of network proximity.
In contrast, we regularize GNNs to learn attributed structural roles based on the co-variance of attributes in motifs, thus simultaneously enhancing the distinguishability of node representations and identifying correspondences between distant nodes.

\textbf{Heterogeneous Graphs:}
Representation learning techniques over heterogeneous graphs primarily focus on preserving structural information indicated by the type semantics in meta-path or meta-graph structures.
A few popular approaches include node representation learning by capturing proximities between node pairs connected via meta-graphs~\cite{metagraph2vec, aspem} and meta-path guided random walks~\cite{hin2vec, metapath2vec}.
Recently, graph neural networks have been generalized to heterogeneous graphs through message passing aggregation over local neighborhoods induced via specific node types~\cite{hetgnn, hgt}, meta-paths~\cite{graphinception, han, magnn} and meta-graphs~\cite{metagnn}.
While these advances effectively incorporate rich heterogeneous semantics into message-passing GNNs, the key limitation due to localization remains.
To our knowledge, structure role learning in heterogeneous graphs is unexplored and ours is the first to examine structural role learning in GNNs with rich type semantics and attributes.

\textbf{Self-supervised Learning:}
The emerging paradigm of self-supervised learning~\cite{dim} aims to alleviate the need for large volumes of labeled examples by extracting supervision signals from the intrinsic structure of the raw data.
For instance, auxiliary supervision signals for images are created by rotating, cropping and colorizing images, followed by new training objectives to facilitate representation learning~\cite{s4l}.
One empirically effective strategy is mutual information maximization~\cite{bachman2019learning} to maximize agreement across different views of the data.
A few recent advances extend self-supervised learning to graph representation learning by exploiting structural properties such as node degree, proximity~\cite{peng2020self}, and attributes~\cite{simpgcn}, for model pre-training~\cite{graphcl, gcc, graph_ssl}.
In our work, we design self-supervised learning objectives to regularize graph neural networks for node classification by learning attribute correlations in higher-order connectivity patterns (typed and untyped motif structures).

\section{Conclusion}
\label{sec:infomotif_conclusion}
This chapter presents a new class of motif-regularized GNNs with an architecture-agnostic framework~\infomotif~for semi-supervised learning on graphs.
To overcome limitations of prior GNNs due to localized message passing,
we introduce attributed structural roles to regularize GNNs by learning statistical dependencies between structurally similar nodes with co-varying attributes, independent of network proximity.
~\infomotif~maximizes motif-based mutual information, and dynamically prioritizes the significance of different motifs.
Our experiments on nine real-world datasets spanning homogeneous and heterogeneous networks, show substantial consistent gains for~\infomotif~over state-of-the-art methods.

In this chapter, we have explored a transductive learning scenario of semi-supervised learning over graph-structured entity interactions given sparse training labels. We leveraged higher-order connectivity structures (network motifs) between nodes to compensate for the lack of sufficient labeled information in the local neighborhoods.
In the next chapter, we examine another central transductive scenario of collaborative filtering in the bipartite user-item interaction setting. Our key focus is to enable accurate personalized recommendations for data-poor tail entities (such as long-tail items) with severe interaction sparsity.

\chapter{\textsc{ProtoCF}: Few-shot Collaborative Filtering via Meta-Transfer}
\label{chap:protocf}

\section{Introduction}
\label{sec:protocf_intro}
Neural Collaborative Filtering (NCF) methods have recently enabled substantial advances in modern recommender systems that are critical to diverse e-commerce applications.
Collaborative Filtering (CF) methods generate personalized recommendations by learning patterns from historical user-item interactions. %
However, a close examination of neural recommenders' performance reveals a \textit{paradox}: while the overall recommendation accuracy is high, accuracy levels are poor for most items in the inventory.
A majority of recommendations are biased towards popular items~ %
\cite{popularity_bias}, while ignoring \textit{long-tail} items in under-represented categories.
\textit{Popularity bias} restricts personalization 
and impedes suppliers of long-tail items, who struggle to attract consumers given the low exposure.
Targeting long-tail items can enhance recommendation diversity and bring relatively larger marginal profits, compared to popular items with competitive markets.
The increasing impact of recommendations (\textit{e.g.}, 80\% of Netflix activity is a result of recommendations~\cite{netflix}), also raises \textit{ethical} questions: can items that are never recommended by a system be considered an instance of \textit{discrimination}~\cite{holzapfel2018ethical}?
Thus, %
we focus on learning robust models for long-tail item recommendations.

\textbf{Empirical evidence of long-tail challenges}: 
We highlight two key observations on prior neural recommendation models~\cite{ncf, vae-cf, cdae}: %

\begin{figure}[ht]
\centering
\includegraphics[width=0.9\linewidth]{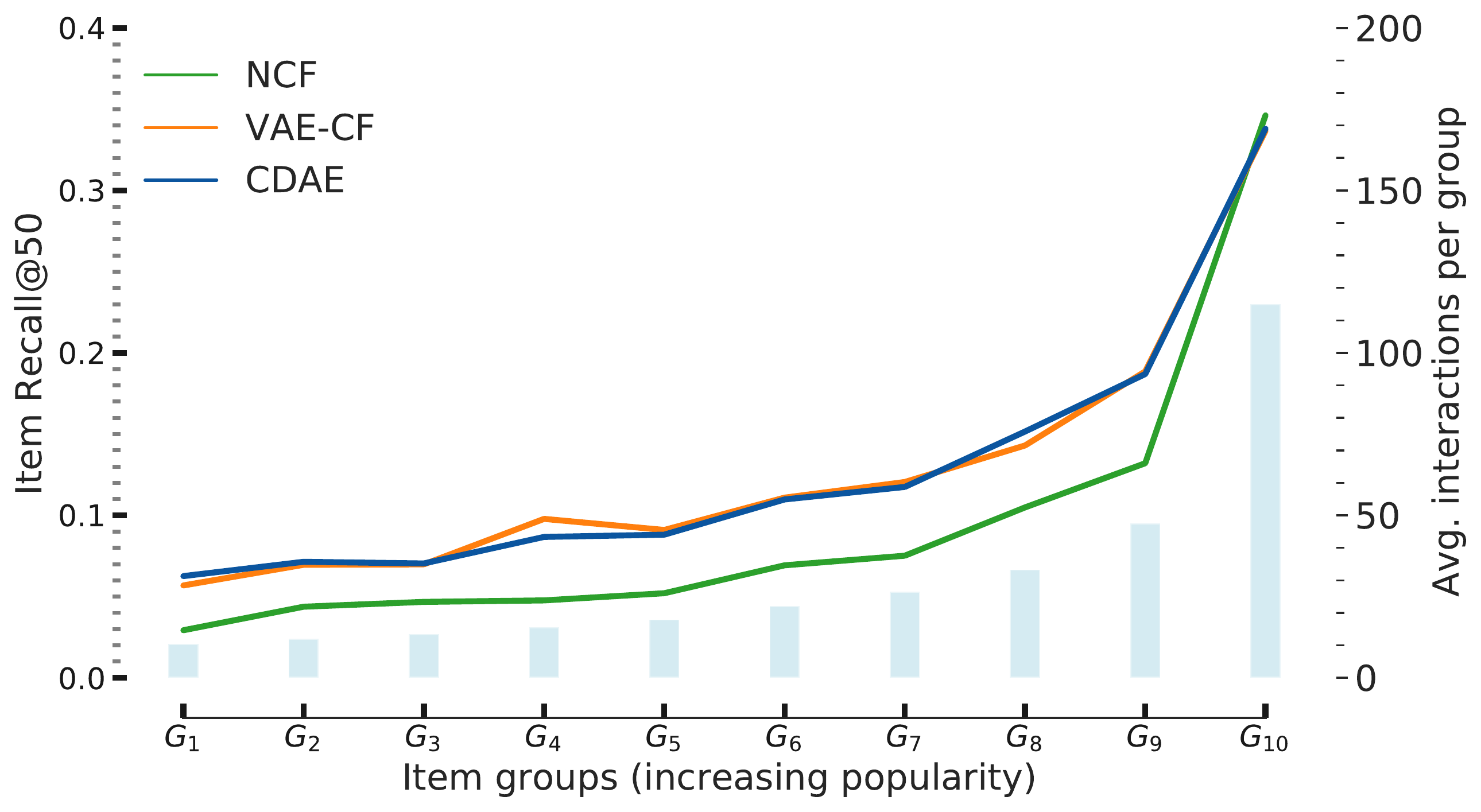}
\caption{Item Recall@50 of three popular neural recommendation models across different item-groups (increasing popularity) in Epinions.
Neural recommenders \textit{overfit} to \textit{popular} items with considerably lower performance for \textit{long-tail} items.}
\label{fig:intro_item_recall}
\end{figure}

\begin{itemize}
\item %
Performance gains are \textit{skewed} towards popular items with abundant historical interactions.
Figure~\ref{fig:intro_item_recall} depicts Item Recall$@K$ (fraction of correctly ranked items within top-$K$) of three neural recommenders for different item-groups ordered by increasing item interaction count.
Model performance is considerably lower for long-tail items (low popularity), indicating a clear \textit{popularity bias}.
\item Prior recommenders lack the \textit{resolution power} to accurately rank relevant items within the long-tail. To empirically illustrate this observation, we split the item inventory into two equal-sized sets by interaction count (popular and long-tail), and evaluate personalized ranking performance independently within each set.
Table~\ref{tab:intro_table} shows a big performance gap between ranking within the popular and long-tail item sets,  which reflects on the poor quality of their long-tail item representations.

\end{itemize}

\renewcommand*{\factor}{0.13}

\begin{table}[hbtp]
\centering
\begin{tabular}{@{}p{0.25\linewidth}K{\factor\linewidth}K{\factor\linewidth}
K{\factor\linewidth}@{\hspace{13pt}}K{\factor\linewidth}K{\factor\linewidth}K{\factor\linewidth}@{}} \\
\toprule
{\textbf{Item Subset}} &  \multicolumn{2}{c}{\textbf{Top 50\% Items} } & \multicolumn{2}{c}{\textbf{Bottom 50\% Items}} \\
\cmidrule(lr){2-3} \cmidrule(lr){4-5}
\textbf{Metric} & \textbf{N@50} & \textbf{R@50} & \textbf{N@50} &  \textbf{R@50} \\
\midrule
NCF~\cite{ncf}  & 0.0906 &  0.1874 &  0.0352 & 0.0973  \\
VAE-CF~\cite{vae-cf} & 0.1055 & 0.2106 & 0.0457  & 0.1125 \\
CDAE~\cite{cdae} & 0.1050 & 0.2102 & 0.0471 & 0.1149 \\
\bottomrule
\end{tabular}
\caption{
Recommendation performance within top-50\%  \textit{head} and bottom-50\%  \textit{tail} items (ordered by item popularity) on Epinions. R@50 and N@50 denote Recall@50 and NDCG@50 metrics respectively. Compared to \textit{head} items, we observe poor item ranking \textit{resolution} within the long-tail.
}
\label{tab:intro_table}
\end{table}

Prior efforts towards the long-tail have been two-fold. First, regularizing an existing recommender via inter-item associations (derived from item co-occurrences)~\cite{challenging_long_tail, cofactor, efm, cikm18adv} or distributional priors over the latent space~\cite{vae-cf, enhanced_vae}. %
While regularization strategies can partially alleviate overfitting, 
they typically involve static hypotheses that may be counter-productive for massive item inventories with a diverse collection of long-tail items. 
Second, incorporating external side information~\cite{cvae}, \textit{e.g.}, item attributes~\cite{esam}, social networks~\cite{social_rec}, and knowledge graphs~\cite{kgat} to overcome interaction sparsity.
Note that exploiting side information is typically application-dependent, thus lacking generalizability to diverse scenarios.
In contrast, we exploit the capabilities of any neural base recommender for long-tail item recommendation without any requirement of side information.

\begin{figure}[th]
  \centering
  \includegraphics[width=0.9\linewidth]{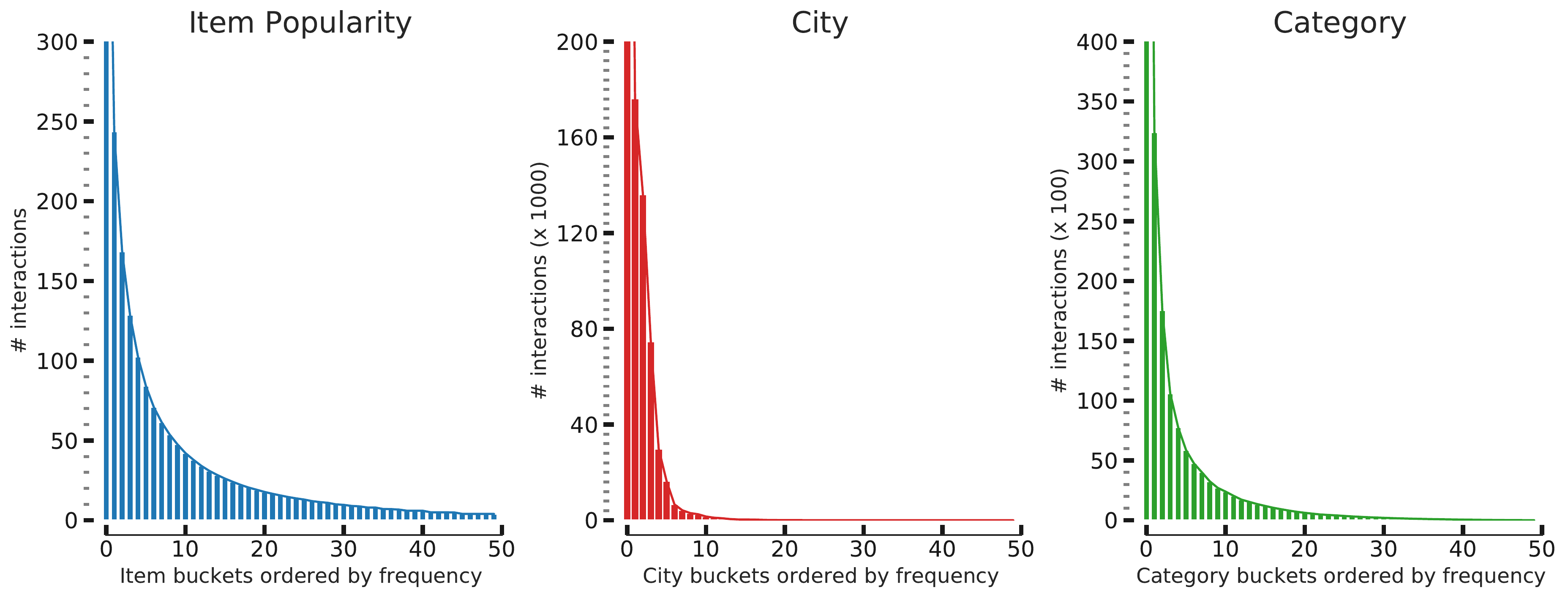}
  \caption{Multiple long-tailed distributions along different item dimensions. (a) shows interaction frequency distribution over unique \textit{items}. (b) and (c) show frequency distributions over unique \textit{cities} and \textit{categories} of Yelp businesses.}
  \label{fig:multi_tail}
\end{figure}

Learning a latent space tailored to long-tail items poses two key challenges:
first, \textit{sparsity and heterogeneity}: although tail items have sparse interactions, they belong to diverse item categories, \textit{e.g.}, Figure~\ref{fig:multi_tail} shows long-tailed distributions along different dimensions described by item attributes, location and category of Yelp businesses.
Thus, we need sufficient resolution power to learn discriminative representations for tail items, while being careful to avoid overfitting.
Second, \textit{distribution mismatch} between head items (substantial interactions) and tail items (sparse interactions).
Specifically, neural recommenders typically sample user-item pairs from the overall interaction distribution (biased to head items) for model training; this results in overfitting and degenerate representations for tail items. %

\textbf{Present Work:} To overcome item interaction sparsity, we \textit{extract, relate, and transfer} the knowledge learned by a neural base recommender over \textit{head} items, to learn robust and discriminative tail item representations.
We eliminate the distribution inconsistency between head and tail items with an \textit{episodic few-shot learning} setup~\cite{fsl_opt} to \textit{simulate} the distribution of tail items during model training by \textit{sub-sampling} interactions from data-rich head items.

In this work, we present a novel \textit{metric-based} few-shot learning framework~\protocf~for long-tail item recommendation.
First, we pretrain a \textit{base recommender} to extract preferences over head items and item-item relationships.
Then, we design a \textit{few-shot recommender} that learns a \textit{shared metric space} of users and items by extracting \textit{meta-knowledge} across a collection of meta-training tasks designed to \textit{mimic} long-tail item recommendation.
~\protocf~\textit{learns-to-compose} a representative \textit{prototype} for each item from a small set of user interactions and recommends relevant items by finding the nearest item prototypes to each user.
To transfer knowledge from the base recommender, we introduce a \textit{knowledge distillation} strategy to distill the discovered \textit{item-to-item} relationships into a \textit{compact} set of \textit{group embeddings}; we compose \textit{discriminative} item prototypes via learnable mixtures of group embeddings.
We summarize our key contributions below:
\begin{description}
\item \textbf{Few-shot Item Recommendation}:
To our knowledge, ours is the first to formulate long-tail item recommendation as \textit{learning-to-recommend} items with few interactions.
We eliminate the distribution mismatch between head and tail items via episodic few-shot learning. Our problem formulation applies to diverse recommendation scenarios without requiring any auxiliary side information.

\item \textbf{Discriminative Prototype Learning}: We learn to compose \textit{discriminative prototypes} for tail item \textit{only}  from their \textit{sparse} user interactions.
Unlike prior gradient based meta-learning for cold-start recommendation scenarios~\cite{melu, MetaHIN, ml2e}, we learn a metric space where the item prototypes directly cluster the interacted users in a \textit{metric space}, which further eliminates expensive online adaptation costs.

\item \textbf{Architecture-agnostic Knowledge Transfer}:
We enhance item prototypes by knowledge transfer from neural base recommenders.
In contrast to layer transfer or adaptation methods~\cite{cross-domain}, our \textit{knowledge distillation} strategy extracts a compact representation of the item relationships discovered by arbitrary base recommenders.
Significantly,~\protocf~is complementary to architectural advances in neural base recommenders and enables flexible adaptation to items in the long tail.

\end{description}

We instantiate~\protocf~by transferring meta-knowledge from three base recommenders.
Our experiments 
show that~\protocf~outperforms state-of-the-art baselines (by 5\%  Recall@50) in \textit{overall} recommendation, with notably significant \textit{few-shot} gains (of 60-80\% Recall@50) on tail items with less than 20 training interactions.

We organize the rest of the chapter as follows. In Section~\ref{sec:protocf_problem}, we formally define the problem of long-tail item recommendation.
In Section~\ref{sec:protocf_methods}, we first outline the components of a neural base recommender, and then introduce our proposed framework~\protocf~for few-shot item recommendation.
We then present our experimental results in Section~\ref{sec:protocf_experiments}, and finally conclude in Section~\ref{sec:protocf_conclusion}.

\section{Problem Definition}
\label{sec:protocf_problem}

We consider the implicit feedback setting (only clicks, no explicit ratings) with a user set $\gU = \{ u_1, \dots, u_{U} \}$, an item set $\gI = \{ i_1, \dots, i_{I} \}$, and a binary $U \times I$ user-item interaction matrix $\mX $.
Let $N_i$ denote the set of all users who have interacted with item $i \in \gI$.
Prior neural recommenders learn a scoring function $f (i \mid u, \mX), i \in \gI$ personalized to each user $u \in \gU$ for item ranking over the item set $\gI$.
From Table~\ref{tab:intro_table}, we find their performance for tail items (bottom 50\%) to be considerably inferior to head items (upper 50\%).
Thus, our focus is to develop personalized recommenders tailored to the \textit{long-tail} items (with sparse interactions), while ensuring reasonable performance over the entire item set $\gI$.

\begin{problem}[{Long-Tail Item Recommendation}] Given user-item interaction matrix $\mX$, learn a scoring function $f_T (i \mid u, \mX),  i \in \gI$ to generate a ranked list of items personalized to each user $u \in \gU$ that improves recommendation quality on the long-tail items without compromising overall model performance on the entire item set $\gI$.
\end{problem}

\begin{table}[t]
\centering
\begin{tabular}{@{}rl@{}}
        \toprule
        Symbol                & Description                                                               \\
        \midrule        
$p_B(i, j)$ & proximity induced by $\mR_B$ for items $i, j \in \gI$ \\
$p(\sT)$ & meta-training task distribution over items $\gI$ \\
$\gT$ &  meta-training task sampled from $p(\sT)$ \\
$\gI_{\gT, N}$ & set of $N$ items sampled from $\gI$ for task $\gT$ \\
$\gS_i$ & support set for item $i \in \gI_{\gT, N}$ in task $\gT$ \\
$\gQ_i$ & query set for item $i \in \gI_{\gT, N}$ in task $\gT$ \\
$G_U (\cdot \mid \theta)$  & user encoder in few-shot recommender $\mR_F$ \\
$\gZ_M$ & set of $M$ group embeddings $\{\rvz_m \in \sR^D \}_{m=1}^M$ \\
$\text{sim}_m$ & similarity metric for meta-recommender $\mR_F$ \\
$\rvp_i$ & initial item prototype for item $i \in \gI$ \\
$\rvg_i$ & group-enhanced item prototype for item $i \in \gI$ \\
$\rve_i$ & final gated item prototype item for item $i \in \gI$ \\     
        \bottomrule
    \end{tabular} \label{tab:protocf_notations}
        \caption{Notations}
    \end{table}

\section{Related Work}
\label{sec:protocf_related}
We briefly review a few related lines of prior work on neural collaborative filtering and long-tail item recommendations with interaction sparsity.

\textbf{Neural Collaborative Filtering}:
The core paradigm of latent-factor CF models is to parameterize users and items with latent representations learned from historical user-item interactions.
Recent neural recommenders enhance the representational capacity of CF models via non-linear latent representations~\cite{cdae, vae-cf}, neural interaction modeling~\cite{ncf, lrml}, and graph-based representation learning~\cite{ngcf}.
While these neural recommenders learn expressive models to significantly outperform conventional CF approaches, sparsity concerns owing to long-tail items remain a critical challenge.

\textbf{Sparsity-aware Recommendation:}
Clustering is one popular way to address interaction sparsity by modeling group-level purchasing behaviors; early methods generate recommendations for tail items at the granularity of item clusters~\cite{clustering}, \textit{e.g.}, cluster-based smoothing~\cite{smoothing}, user-item co-clustering~\cite{cccf} and joint clustering and collaborative filtering~\cite{dbrec}. However, clustering in the presence of distributional skew can lead to uninformative results~\cite{cobafi} and degrades the extent of personalization.

Recent techniques leverage \textit{item-item co-occurrence} statistics and \textit{distributional priors} to regularize latent-factor CF methods. One common approach is \textit{regularization} of recommendation models, \textit{e.g.}, factorization of an item co-occurrence matrix that shares latent factors with a CF model~\cite{cofactor, efm}. Variational auto-encoders (VAEs)~\cite{vae-cf, enhanced_vae, infvae} employ multivariate Gaussian priors as regularizers on the representational space to handle sparsity.
Another strategy to alleviate sparsity is \textit{data augmentation} for items (or users) in the tail via \textit{adversarial} regularization~\cite{cikm18adv}  or rating generation~\cite{rating_gan, arcf} techniques.
However, regularization techniques typically impose static hypotheses with restrictive assumptions, while adversarial learning is computationally expensive and does not scale to massive item inventories.

Cold-start recommendation is a related problem that targets \textit{new} users or \textit{new} items with \textit{no} historical interactions.
In such a scenario, models rely on auxiliary information, \textit{e.g.}, user profiles~\cite{melu}, item content~\cite{esam}, social connections~\cite{social_rec, groupim, grafrank}, and knowledge graphs~\cite{kgat}.
An effective strategy is randomized feature dropouts to enable generalization to missing inputs~\cite{dropoutnet, randomized}.
In contrast, we focus on \textit{few-shot} recommendations over long-tail items with very few interactions, \textit{without} any auxiliary information.

\textbf{Few-shot Learning:} this learning paradigm involves designing models capable of learning new tasks rapidly given 
a limited number of training examples. 
Here, the \textit{meta-learning} (learning to learn) paradigm has achieved considerable success in several domains including computer vision~\cite{matching_nets, protonet}, natural language~\cite{fewrel}, and data mining~\cite{melu, meta_rec}.

A few recent methods adapt meta-learning to cold-start recommendation~\cite{selection}.
Prior meta-learning work have examined gradient-based parameter~\cite{melu, ml2e, MetaHIN, mamo_kdd, dual_tales} and hyper-parameter initialization~\cite{s2meta}, and layer sharing with user-specific (or item) parameter adaptation~\cite{meta_rec, wei2020fast, adaptive}. However, these methods mostly address the cold-start (zero-shot) scenario with side information, thus inapplicable to our general setting of few-shot recommendation given limited interactions. %

Our work is conceptually related to \textit{metric-learning} approaches \cite{matching_nets, protonet, sreepada2020mitigating} for few-shot learning, which \textit{learn-to-learn} a metric space that generalizes to new tasks without any need for adaptation.
A few recent explorations address task heterogeneity~\cite{tadanet} in few-shot classification settings, by designing architectures for explicit knowledge transfer from data-rich classes~\cite{l2_tail, hierarchy, large_tail} to construct few-shot classifiers.
However, such techniques are suited to learning classifiers from a \textit{limited} number of classes, hence cannot scale to handle \textit{massive} item inventories in recommendation applications.
To our knowledge, ours is the \textit{first} investigation of metric-based few-shot learning for long-tail item recommendations.

\section{ProtoCF Framework}
\label{sec:protocf_methods}

Learning informative representations for tail items poses the key challenge of \textit{interaction sparsity}, \textit{i.e.}, each tail item only has a few observed interactions.
Without access to side information (\textit{e.g.}, item content or contextual attributes), we critically remark that prior neural recommenders~\cite{vae-cf, cdae} learn informative representations for head items with abundant training interactions.
Thus, to learn discriminative tail item representations, we propose a novel few-shot learning framework~\protocf~with two key steps:

First, we train a \textit{base recommender} $\mR_B$ to learn high-quality user representations that mainly capture preferences over head items and infer item-item relationships (Section~\ref{sec:protocf_base_recommender}).
Then, we present a \textit{few-shot recommender} $\mR_F$ that \textit{extracts and transfers} knowledge from the base recommender $\mR_B$ to \textit{learn-to-recommend} few-shot items (Sections~\ref{sec:protocf_few_shot}, \ref{sec:protocf_model_details}).

\begin{figure}[t]
    \centering
    \includegraphics[width=0.9\linewidth]{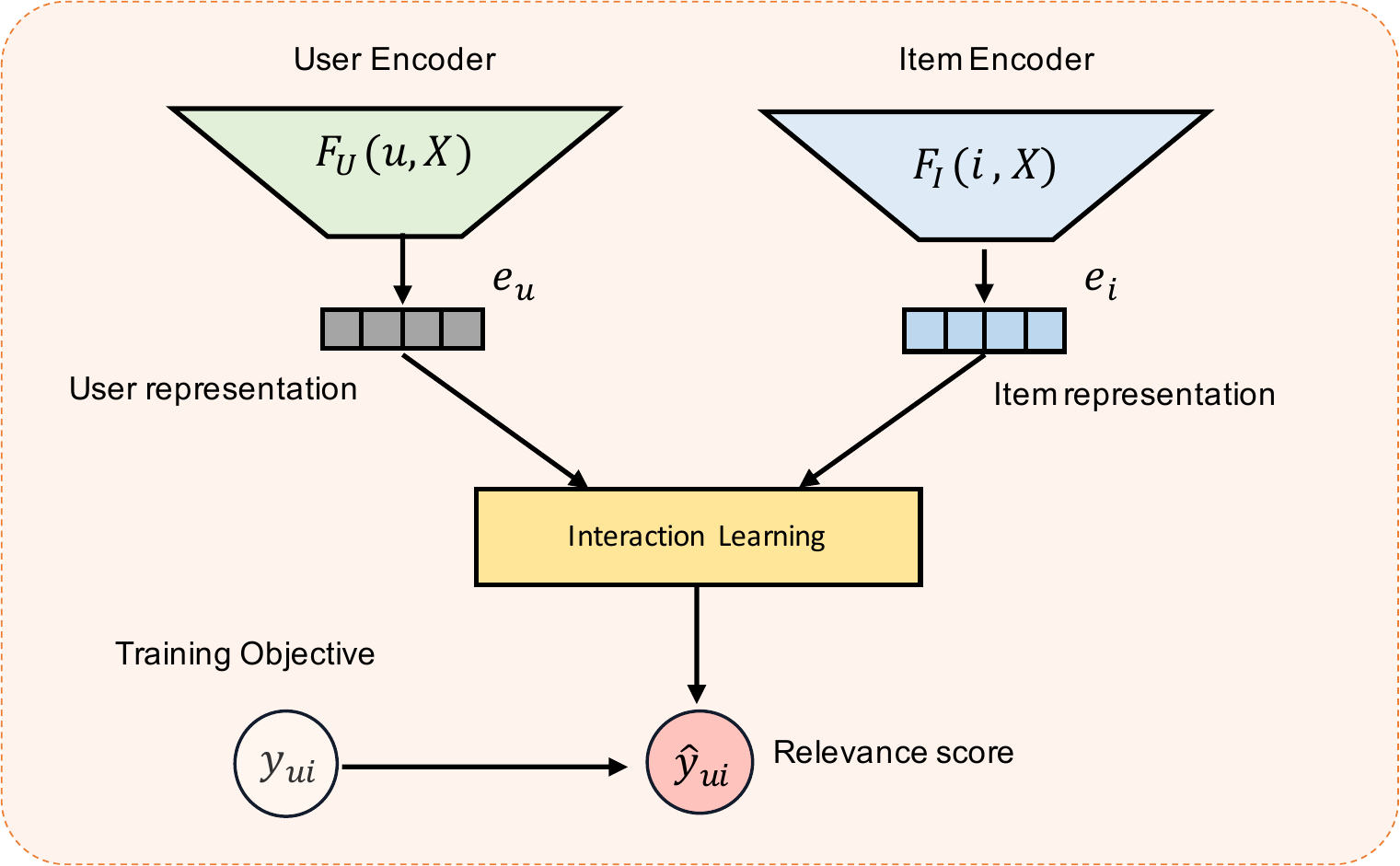}
    \caption{Outline of the different components of a neural base recommender based on collaborative filtering.}
    \label{fig:meta_setup}
\end{figure}

\subsection{Neural Base Recommender}
\label{sec:protocf_base_recommender}
In this section, we outline the architecture of a differentiable base recommender $\mR_B$ to learn a ranking $f_{\phi} (i  \mid u) , i \in \gI$ (with parameters $\phi$) over the interactions $\mX$. %
Neural CF models~\cite{vae-cf, ncf, cdae} typically learn latent representations for users and
items, followed by an interaction function and learning objective for model training.

\subsubsection{\textbf{User and Item Representations}}
Latent-factor CF methods adopt a variety of representation learning strategies, including matrix factorization~\cite{mf, bpr}, autoencoders~\cite{vae-cf, cdae} and graph neural networks~\cite{ngcf}. %
We define preference encoders $F_{U} (\cdot \mid \phi) $ and $F_{I} ( \cdot \mid \phi)$ to learn user $ \rvh_u \in  \sR^D $ and item $ \rvh_i \in  \sR^D $ embeddings for user $u \in \gU$ and item $i \in \gI$, which can be described by:

\begin{equation}
 \rvh_u  = F_{U} (u, \mX \mid \phi) \hspace{15pt} \rvh_i  = F_{I} (i, \mX \mid \phi) 
 \label{eqn:protocf_base_encoder}
\end{equation}

\subsubsection{\textbf{Training Objective}}
We define a user-item interaction function $F_{\textsc{int}} ( \cdot \mid \phi)$ to compute a prediction score $\hat{y}_b (u, i)$ that indicates the relevance of item $i$ to user $u$. The training objective $L_B$ of the base recommender $\mR_B$ is given by:

\begin{equation}
\hat{y}_{b} (u, i) = F_{\textsc{int}}(\rvh_u, \rvh_i \mid \phi) \hspace{15pt} L_B = l(\hat{y}_b (u, i), y _{ui}))
\label{eqn:protocf_base_interaction}
\end{equation}
where $y_{ui} = 1$ for observed $(u, i)$ pairs and $l(\cdot)$ is a pairwise~\cite{bpr} or pointwise~\cite{ncf} loss function. over the user-item interactions.
The interaction function $F_{\textsc{int}}$ measures user-item relevance and is typically modeled using an inner product~\cite{bpr, cdae, vae-cf, ngcf}.

We train the base recommender $\mR_B$ using cosine similarity since normalized representations generalize better to few-shot settings~\cite{gfsl} (compared to unnormalized inner products), and facilitate unified recommendation of head and tail items during model inference. We denote the \textit{item-item proximity} $\text{sim}_b (\cdot)$ in the latent space of $\mR_B$ by:

\begin{equation}
p_{B} (i, j) \propto  \text{sim}_b (\rvh_{i}, \rvh_{j}) = \text{cos} (\rvh_{i}, \rvh_{j})  \hspace{15pt} i, j \in \gI
\label{eqn:protocf_base_item_proximity}
\end{equation}

We pre-train $\mR_B$ to extract knowledge of user preferences over head items via encoder $F_{U} (\cdot)$ and item-item proximities among different items via $p_B(i, j)$.
Below, we present our proposed framework~\protocf~that transfers knowledge from the \textit{data-rich} head to the \textit{data-poor} tail to enable robust few-shot recommendations.

\subsection{Few-shot Item Recommendation}
\label{sec:protocf_few_shot}
In this section, we formulate long-tail item recommendation as \textit{few-shot item representation learning} for items given a small \textit{support set} of upto $K$ interacted users for each tail item (typically $K \approx 5 \text{ to } 20$ interactions).

\subsubsection{\textbf{Few-shot Task Formulation}}
Our framework is grounded on \textit{episodic} learning~\cite{matching_nets} with a collection of \textit{meta-training} tasks or episodes.
One approach~\cite{protonet, gfsl} is to construct meta-training tasks from the interactions of data-rich head items; 
however, excluding the diverse collection of tail items may impede generalization.
Thus, our meta-training tasks operate on the entire item set $\gI_T$.

\begin{figure}[t]
    \centering
    \includegraphics[width=0.9\linewidth]{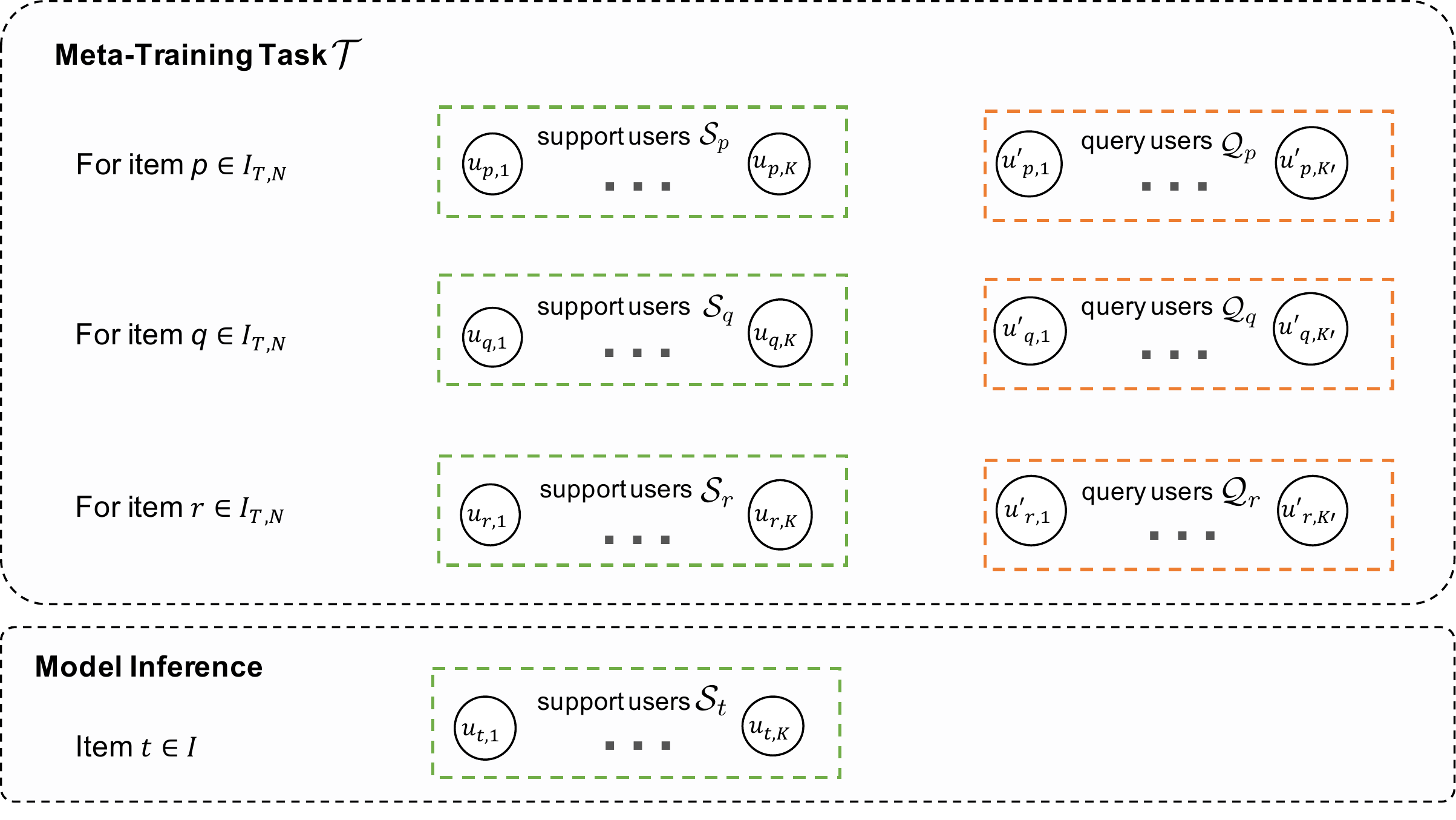}
    \label{fig:meta_setup}
    \vspace{-5pt} 
\caption{Episodic few-shot learning with meta-training task $\gT$ and item embedding inference at meta-testing.}
\end{figure}

Each meta-training task $\gT$ is a personalized ranking problem over a subset of $N$ items $\gI_{\gT, N}$ randomly sampled from $\gI$. %
We \textit{simulate the interaction distribution of few-shot items during meta-training} by sampling $K$ training interactions (from $N_{i}$) for each item $i \in \gI_{\gT, N}$ in task $\gT$.
During inference, we generate \textit{few-shot} item recommendations from samples of upto $K$ training interactions per item.
By ensuring consistency between meta-training and inference, we bridge the \textit{distribution mismatch} between head and tail items. %
The meta-knowledge extracted across diverse meta-training tasks benefits tail items with sparse interactions.

The episodic few-shot training process operates over a collection of meta-training tasks $\{ \gT_1, \gT_2, \dots \}$ sampled from a task distribution $p (\sT)$ over the item set $\gI$. %
Specifically, a \textit{$K$-shot, $N$-item} meta-training task $\gT$ sampled from $p(\sT)$ consists of support $\gS$ and query $\gQ$ user sets over $N$ items $\gI_{\gT, N} \subset \gI$ (analogous to the usual sense of training and testing sets respectively). Each meta-training task $\gT \sim p(\sT)$ is defined by:
\begin{align}
 \gT &= \{\gI_{\gT, N} , \gS, \gQ \}   \qquad \hspace{8pt} \gI_{\gT, N} \subset \gI  \\
 \gS &= \{ \gS_{i} : i \in \gI_{\gT, N} \}   \qquad  \gS_{i} = \{ u_{i, 1}, \dots, u_{i, K} \}  \qquad  & u_{i, k} \in N_i  \quad \nonumber   \\ \nonumber
 \gQ &= \{ \gQ_{i} : i \in \gI_{\gT, N}  \}  \qquad   \gQ_{i} = \{ u^{\prime}_{i, 1}, \dots, u^{\prime}_{i, K^{\prime}}  \}    &  u^{\prime}_{i, k^{\prime}} \in N_i \quad 
\end{align}

where the support set $\gS_i$ contains $K$ interacted users sampled for each item $i \in \gI_{\gT, N}$ and the corresponding query set $\gQ_i$ includes $K^{\prime}$ interacted users sampled for each of the $N$ items.

We learn a \textit{few-shot recommender} $\mR_F$ that takes as input the support users $\gS$ to \textit{learn-to-compose} representations for items $i \in \gI_{\gT, N}$ in task $\gT$.
The few-shot recommender $\mR_F$ is trained by optimizing a learning objective designed to match the item recommendations generated by $\mR_F$ for its query users $\gQ$ with their corresponding ground-truth interactions over the item $\gI_{\gT, N}$ in each meta-training task $\gT$.

\subsubsection{\textbf{Initial Item Prototype}}
We learn a shared \textit{metric space} of users and items to synthesize a representation for each item $i \in \gI_{\gT, N}$ based on its support set $\gS_i$ (with $K$ interactions) in task $\gT$.
The metric space (with similarity $\text{sim}_m (\cdot)$) compactly clusters the interacted users $\gS_i$ of each item $i \in \gI_{\gT, N}$ around a \textit{prototype} representation $\rvp_i \in \sR^D$~\cite{protonet}.

We first define a user preference encoder $G_{U} (\cdot \mid \theta) $ that maps each user $u \in \gU$ into a latent embedding space.
To directly transfer knowledge from the base recommender $\mR_B$, the few-shot user encoder $G_{U} (\cdot \mid \theta)$ has model parameters initialized from its pre-trained encoder $F_U (\cdot \mid \phi)$ in the base recommender $\mR_B$ (Equation~\ref{eqn:protocf_base_encoder}), but is parameterized with learnable parameters $\theta$.
The prototype $\rvp_i$ for item $i \in \gI_{\gT, N}$ is computed as the mean vector of the embedded support user set $\gS_i$, defined by:
\begin{equation}
\rvp_i = \frac{1}{\gS_i} \sum\limits_{u_{i, k} \in \gS_i} G_U (u_{i, k}, \mX_H  \mid \theta) \qquad i \in \gI_{\gT, N}
\label{eqn:protocf_initial_prototype}
\end{equation}
Equation~\ref{eqn:protocf_initial_prototype} encourages the prototype $\rvp_i$ to learn a representative vector summarizing its cluster of interacted users.
However, we face two critical challenges in handling tail items:
first, due to sparse support sets, the prototypes are often \textit{noisy} and sensitive to \textit{outliers};
second, due to substantial \textit{heterogeneity} in the tail, simplistic averaging may lack the \textit{resolution} to discriminate across diverse tail items.
Thus, the few-shot recommender $\mR_F$ needs a strong \textit{inductive bias} during prototype learning to avoid overfitting, and yet have sufficient expressivity to learn \textit{discriminative} prototypes.

\subsubsection{\textbf{Head-Tail Meta Knowledge Transfer}}
We now exploit the \textit{item-to-item} relationship knowledge acquired by the neural base recommender $\mR_B$ (Section~\ref{sec:protocf_base_recommender}) as an inductive bias to enhance the item prototype $\rvp_i$ (Equation~\ref{eqn:protocf_initial_prototype}).
Given an item $i \in \gI_{\gT, N}$ with limited support users, extracting knowledge from the most related items to item $i$ (based on the item-to-item proximities) can provide valuable guidance to synthesize robust item prototypes.

However, direct knowledge transfer from the base recommender $\mR_B$ to the few-shot recommender $\mR_F$ is technically challenging: dynamically identifying related items for each few-shot item during prototype construction is not scalable due to the arbitrary complexity of $\mR_B$ and the high cardinality of item sets in massive inventories.
Thus, we extract a \textit{compact} representation of the \textit{item-item proximity} knowledge (Equation~\ref{eqn:protocf_base_item_proximity}) discovered by $\mR_B$ and transfer this knowledge to enhance item prototypes.

\begin{figure}[t]
    \centering
    \includegraphics[width=0.9\linewidth]{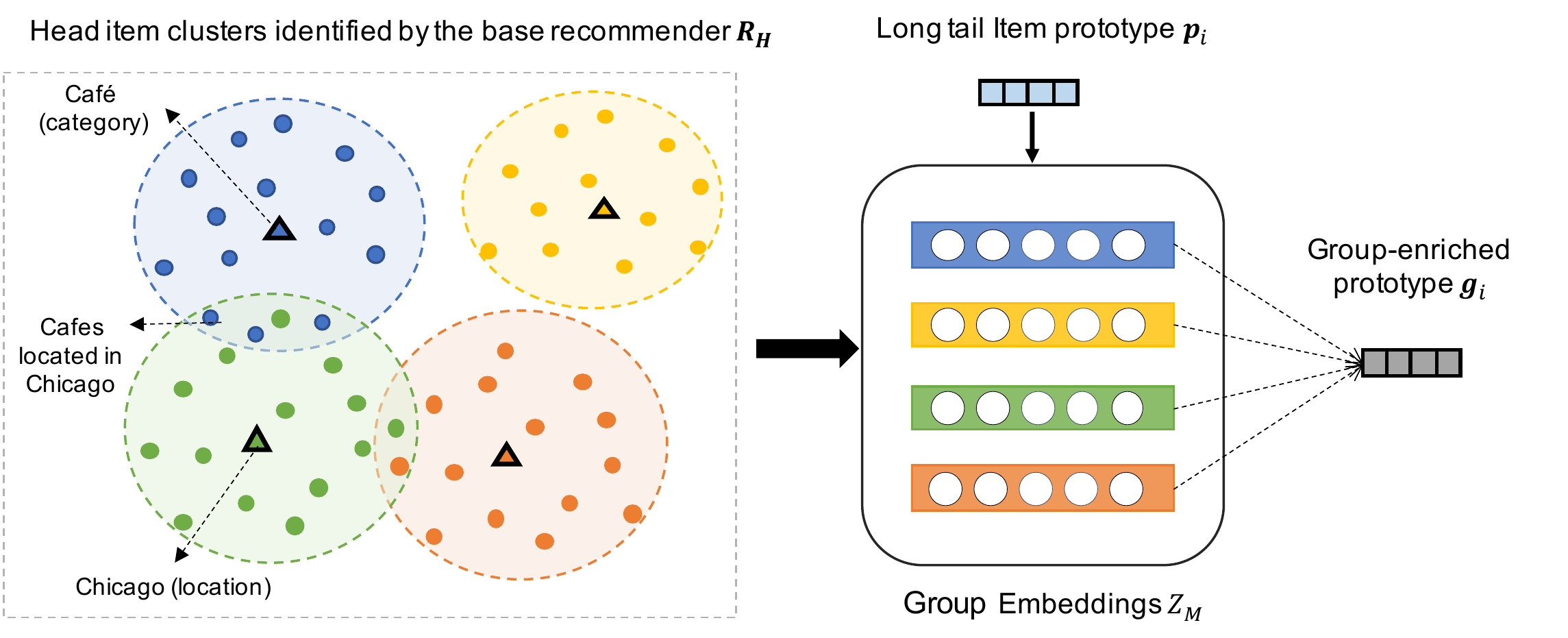}
    \caption{Group embeddings can be viewed as centroids of overlapping item clusters, \textit{e.g.}, Cafe (category) or Chicago (location); items (Cafes located in Chicago) may often belong to overlapping item clusters. }
    \label{fig:protocf_group_vectors}
\end{figure}

\textbf{Group-Enhanced Item Prototype Learning.}
We learn a set of $M$ group embeddings $\gZ_M$ as \textit{basis vectors} modeling \textit{item-item proximity} in the latent space of the base recommender $\mR_B$, \textit{i.e.}, if $p_B (i, j)$ is high for a pair of items $i, j \in \gI$, then items $i$ and $j$ are likely to exhibit similar characteristics. 
Formally, the set of group embeddings $\gZ_M$ are defined by the following equation below:
\begin{equation}
\gZ_M = \{z_m \in \sR^{D}:  1 \leq m \leq M \} \qquad M \ll |\gI|
\label{eqn:protocf_group_vectors}
\end{equation}

As depicted in Figure~\ref{fig:protocf_group_vectors}, we intuitively visualize the group embeddings as \textit{discriminative centroids} of \textit{overlapping} clusters of items identified by $\mR_B$, and conversely view items as \textit{mixtures} over the group embeddings, \textit{e.g.}, each centroid may represent a contextual factor such as a restaurant type (Cafe) or location (Chicago), while restaurants are mixture of multiple centroids (Cafe located in Chicago) and belong to overlapping item clusters.

To enhance the prototype $\rvp_i$ of item  $i \in \gI_{\gT, N}$, we synthesize a \textit{group-enhanced prototype} $\rvg_i \in \sR^D$ as a mixture over the $M$ group embeddings.
The mixture coefficients are estimated by a learnable attention mechanism~\cite{attention} to measure \textit{compatibility} $\alpha_{im}$ between the prototype $\rvp_i$ and each centroid $z_m  \in \gZ_M $.
We parameterize the attention function with a lightweight network comprised of an auxiliary set $\gK_M = \{ \rvk_m \in \sR^{D}:  1 \leq m \leq M \} $ of trainable keys to index the group embeddings.
We implement the attention function with an inner product followed by softmax normalization, as:
\begin{equation}
\rvg_i = \sum\limits_{m=1}^M \alpha_{im} \rvz_{m} \qquad
\alpha_{im} =\frac{\exp \big(\mW_q \rvp_i \cdot \rvk_m \big) }{ \sum_{m^{\prime} = 1}^M  \exp \big(\mW_q \rvp_i \cdot \rvk_{m^{\prime}} \big)}
\label{eqn:protocf_group_enhanced_prototype}
\end{equation}
where $\mW_q \in \sR^{D \times D} $ is projects the prototype $\rvp_i$ into a query to index the centroids.
The group-enhanced item prototype $\rvg_i$ relates the different centroids via attention,  transferring knowledge to few-shot items with sparse support sets.

\textbf{Task-level Stochastic Knowledge Distillation.}
We present a \textit{knowledge distillation} strategy~\cite{distillation} to learn compact \textit{group embeddings} $\gZ_M$ that capture \textit{item-item relationships} in $\mR_B$.
We transfer knowledge from a high-capacity \textit{teacher} model (base recommender $\mR_B$) to a compact \textit{student} model (group embeddings $\gZ_M$) by encouraging the student to emulate predictions of the teacher; $\gZ_M$ is trained to emulate the item proximity distribution in $\mR_B$.

Aligning pairwise item proximities over all items in $\gI$ is not scalable; thus, we operate at the granularity of each meta-training task $\gT$.
For each item $i \in \gI_{\gT, N}$, we compute a soft probability distribution $p_B( j \mid i, \mR_B)$ for the teacher model $\mR_B$ (based on equation~\ref{eqn:protocf_base_item_proximity}) over other items $j \in \gI_{\gT, N}$ in the task $\gT$, described by:
\begin{equation}
 p_B( j \mid i, \mR_B) =  \frac{ \exp \big( p_B (i, j) / T \big)} {\sum_{k \in \gI_{\gT, N}}   \exp \big( p_B (i, k) / T  \big)} \quad i, j \in \gI_{\gT, N}
\label{eqn:protocf_teacher_softmax}
\end{equation}

where $T > 0$ is a temperature scaling hyper-parameter to regulate the rate of knowledge transfer.
We analogously define the item similarity distribution $p_F( j \mid i, \mZ_M)$ for the student model $\gZ_M$ based on the metric space proximity $\text{sim}_m (\cdot)$ of group-enhanced prototypes $\rvg_i$ and $\rvg_j$ for items $ i, j \in \gI_{\gT, N}$, defined by:
\begin{equation}
p_F( j \mid i, \mZ_M) =  \frac{ \exp \big( \text{sim}_m (\rvg_i, \rvg_j)\big)} {\sum_{k \in \gI_{\gT, N}}   \exp \big(  \text{sim}_m (\rvg_i, \rvg_k)  \big)} \;  i, j \in \gI_{\gT, N}
\label{eqn:protocf_student_softmax}
\end{equation}

We align the two distributions by minimizing cross-entropy between their task-level similarities~\cite{distillation}.
Since each item is typically related to very few items within the task $\gT$, our \textit{stochastic knowledge distillation} loss $L_G$ minimizes distribution divergence over the top-$n$ ($n \approx 10$) related items (out of $N$) identified by the teacher $\mR_B$, by:
\begin{equation}
L_G = - \frac{1}{nN}\sum\limits_{i \in \gI_{\gT, N} }  \sum\limits_{j \in \Top_{B,n} (i) }  p_B (j \mid i, \mR_B) \log p_F (j \mid i, \mZ_M) 
\label{eqn:protocf_group_loss}
\end{equation}
\normalsize
where ${\Top}_{B,n} (i) = \text{Top}_{n} \big( p_B( \cdot \mid i, \mR_B) \big)$ denotes the top-$n$ most related items to item $i$ within $\gI_{\gT, N}$ based on the teacher $\mR_B$.
Distinct from prior distillation approaches~\cite{distillation, relational_distillation} that match student and teacher predictions over a fixed set of classes, $L_G$ is stochastic and thereby more efficient since it matches the top-$n$ item proximity distributions over different sets of sampled items in each task.
The distillation loss $L_G$ transfers knowledge from $\mR_B$ to the group embeddings $\gZ_M$ and is trained jointly with the rest of the framework.

\subsubsection{\textbf{Item Prototype Fusion via Neural Gating}}
The initial prototype $\rvp_i$ for item $i \in \gI_{\gT, N}$ direct encodes its support users $\gS_i$, while the group-enhanced prototype $\rvg_i$ captures the knowledge transferred  from related items (via base recommender $\mR_B$).
We design a \textit{gating mechanism}~\cite{gating} that adaptively selects salient feature dimensions from $\rvp_i$ and $\rvg_i$ to infer the final \textit{gated item prototype} $\rve_i \in \sR^D$.
Specifically, we introduce a neural gating layer to merge $\rvp_i$ and $\rvg_i$ by learning a \text{non-linear gate} to flexibly modulate information flow via dimension re-weighting, defined by:
\begin{align}
\bf{gate} &= \sigma \big(\mW_{g_1} \rvp_i + \mW_{g_2} \rvg_i + \rvb_g \big) \quad i \in \gI_{\gT, N} \nonumber \\
\rve_i  &= \bf{gate}  \odot \vp_i   +  (1 - \bf{gate}) \odot \vg_i
\label{eqn:protocf_prototype_gating}
\end{align}

where $\mW_{g_1} \in \sR^{D \times D}$,  $\mW_{g_2} \in \sR^{D \times D}$, and $\rvb_g \in \sR^{D}$  are learnable weight parameters in the neural gating layer, $\odot$ denotes the element-wise (or hadammard) product operation, and $\sigma$ is the sigmoid non-linearity function.

\begin{figure*}[t]
    \centering
    \includegraphics[width=0.9\linewidth]{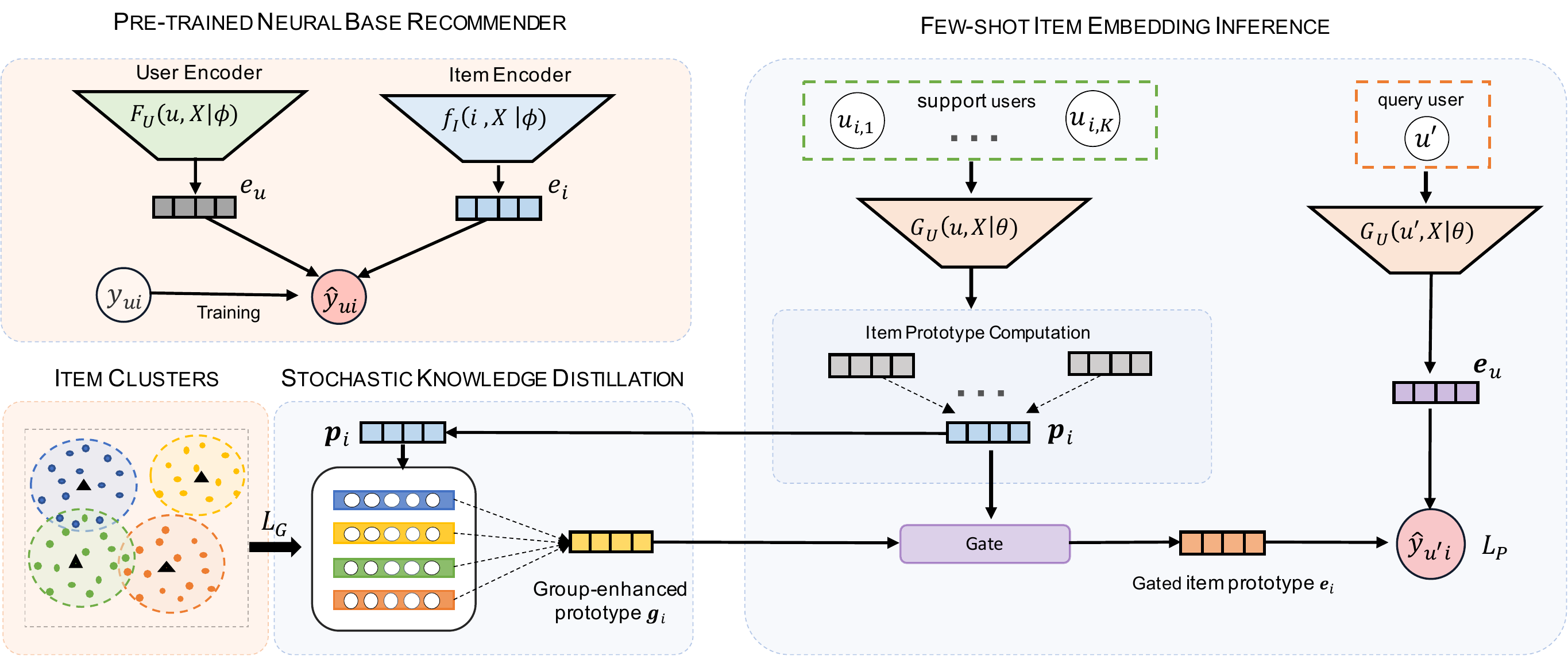}
    \caption{Architecture diagram of~\protocf~depicting the different model components: pre-trained neural base recommender $\mR_B$ (top left), group embedding learning via stochastic knowledge distillation $L_G$ (bottom left), initial item prototype construction via support set averaging followed by group-enrichment and adaptive gating to construct gated item prototype $\rve_i$ (right).}
    \label{fig:protocf_arch}
\end{figure*}

\subsubsection{\textbf{Few-shot Recommender Training}}
We induce few-shot item rankings for the query users $\gQ$ using the gated item prototypes $\{ \rve_{i} : i \in \gI_{\gT, N} \}$ in meta-training task $\gT$.
We generate item recommendations for each query user $u^{\prime} \in \gQ$ (over the task items $\gI_{\gT, N}$) by measuring the similarity $\text{sim}_m (\cdot)$ with each item prototype $\rve_i$ %
Each meta-training task $\gT$ minimizes a negative log-likelihood $L_P$ between the few-shot item recommendations for query users $\gQ$ and their ground-truth interactions in $\gT$, which is defined by:

\begin{equation}
 L_P = -\frac{1}{KN} \sum\limits_{i \in \gI_{\gT, N}} \sum\limits_{u^{\prime}_{i, k^{\prime}} \in Q_i} \log p_F (i \mid u^{\prime}_{i, k^{\prime}}, \theta)
\label{eqn:protocf_few_shot_loss} 
\end{equation}
where $p_F (i \mid u^{\prime}_{i, k^{\prime}}, \theta)$ is computed based on cosine similarity and the choice of likelihood function for few-shot training.

The overall loss $L$ is composed of two terms, the few-shot recommendation loss $L_P$ (Equation~\ref{eqn:protocf_few_shot_loss}) and the knowledge distillation loss $L_G$ for group embedding learning (Equation~\ref{eqn:protocf_group_loss}), given by:
\begin{equation}
L = L_P + \lambda L_G
\label{eqn:protocf_overall_loss}
\end{equation}
where $\lambda$ is a tunable hyper-parameter that balances the two loss terms. Algorithm~\ref{alg:protocf_opt} summarizes the training procedure of our entire framework~\protocf.

\subsubsection{\textbf{Model Inference}}
We infer the \textit{gated prototype} $\rve_i$ for each item $i \in \gI$ by sub-sampling $K$ interactions from its historical interactions $N_i$ as the the support set (Equation~\ref{eqn:protocf_prototype_gating}).
We generate personalized item recommendations for each user $u \in \gU$ by:
\begin{equation}
  \hat{y}_f (u,i) = \text{sim}_m (\rve_u, \rve_i )  \quad i \in \gI \qquad \rve_u = G_U (u, \mX \mid \theta )
  \label{eqn:protocf_meta_inference}
\end{equation}

The few-shot recommender $\mR_F$ is designed for recommendations over tail items (sparse interactions), while the base recommender $\mR_B$ is effective for head items (abundant interactions).
In~\protocf, we compute rankings over the entire item set $\gI$ by ensembling predictions from $\mR_B$ and $\mR_F$. One simple interpolation approach is given by:

\begin{equation}
\hat{y}(u,i) =  (1 - \eta) \cdot \hat{y}_b (u,i) + \eta  \cdot \hat{y}_f (u,i)
  \label{eqn:protocf_inference}
\end{equation}
where $0 <\eta < 1$ balances $\mR_F$ and $\mR_B$. We empirically show effective overall item recommendations with $\eta=0.5$. %

\subsection{Model Details}
\label{sec:protocf_model_details}
We now discuss different choices of likelihood functions for training the few-shot recommender $\mR_F$ and architectural details of the base recommender $\mR_B$.

\subsubsection{\textbf{Few-shot Likelihood}}
We examine two likelihood functions in neural CF models: \textit{multinomial} and \textit{logistic}.

\begin{itemize}
\item \textbf{Multinomial log-likelihood:}
The  scores $\text{sim}_m (\rve_{u^{\prime}}, \rve_i)$ for each query user $u^{\prime} \in \gQ_i$, over the $N$ gated item prototypes in meta-training task $\gT$, are normalized to produce probabilities $p_F (i \mid u^{\prime}, \theta)$ over the $N$ items $\gI_{\gT, N}$, defined by:
\begin{equation}
p_F (i \mid u^{\prime}, \theta) = \frac{\exp \big(\text{sim}_m (\rve_{u^{\prime}}, \rve_i) \big) }{ \sum_{j \in \gI_{\gT, N}} \exp \big( \text{sim}_m (\rve_{u^{\prime}}, \rve_j)  \big) }  \qquad u^{\prime} \in  \gQ _i
\label{eqn:protocf_multinomial}
\end{equation}
The resulting loss $L_G$ (with Equation~\ref{eqn:protocf_few_shot_loss}) is also known as the \textit{cross-entropy} loss, defined over the $N$ items of meta-training task $\gT$.

\item \textbf{Logistic log-likelihood:}
The relevance $\hat{y}_{u^{\prime}i} = \text{sim}_m (\rve_{u^{\prime}}, \rve_i)$ for item $i \in \gI_{\gT, N}$ to query user $u^{\prime} \in \gQ_i$, is transformed into a probability using the sigmoid function $\sigma$.
We use a \textit{confidence} weight $\beta >0$ to re-weight the likelihood of observed 1’s which are far fewer than the unobserved 0's in implicit feedback. The logistic log-likelihood is given by:
\begin{equation}
\log  p_F(i \mid u^{\prime}, \theta) = \beta \log  \sigma (\hat{y}_{u^{\prime}i})   + \sum\limits_{ j \in \gI_{\gT, N}, u^{\prime} \notin N_j}  \log (1 -  \sigma (\hat{y}_{u^{\prime}j} ) )
\label{eqn:protocf_logistic}
\end{equation}
\end{itemize}

\begin{algorithm}[t]
\small
\caption{\protocf:~Prototypical Collaborative Filtering}
\begin{algorithmic}[1]
\Require User-item interactions $\mX$,  meta-training task distribution $p(\sT)$.
\Ensure  Function $f_T (i \mid u), i \in \gI_T$ for few-shot item recommendations.
\State Pre-train the neural base recommender $\mR_B$ on the training interactions $\mX$ to learn $f_{H, \phi} (i \mid u)$ (eqn~\ref{eqn:protocf_base_interaction}) and freeze parameters $\phi$.
\State Initialize user preference encoder $G_U (\cdot \mid \theta)$ of few-shot recommender $\mR_F$ with parameters $F_U (\cdot \mid \phi)$ from the base recommender $\mR_B$.
\LeftComment  \textbf{\textit{Meta-training: learn-to-recommend few-shot items.}}

\While{\textit{not converged}}
\State Sample a meta-training task $\gT = \{ I_{\gT, N}, \gS, \gQ \} \sim p(\sT)$.
\parState{Calculate the initial $\rvp_i$ and group-enhanced $\rvg_i$ prototypes for items $i \in \gI_{\gT, N}$ (eqn~\ref{eqn:protocf_initial_prototype} and eqn~\ref{eqn:protocf_group_enhanced_prototype}).}
\parState {Compute the knowledge distillation loss $L_G$ (eqn~\ref{eqn:protocf_group_loss}).}
\parState {Compute gated prototype $\rve_i$ for items $i \in \gI_{\gT, N}$ (eqn~\ref{eqn:protocf_prototype_gating}).}
\parState {Estimate few-shot recommendation loss $L_P$ (eqn~\ref{eqn:protocf_few_shot_loss}) with \textit{multinomial} (eqn~\ref{eqn:protocf_multinomial}) or \textit{logistic} (eqn~\ref{eqn:protocf_logistic}) log-likelihoods.}
\parState{Minimize overall loss $L$ (eqn~\ref{eqn:protocf_overall_loss}) using mini-batch gradient descent.}
\EndWhile

\LeftComment  \textbf{\textit{Model Inference: few-shot and overall item recommendations}.}
\State Generate item recommendations over $\gI$ for user $u \in \gU$ (eqn~\ref{eqn:protocf_inference})

\end{algorithmic}
\label{alg:protocf_opt}
\end{algorithm}

The cosine similarity scores are also scaled to match the non-saturating regimes of the softmax and sigmoid functions in the multinomial and logistic log-likelihoods respectively.

\subsubsection{\textbf{Neural Base Recommender Architecture}}
\label{sec:protocf_base_models}
We consider three neural collaborative filtering methods, matrix factorization (MF)~\cite{bpr}, denoising (CDAE)~\cite{cdae} and variational (VAE-CF)~\cite{vae-cf} autoencoders, as base recommenders $\mR_B$ (Equation~\ref{eqn:protocf_base_encoder}) for our framework~\protocf.  We briefly describe their architectures:

\begin{itemize}
\item \textbf{Matrix Factorization (MF)~\cite{bpr}}: The user $F_{U} $ and item $F_{I}$ encoders are learnable latent embeddings for each user and item respectively, which are trained using a logistic log-likelihood training objective.

\item \textbf{Variational AutoEncoder (VAE-CF)~\cite{vae-cf}}: The user encoder $F_{U} $ is a two-layer Multi-Layer Perceptron transforming the binary user preference vector $\rvx_{u} \in \sR^{I}$ into a $D$-dimensional user embedding $\vh_u \in \sR^D$, defined by:
\begin{equation}
\vh_u = F_{U} (\vx_{u} \mid \phi) = \sigma (\mW_2^T (  \sigma  (\mW_1^T \vx_{u} + b_1 ) + b_2  )
\label{eqn:protocf_vae-cf}
\end{equation}
with learnable weight matrices $\mW_1 \in \sR^{I_H \times D}$ and $\mW_2 \in \sR^{D \times D}$, biases $b_1, b_2 \in \sR^D$, and $tanh(\cdot)$ activations for non-linearity $\sigma$.
The item encoder $F_{I}$ is a latent embedding. VAE-CF uses a multinomial likelihood with KL-divergence regularization.

\item \textbf{Denoising AutoEncoders (CDAE)~\cite{cdae}}:~The user encoder $F_{U}$ operates on partially corrupted inputs $\rvx_{u} $ and adds an auxiliary per-user embedding to the encoder output from Equation~\ref{eqn:protocf_vae-cf}. 
We train CDAE using a multinomial log-likelihood, albeit without the KL-divergence prior regularization.
\end{itemize}

\section{Experiments}
\label{sec:protocf_experiments}

We present experiments on four real-world datasets to evaluate our framework~\protocf.  %
We introduce datasets, baselines and experimental setup in Sections~
\ref{sec:protocf_datasets}, \ref{sec:protocf_baselines} and~\ref{sec:protocf_setup}.
We propose four research questions to guide our experiments:

\begin{enumerate}[label=(\subscript{\textbf{RQ}}{{\textbf{\arabic*}}}),leftmargin=*]

\item Does~\protocf~beat state-of-the-art NCF and sparsity-aware methods on \textit{overall} recommendation performance?
\item What is the impact of item \textit{interaction sparsity} on the \textit{few-shot} recommendation performance of~\protocf?
\item How do the different \textit{architectural} choices impact the few-shot and overall performance of~\protocf? 
\item How do the \textit{hyper-parameters} (distillation loss balance factor $\lambda$ and meta-training task size $N$) affect~\protocf? 
\end{enumerate}

Finally, we discuss the limitations of our~\protocf~model and explore future directions in Section~\ref{sec:protocf_discussion}.

\subsection{{Datasets} }
\label{sec:protocf_datasets}
We conducted experiments on four publicly available benchmark datasets, including two online product review platforms (\textit{Epinions}, \textit{Yelp}) and two check-in based social networks (\textit{Weeplaces}, \textit{Gowalla}).
\begin{itemize}
\item \textbf{Epinions\footnote{https://www.cse.msu.edu/ tangjili/datasetcode/truststudy.htm}:}  product ratings from an e-commerce platform;
we retain interactions with ratings higher than two.
\item \textbf{Yelp\footnote{https://www.yelp.com/dataset}:} user ratings on local businesses located in the state of Arizona, obtained from Yelp dataset challenge round 13.
\item \textbf{Weeplaces\footnote{\url{https://www.yongliu.org/datasets/}}:}  we extract business check-ins from Weeplaces of different categories, including Nightlife, Outdoors, Entertainment, Travel and Food, across all cities in the United States.
\item \textbf{Gowalla~\cite{gowalla}:}  restaurant check-ins in Gowalla by users across different cities in the United States.
\end{itemize}

We pre-process the datasets to retain users and items with at least ten interactions using the 10-core setting~\cite{10-core} (Table~\ref{tab:protocf_stats}).

 \begin{table}[t]
 \centering
 \small
 \begin{tabular}{@{}p{0.32\linewidth}K{0.13\linewidth}K{0.13\linewidth}K{0.13\linewidth}K{0.13\linewidth}@{}}
 \toprule
 \textbf{Dataset} & \textbf{Epinions} & \textbf{Yelp} & \textbf{Weeplaces} & \textbf{Gowalla}\\
 \midrule
  \textbf{\# Items} & 9,035 & 10,451& 11,679 & 27,920 \\
 \textbf{\# Users} & 9,729 & 13,926 & 6,167 & 17,848 \\
 \textbf{\# Interactions} & 260,263 & 465,386 & 427,236 & 907,351 \\
 \textbf{\# Interactions per item} &28.81 & 44.53 & 36.58 & 32.50 \\
 \bottomrule
 \end{tabular}
 \caption{Dataset statistics}
\label{tab:protocf_stats}
\end{table}

\newcommand*{\newfactor}{0.056}
\subsection{{Baselines}}
\label{sec:protocf_baselines}
We present comparisons against prior work that broadly fall into two categories:
standard \textit{neural collaborative filtering} models, and \textit{sparsity-aware long-tail item recommendation} models,  including regularization and meta-learning techniques.
\begin{itemize}
\item \textbf{Neural Base Recommenders}: neural CF methods with matrix factorization (BPR) \cite{bpr}, and autoencoder models VAE-CF~\cite{vae-cf} and CDAE~\cite{cdae} (described in Section~\ref{sec:protocf_base_models}).
\item \textbf{NCF ~\cite{ncf}}: neural CF model with non-linear interactions (via neural layers) between the user and item embeddings.
\item \textbf{NGCF ~\cite{ngcf}}: state-of-the-art graph-based NCF with embedding propagation layers on the user-item interaction graph.
\item \textbf{Cofactor~\cite{cofactor}}: regularized MF to capture inter-item associations by jointly decomposing the user-item interaction matrix and the item-item co-occurrence matrix, with shared latent item factors.
\item \textbf{EFM~\cite{efm}}: embedding factorization model that uses item-item co-occurrence with bayesian personalized ranking.
\item \textbf{DropoutNet~\cite{dropoutnet}:} randomly dropout latent  CF embedding for regularization in content-based CF for cold-start (auxiliary content). We adapt it to long-tail items by replacing content embeddings with prototypes (over support set).
\item \textbf{MetaRec-LWA}~\cite{meta_rec} meta-learning method to construct recommendation models for cold-start users with user-specific linear transforms; we adapt it to construct few-shot item embeddings.
\item \textbf{MetaRec-NLBA}~\cite{meta_rec}: meta-learning recommendation model that learns user-specific biases parameterized by non-linear layers, to replace the user-specific weights in MetaRec-LWA.
\end{itemize}

Note that we omit experimental baseline comparisons with gradient-based meta-learning recommenders~\cite{melu, MetaHIN}, since they are designed for cold-start recommendation in the presence of auxiliary attributes.
We test~\protocf~by adopting BPR~\cite{bpr} (matrix factorization), VAE-CF~\cite{vae-cf},  and CDAE ~\cite{cdae} as base recommenders of our framework.

\renewcommand*{\factor}{0.056}
\begin{table}[ht]
\centering
\footnotesize
\begin{tabular}{@{}p{0.24\linewidth}@{\hspace{18pt}}
K{\factor\linewidth}K{\factor\linewidth}@{\hspace{18pt}}
K{\factor\linewidth}K{\factor\linewidth}@{\hspace{18pt}}
K{\factor\linewidth}K{\factor\linewidth}@{\hspace{18pt}}
K{\factor\linewidth}K{\factor\linewidth}K{\factor\linewidth}@{}} \\
\toprule
\multirow{1}{*}{\textbf{Dataset}} &  \multicolumn{2}{c}{\textbf{Epinions}} & \multicolumn{2}{c}{\textbf{Yelp}}  & \multicolumn{2}{c}{\textbf{Weeplaces}} & \multicolumn{2}{c}{\textbf{Gowalla}} \\
\cmidrule(lr){2-3} \cmidrule(lr){4-5} \cmidrule(lr){6-7} \cmidrule(lr){8-9} 
\multirow{1}{*}{\textbf{Metric}} & \textbf{N@50} & \textbf{R@50} &
\textbf{N@50} & \textbf{R@50} & \textbf{N@50}& \textbf{R@50} & \textbf{N@50} & \textbf{R@50}\\
\midrule
\multicolumn{9}{c}{\textsc{Standard Neural Collaborative Filtering Methods}} \\
\midrule[0pt]
\textbf{BPR~\cite{bpr}}  & 0.0860  & 0.1666 &
0.0749 & 0.1416 &
0.2537 & 0.3778 &
0.1661 & 0.2703 
\\
\textbf{NCF~\cite{ncf}} & 0.0878 & 0.1694 &
0.0752 & 0.1429 & 
0.2462 & 0.3694  & 
0.1702 & 0.2745  \\
\textbf{NGCF ~\cite{ngcf}} & 0.0913 & 0.1725 & 
0.0826 & 0.1579 & 
0.2533 & 0.3764 & 
0.1696 & 0.2758 \\
\textbf{VAE-CF~\cite{vae-cf}} & 0.0938 & 0.1778 & 
0.0854 & 0.1602 & 
0.2482 & 0.3730 & 
0.1710 & 0.2769\\
\textbf{CDAE ~\cite{cdae}} &0.0927 & 0.1774& 
0.0870 &0.1611  &
0.2570 & 0.3760  &
 0.1634& 0.2644 \\
\midrule[0pt]
\multicolumn{9}{c}{\textsc{Sparsity-aware Long-tail Item Recommendation Methods}} \\
\midrule[0pt]
\textbf{DropoutNet~\cite{dropoutnet}} &  0.0881& 0.1697&
0.0761 & 0.1435 &
0.2516 & 0.3751  &
0.1697 & 0.2768 \\
\textbf{Cofactor}~\cite{cofactor} & 0.0845  & 0.1639 & 
0.0734 & 0.1402 &
0.2342 & 0.3539  & 
0.1596 & 0.2642 
\\
\textbf{EFM}~\cite{efm} & 0.0742  & 0.1534&
0.0741 & 0.1403 & 
0.2306 & 0.3429  &
0.1532 & 0.2584 \\
\textbf{MetaRec-NLBA ~\cite{meta_rec}} &  0.0453& 0.0937& 
0.0381 & 0.0875 &
 0.1698& 0.2889  &
 0.0753& 0.1384  \\
\textbf{MetaRec-LWA ~\cite{meta_rec}}  &  0.0467&0.0943 &
 0.0392&0.1425  &
 0.1702&  0.2997 & 
 0.0722&0.1391  \\
\midrule[0pt]
\multicolumn{9}{c}{\textsc{Prototypical Collaborative Filtering Recommenders (\protocf)}} \\
\midrule[0pt]
\textbf{\protocf~+ BPR} & 0.0964&0.1812 &
0.0815 &0.1533   &
0.2576 & 0.3879 &
 0.1737&0.2800  
 \\
\textbf{\protocf~+ VAE} &  \textbf{0.0977}   &\textbf{0.1830} &
 0.0857& 0.1605 &
 \textbf{0.2725}& \textbf{0.4035}  &
 \textbf{0.1899}& \textbf{0.3004} \\
\textbf{\protocf~+ CDAE} &0.0972 &0.1824 &
 \textbf{0.0883 }& \textbf{0.1623} &
 0.2697& 0.4011  &
0.1786 & 0.2875\\
\textbf{Percentage Gains} &  4.16\%& 2.92\%& 
 1.50\% & 0.75\%  & 
 6.03\%&  6.80\%&
11.05\% &  8.49\%\\
\bottomrule
\end{tabular}
\caption{Overall item recommendation results on four datasets, R@K and N@K denote Recall@K and NDCG@K metrics at $K = 50$. Sparsity-aware models are generally outperformed by standard NCF methods on overall item recommendation;~\protocf~achieves  \textit{overall} NDCG@50 gains of 6\% and Recall@50 gains of 4\% (over the best baseline) across all datasets.}
\label{tab:protocf_7030result}
\end{table}
\subsection{{Experimental Setup}}
\label{sec:protocf_setup}
In each dataset, we randomly split the user interactions of all items into train (70\%) and test (30\%) sets, which ensures consistent interaction distributions across train and test sets.
We also use 10\% of training interactions as validation for hyper-parameter tuning.
We use NDCG@K and Recall@K as evaluation metrics, computed based on the rank of ground-truth test interactions in top-$K$ ranked item lists.

We design a unified model to enhance few-shot recommendation on the tail without affecting the overall performance.
We first evaluate \textit{overall} recommendations over the entire item set $\gI$ followed by \textit{few-shot} performance analysis over the long-tail. %

All experiments are conducted on a Tesla K-80 GPU using PyTorch.~\protocf~is trained using episodic learning for a maximum of 300 episodes with a meta-training task size of 512 items using Adam optimizer.
~\protocf~uses a multinomial log-likelihood for few-shot training and uses support sets of upto $K=10$ and query sets of $K^{\prime}=5$ interactions per item.
We learn $M=100$ group embeddings $\mZ_M$, tune the learning rate and balance hyper-parameter $\lambda$ (Equation~\ref{eqn:protocf_overall_loss}) in the range $\{ 10^{-4}, 10^{-3}, 10^{-2}\}$, and use dropout regularization with a rate of 0.5.
We tune baselines in hyper-parameter ranges centered at the author-provided values on each dataset and set the latent embedding dimension to 128 for consistency.
Our implementation of~\protocf~and datasets are publicly available\footnote{\url{https://github.com/aravindsankar28/ProtoCF}}.

\subsection{{Overall Recommendation Results} ($\text{RQ}_1$)}
\label{sec:protocf_main_results}
Our experimental results comparing the \textit{overall} recommendation performance of~\protocf~with competing baselines are shown in Table~\ref{tab:protocf_7030result}. We summarize our key empirical observations below:

\begin{itemize}
\item Neural CF models based on autoencoders (VAE-CF, CDAE) and graph neural networks (NGCF) outperform other latent-factor models (NCF, BPR) on overall model performance (Table~\ref{tab:protocf_7030result}).
\item Model regularization strategies that use item co-occurrence information (CoFactor, EFM) to improve long-tail item recommendations, are noticeably worse than BPR in overall performance. %
\item Sparsity-aware meta-learning models (MetaRec) perform poorly in overall item rankings.
One potential reason is their inability to effectively exploit or transfer knowledge from head items. %
\item \protocf~outperforms state-of-the-art baselines (NDCG@50 gains of 5\% on average) on \textit{overall} item ranking, with consistent gains for the variants of all neural base recommenders. 

Next, we examine the few-shot performance of~\protocf.

\end{itemize}

\subsection{{Few-Shot Recommendation Results} ($\text{RQ}_2$)}
\begin{figure*}[ht]
     \centering
      \includegraphics[width=0.9\linewidth]{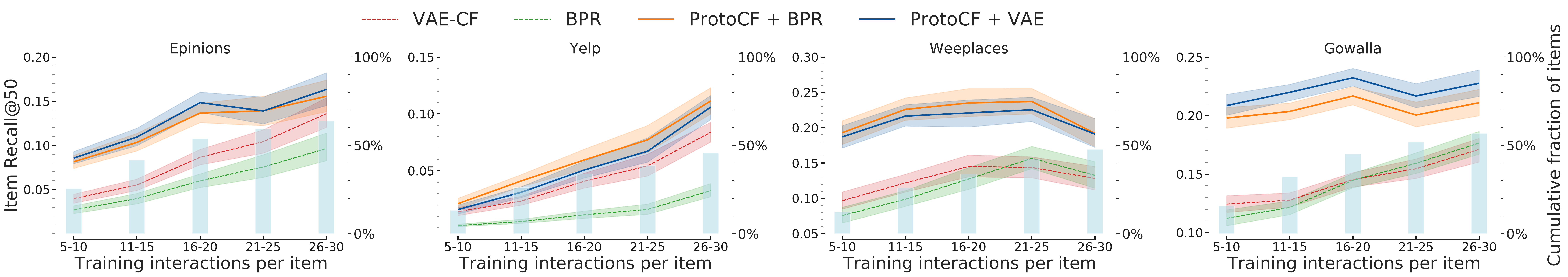}           
      \caption{Few-shot item recommendation results: Performance comparison for long-tail items with varying number of training interactions $K$ (5 to 30); lines denote model performance (Recall@50) and background histograms indicate the cumulative fraction of the item inventory covered by tail items with $\leq K$ impressions. Overall performance generally increases with $K$ for all models;~\protocf~achieves notably stronger gains (over baselines) for items with few training interactions (small $K$).}
      \label{fig:protocf_item_recall_counts}
      \vspace{-10pt}
  \end{figure*}

To evaluate few-shot results, we qualitatively analyze performance for long-tail items with sparse interactions. Specifically, we compare recommendation results for long-tail items with varying number of training interactions $K$ (5 to 30) by computing Recall@50 metrics only on their associated test interactions (Figure~\ref{fig:protocf_item_recall_counts}).
We only include the base recommenders BPR and VAE-CF here for comparison since they consistently outperform other sparsity-aware variants.

We find that model performance generally increases with item interaction count;~\protocf~achieves significant performance gains for items with less than 20 interactions. 
Episodic training with knowledge transfer is one of the key factors responsible for our higher gains over items with sparse interactions (small values of $K$).

\begin{figure*}[ht]
     \centering
      \includegraphics[width=0.9\linewidth]{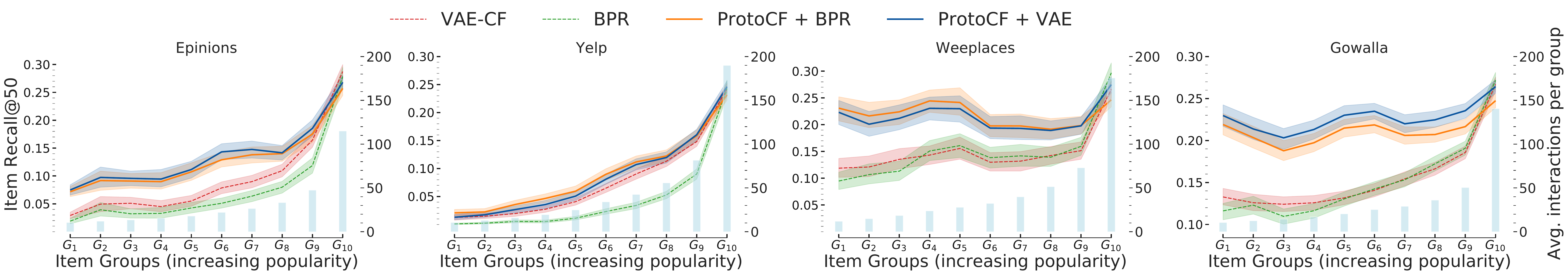}           
      \caption{Impact of item interaction sparsity: Performance comparison for item-groups sorted in increasing order by their average training interaction counts; lines denote model performance (Recall@50) and background histograms indicate the average number of interactions in each item-group.~\protocf~has significant performance gains (over baselines) on the tail items (item-groups $G_1$ to $G_8$) while maintaining comparable performance on the head items (item-groups $G_9$ to $G_{10}$).}
      \label{fig:protocf_item_recall}
  \end{figure*}

\label{sec:protocf_analysis}
To evaluate the impact of \textit{interaction sparsity} across the entire item set, we compare overall recommendation performance (Recall@K) across item-groups with different sparsity levels. 
We divide the test set into ten equal-sized item-groups, sorted in increasing order by the average number of interactions per item.
Figure~\ref{fig:protocf_item_recall} compares~\protocf~against two base recommenders BPR and VAE-CF. %

From figure~\ref{fig:protocf_item_recall}, model performance is lower on the \textit{long-tail} for base recommenders BPR and VAE due to severe interaction sparsity.
Notably,~\protocf~achieves~much higher item recall scores on the \textit{tail} items with significant gains over the corresponding base recommenders, while ensuring comparable performance on the \textit{head} items. %
Knowledge transfer of \textit{item-to-item relationships} via group embeddings and pre-trained user encoder enables~\protocf~to learn discriminative prototypes for tail items with sparse interactions.

\subsection{{Model Ablation Study} ($\text{RQ}_3$)}
\label{sec:protocf_model_analysis}
We examine the impact of different \textit{architectural} design choices in~\protocf~on its \textit{overall} and \textit{few-shot} performance (Table~\ref{tab:protocf_ablation_study}). %
We choose VAE-CF as the base recommender to instantiate~\protocf~due to its consistent results. %
Here, we report few-shot performance by only considering the test interactions of \textit{long-tail} items with less than 20 training interactions.
Note that most ablation variants do not have a significant impact on overall results since all predictions are computed by ensembling $\mR_F$ and $\mR_B$ (Equation~\ref{eqn:protocf_inference}).
Our key design choices and empirical insights are shown below:

\renewcommand*{\factor}{0.13}
\newcolumntype{K}[1]{>{\centering\arraybackslash}p{#1}}
\newcolumntype{R}[1]{>{\RaggedLeft\arraybackslash}p{#1}}

\begin{table}[h]
\small
\centering\noindent
\noindent\setlength\tabcolsep{1.2pt}
\begin{tabular}{@{}p{0.35\linewidth}K{\factor\linewidth}K{\factor\linewidth}
@{\hspace{8pt}}K{\factor\linewidth}K{\factor\linewidth}@{}} \\
\toprule
{\textbf{Dataset}} &  \multicolumn{2}{c}{\textbf{Epinions}} & \multicolumn{2}{c}{\textbf{Gowalla}} \\
\cmidrule(lr){2-3} \cmidrule(lr){4-5}
\textbf{Metric} & \textbf{Overall R@50} & \textbf{Few-shot R@50} & \textbf{Overall R@50} &  \textbf{Few-shot R@50} \\
\midrule
ProtoCF & 0.1830 & 0.1070  &  0.3004 & 0.2195  \\
w/o Prototype Gating & 0.1823 & 0.0948 & 0.2992 & 0.2082 \\ 
w/o Knowledge Distillation & 0.1805 & 0.0869 &  0.2923 & 0.1983 \\
ProtoCF-Avg & 0.1801 &  0.0712 &  0.2898 & 0.1696 \\
\protocf-logistic & 0.1804 & 0.0896 & 0.2853 & 0.1843 \\
\midrule
VAE-CF~\cite{vae-cf} & 0.1778 &  0.0549 & 0.2769 & 0.1316 \\
MetaRec-LWA~\cite{meta_rec} &  0.0943&  0.0898&0.1391 & 0.1804\\
\bottomrule
\end{tabular}
\caption{Model ablation study of~\protocf; few-shot performance is reported for tail items (less than 20 training interactions). Knowledge transfer and prototype gating contribute 10-19\% and 5-11\% to few-shot gains respectively.}
\label{tab:protocf_ablation_study}
\end{table}

\begin{itemize}
\item \textbf{Remove Prototype Gating:} We replace the gating layer (Equation~\ref{eqn:protocf_prototype_gating}) that adaptively fuses the initial and group-enhanced item prototypes, with a simpler additive operation.
Adaptive gating contributes 5-11\% few-shot performance gains.
\item \textbf{Remove Knowledge Distillation:} We test the importance of item-to-item relationships transferred from the base recommender via distillation loss $L_G$; here, we exclude knowledge transfer from group embeddings $\mZ_M$ and only 
train on the few-shot loss $L_P$. %
Removing distillation loss $L_G$ reduces few-shot results by 10-19\%.

\item \textbf{Averaging-based Prototype (\protocf-AVG):} We directly use the averaging-based initial item prototype (Equation~\ref{eqn:protocf_initial_prototype}) for inference; here, we also exclude the pre-trained parameter initialization for the user encoder $G_U$ (from $F_U$ in the base recommender).
~\protocf-AVG~(without any knowledge transfer) is worse than~\protocf~by a margin of 20-30\%, yet outperforms the base recommender VAE-CF by nearly 30\% on few-shot items.

\item \textbf{Logistic log-likelihood (\protocf-logistic):} We train~\protocf~using \textit{logistic} log-likelihood (Equation~\ref{eqn:protocf_logistic}).~\protocf~with multinomial log-likelihood (Equation~\ref{eqn:protocf_multinomial}) outperforms \protocf-logistic by a considerable margin; this validates prior findings~\cite{vae-cf} on the efficacy of multinomal for top-$N$ recommendation.
\end{itemize}

Note that the meta-learning baseline MetaRec-LWA improves few-shot results (compared to VAE-CF), but forgets knowledge of head items resulting in poor overall recommendations.

\subsection{{Parameter Sensitivity} ($\text{RQ}_4$)}
\label{sec:protocf_sensitivity}
\begin{figure}[th]
      \includegraphics[width=0.9\linewidth]{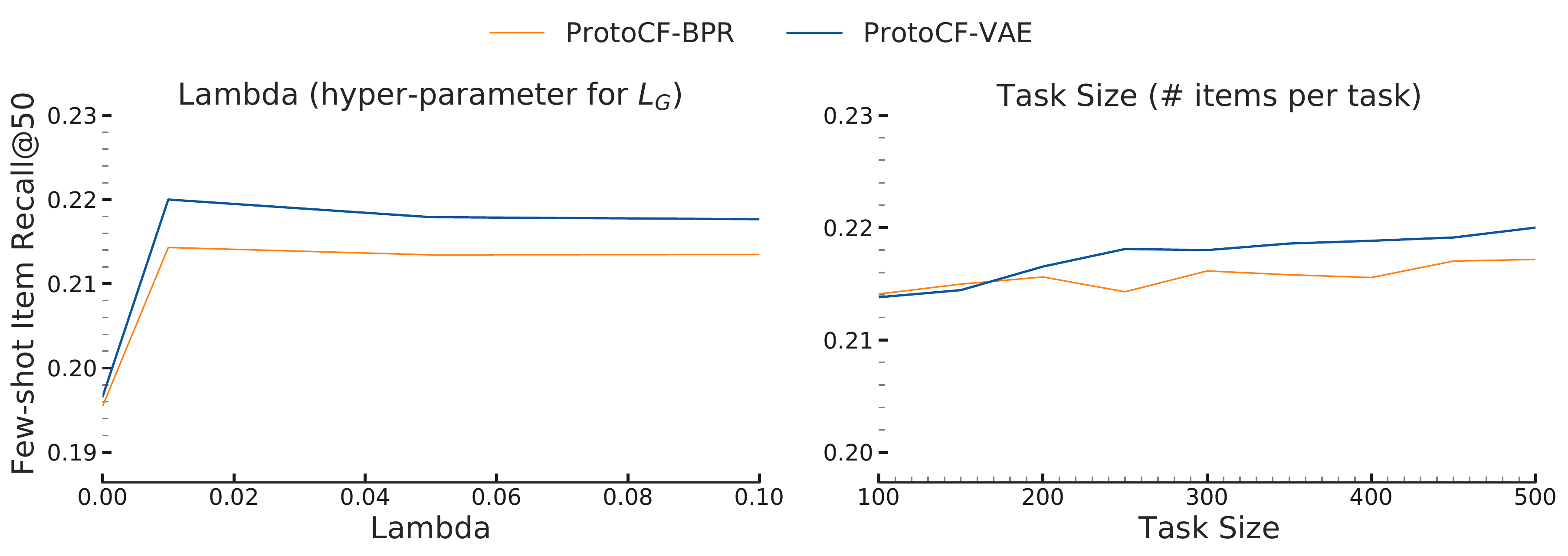}
      \caption{Few-shot performance on Gowalla (for tail items with less than 20 training interactions) is higher for larger meta-training tasks; the empirically optimal value of balance factor $\lambda = 0.01$ also transfers across all datasets.}
      \label{fig:protocf_sensitivity}
  \end{figure}

We analyze sensitivity to the hyper-parameter $\lambda$ that weights the knowledge distillation loss $L_G$ (\Cref{eqn:protocf_overall_loss}).
In Figure ~\ref{fig:protocf_sensitivity} (a), we show the effect of $\lambda$ on few-shot results with base recommenders BPR and VAE-CF on Gowalla.
We empirically find the optimal value of $\lambda$ to be 0.01 for both models, which also transfers across datasets.

We investigate the impact of \textit{task size}, which is the number of items $N$ sampled in each meta-training task $\gT$. In Figure~\ref{fig:protocf_sensitivity} (b), performance typically increases with task size (and stabilizes $\sim 400$), due to a larger set of sampled items available for ranking; this observation is consistent with prior few-shot learning studies~\cite{protonet}.  %

\subsection{{Discussion}}
\label{sec:protocf_discussion}
Our framework~\protocf~is orthogonal to architectural advances in neural recommendation models that enhance representational capacity to learn from massive interaction data. We adapt expressive neural recommenders to develop light-weight few-shot models tailored to the long-tail.
Furthermore, our few-shot formulation requires no side information and can further be easily adapted to address the long-tail of users.

~\protocf~learns a metric space predicated on knowledge transfer over a shared user space. Few-shot transfer learning across domains with disjoint feature spaces is a potential future direction.

\section{Conclusion}
\label{sec:protocf_conclusion}
We formulate long-tail item recommendation as learning-to-embed items with sparse interactions.
A few-shot learning framework~\protocf~is introduced to extract meta-knowledge across a collection of training tasks designed to simulate tail item ranking.
~\protocf~efficiently transfers knowledge from arbitrary base recommenders to construct discriminative prototypes for items with few interactions.
Our experiments indicate 5\% overall performance gains (Recall@50) for~\protocf~over the state-of-the-art, with notable 60-80\% few-shot performance gains (Recall@50) on long-tail items with less than 20 training interactions.

In this chapter, we examined another common transductive learning setting of collaborative filtering over bipartite user-item interactions that are prevalent in several online interaction platforms.
To overcome entity-level (user/item) interaction sparsity challenges that arise due to heavy-tailed distributions in user interests and interaction patterns, we introduced  a meta-learning framework for few-shot recommendation.
In the next part of this dissertation, we examine several inductive learning applications targeted at behavior modeling for new (or unseen) entities that are only observed during inference.

\chapter{\textsc{InfVAE}: Social Regularization to Predict Information Diffusion}
\label{chap:infvae}

\section{Introduction}
\label{sec:infvae_intro}
In social media, information disseminates or \emph{diffuses} to a large number of users through content posting or re-sharing behavior, resulting in a \textit{cascade} of user activations, \textit{e.g.}, a user voting a news story on Digg (a social news sharing website) triggers a series of votes from multiple users, who may be his friends or other users interested in the same story.
Given a set of \textit{activated} seed users who have posted a piece of user-generated content, diffusion models aim to predict the set of all influenced users who re-share it.
Diffusion modeling has widespread social media applications, including viral marketing~\cite{viral}, limiting misinformation spread, personalized recommendations~\cite{social_adv}, and popularity prediction~\cite{tweet_popularity}.

The diffusion prediction problem has received significant attention in the research community. The earliest methods assume that the mechanism of social influence propagation is known a priori, \textit{e.g.}, the Independent Cascade (IC)~\cite{ic,ctic} and Linear Threshold (LT) models~\cite{lt} that associate each link with a non-negative weight indicative of pairwise user-user influence.
Unlike pre-defined propagation hypotheses~\cite{ic}, recent methods learn data-driven diffusion models from collections of user activation sequences (\textit{diffusion cascades}).
Prior diffusion models broadly fall into two categories.

\textit{Probabilistic generative cascade} models use hand-crafted features including social roles~\cite{rain}, communities~\cite{cic-icdm}, topics~\cite{topic-ic}, and structural patterns~\cite{structinf}. Such methods
rely on feature engineering that requires manual effort and extensive domain knowledge, and are limited by the modeling capacity of carefully chosen probability distributions.

\textit{Representation learning} methods avoid feature extraction by learning user representations characterizing their influencing ability and conformity~\cite{wsdm16,inf2vec}.
Prior work mainly consider the impact of temporal user-user influence learned from historical diffusion cascades; prior methods project cascades onto local social neighborhoods to generate Directed Acyclic Graphs (DAGs), and propose extensions of Recurrent Neural Networks (RNNs) to model \textit{temporal influence}.
State-of-the-art models include DAG-structured LSTMs~\cite{topolstm} that explicitly operate on induced DAGs and attention-based RNNs~\cite{cyanrnn,deepdiffuse,cikm18_attention} that implicitly consider cross-correlations across activated seed users for diffusion prediction.

Prior work only model the sequence or projected social structure (induced DAG) of previously influenced users while ignoring \textit{social structures that do not manifest in cascades}; they only capture the temporal correlation in diffusion behaviors known as \textit{temporal influence} or \textit{contagion}~\cite{shalizi}.
Consider a Twitter user interested in politics who follows famous political leaders and joins interest groups; this induces transitive connections to other users that however may not manifest in cascades unless she re-tweets or posts content.
\textit{Social homophily}~\cite{homophily} suggests \textit{stronger ties} between users with shared traits or interests, which induces correlated diffusion behaviors without direct causal influence.
Social graph connectivity information reveals insights into \textit{homophily} and is critical to model diffusion behaviors for the vast majority of users who rarely post content and appear in cascades.

However, homophilous diffusion and contagion can result in differing dynamics, \textit{e.g.}, contagions are self-reinforcing and viral while homophily hinges on users' preferences or traits.
Indeed, social homophily and temporal influence are fundamentally confounded in observational studies of diffusion processes~\cite{shalizi}, which makes it challenging to contextually model the impact of both factors.
Thus, our key objective is to \textit{develop a principled neural framework to contextually the model the co-variance of temporal influence (recent posting activities) with social hompohily (structural graph connectivity) for diffusion prediction}.

\textbf{Present Work:} Our \textit{social regularization} framework~\infvae~jointly models \emph{homophily} through \emph{social} embeddings preserving social network proximity and \emph{influence} through \emph{temporal} embeddings encoding the relative sequential order of user activations.
Motivated by the capabilities of variational autoencoders (VAEs)~\cite{vae} in alleviating sparsity via Gaussian priors~\cite{vae-cf} and the expressive power of Graph Neural Networks (GNNs) ~\cite{gcn, graph_enc_dec}, we adopt graph-VAEs to model social homophily.
We learn structure-preserving social embeddings for each user through VAEs, and differentiate their social roles (influential versus susceptible) towards diffusion modeling.
Given an initial set of seed user activations,~\infvae~utilizes an expressive \textit{co-attentive fusion network} that captures non-linear correlations between social and temporal embeddings, to contextually model the co-variance of homophily and influence on predicting the set of all influenced users. We make the following contributions:

\begin{description}

\item \textbf{Generalizable Variational Autoencoder Framework}:
Unlike existing diffusion prediction methods that only consider temporal influence via induced propagation structures,~\infvae~models homophily and distinguishes social roles of users towards diffusion prediction through a novel generalizable VAE framework that can be instantiated with a wide variety of GNN architectures of arbitrary model complexity.

\item \textbf{Efficient Homophily and Influence Integration:}
To the best of our knowledge, ours is the first neural framework to exploit the co-variance of social homophily and temporal influence for predicting diffusion behaviors. Given a sequence of seed user activations,~\infvae~employs an expressive \textit{co-attentive fusion network} to jointly attend over their social and temporal representations to predict the set of all influenced users.~\infvae~is faster than state-of-the-art recurrent methods by an order of magnitude.

\item \textbf{Robust Experimental Results:}
Our experiments on multiple real-world social networks, including Digg, Weibo, and Stack-Exchanges, demonstrate significant gains for~\infvae~over state-of-the-art models.
Modeling social homophily through VAEs enables massive gains for users with \textit{sparse activities}, and users who \textit{lack direct social connections in seed sets}. An ablation analysis of various modeling choices further justifies the benefits of modeling the co-variance of social homophily and temporal influence.

\end{description}

We organize the rest of the chapter as follows. In Section~\ref{sec:infvae_defn}, we formally define the problem of diffusion prediction in social networking platforms. In Section~\ref{sec:infvae_model}, we describe our proposed Influence Variational AutoEncoder framework (\infvae). We present experimental results in Section~\ref{sec:infvae_experiments}, and finally conclude in Section~\ref{sec:infvae_conclusion}.

\section{Related Work}
\label{sec:related}
We discuss existing work on diffusion modeling followed by
related work on network representation learning, variational auto-encoders and co-attentions.

\textbf{Information diffusion overview.}
Historically, information diffusion has been studied through two seminal models: Independent Cascade (IC)~\citep{ic} and Linear Threshold (LT)~\citep{lt}.
Three distinct applications emerged, namely: \textit{network inference}~\cite{netinf}, which infers the underlying social network that best explains the observed cascades; \textit{cascade prediction}~\cite{deepcas}, which predicts macroscopic properties of cascades, including size, growth, and shape; and \textit{diffusion prediction}~\cite{topolstm}, which learns a model from social links and cascade sequences, to predict the set of influenced users given a seed set of activated users.
In this work, we focus on diffusion prediction.

\textbf{Diffusion prediction.} 
The earliest data-driven methods propose several extensions of IC and LT incorporating topics~\cite{topic-ic}, continuous timestamps~\cite{ctic}, user profiles~\cite{node_attribute}, and community structure~\cite{cic-icdm}. A few techniques explore probabilistic generative models via latent topics and communities~\citep{cold,hcid}. 
Most recent studies focus on learning representations to overcome feature engineering or pre-defined hypotheses in diffusion modeling~\cite{cdk,aaai15,wsdm16,topolstm,cyanrnn,deepinf,inf2vec,cikm18_attention}.
Emb-IC~\cite{wsdm16}, Inf2vec~\cite{inf2vec} embed user influencing capability and susceptibility in diffusion.
Topo-LSTM~\cite{topolstm}, CYAN-RNN~\cite{cyanrnn}, SNIDSA~\cite{cikm18_attention}, and DeepDiffuse~\cite{deepdiffuse} project the diffusion cascades on local social neighborhoods and model the resulting DAG propagation structures with RNNs.
These techniques outperform classical approaches by significant margins in diffusion prediction. 
Our key observation is that these projected DAGs could ignore social structures that do not appear in any observed cascade. In contrast, our model~\infvae~can account for unobserved social connections in the user activation process by modeling social homophily through VAEs.

A related problem is social influence prediction, which aims to classify social media users based on the activation status of their ego-network~\cite{locality, deepinf}.
Direct extensions to predict the set of all influenced users (diffusion prediction) entails reapplying their models on each candidate inactive user in the social network, resulting in prohibitive inference costs, hence preventing a comparison. 

\textbf{Network representation learning:}
This line of work captures varied notions of structural node proximity~\cite{node2vec, rase} in networks via low-dimensional vectors.
Notably, graph neural networks have achieved great success
in node classification and link prediction~\cite{gcn,graphsage,gat, motifcnn, metagnn, dysat_arxiv, Narang2019induced}.
Graph Autoencoders~\cite{sdne,vgae} employ various encoding and decoding architectures to embed network structure and learn unsupervised node embeddings.
~\cite{graph_enc_dec} unify a large family of network embedding methods
in an autoencoder framework.
However, general-purpose embeddings modeling structural proximity are not directly suited to diffusion modeling.

\textbf{Co-attentional models:}
Our work also leverages recent advances in neural attention mechanisms, especially in Natural Language Processing~\cite{attention}. 
Specifically, co-attention has achieved great success in modeling relationships between pairs of sequences, \textit{e.g.}, question-answer~\cite{dynamic_coattention}, etc.
Co-attentional methods compute interaction weights between data modalities, learning fine-grained non-linear correlations. In our work, we develop a co-attentive fusion network to capture the contextual interplay of users' social and temporal representations for diffusion prediction.

\section{Problem Definition}
\label{sec:infvae_defn}
We study diffusion prediction in social networks where the goal is to predict the set of all influenced users, given temporally ordered seed user activations for user-generated content.
\begin{definition}\textbf{Social Network:} The social network is represented as a graph $\gG = (\gV,\gE)$ where $\gV = \{ v_i\}_{i=1}^N$ is the set of $N$ users and $\gE = \{ e_{ij} \}_{i,j=1}^N$ is the set of links. We denote the adjacency matrix of $\gG$ by $A \in \sR^{N \times N}$ where $A_{i,j} = 1$ if $e_{i,j} \in \gE$ otherwise 0.
\end{definition}

\begin{definition}\textbf{Diffusion cascade:} A diffusion cascade $D_i$ is an ordered sequence of user activations in ascending time order denoted by: $D_i = \{ (v_{i_k}, t_k) \mid v_{i_k} \in \gV, t_k \in [0, \infty), \; k = 1\dots K\}$, each $v_{i_k}$ is a distinct user in $\gV$ (no repeats) 
and $t_k$ is non-decreasing, \textit{i.e.}, $t_k \leq t_{k+1}$. The $k^{th}$ user activation has a tuple $(v_{i_k}, t_k)$, referring the activated user and activation time. 
\end{definition}

We represent cascades by delay-agnostic relative activation orders similar to~\cite{inf2vec,topolstm,wsdm16}, \textit{i.e.}, a cascade is equivalently written as $D = \{(v_{i_k} , k) \mid v_{i_k} \in \gV\}_{k=1}^K$. We do not assume the availability of explicit re-share links between users in cascades; this corresponds to the simplest yet most general setting of diffusion~\cite{topolstm,inf2vec}.
Though timestamps may be easily used as input features, we leave generation of continuous timestamps as future work.

\begin{definition} \textbf{Diffusion prediction:}
Given a social network $G$ and a collection of cascade sequences $\sD = \{D_i, 1 \leq i \leq |\sD|\}$, learn diffusion model $M$ to predict the future set of influenced users in a cascade with seed activation sequence $I = \{(v_{i_1}, 1), \dots, (v_{i_k},k) \}$ of $k$ seed users. %
Diffusion prediction estimates the probability of influencing each inactive user: $P_\Theta(v \mid I) \; \forall v \in \gV -  I$, inducing a ranking of activation likelihoods over the inactive user set.
\end{definition}

We create a training set $\sT$ of diffusion \emph{episodes} containing (seed activations, activated users) tuples from the cascade collection $\sD$, by randomly splitting each cascade $D \in \sD$ of length $K$ at each time step $ 2 \leq k \leq K-1$.
Specifically, a split at time step $k \geq 2$, creates a training episode $(I_{k}, C_{k})$ where $I_k = \{ (v_{i_j}, j);  1 \leq j \leq k \}$ is the seed set consisting of the cascade sliced at $k$ and $C_{k} = \{ v_{i_{k+1}}, \dots, v_{i_K} \}$ is the set of influenced users after time step $k$. Thus, we denote the training set by $\sT = \{ (I_i, C_i ) \; 1 \leq i \leq |\sT| \}$.

\section{\infvae~Framework}
\label{sec:infvae_model}
In this section, we describe our proposed Influence Variational  Autoencoder (\infvae) framework for predicting information diffusion in social networks.
We first introduce the different interacting latent variables in our deep generative model, then describe the process of generating diffusion cascades and finally present model training details.

\begin{table}[t]
    \centering
    \begin{tabular}{@{}c|l@{}}
    \toprule
        Symbol &  Description \\
        \midrule
         $\mZ$ &  Social variables modeling network proximity, for all users $\gV$ \\
         $\mV_S$ &  Sender variables for all users $\gV$ \\
         $\mV_R$ &  Receiver variables for all users $\gV$ \\
         $\mV_{T}$ & Temporal influence variables for all users $\gV$ \\
         $\mV_{P}$ & User-specific popularity variables for all users $\gV$ \\
         $\mP_K$ & Position-encoded temporal embeddings for all time steps $K$ \\
         \bottomrule
    \end{tabular}
    \caption{Notations}
    \label{tab:notations}
\end{table}
\subsection{{Generative Model Description}}
\label{sec:infvae_model_description}
\infvae~is a neural latent variable model that jointly capture social homophily and temporal influence for diffusion prediction.
We describe the latent variables modeling  homophily and  influence, followed by the structure of our generative network~\infvae.

\subsubsection{\textbf{Social Homophily.}}
\label{sec:infvae_structural}
Our objective is to define latent \textit{social variables} for users that capture social homophily.
The homophily principle stipulates that users with similar interests are more likely to be connected~\cite{homophily}. In the absence of explicit user attributes (\textit{e.g.}, demographics or interests), we posit that 
highly interconnected users in social communities share homophilous relationships.
We model social homophily through latent \textit{social} variables designed to encourage users with shared social neighborhoods to have similar latent representations.

\begin{figure}[t]
    \centering
    \includegraphics[width=0.9\linewidth]{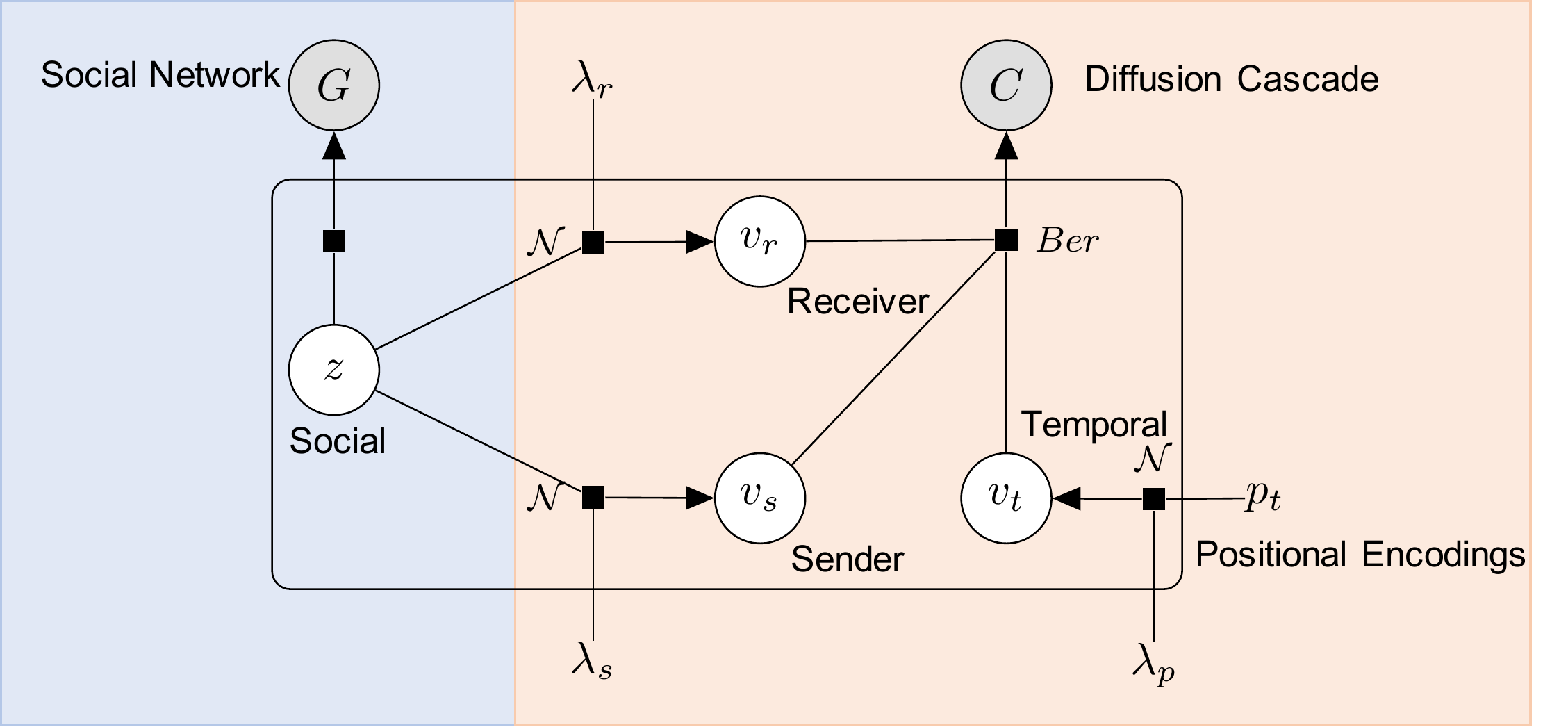}
    \caption{Plate Diagram of~\infvae~depicting interactions between different latent variables: social, sender, receive and temporal variables.}
    \label{fig:plate}
\end{figure}

Specifically, we assign a latent \emph{social} variable $\rvz_i$ for user $v_i$, where the prior for $\rvz_i$ is chosen to be a unit normal distribution, in line with standard assumptions in VAEs.
Normal distributions are chosen in VAE frameworks due to their flexibility to support arbitrary functional parameterizations by isolating sampling stochasticity to facilitate back-propagation~\cite{vae}.
We assume the latent social variables $\mZ$ to collectively generate the social network $\gG$, through a graph generation neural network $f_{\textsc{dec}} (\mZ)$ parameterized by $\theta$. The corresponding graph generative process is given by:
\begin{equation}
\rvz_i \sim \gN (0, I_D) \hspace{10pt} \gG \sim p_{\theta}(\gG \mid \mZ) =  p_{\theta}(\gG \mid  f_{\textsc{dec}} (\mZ))
\label{eqn:infvae_struct_gen}
\end{equation}
where $I_D \in \sR^{D \times D}$ is an identity matrix of $D$ dimensions. Here, the graph generation neural network $f_{\textsc{dec}} (\mZ)$ can be instantiated to preserve an arbitrary notion of structural proximity in the social network $\gG$ (Sec~\ref{sec:infvae_neural_vgae}).
In the above equation, we abuse the notation of $\gG$ to denote an appropriate representational form of the social network structure, which can take multiple forms, including the adjacency matrix, random walks sampled from $\gG$, etc.

While homophily characterizes peer-to-peer interest similarity, its impact on diffusion behaviors of users is asymmetric since users who share interests may drastically differ in their posting rates; \textit{e.g.}, certain users are naturally predisposed to be socially active and hence more \textit{influential} in comparison to others.
Thus, it is necessary to differentiate social user \textit{roles} (influential versus susceptible) when modeling the impact of social homophily on diffusion behaviors.
Similar concepts have been examined in social influence literature to characterize users by their influencing capability and conformity~\cite{conformity,aaai15,wsdm16,topolstm,inf2vec}.

We associate each user $v_i \in \gV$ with a \textit{sender} $\rvv^{s}_i \in \sR^{D}$ and \textit{receiver} $\rvv^{r}_i \in \sR^{D}$ latent variable.
In contrast to prior VAEs over graphs~\cite{graph_enc_dec} that preserve structural graph connectivity, our key innovation lies in conditioning the information sending and receiving capabilities of users on their homophilous traits through their social latent variables.
We use normal distributions centered at $\rvz_i$ 
to define the \emph{sender} and \emph{receiver} variables for user $v_i$ as: 

\begin{align}
\rvv^{s}_i &\sim \gN (\mathbf{z}_i, \lambda_s^{-1} I_D) \hspace{10pt} \rvv^{r}_i \sim \gN (\mathbf{z}_i , \lambda_r^{-1} I_D) 
\end{align}
where $\lambda_s$ and $\lambda_r$ are tunable hyper-parameters controlling the degree of variation or uncertainty for $\rvv^{s}_i$ and $\rvv^{r}_i$ \textit{w.r.t.} $\rvz_i$ for user $v_i$.
Let $\mV_S$ and $\mV_R $ denote the set of all sender and receiver variables respectively for all users in the social network, which are independently conditioned on the corresponding set of latent social variables $\mZ$.

\begin{figure}[t]
    \centering
    \includegraphics[width=0.9\linewidth]{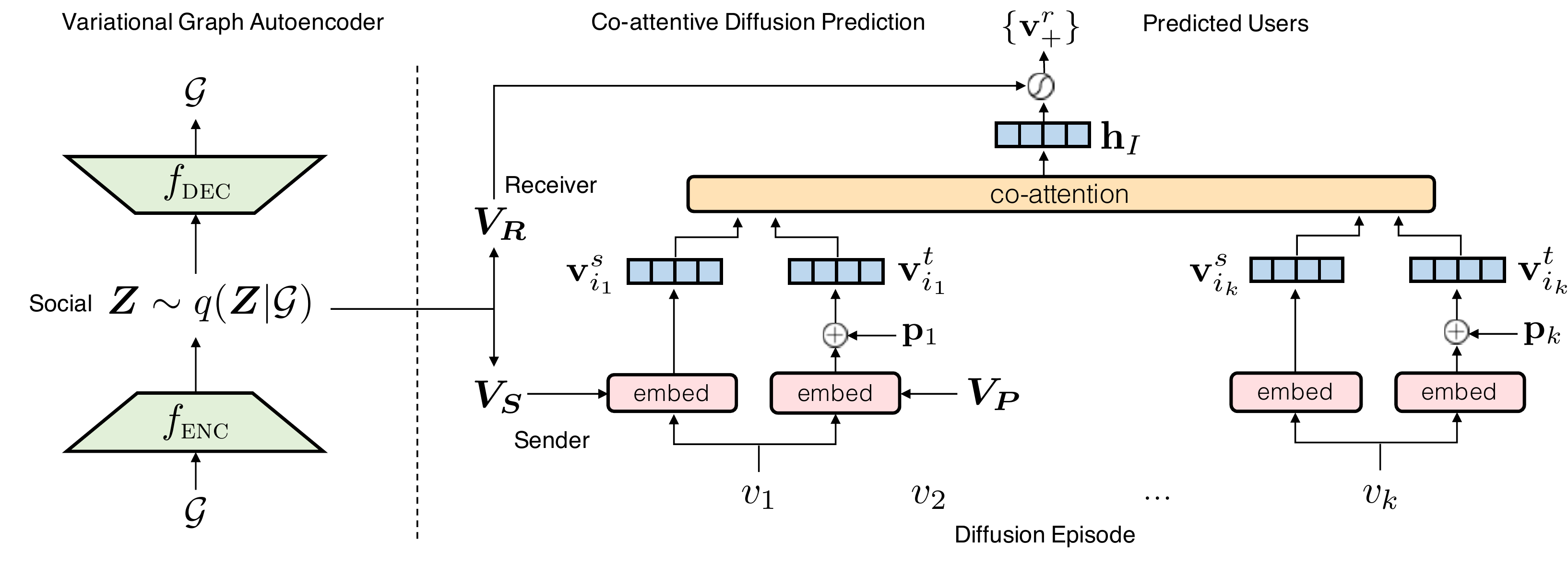}
    \caption{Neural Architecture of~\infvae~depicting latent variable interactions. The left side indicates the VAE framework to model social homophily; right side denotes the co-attentive fusion network to integrate the social and temporal variables.}
    \label{fig:arch}
\end{figure}

\subsubsection{\textbf{Temporal Influence.}} Now, we define latent \textit{temporal influence} variables to describe the varying influence effects of seed users depending on the relative sequential order of activations.
There are two interesting factors at play: activation orders and popularity effects.
A majority of social media users adopt more recent information while often ignoring old and obsolete content~\cite{implicit_diffusion}.
On the other hand, social status impacts the influencing power of seed users independent of their activation order and social neighbors,
\textit{e.g.}, famous media figures naturally exert significant influence.
Thus, we consider both the relative sequential order of user activations and popularity effects of seed users to model temporal influence.

To quantify the temporal influence exerted by a seed user activation $(v_{i_k}, k)$ of user $v_{i_k}$ at time step $k$ ($ 1\leq k \leq K$), we first encode the relative position $k$ through positional-encodings~\cite{posn-enc} to obtain temporal embeddings $\rvp_k$.
Since we expect the variation in popularity effects to be quite small, we draw
user-specific popularity variables from a zero-mean normal distribution to serve as offsets to the temporal embeddings.
Specifically, the \emph{temporal influence} variable for activation $(v_{i_k}, k)$ denoted by $\rvv^{t}_{i_k}$, is given by:

\begin{align}
    \rvv^{p}_{i_k} \sim \gN (0, \lambda_{p}^{-1} I_D) \hspace{15pt}  \rvp_k = PE(k) \hspace{15pt} \rvv^{t}_{i_k} =  \rvv^{p}_{i_k} + \rvp_k \\
    PE(k)_{2d} = sin(k/10000^{2d/D}) \hspace{5pt} PE(k)_{2d+1} = cos(k/10000^{2d/D}) \nonumber
\end{align}
where $\lambda_{p}$ is a hyper-parameter to control the popularity effects, and $1 \leq d \leq D/2$ denotes the dimension in the temporal embedding $\rvp_k$.
Note that the popularity variable $\rvv^{p}_{i_k}$ is user-specific, while temporal embedding $\rvp_k$ only depends on the activation step $k$.
The set of all latent user popularity variables are denoted by $\mV_{P}$, while $\mP_K$ represents the set of position-encoded \emph{temporal} embeddings.
\subsubsection{\textbf{Co-attentive Diffusion Episode Generation.}}
Let us consider a single diffusion episode $(I, C) \in \sT$, with initial seed user activations $I = \{ (v_{i_1}, 1), \dots, (v_{i_k}, k)\}$ and influenced users $C = \{ v_{i_{k+1}}, \dots, v_{i_K}\}$.
A diffusion model aims to predict the set of influenced users $C$ given seed activations $I$.
Since information diffusion is always conditioned on the seed user $I$, we conditionally sample $C$ given $I$. %

Let us denote the set of seed users by $I_U = \{ v_{i_1}, \dots, v_{i_k} \}$.
Our objective is to model the co-variance of social homophily and temporal influence exerted by seed users $I_U$, which can be summarized by:
\textit{sender variable sequence} $(\rv^s_{i_1}, \rv^s_{i_2}, \dots, \rv^s_{i_K})$; and \textit{temporal influence variable sequence} $(\rv^{t}_{i_1}, \rv^{t}_{i_2}, \dots, \rv^{t}_{i_K})$.
To model complex correlations between the sender and temporal influence variable sequences, 
we propose an expressive \textit{co-attentive} fusion strategy to learn attention scores for each seed user by modeling interactions between the two sequences.
We describe the conditional generative process in two steps:

\begin{itemize}
\item The social homophily and temporal influence aspects of seed users, are integrated into an aggregate seed set representation $\rvh_I$.
The co-attentive fusion network $G_{\textsc{diff}} (\cdot)$ performs homophily-guided temporal attention, \textit{i.e.}, attends over the temporal influence variables by computing co-attentional weights that jointly depend on both homophily and temporal influence characteristics.
As illustrated in Figure~\ref{fig:arch}, the sender and temporal influence variables of seed users feed into a fusion network $G_{\textsc{diff}}(\rvv^s_{i_k}, \rvv^{t}_{i_k})$.
The aggregate seed representation $\rvh_I$ is computed as:
\begin{align}
\alpha_{k} = \frac{ \exp( G_{\textsc{diff}}(\rvv^s_{i_k}, \rvv^{t}_{i_k} (k))) }{\sum\limits_{j=1}^K \exp(G_{\textsc{diff}}(\rv^s_{i_j}, \rv^{t}_{i_j} (j))) }  \hspace{10pt} \rvh_I  = \sum\limits_{j=1}^K \alpha_j \rvv^{t}_{i_j} (j)
\end{align}
Each $\alpha_j$ is the normalized co-attentional coefficient for seed user $v_{i_k}$ denoting its  contribution in computing the aggregate representation $\rvh_I$.
To model the co-dependence between $\rvv^s_i$, $\rvv^{t}_i$, we define the co-attentive function $G_{\textsc{diff}}(\rvv^s_{i_k}, \rvv^{t}_{i_k}) = tanh({\rvv^s_{i_k}}^T \mW \rvv^{t}_{i_k})$ as a bi-linear product parameterized by $\mW \in \sR^{D \times D}$. %

\item The probability of influencing an inactive user $v_j$ depends on the sending capacity of seed users (embedded in $\rvh_I$) and her receiving capability (encoded by \emph{receiver} variable $\rvv_j^r$).
We quantify the likelihood of influencing $v_j$ by the inner product $\rvh_{I}^T \rvv_j^r$.
For each inactive user $v_j \in \gV- I_U$, we draw a binary variable $\gC_j \in \{0,1\}$ indicating whether user $v_j$ is influenced by the seed users $I_U$ or not, given by:
\begin{equation}
\gC_j \sim %
Ber (\sigma(\rvh_{I}^T \rvv_j^r)) \; \forall v_j \in \gV- \{ v_{i_1}, \dots, v_{i_K}\}
\label{eqn:infvae_diffusion_gen}
\end{equation}
\noindent where $\sigma(\cdot)$ is the sigmoid function and $Ber(\cdot)$ is the Bernoulli distribution.
The corresponding logistic log-likelihood of generating a diffusion episode $(I, C)$ is given by:
\begin{align}
\gL^{\textsc{diff}}_{I,C} &= \log p_{\theta} (C \mid I, \mV_S, \mV_R, \mV_{P})  \\ \nonumber \vspace{-3pt}
&=  \sum\limits_{v \in C} \eta \log (\sigma(\rvh_{I}^T \rvv_i^r)) +  \sum\limits_{v_n \in \gV - C - I_U} \log (1-  \sigma(\rvh_{I}^T \rvv_n^r ))
\end{align}
Here, the hyper-parameter $\eta$ re-weights the observed positive examples since the actual number of influenced users is much smaller than the total number of users.
\end{itemize}

\subsection{{Model Likelihood}}
\label{sec:infvae_likelihood}

An analytical computation of the latent posterior distribution $p (\mV_S, \mV_R, \mV_{P}, \mZ | \gG, \sT)$ is intractable due to the learnable neural layers in the generative process. Thus, we use variational inference to factorize the latent posterior with a mean-field approximation as:

\begin{align}
q(\mV_S, \mV_R, \mV_P, \mZ | \gG) =   q(\mV_S ) q(\mV_R) q(\mV_P) q(\mZ | \gG)
\label{eqn:infvae_mean_field}
\end{align}

The variational distributions of variables $\mV_S, \mV_R,$ and $\mV_P$ follow normal distributions while the 
social variables $\mZ$ are conditioned on $\gG$ through a structure-encoding inference network~\cite{vae}.
Specifically, the variational distribution of $\mZ$ denoted by $q_{\phi}(\mZ | \gG)$, is a diagonal normal distribution parameterized by $f_{\textsc{enc}} (\gG)$ defined as:

\begin{equation}
f_{\textsc{enc}} (\gG) \equiv [\mu_{\phi}(\gG),  \log \sigma_{\phi}^2 (\gG)]  \hspace{5pt} q_{\phi} (\mZ | \gG) = \gN \left(\mu_{\phi}(\gG), diag( \sigma_{\phi}^2 (\gG))\right)
\end{equation}
The inference network outputs the parameters, $\mu_{\phi}(\gG), \sigma_{\phi} (\gG)$ of the variational distribution $q_{\phi} (\mZ | \gG)$, which is designed to approximate the corresponding posterior $p(\mZ | \gG)$.
The inference network $f_{\textsc{enc}} (\gG)$
endows the model with added flexibility to incorporate arbitrary neighborhood aggregation functions such as graph convolutions~\cite{gcn}, attentions~\cite{gat}, etc.
The variational structure distribution $q_{\phi} (\mZ | \gG)$  and the structure generative model $p_{\theta} (\gG | \mZ)$ (Eqn.~\ref{eqn:infvae_struct_gen}) together constitutes a \emph{variational graph autoencoder}~\cite{vgae}.

 \begin{figure}[t]
 \centering
 \includegraphics[width=0.85\linewidth]{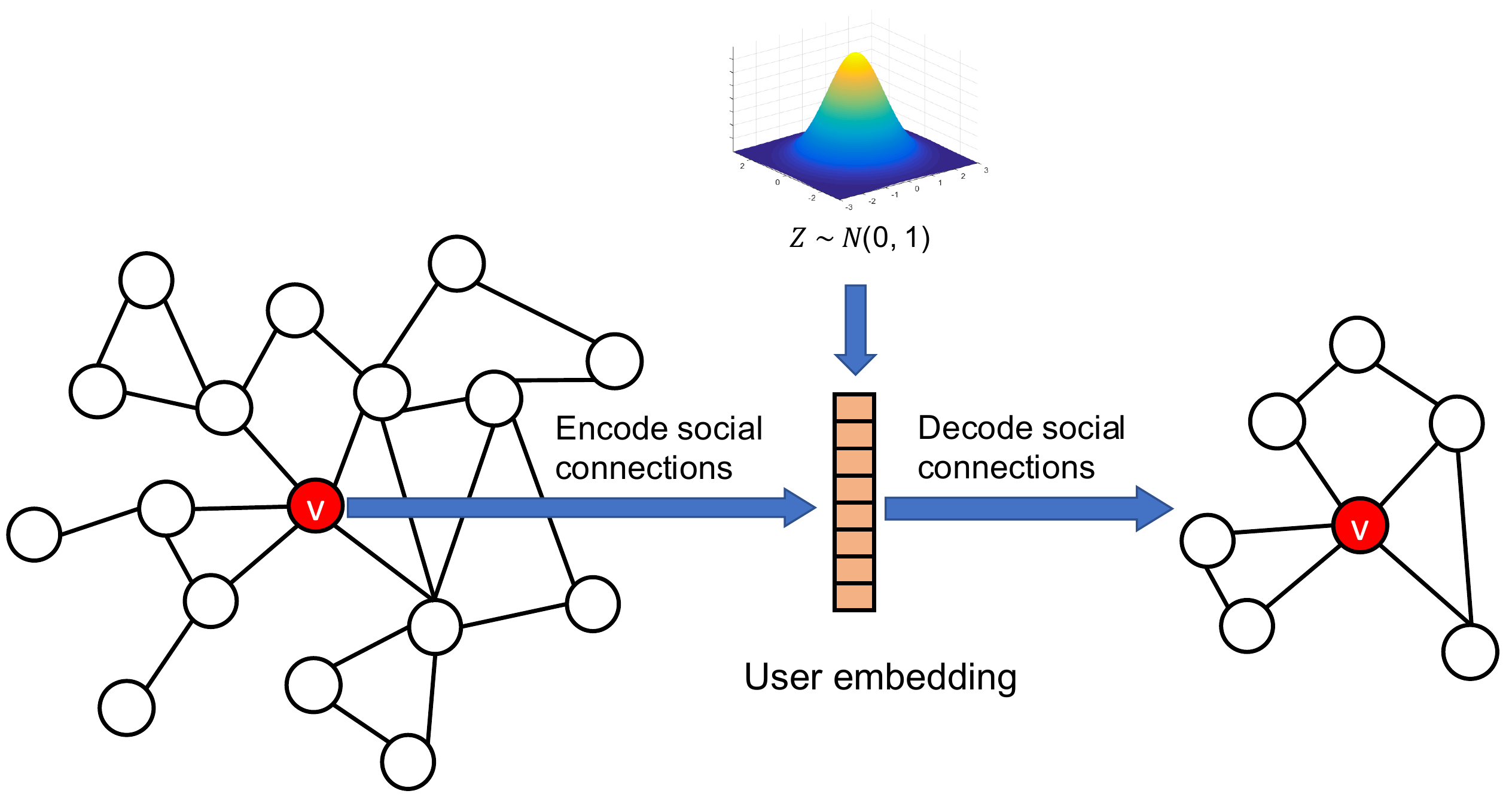}
 \caption{Variational Graph Encoder-Decoder framework showing the social latent variables in~\infvae~that preserve structural proximity in the social network.}
 \label{fig:infvae_ae}
 \end{figure}
 
\subsection{{Neural Graph Autoencoder Details}}
\label{sec:infvae_neural_vgae}
In this section, we describe functions $f_{\textsc{enc}} (\gG)$ and $f_{\textsc{dec}} (\mZ)$ which describe the graph structure inference and generative networks of~\infvae.
The \emph{encoder} summarizes local social
neighborhoods into latent vectors, which are subsequently transformed by the \emph{decoder} into high-dimensional structural information (\textit{e.g.}, adjacency matrix).
~\cite{graph_enc_dec} present an encoder-decoder framework to conceptually unify a large family of graph embedding methods.
Encoder architectures fall into three major categories: embedding lookups~\cite{deepwalk,node2vec}, neighborhood vector encoding~\cite{sdne}, and neighborhood aggregation~\cite{graphsage}, while decoders comprise unary and pairwise variants.
In~\infvae, we explore two representative choices:

\begin{description}
\item \textbf{MLP + MLP}: We use a Multi-Layer Perceptron (MLP) to both encode and decode the laplacian matrix of $\gG$, given by $\mL = \mD^{-1/2} \mA \mD^{-1/2}$. %
The neighborhood vector for user $v_i$, denoted by $\rva_i$, is the $i^{th}$ row of $\mL =  [\rva_1, \dots, \rva_N ]^T$.
The encoder is an MLP network $f_{\textsc{enc}} (\rva_i)$ which encodes $\rva_i$ into $\rvz_i$, while the decoder $f_{\textsc{dec}} (\rvz_i)$ strives to reconstruct $\rva_i$ from $\rvz_i$.
We introduce a re-weighting vector $\rvb_i = \{ b_{ij} \}_{j=1}^N$ where $b_{ij} = 1$ if $L_{ij} =0$ and $b_{ij} = \beta >1$ when $L_{ij} >0 $.
$\beta$ is a confidence parameter that re-weights the positive terms ($L_{ij} > 0$) to balance the unobserved $0's$ which far outnumber the observed links in real-world networks.
The generative process to obtain $\rva_i$ from $\rvz_i$ is given by:

\begin{equation}
\rva_i \sim p_\theta (\rva_i | \rvz_i) = \gN (f_{\textsc{dec}} (\rvz_i),  diag(\rvb_i))
\end{equation}
where $diag(\rvb_i)$ is a diagonal matrix with non-zero entries from vector $b_i$.
The corresponding Gaussian log-likelihood is given by:

\begin{equation}
\log p_{\theta} (\mA | \mZ) = \sum\limits_{i=1}^N \log p_{\theta} (\rva_i | \rvz_i)  =  \sum\limits_{i=1}^N \big\Vert \rvb_i \odot (\rva_i - f_{\textsc{dec}}(\rvz_i))\big\Vert^2 
\end{equation}
\item \textbf{GCN + Inner Product}: We use a Graph Convolutional Network (GCN) as the encoder and an inner product decoder that maps pairs of user embeddings to a binary indicator of link existence in $\gG$. The GCN network comprises multiple stacked graph convolutional layers to extract features from higher-order structural neighborhoods.
The input to a layer is 
a user feature (or embedding) matrix $X \in \sR^{N \times F}$ and a normalized adjacency matrix $\hat{\mA}$, where each GCN layer computes the function:

\begin{equation}
f_{\textsc{enc}} (\mA) = \sigma(\hat{\mA} \mX \mW) \hspace{10pt} \hat{\mA} = \mD^{-1/2} \mA \mD^{-1/2} + \mI_N
\end{equation}

where $\mX$ is an identity matrix encoding user identities. Each entry $A_{ij}$ of adjacency matrix $\mA$ is generated according to:

\begin{equation}
A_{ij} \sim p_\theta (A_{ij} | \rvz_i, \rvz_j) = Ber(\sigma(\rvz_i^T \rvz_j))
\end{equation}
Similar to above, we re-weight the positive entries of $\mA$ with a confidence parameter $\beta$. The logistic log-likelihood is given by:

\begin{equation}
\log p_{\theta} (\mA | \mZ) = \sum\limits_{(i,j) \in \gE} \beta \log (\sigma(\rvz_i^T \rvz_j)) + \sum\limits_{(i,j) \notin \gE} \log (1 - \sigma(\rvz_i^T \rvz_j))
\end{equation}
As an alternative to re-weighting positive entries, negative sampling~\cite{deepwalk} can scale this training objective to large-scale social networks.
\end{description}

\begin{algorithm}[t]
\caption{\infvae~ training with block coordinate ascent.}
\begin{algorithmic}[1]
\Require Social Network ($\gG$), Training episodes ($\sT$)
\Ensure MAP estimates of $\mV_S, \mV_R, \mV_{P}$ and parameters $\theta, \phi$.
\State Initialize latent variables from a standard normal distribution.
\State \textbf{Pre-training}: Train $f_{\textsc{dec}}(\gG|\mZ)$ and $f_{\textsc{enc}}(\mZ| \gG)$ on $\gG$ using a VAE with log-likelihood:
$$L^{\textsc{VAE}} =  \E_{\scriptscriptstyle q_{\phi}(\mZ | \gG)} \log p_\theta (\gG | \mZ) - \KL{(q_{\phi} (\mZ | \gG), p(\mZ))}$$
\While{\textit{not converged}}
\LeftComment \textbf{\textit{Optimize over social network $\gG$ }}
\For{each batch of users $\gU \subseteq \gV$}
\parState {Fix $\mV_S, \mV_R, \mV_{P}, G_{\textsc{diff}} (\cdot)$ and update weights of
$f_{\textsc{enc}} (\gG)$ and $f_{\textsc{dec}} (\mZ)$ using mini-batch gradient ascent (Equation~\ref{eqn:infvae_objective})}
\EndFor
\LeftComment \textbf{\textit{Optimize over diffusion episodes $\sT$ }}
\For {each batch of diffusion episodes $B\subseteq \sT$}
\parState {Fix $\mZ$, $f_{\textsc{enc}} (\gG)$, $f_{\textsc{dec}}(\mZ)$ and update $\mV_S, \mV_R, \mV_{P}$, and $G_{\textsc{diff}} (.)$ using mini-batch gradient ascent. (Equation~\ref{eqn:infvae_objective})}
\EndFor
\EndWhile
\end{algorithmic}
\label{alg:infvae_opt}
\end{algorithm}

\subsection{{Model Inference}}
The overall model training objective maximizes a lower bound on the marginal log likelihood, also named evidence lower bound (ELBO)~\cite{variational_inf}, which is given by:

\begin{align}
L_q &= \E_{q} [ \log p(\gG,\sT, \mV_S, \mV_R , \mV_{P}, \mZ) -   \log q(\mV_S, \mV_R, \mV_P, \mZ | \gG)] 
\label{eqn:infvae_likelihood}
\end{align}

Note that $L_q$ is a function of both generative ($\theta$) and variational ($\phi$) parameters.
However, an analytical computation of the expectation with respect to $q_\phi(\mZ |\gG)$ is intractable, while Monte Carlo sampling prevents gradient back-propagation to the neural parameters of $f_{\textsc{enc}}(\gG)$. 
With the reparametrization trick~\cite{vae}, we instead sample $\mathbf{\epsilon} \sim \gN(0, I_{N \times D})$ and form samples of $\mZ = \mu_{\phi} (\gG)  + \epsilon \odot \sigma_{\phi} (\gG)$.
This isolates the stochasticity during sampling and the gradient with respect to $\phi$ can be back-propagated through the sampled $\mZ$.

\subsubsection{\textbf{Optimization}}
Since bayesian learning methods to infer latent posterior distributions incur high computational costs, and considering our goal of making good diffusion predictions rather than explanations, we resort to MAP (Maximum A Posteriori) estimation.
Thus, we sample $\mZ$ from $q_{\phi}(\mZ|\gG)$ using point estimates for $\mV_S, \mV_R$ and $\mV_{P}$.
We maximize the joint log-likelihood with MAP estimates of latent variables $\mV_S, \mV_R, \mV_{P}$, inference and generative network parameters $\theta,\phi$, and observations $\sT$ and $\gG$, given hyper-parameters $\lambda_s, \lambda_r, \lambda_p$, as defined by:

\begin{align}
\gL^{ \textsc{MAP}} &= \E_{\scriptscriptstyle q_{\phi}}  [ \log p_\theta (\gG | \mZ)] - \KL{(q_{\phi}, p(\mZ))} + \sum\limits_{\scriptscriptstyle (I,C) \in \sT} \gL^{\textsc{diff}}_{I,C} \label{eqn:infvae_objective} \\ 
&- \sum\limits_{i=1}^N  \left( \frac{\lambda_s}{2} \E_{\scriptscriptstyle q_{\phi}} \Vert \rvv^s_i - \rvz_i \Vert^2 + \frac{\lambda_r}{2} \E_{\scriptscriptstyle q_{\phi}} \Vert \rvv^r_i - \rvz_i \Vert^2 + \frac{\lambda_p}{2} \Vert \rvv^p_i \Vert^2 \right) \notag
\end{align}

where $q_\phi$ is a shorthand for the variational distribution of network structure generation $q_{\phi}(\mZ | \gG)$, and $\E_{\scriptscriptstyle q_{\phi}(\mZ | \gG)} [\mZ]$ is equal to $\mu_\phi(\gG)$ output by the inference network.
To optimize this objective, we employ block coordinate ascent with two sets of variables, $\{ f_{\textsc{enc}}(G), f_{\textsc{dec}} (\mZ) \}$ and $\{ \mV_S, \mV_R, \mV_{P}, G_{\textsc{diff}}\}$. 
As illustrated in Alg~\ref{alg:infvae_opt}, each iteration of the algorithm proceeds in two steps, by alternating optimization over the social network and diffusion cascades.

\subsubsection{\textbf{Diffusion Prediction}}
Our goal is to design a deep generative model towards diffusion prediction in a social network.
Thus, after learning the (locally) optimal model parameters and MAP estimates of latent variables in our model~\infvae, we compute the likelihood of influencing a candidate user $v_j$ given a sequence of seed activations $I$ as:
\begin{equation}
p(v_j | I) =  \sigmoid( h_I^T \rvv^r_j)
\label{eqn:infvae_user_prediction}
\end{equation}

\subsubsection{\textbf{Complexity}}
The cost per iteration comprises two parts:
(a) optimizing over social network $\gG$ gives $O(|\gE| \cdot F^2 + |\gE| \cdot D)$ assuming GCN + Inner Product (b) optimizing over diffusion episodes is $O(|\sT| \cdot D \cdot N)$ where $F$ is the maximum layer dimension in $f_{\textsc{enc}}$. 
The overall complexity per iteration is $O(|\gE| \cdot F^2 + |\gE| \cdot D + |\sT| \cdot D \cdot N)$.

\subsection{{Model Objective Derivation}}
In this section, we start from the model likelihood and derive the final objective function (Equation 9) by following the steps of variational inference and MAP estimation.
The joint probability distribution of both the observed data and latent variables in our deep generative framework~\infvae~is given by:

\begin{align}
p(\gG, \sT, \mV_S, \mV_R, \mV_{P}, \mZ) &= p(\gG,  \sT | \mV_S, \mV_R, \mV_{P}, \mZ) p (\mV_S, \mV_R, \mV_{P}, \mZ | \gG, \sT)
\end{align}
The conditional likelihood of generating the observed data $ \{ \gG, \sT \}$ given latent variables $\{ \mV_S, \mV_R, \mV_{P}, \mZ \}$ is given by: 
\begin{align}
p(\gG,  \sT | \mV_S, \mV_R, \mV_{P}, \mZ) &= p(\gG |\mZ) p(\sT |  \mV_S, \mV_R, \mV_{P}) \\
&= p_{\theta}(\gG |\mZ) \times \prod_{\scriptscriptstyle (I,C) \in \sT} p_{\theta} (C | \mV_S, \mV_R, \mV_{P}, I)
\end{align}
The latent posterior distribution, given by $p(\mV_S ,\mV_R ,\mV_P , \mZ |\gG, \sT)$, can be factorized as:
\begin{align}
p(\mV_S ,\mV_R ,\mV_P , \mZ |\gG, \sT) = \prod_{i=1}^N p(\rv^s_i | \rvz_i) P(\rvv^t_i | \rvz_i) p(\rvv^{P}_i) p(\mZ  | \gG)
\label{eqn:posterior}
\end{align}
Due to the intractability of computing an analytical form for the posterior, we use variational inference to factorize the posterior with a mean-field approximation:
\begin{align}
q(\mV_S, \mV_R, \mV_P, \mZ | \gG) &=   q(\mV_S ) q(\mV_R) q(\mV_P) q(\mZ | \gG) \label{eqn:mean_field_expand} \\ \nonumber
& = \prod_{i=1}^N \left( q(\rvv^s_i | \phi^s_i) q(\rvv^r_i | \phi^r_i) q(\rvv^{p}_i | \phi^{p}_i) \right) q_{\phi}(\mZ | \gG)
\end{align}

where the variational distributions of latent variables $\mV_S, \mV_R,$ and $\mV_P$ follow normal distributions, parameterized by $\theta^s_i, \theta^r_i$, and $\theta^{p}_i \; \forall 1 \leq i \leq N$ respectively.
The variational distribution of $\mZ$, while being conditioned on $\gG$ is modeled through a normal distribution whose parameters are given by an inference network $f_{\textsc{enc}} (\gG)$%
Following the standard conventions of variational inference, the objective function to optimize, is given by the variation lower bound or evidence lower bound (ELBO):
\begin{align}
L_q =\; & \E_{q} [ \log p(\gG,\sT, \mV_S, \mV_R , \mV_{P}, \mZ) -   \log q(\mV_S, \mV_R, \mV_P, \mZ | \gG)] \\
 =\; & \E_{q} [ \log p_\theta(\gG | \mZ) + \log p_\theta(\sT | \mV_S, \mV_R , \mV_{P})  \nonumber\\
 & + \log p( \mV_S | \mZ) + \log p(\mV_R | \mZ)  +  \log p(\mV_{P}) + \log p(\mZ)]\\
 & - \E_{q} [\log q(\mV_S) + \log q(\mV_R) + \log q(\mV_P) +  \log q(\mZ) )]
 \end{align}
By factorizing the joint variational distribution into separate components, we obtain:
\begin{align}
L_q = \; &\E_{\scriptscriptstyle q_\phi(\mZ | \gG)} [ \log p_\theta(\gG | \mZ)] + \E_{\scriptscriptstyle q(\mV_S) q(\mV_R) q(\mV_{P})} [\log p_\theta(\sT | \mV_S, \mV_R , \mV_P)] \\ 
& +\E_{\scriptscriptstyle q(\mV_{S}) q_{\phi}(\mZ | \gG)} \log p(\mV_{S} | \mZ) +   \E_{\scriptscriptstyle q(\mV_{R}) q_{\phi}(\mZ | \gG)} \log p(\mV_{R} | \mZ) \\
& + \E_{\scriptscriptstyle q(\mV_{P})} \log p(\mV_{P})  +   \E_{\scriptscriptstyle q_{\phi}(Z | X)} \log p(\mZ)  -  \E_{\scriptscriptstyle q_{\phi}(\mZ | \gG)} \log q(\mZ)\\
& - [\E_{\scriptscriptstyle q(\mV_{S})} \log q(\mV_S) + \E_{\scriptscriptstyle q(\mV_{R})}  \log q(\mV_R) + \E_{\scriptscriptstyle q(\mV_{P})} \log q(\mV_P)  ]
\end{align}

Instead of conducting full variational inference to obtain the variational distributions of each latent variable, we instead 
use MAP estimates for the variational distributions $q(\mV_S), q(\mV_R)$, and $q(\mV_P)$ of latent variables $\mV_S, \mV_R$, and $\mV_P$.
Specifically, we denote the point estimates by $\{ \rvv^s_i, \rvv^r_i$, $\rvv^p_i \; \forall 1 \leq i \leq N\}$, which are the means of the normal distributions parameterized by  $\{ \theta^s_i, \theta^r_i$, $\theta^{p}_i \; \forall 1 \leq i \leq N \}$ respectively.
However, we retain the variational distribution $q(\mZ)$ for $\mZ$ since it is sampled from $\gN(0, I_{N \times D})$ and $f_{\textsc{enc}} (\gG)$ through the reparameterization trick.
Using these assumptions, we can simplify each of the terms in the above equation as follows:

\begin{align}
\mathcal{L}^{MAP} = \; &\E_{\scriptscriptstyle q_\phi(\mZ | \gG)} [ \log p_\theta(\gG | \mZ)] +  \sum\limits_{(I,C) \in \sT} \gL^{\textsc{diff}}_{I,C} \\
& + \sum\limits_{i=1}^N \E_{\scriptscriptstyle q_{\phi}(\mZ | \gG)} \log p(\rvv^s_i | \rvz_i) +  \sum\limits_{i=1}^N  \E_{\scriptscriptstyle q_{\phi}(\mZ | \gG)} \log p(\rvv^r_i | \rvz_i) \\
& +  \sum\limits_{i=1}^N \log p (\rvv^p_i) + \E_{\scriptscriptstyle q_{\phi}(Z | X)} \log \frac{p(\mZ)}{q(\mZ)} 
\end{align}

Here, it can be easily seen that $\E_{\scriptscriptstyle q(\mV_S) q(\mV_R) q(\mV_{P})} [\log p_\theta(\sT | \mV_S, \mV_R , \mV_P)]$ reduces to $\gL^{\textsc{diff}}_{I,C}$ under point-estimates for $\mV_S, \mV_R$ and $\mV_P$.
A key point to note is the absence of three terms: $\E_{\scriptscriptstyle q(\mV_{S})} \log q(\mV_S)$, $\E_{\scriptscriptstyle q(\mV_{R})} \log q(\mV_R)$ and $ \E_{\scriptscriptstyle q(\mV_{P})} \log q(\mV_P)$. These expectations integrate to 1, since the respective variational distributions reduce to point-estimates during MAP estimation.
On substituting the probability densities of remaining probability distributions, we get our final objective as:

\begin{align}
    \mathcal{L}^{MAP} = \; & \E_{\scriptscriptstyle q_\phi(\mZ | \gG)} [ \log p_\theta(\gG | \mZ)] +  \sum\limits_{(I,C) \in \sT} \gL^{\textsc{diff}}_{I,C} \\
     &-   \frac{\lambda_s}{2} \sum\limits_{i=1}^N  \E_{\scriptscriptstyle  q_\phi(\mZ | \gG)} \Vert \rvv^s_i - \rvz_i \Vert^2 - \frac{\lambda_r}{2}  \sum\limits_{i=1}^N  \E_{\scriptscriptstyle  q_\phi(\mZ | \gG)} \Vert \rvv^r_i - \rvz_i \Vert^2  \\
     &-  \frac{\lambda_p}{2} \sum\limits_{i=1}^N   \E_{\scriptscriptstyle  q_\phi(\mZ | \gG)}  \Vert \rvv^p_i \Vert^2  - \KL{(q_{\phi}(\mZ | \gG), p(\mZ))}
\end{align}
where $\KL{(x, y)}$ refers to the KL-divergence between probability distributions $x$ and $y$.

\section{Experiments}
\label{sec:infvae_experiments}

In this section, we present our experimental results on multiple datasets from real-world social networks and public Stack-Exchanges\footnote{\url{https://archive.org/details/stackexchange}}. We examine two popular social networking platforms, \textit{Digg} and \textit{Weibo} and three stack-exchange networks, \textit{Android}, \textit{Christianity} and Travel, to demonstrate the effectiveness of~\infvae~for diffusion prediction.
\begin{itemize}
\item \textbf{Digg}~\cite{digg}: A social news aggregation platform where users vote on news stories.
The sequence of votes on each news story constitutes a diffusion cascade, while the social network comprises friendship links among voters.
We retain only users who have voted on at least 40 stories.

\item \textbf{Weibo}~\cite{locality}: A Chinese micro-blogging platform,
where the social network consists of follower links, and cascades reflect re-tweeting behavior of users. %
We use the posts and social connections of the 5000 most popular users in our experiments.

\end{itemize}
\textbf{Stack-Exchanges:} Community Q\&A websites where users post questions and answers on a wide range of topics.
The inter-user knowledge-exchanges on various interaction channels (\textit{e.g.}, question, answer, comment, upvote, etc.), constitute the social network.
Cascades correspond to chronologically ordered series of posts associated with the same tag, \textit{e.g.}, ``google-pixel-2" on Android.
We choose three Stack-Exchanges, Android, Christianity and Travel, spanning diverse themes.
Dataset statistics are provided in Table~\ref{tab:infvae_dataset_stats}.

\subsection{{Baselines}}
We compare~\infvae~against state-of-the-art representation learning methods for diffusion prediction since they have been shown to significantly outperform classical approaches (\textit{e.g.}, IC and LT) and probabilistic generative models~\cite{topolstm, inf2vec}.

\begin{itemize}
    \item \textbf{CDK}~\cite{cdk}: an embedding method that models information spread as a heat diffusion process in the representation space of users.
    \item \textbf{Emb-IC}~\cite{wsdm16}: an embedded cascade model that generalizes IC to learn user representations from partial orders of user activations.
    \item \textbf{Inf2vec}~\cite{inf2vec}: an influence embedding method that combines local propagation structure and user co-occurrence in cascades. 
    \item \textbf{DeepDiffuse}~\cite{deepdiffuse}: an attention-based RNN that operates on just the sequence of previously influenced users, to predict diffusion.
    \item \textbf{CYAN-RNN}~\cite{cyanrnn}: a sequence-based RNN that uses an attention mechanism to capture cross-dependence among seed users.
    \item \textbf{SNIDSA}~\cite{cikm18_attention}: an RNN-based model to compute structure attention over the local propagation structure of a cascade.    
    \item \textbf{Topo-LSTM}~\cite{topolstm}: a recurrent model that exploits the local propagation structure of a cascade through a dynamic DAG-LSTM.
\end{itemize}

\begin{table}[t]
\centering
\small
\begin{tabular}{@{}p{0.24\linewidth}K{0.08\linewidth}K{0.08\linewidth}K{0.11\linewidth}K{0.14\linewidth}K{0.13\linewidth}@{}}
\toprule
\multirow{2}{*} & \multicolumn{2}{c}{\small \textbf{Social Networks}}  & \multicolumn{3}{c}{\small \textbf{Stack-Exchange Networks}}\\
\cmidrule(lr){2-3} \cmidrule(lr){4-6}
\textbf{Dataset} & \small \textbf{Digg}  & \small \textbf{Weibo} & \small \textbf{Android} & \small\textbf{Christianity}  & \small\textbf{Travel}\\
\midrule
\textbf{\small \# Users} & 8,602 & 5,000 & 9,958 & 2,897 & 8,726\\
\textbf{\small \# Links} & 173,489 & 123,691 & 48,573 &  35,624 &  76,555\\
\textbf{\small \# Cascades} & 968 & 23,475 & 679 &  589 & 711\\
\textbf{\small Avg. cascade len} & 100.0 & 23.6 & 33.3 &  22.9 & 26.8\\
\bottomrule
\end{tabular}
\caption{Statistics of datasets used in our experiments}
\label{tab:infvae_dataset_stats}
\end{table}

\subsection{{Experimental Setup}}
We denote our two model variants with GCN and MLP  architectures, by ~\infvae+\textsc{GCN} and ~\infvae+\textsc{MLP} respectively.
We randomly sample 70\% of the cascades for training, 10\% for validation and remaining 20\% for testing.
We consider the task of predicting the set of all influenced users as a retrieval problem~\cite{topolstm,wsdm16,cyanrnn,inf2vec}.
The fraction of users sampled from each test cascade to serve as the seed set is defined as \textit{\textbf{seed set percentage}}, which is varied from 10\% to 50\% to create a large evaluation test-bed spanning diverse cascade lengths.
The likelihood of influencing an inactive user determines its rank (Equation~\ref{eqn:infvae_user_prediction}).
We use MAP$@K$ (Mean Average Precision) and Recall$@K$ as evaluation metrics. Note that MAP$@K$ considers both the \textit{existence} and \textit{position} of ground-truth target users in the rank list, while Recall$@K$ only reports occurrence within top-$K$ ranks.

Hyper-parameters are tuned by evaluating MAP@10 on the validation set.
Since Emb-IC generalizes IC, we use 1000 Monte Carlo simulations to estimate influence probabilities.
Since the recurrent neural models (\textit{e.g.,} Topo-LSTM) are trained for next user prediction, 
we use the ranking induced by user activation probabilities for diffusion prediction, which we found to significantly outperform a similar simulation approach.
For Inf2vec, we examine several seed influence aggregation functions (Ave, Sum, Max, and Latest) to report the best results.
Our reported results are averaged over 10 independent runs with different random weight initializations. Our implementation of~\infvae~is publicly available\footnote{\url{https://github.com/aravindsankar28/Inf-VAE}}.

\subsection{{Evaluation Metric Definitions}}
\label{sec:infvae_metrics}
We consider the task of predicting the set of all influenced users as a retrieval problem due to the large number of potential targets~\cite{topolstm,wsdm16,cyanrnn,inf2vec}. The likelihood of influencing a user is used as a ranking score for evaluation (Equation~\ref{eqn:infvae_user_prediction}).
We use MAP$@K$ (Mean Average Precision) as our primary metric to evaluate diffusion prediction performance.\\

\textbf{Mean Average Precision (MAP):}
Average precision (AP) is a ranked precision metric that gives larger credit to correctly predicted users in top ranks. Given a ranked list with $K$ user predictions, $AP@K$ is defined as:

\begin{equation}
 \text{AP@K}  =  \frac{\sum\limits_{k=1}^K P(k) \times rel(k)}{ \# \text{ Influenced Users}}
\end{equation}

where $P(k)$ is the precision at cut-off $k$ in the top-$K$ list, and $rel(k)$ is an indicator function equaling 1 if the user at rank $k$ has been influenced, otherwise 0.
Finally, $MAP@K$ is defined as the mean of the AP scores over all diffusion episodes.

\textbf{Recall:}
Recall measures the fraction of correctly predicted influenced users within the top-$K$ ranks.
\begin{equation}
 \text{Recall@K} = \frac{ \sum\limits_{k=1}^K rel(k)}{\# \text{Influenced Users}}
\end{equation}

\renewcommand*{\factor}{0.075}
\begin{table}[t]
\centering
\small
\begin{tabular}{@{}p{0.25\linewidth}K{\factor\linewidth}K{\factor\linewidth}K{\factor\linewidth}K{\factor\linewidth}K{\factor\linewidth}K{\factor\linewidth}@{}} \\
\toprule
\multirow{1}{*}{\textbf{Method}} & \multicolumn{3}{c}{\textbf{Digg}}  &   \multicolumn{3}{c}{\textbf{Weibo} } \\
\cmidrule(lr){2-4} \cmidrule(lr){5-7} 
\multirow{1}{*}{\textbf{MAP}} & \textbf{@10} & \textbf{@50} & \textbf{@100}& \textbf{@10} & \textbf{@50} & \textbf{@100}\\
\midrule
\textbf{CDK} &  0.0437 & 0.0222 & 0.0228 & 0.0130 & 0.0106 & 0.0123 \\
\textbf{Emb-IC} & 0.0862 & 0.0431 & 0.0431 & 0.0140 & 0.0116 & 0.0131 \\
\textbf{Inf2vec} & 0.1189 & 0.0554 & 0.0546 & 0.0156 & 0.0103 & 0.0121 \\
\textbf{DeepDiffuse} & 0.0919 & 0.0460 & 0.0471 & 0.0291 & 0.0186 & 0.0213 \\
\textbf{CYAN-RNN} & 0.1188 & 0.0479 & 0.0427 & 0.0296 & 0.0207 & 0.0234 \\
\textbf{SNIDSA} & 0.0941 & 0.0363 & 0.0348 & 0.0224 & 0.0146 & 0.0169    \\
\textbf{Topo-LSTM} & 0.1193 & 0.0577 & 0.0587 & 0.0325 & 0.0226 & 0.0247  \\
\midrule
\textbf{Inf-VAE+MLP} & 0.1587 & 0.0774 & 0.0719 & 0.0322 & 0.0211 & 0.0234  \\
\textbf{Inf-VAE+GCN} & \textbf{0.1642} & \textbf{0.0779} & \textbf{0.0724} & \textbf{0.0373} & \textbf{0.0230} & \textbf{0.0257}  \\
\bottomrule
\end{tabular}
\caption{Experimental results for diffusion prediction on two social network datasets ($MAP@K$ scores for $K = 10, 50$ and $100$), the \textit{seed set percentage} varies in the range to 10 to 50\% users in each test cascade.~\infvae~achieves 26\% relative gains in MAP@10 (on average) over the best baseline}
\label{tab:infvae_social_results}
\end{table}

\renewcommand*{\factor}{0.056}
\begin{table}[t]
\centering
\footnotesize
\begin{tabular}{@{}p{0.17\linewidth}K{\factor\linewidth}K{\factor\linewidth}K{\factor\linewidth}K{\factor\linewidth}K{\factor\linewidth}K{\factor\linewidth}K{\factor\linewidth}K{\factor\linewidth}K{\factor\linewidth}@{}} \\
\toprule
\multirow{1}{*}{\textbf{Method}} &  \multicolumn{3}{c}{\textbf{Android}}  & \multicolumn{3}{c}{\textbf{Christianity}} & \multicolumn{3}{c}{\textbf{Travel}} \\
\cmidrule(lr){2-4} \cmidrule(lr){5-7} \cmidrule(lr){8-10}
\multirow{1}{*}{\textbf{MAP}}  & \textbf{@10} & \textbf{@50} & \textbf{@100}& \textbf{@10} & \textbf{@50} & \textbf{@100}& \textbf{@10} & \textbf{@50} & \textbf{@100}\\
\midrule
\textbf{CDK}  & 0.0319 & 0.0121 & 0.0125 & 0.0876 & 0.0531 & 0.0578 & 0.0650 & 0.0333 & 0.0341 \\
\textbf{Emb-IC}  & 0.0505 & 0.0248 & 0.0267 & 0.1340 & 0.0905 & 0.0962 & 0.0924 & 0.0584 & 0.0609 \\
\textbf{Inf2vec} & 0.0412 & 0.0141 & 0.0150 & 0.1824 & 0.0790 & 0.0852 & 0.1245 & 0.0495 & 0.0529\\
\textbf{DeepDiffuse} & 0.0437 & 0.0228 & 0.0250 & 0.1632 & 0.0828 & 0.0831 & 0.1220 & 0.0675 & 0.0693  \\
\textbf{CYAN-RNN}  & 0.0520 & 0.0276 & 0.0296 & 0.1971 & 0.1229 & 0.1304 & 0.1551 & 0.0791 & 0.0799\\
\textbf{SNIDSA} & 0.0397 & 0.0207 & 0.0222 & 0.1233 & 0.0699 & 0.0781 & 0.0857 & 0.0562 & 0.0585   \\
\textbf{Topo-LSTM}  & 0.0595 & 0.0283 & 0.0289 & 0.1811 & 0.0989 & 0.0991 & 0.1393 & 0.0773 & 0.0783 \\
\midrule
\textbf{Inf-VAE+MLP} & 0.0584 & 0.0272 & 0.0285 & 0.2549 & 0.1355 & 0.1402 & 0.1865 & 0.0897 & \textbf{0.0913} \\
\textbf{Inf-VAE+GCN} & \textbf{0.0601} & \textbf{0.0290} & \textbf{0.0304} & \textbf{0.2594} & \textbf{0.1413} & \textbf{0.1461} & \textbf{0.1924} & \textbf{0.0906} & 0.0910 \\
\bottomrule
\end{tabular}
\caption{Experimental results for diffusion prediction on three stack-exchange datasets ($MAP@K$ scores for $K = 10, 50$ and $100$), the \textit{seed set percentage} varies in the range to 10 to 50\% users in each test cascade.~\infvae~achieves 16\% relative gains in MAP@10 (on average). over the best baseline.}
\label{tab:infvae_stack_results}
\end{table}

\subsection{{Experimental Results}}
We note the following key observations from our experimental results comparing~\infvae~against competing baselines (Tables~\ref{tab:infvae_social_results} and~\ref{tab:infvae_stack_results}).

Methods that do not explicitly model sequential activation orders (\textit{e.g.}, CDK and Emb-IC), perform markedly worse than their counterparts.
Modeling local projected cascade structures with neural recurrent models results in improvements (\textit{e.g.}, Topo-LSTM and others).
Jointly modeling social homophily derived from global network structure and temporal influence by our model 
~\infvae~yields significant relative gains of 22\% ($MAP@10$) on average across all datasets.
~\infvae+GCN consistently beats the MLP variant, validating the power of graph convolutional networks in effectively propagating higher-order local neighborhood features.

\begin{figure}[t]
    \centering
    \includegraphics[width=0.9\linewidth]{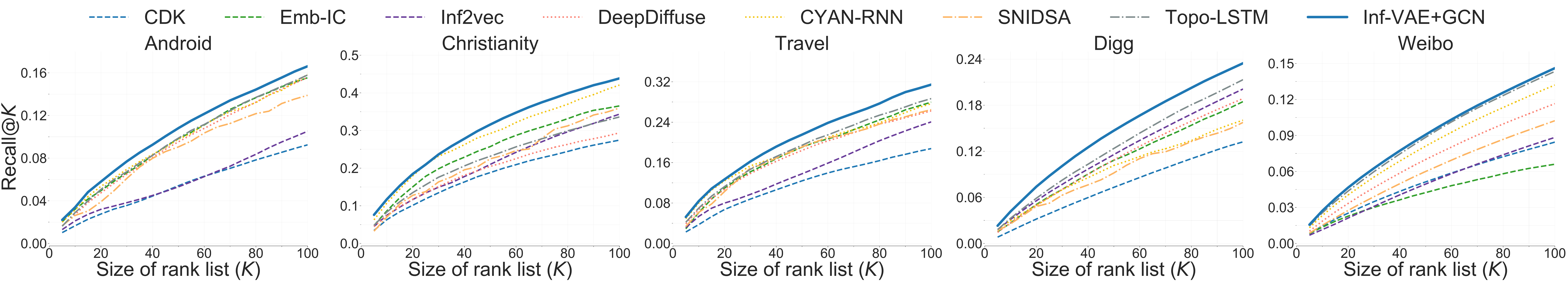}
    \caption{Experimental results for diffusion prediction on 5 datasets, Recall$@K$ scores on varying size of the rank list $K$}
    \label{fig:infvae_recall}
\end{figure}

Figure~\ref{fig:infvae_recall} depicts the variation in recall with size of rank list $K$.
As expected, recall increases with $K$, however, the relative differences across methods is much smaller.
~\infvae~consistently outperforms baselines across a wide range of $K$ values.
For instance, 
the Christianity dataset has seed sets with 2-10 users, and corresponding target sets with 10-15 users out of a possible 3000.
Here, a recall$@100$ of 0.45 for~\infvae~is quite impressive, especially considering the absence of explicit re-share links and the noise associated with real-world diffusion processes.
We restrict our remaining analyses to~\infvae+GCN~since it consistently beats the MLP variant.

\subsection{{Impact of Social and Interaction Sparsity}}
In this section, we analyze the benefits of explicitly modeling social homophily through VAEs, compared to the best baseline (Topo-LSTM) that only considers local propagation structures.
\begin{description}
\item \textbf{{Users with sparse diffusion activities.}}
We divide users into quartiles by their \textit{activity levels}, which is the number of participating cascades per user. 
We evaluate \textit{target recall}$@100$ for each user $u$, defined as the fraction of times $u$ was predicted correctly within top-$100$ ranks.
In Figure \ref{fig:infvae_social_activity}(a), we depict 
both recall scores and relative performance gains of~\infvae~(over Topo-LSTM) across diffusion activity quartiles.

While target recall increases with activity levels,~\infvae~significantly improves performance for inactive users (quartiles Q1-Q3).
Thus, modeling social homophily through VAEs contributes to massive gains for users with \textit{sparse diffusion activities}.
Interestingly, Topo-LSTM performs comparably on the most active users (quartile Q4), which indicates the potential of pure sequential modeling techniques for highly active users.
\begin{figure}[t]
    \centering
    \includegraphics[trim={0 0 0 0.8cm},clip,width=0.9\linewidth]{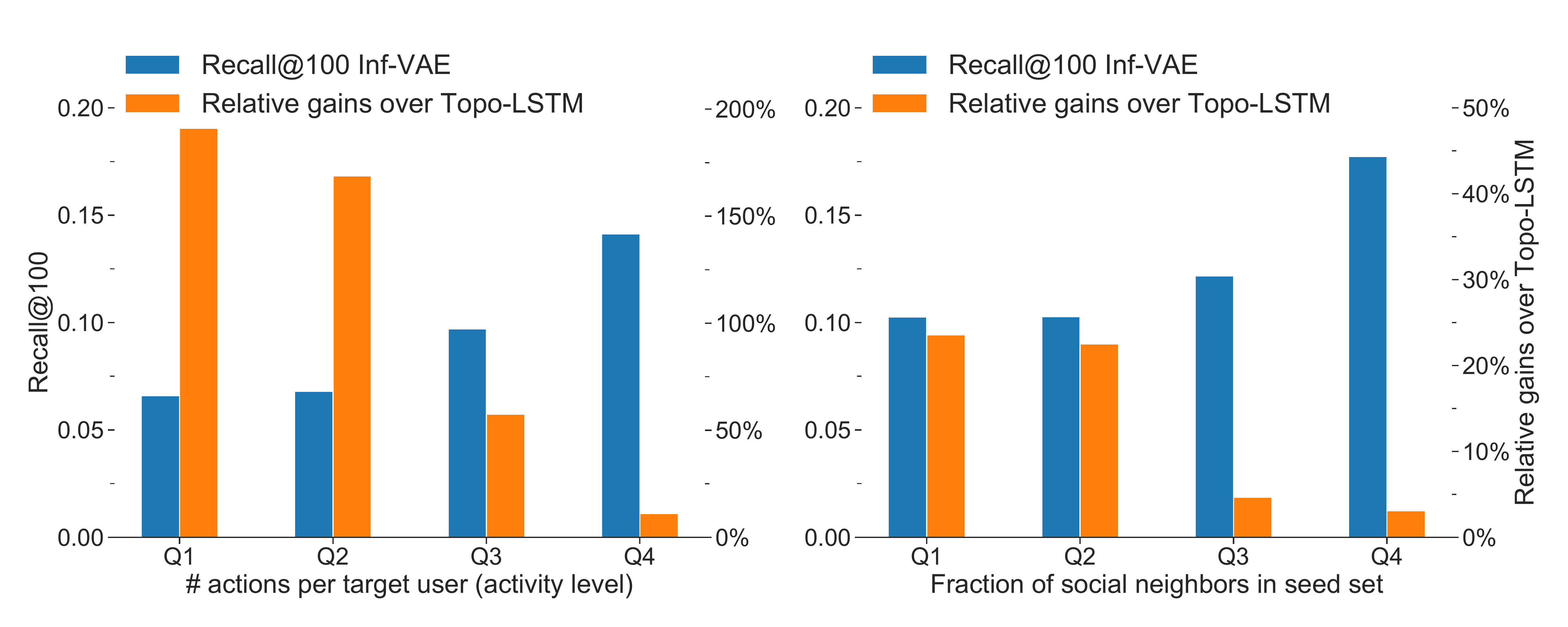}
    \caption{Performance across user quartiles on \textit{diffusion activity level}, and \textit{seed neighbor fraction} (Q1: lowest, Q4: highest).~\infvae~has higher gains for users with sparse activities and lacking direct neighbors in seed sets (quartiles Q1-Q3).}
    \label{fig:infvae_social_activity}
\end{figure}

\item  \textbf{Users that lack direct social connections in seed sets.}
We separate users into quartiles by \textit{seed neighbor fraction}, which is computed as the fraction of seed users that are direct social neighbors, averaged over the training examples. We similarly report target recall$@K$ and relative gains across quartiles (Figure~\ref{fig:infvae_social_activity}(b)).

As expected, performance increases with seed neighbor fraction. Note higher relative gains over Topo-LSTM for users that lack direct neighbors in the seed set (quartiles Q1-Q3). 
This demonstrates the ability of~\infvae~to implicitly regularize seed user representations based on higher-order social neighborhoods captured by GCN-based autoencoders.
Again, we find that local sequential models suffice for users with 
large seed neighbor fractions, as evidenced by the results of Topo-LSTM in quartile Q4.

\end{description}

\subsection{{Model Analysis}}
In this section, we first present an \textit{ablation study} on the architectural design choices in~\infvae, followed by a sensitivity analysis on \textit{seed set percentage} and \textit{hyper-parameters}.

\subsubsection{\textbf{Ablation Study}}
We analyze model design choices including modeling social homophily via VAEs and co-attention to capture co-variance of homophily with infuence, on Android and Weibo datasets.

\noindent \textbf{Social Homophily:} We examine ways to condition the sender $\mV_S$ and receiver $\mV_R$ variables on $\mZ$:

\begin{enumerate}
    \item $\mV_S$ and $\mV_R$ are identical and are conditioned on $\mZ$ through hyper-parameter $\lambda_s (=\lambda_r)$, \textit{i.e.}, $\mV_S = \mV_R \not \independent \mZ$ (note that this is different from setting $\lambda_s = \lambda_r$ without enforcing $\mV_S=\mV_R$).
    \item $\mV_S$ is a free variable conditionally independent of $\mZ$, \textit{i.e.}, $\mV_S \independent \mZ$, which is equivalent to setting $\lambda_s=0$.
    \item $\mV_R$ is a free variable, \textit{i.e.}, $\mV_R \independent \mZ$, which is the inverse of (3).
    \item $\mV_S$ and $\mV_R$ are both free variables conditionally independent of $\mZ$ ($\lambda_s= \lambda_r =0$), \textit{i.e.}, $\mV_S \independent \mZ, \mV_R \independent \mZ$.
\end{enumerate}

\noindent Independent conditioning of $\mV_S$ and $\mV_R$ on $\mZ$ (default) achieves best results. %
Enforcing $\mV_S=\mV_R$ (row 1) is clearly inferior,
which validates the choice of differentiating user roles.
Notably, allowing $\mV_S$ to be a free variable 
results in minor performance degradation (row 2), while the drop is significant when $\mV_R$ is independent of $\mZ$ (row 3).
As expected, setting both $\mV_S$ and $\mV_R$ as free variables (row 4),
performs the worst due to lack of social homophily signals.

\renewcommand*{\factor}{0.071}
\begin{table}[ht]
\centering
\small
\begin{tabular}{@{}p{0.31\linewidth}K{\factor\linewidth}K{\factor\linewidth}K{\factor\linewidth}K{\factor\linewidth}K{\factor\linewidth}K{\factor\linewidth}K{\factor\linewidth}K{\factor\linewidth}@{}} \\
\toprule
{\textbf{Metric}} &  \multicolumn{3}{c}{\textbf{Weibo} } & \multicolumn{3}{c}{\textbf{Android}} \\
\cmidrule(lr){2-4} \cmidrule(lr){5-7}
MAP & @10 & @50 & @100 & @10 & @50 & @100 \\
\midrule
(0) Default  & \textbf{0.0373} & \textbf{0.0230} & \textbf{0.0257} & \textbf{0.0601} & \textbf{0.0290} & \textbf{0.0304}\\
(1)$\mV_S = \mV_R \not \independent \mZ$ & 0.0353 & 0.0220 & 0.0248 & 0.0558 & 0.0275 & 0.0287\\
(2)$\mV_S \independent \mZ$ & 0.0351 & 0.0213 & 0.0240 & 0.0595 & 0.0285 & 0.0301\\
(3)$\mV_R \independent \mZ$ & 0.0326 & 0.0217 & 0.0241 & 0.0567 & 0.0276 & 0.0291\\
(4)$\mV_S \independent \mZ, \mV_R \independent \mZ$ & 0.0313 & 0.0205 & 0.0235 & 0.0542 & 0.0274 & 0.0289\\
\midrule
(5) Remove Coattention & 0.0307 & 0.0207 & 0.0233 & 0.0553 & 0.0270 & 0.0284 \\ 
(6) Separate Attentions & 0.0293 & 0.0217 & 0.0192 & 0.0570 & 0.0277 & 0.0291 \\
\midrule
(7) Static-Pretrain & 0.0342 & 0.0203 & 0.0226 & 0.0606 & 0.0281 & 0.0292 \\
\bottomrule
\end{tabular}
\caption{Ablation study on architecture design ($MAP@K$ scores for $K = 10, 50, 100$), $\independent$ denotes variable independence}
\label{tab:infvae_ablation_results}
\end{table}

\noindent \textbf{Co-attention:}
We conduct two ablation studies defined by:

\begin{enumerate}
\setcounter{enumi}{4}
    \item Replace co-attention with meanpool over concatenated sender and temporal influence vectors, followed by a dense layer.
    \item Replace co-attention with two separate attentions on the sender and temporal influence sequences, followed by concatenation.
\end{enumerate}

Learning co-attentional weights (default) consistently outperforms mean pooling (5), illustrating the benefits of assigning variable contributions to seed users. 
Using separate attentions (6) significantly deteriorates results, which indicates 
the existence of complex non-linear correlations between the social and temporal latent factors.

\noindent \textbf{Joint Training:} In (2), we replace joint block-coordinate optimization (Alg~\ref{alg:opt}) with a single step over cascades with pre-trained user embeddings (line 2), \textit{i.e.}, $\mZ$ is not updated based on cascades.

\noindent 
Joint training is beneficial when social interactions are noisy (\textit{e.g.,} Weibo) in comparison to focused stack-exchanges such as Android.

\begin{figure}[t]
    \centering
    \includegraphics[width=0.9\linewidth]{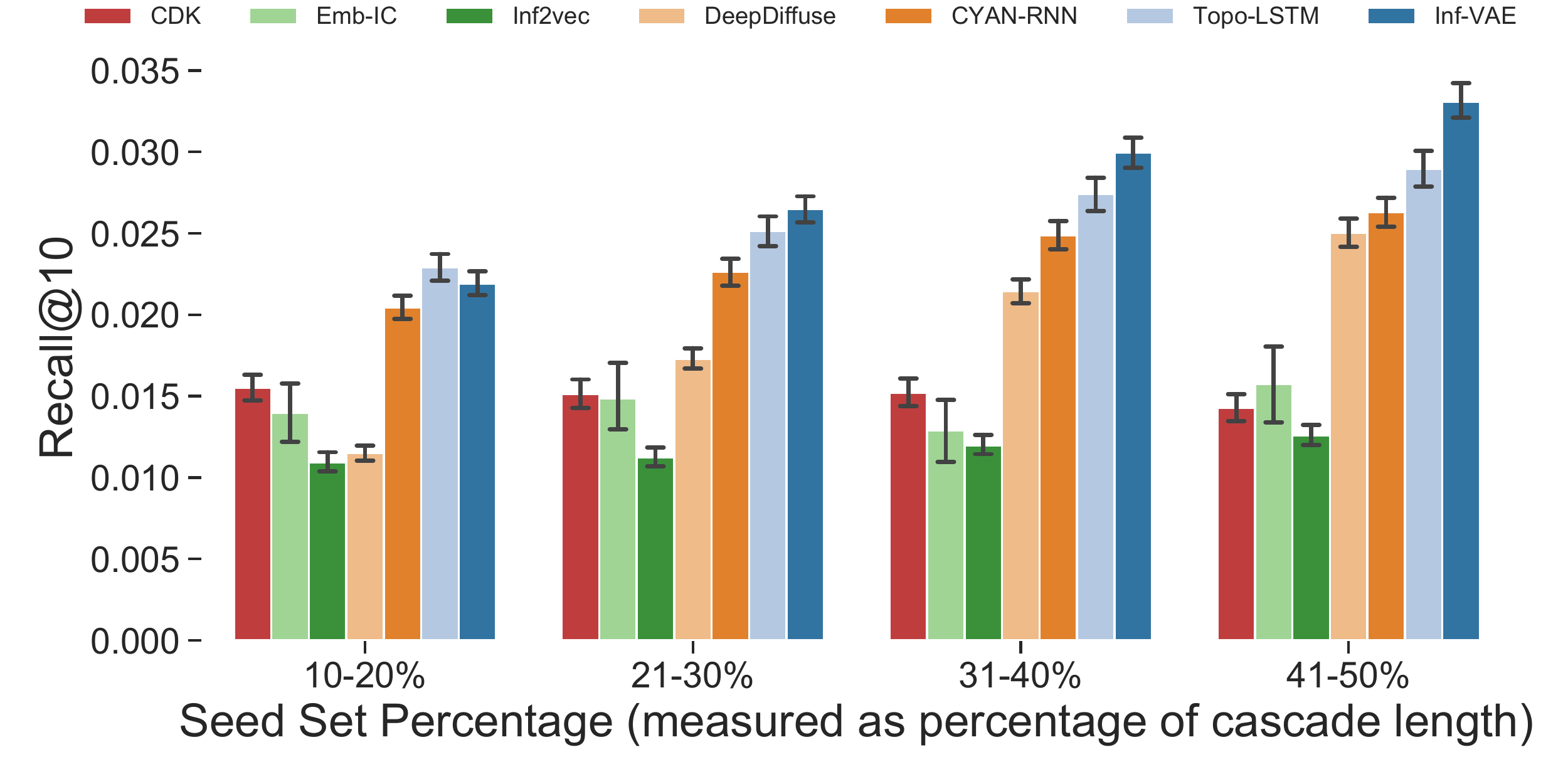}

    \caption{Impact of seed set percentage in Weibo.~\infvae~achieves higher gains 
    for larger seed set fractions.}
    \label{fig:infvae_seed_analysis}

\end{figure}

\subsubsection{\textbf{Impact of Seed Set Percentage.}}
We divide the test set into quartiles based on \textit{seed set percentage}, and report performance per quartile.
Since we require a sizable number of test examples per quartile to obtain unbiased estimates, we use the Weibo dataset. %

Figure~\ref{fig:infvae_seed_analysis} depicts Recall$@10$ scores in different ranges. %
First, recall scores increase with seed set percentage since larger seed sets enable better model predictions; and target set size reduces with increase in seed set percentage.
Second, relative gains of~\infvae~over baselines increase with seed set percentage. %
This highlights the capability of co-attention in focusing on relevant users based on both social homophily and temporal influence factors.

\subsubsection{\textbf{Impact of $\lambda_s$ and $\lambda_r$}}
\begin{figure}[t]
    \centering
    \includegraphics[width=0.9\linewidth]{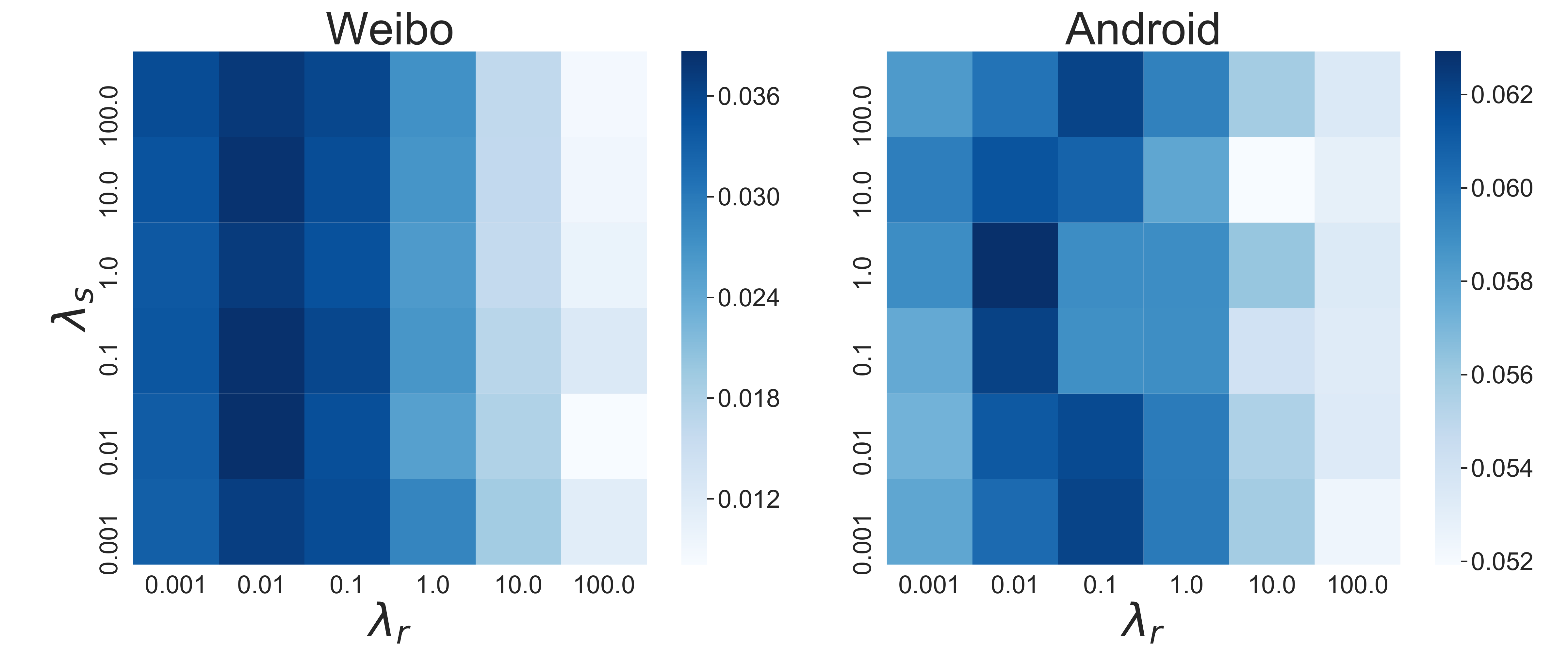}
    \caption{$MAP@10$ on varying $\lambda_s$, $\lambda_r$ over Android and Weibo. Performance is more sensitive to variations in $\lambda_r$ than $\lambda_s$.}
    \label{fig:infvae_lambda_heatmap}
\end{figure}

Hyper-parameters $\lambda_s$ and $\lambda_r$ control the degree of dependence of the sender and receiver variables $\mV_S, \mV_R$ on the social variables $\mZ$.
Figure~\ref{fig:infvae_lambda_heatmap} depicts performance ($MAP@10$) on Android and Weibo datasets.
The performance is sensitive to variations in $\lambda_r$ with best values around 0.01 and 0.1,
while $\lambda_s$ results in minimal variations.
Furthermore, the best values of $\lambda_s, \lambda_r$ are stable in a broad range of values that transfer across datasets, indicating that~\infvae~requires minimal tuning in practice.
Since $\lambda_p$ has minimal performance impact, we exclude it from our analysis.

\begin{figure}[t]
    \centering
        \subfigure[b][Runtime on Weibo and Android
        ]{
        \centering
        \includegraphics[width=0.41\linewidth]{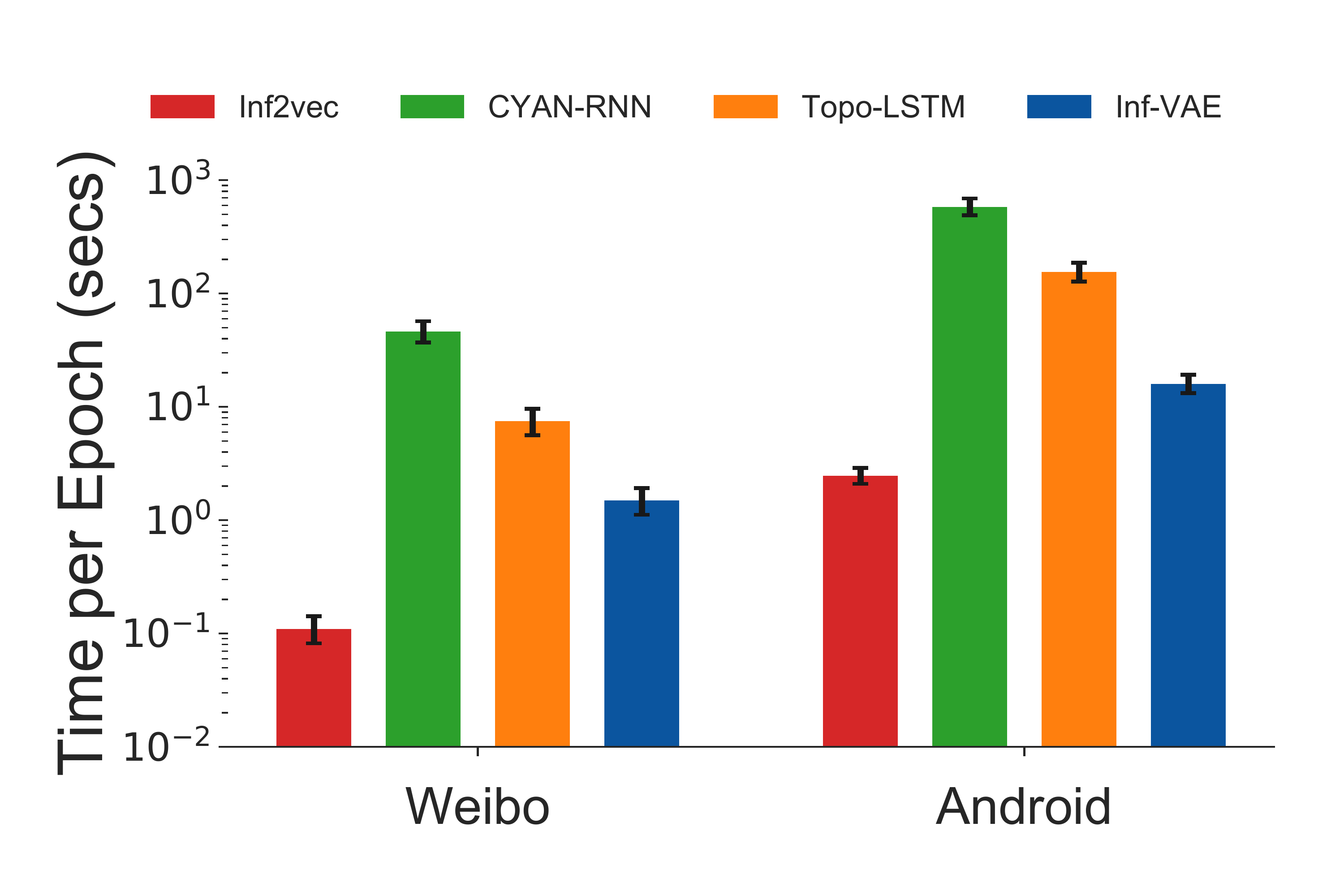}
        \label{fig:infvae_runtime}
        }
        \hspace{2pt}
        \subfigure[b][Scalability on synthetic dataset]{
        \centering
        \includegraphics[width=0.41\linewidth]{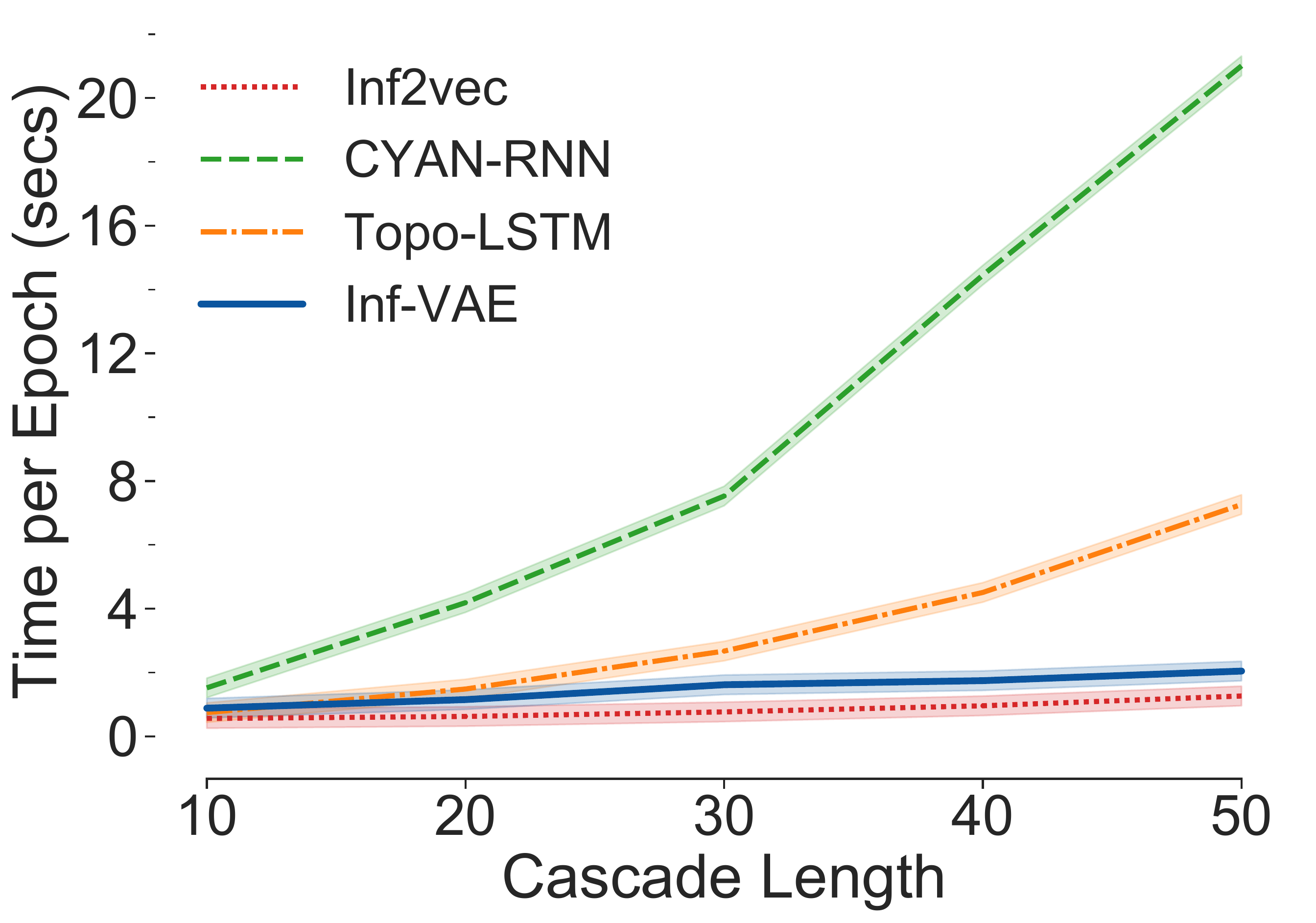}
        \label{fig:infvae_scalability} }        
    \caption{Running time and scalability comparison of~\infvae~with several baselines.~\infvae~is faster than recurrent models (Topo-LSTM, CYANRNN) by an order of magnitude.}
\end{figure}
\subsubsection{\textbf{Runtime Analysis}.} 
In our empirical runtime analysis experiments, all diffusion prediction methods converge within 50 epochs with similar convergence rates.
Thus, for the sake of brevity, we only compare model runtime per training epoch, which includes one optimization step over the social network and diffusion cascades for our model~\infvae.

From figure~\ref{fig:infvae_runtime}, Inf2vec is the fastest while our model~\infvae~ comes second.
Thus,~\infvae~achieves a good trade-off between 
expensive recurrent neural network models (\textit{e.g.}, Topo-LSTM) and simpler embedding methods (\textit{e.g.}, Inf2vec), while achieving consistently superior performance over all of them.
\subsubsection{\textbf{Scalability Analysis}.} We analyze scalability on cascade sequences of varying lengths.
Since real-world datasets possess heavily biased length distributions, we instead 
synthetically generate a 
Barabasi-Albert~\cite{barabasi} network of 2000 users and simulate diffusion cascades using an IC model.
We compare training times per epoch for each cascade length ($l$) in the range of 10 to 50.

Figure~\ref{fig:infvae_scalability} depicts
linear scaling for~\infvae~and Inf2vec \textit{wrt} cascade length.
Recurrent methods scale poorly due to the  sequential nature of back-propagation through time (BPTT), resulting in prohibitive costs for long cascade sequences.
On the other hand,~\infvae~avoids BPTT through efficient parallelizable co-attentions.

\section{Acknowledgements}
Research was sponsored in part by U.S. Army Research Lab. under Cooperative Agreement No. W911NF-09-2-0053 (NSCTA), DARPA under Agreements No. W911NF-17-C-0099 and FA8750-19-2-1004, National Science Foundation IIS 16-18481, IIS 17-04532, and IIS-17-41317, and DTRA HDTRA11810026.
\section{Conclusion}
\label{sec:infvae_conclusion}
In this chapter, we present a novel variational autoencoder framework (\infvae) to jointly embed homophily and influence in diffusion prediction.
Given a sequence of seed user activations,~\infvae~employs an expressive co-attentive fusion mechanism to jointly attend over their social and temporal variables, capturing complex correlations.
Our experimental results on two social networks and three stack-exchanges indicate significant gains over state-of-the-art methods.

In future, ~\infvae~can be extended to include multi-faceted user attributes owing to the generalizable nature of our VAE framework. While the current implementation employs GCN networks, we foresee direct extensions with neighborhood sampling~\cite{graphsage} to enable scalability to social networks with millions of users. 
We also plan to explore neural point processes to predict user activation times. 
Finally, similar frameworks may be examined for joint temporal co-evolution of social network and diffusion cascades.

In this chapter, we have examined our first inductive learning application that models the spread (or diffusion) of user-generated content in social networking platforms. 
Here, our proposed framework effectively overcomes interaction sparsity challenges for the vast population of users who seldom post content (sparse diffusion actions). 
In the next chapter, we explore a new social interaction setting in modern social platforms, involving group interactions; we focus on personalized item recommendations for ephemeral groups with limited or no historical interactions together.

\chapter{\textsc{GroupIM}: Self-supervised Ephemeral Group Interaction Modeling}
\label{chap:groupim}

\section{Introduction}
\label{sec:groupim_intro}

We address the problem of recommending items to \textit{ephemeral}
groups, which comprise users who purchase \textit{very few (or no)} items together~\cite{ephemeral}.
The problem is ubiquitous, and appears in a variety of familiar contexts, \textit{e.g.}, dining with strangers, watching movies with new friends, and attending social events.
We illustrate key challenges with an example: 
Alice (who loves Mexican food) is taking a visitor Bob (who loves Italian food) to lunch along with her colleagues, 
where will they go to lunch? 
There are three things to note here:
first, the group is \textit{ephemeral}, since there is \textit{no}  historical interaction observed for this group.
Second, 
\textit{individual preferences may depend on other group members}.
In this case, the group may go to a fine-dining Italian restaurant. 
However, when Alice is with other friends, they may go to
Mexican restaurants.
Third, groups comprise users with \textit{diverse individual preferences}, and thus the group recommender needs to be cognizant of individual preferences.

Prior work primarily target \textit{persistent} groups which refer to fixed, stable groups where members have interacted with numerous items together \textit{as a group} (\textit{e.g.}, families watching movies).
They mainly
fall into two categories: heuristic pre-defined aggregation (\textit{e.g.}, least misery~\cite{least_misery}) %
that disregards group interactions; 
data-driven strategies such as
probabilistic models~\cite{com, crowdrec} and neural preference aggregators~\cite{agree, agr}.
A key weakness is that these methods either ignore individual user activities~\cite{agr, wang2019group} or assume that users have the same likelihood to follow individual and collective preferences, across different groups~\cite{com, crowdrec, agree}.
Lack of expressive power to distinguish the role of individual preferences across groups results in degenerate solutions for sparse ephemeral groups. %
A few related methods exploit external side information in the form of a social network~\cite{delic2018use, socialgroup,SoAGREE}, user personality traits~\cite{zheng2018identifying} and demographics~\cite{delic2018observational}, for group decision making.
However, such auxiliary side information may often be unavailable.

We train robust ephemeral group recommenders without resorting to any extra side information.
Two observations help: first, while groups are ephemeral, group members may have rich individual interaction histories; this can alleviate group interaction sparsity.
Second, since groups are ephemeral with sparse training interactions, base group recommenders need reliable guidance to learn informative (non-degenerate) group representations,
but the guidance needs to be data-driven and learnable, rather than a heuristic.

\textbf{Present Work:} To overcome group interaction sparsity, we propose two key insights:
First, we propose a self-supervised learning approach to utilize the intrinsic structure in observed group interactions for model training; we exploit the \textit{preference covariance} amongst individuals who are in the same group, to design an auxiliary training objective that enhances the informativeness of user and group representations.
Second, we learn robust estimates of the \textit{contextual relevance} of users' individual preferences in each group and present a novel contextual regularization objective to incorporate knowledge of individual preferences.

We realize the above insights through two novel data-driven model training strategies for group recommendation. \textit{First}, we introduce a \textit{self-supervised learning} objective to contrastively regularize the user-group latent space to capture social user associations and distinctions across groups.
We achieve this by 
maximizing \textit{mutual information} (MI) between representations of groups and group members,
which encourages group representations to encode shared group member preferences while regularizing user representations to capture their social associations.
\textit{Second}, we contextually identify \textit{informative} group members and regularize the corresponding group representation to reflect their personal preferences.
We introduce a novel regularization objective that \textit{contextually} weights users' personal preferences in each group, in proportion to their user-group MI.
\textit{Group-adaptive} preference weighting precludes
degenerate solutions that arise during static regularization over ephemeral groups with sparse activities.
We summarize our key contributions below:

\begin{description}
    \item \textbf{Architecture-agnostic Group Recommendation Framework%
    }: 
    To the best of our knowledge, \textbf{Group} \textbf{I}nformation \textbf{M}aximization (\groupim) is the first recommender architecture-agnostic framework for group recommendation. Unlike prior work~\cite{agr, agree} that design customized group preference aggregators, \groupim~can integrate arbitrary neural preference encoders and aggregators. We show state-of-the-art results with simple efficient aggregators (such as meanpool) that are regularized within our framework.
The effectiveness of meanpool signifies substantially \textit{reduced inference costs} without loss in model expressive power. Thus,~\groupim~facilitates straightforward enhancements to base neural group recommenders. %
    \item \textbf{Group-adaptive Preference Learning and Prioritization}: We learn robust estimates of group-specific member relevance. In contrast, prior work incorporate personal preferences through static regularization~\cite{com, crowdrec, agree}.
    We use \textit{Mutual Information} for self-supervised learning over user and group representations to capture 
    preference covariance across individuals in the same group; and prioritize the preferences of highly relevant members through group-adaptive preference weighting; thus effectively overcoming group interaction sparsity in ephemeral groups.
    An ablation study confirms the superiority of our MI based 
    learning strategies over static alternatives.
    \item \textbf{Robust Experimental Results}: 
    Our experimental results indicate significant performance gains for \groupim~over prior state-of-the-art group recommendation models on four publicly available datasets (relative gains of 31-62\% NDCG@20 and 3-28\% Recall@20). Significantly, \groupim~achieves stronger gains for groups of larger sizes, and groups with diverse member preferences.
\end{description}

We organize the rest of the chapter as follows.
In Section~\ref{sec:groupim_prelim}, we formally define the problem, introduce a base group recommender unifying existing neural methods, and discuss its limitations.
We describe our proposed framework~\groupim~in Section~\ref{sec:groupim_groupim}, present experimental results in Section~\ref{sec:groupim_experiments}, and finally conclude in Section~\ref{sec:groupim_conclusion}.

\section{Related Work}
\label{sec:groupim_related_work}

\textbf{Group Recommendation:} This line of work can be divided into two broad categories based on group types:
\textit{persistent} and \textit{ephemeral}. 
Persistent groups have stable members with rich activity history together, while ephemeral groups comprise users who interact with very few items together~\cite{ephemeral}.
A common approach is to consider persistent groups as virtual users~\cite{dlgr}; thus, personalized user-level recommenders can be directly applied.
However, such methods cannot handle ephemeral groups with sparse item interactions.
We focus on the more challenging scenario---\textit{recommendations to ephemeral groups}.

Prior work either aggregate recommendation results (or item scores) for each member, or aggregate individual member preferences, towards group predictions.
They fall into two classes: \textit{score (or late)} aggregation~\cite{least_misery} and \textit{preference (or early)} aggregation~\cite{com}.

Popular \textit{score aggregation} strategies include least misery~\cite{least_misery}, average~\cite{average}, maximum satisfaction~\cite{max_satisfaction}, and relevance and disagreement~\cite{rd}.
However, these are hand-crafted heuristics that overlook real-world group interactions.
An empirical comparison of different heuristic strategies~\cite{least_misery} has demonstrated the absence of a clear winner, especially with the variance in group sizes and coherence levels.

Early \textit{preference aggregation} strategies~\cite{merging} generate recommendations by constructing a group profile that combines the profiles (raw item histories) of group members.
Recent methods adopt a \textit{model-based} perspective to learn data-driven models.
\textit{Probabilistic} methods~\cite{pit,com,crowdrec} model the group generative process by considering both the personal preferences and relative influence of members, to differentiate their contributions towards group decisions.
However, a key weakness is their assumption that users have the same likelihood to follow individual and collective preferences, across different groups.
\textit{Neural} methods explore attention mechanisms~\cite{attention} to learn data-driven preference aggregators~\cite{agree, agr, wang2019group}.
MoSAN~\cite{agr} models group interactions via sub-attention networks; however, MoSAN operates on persistent groups while ignoring users' personal activities.
AGREE~\cite{agree} employs attentional networks for joint training over individual and group interactions; yet, the extent of regularization applied on each user (based on personal activities) is the same across groups, which results in degenerate solutions when applied to ephemeral groups.

An alternative  approach to tackle interaction sparsity is to exploit external side information, \textit{e.g.}, social network of users~\cite{SoAGREE,socialgroup, infvae,social_rec}, personality traits~\cite{zheng2018identifying}, demographics~\cite{delic2018observational}, and interpersonal relationships~\cite{delic2018use, gartrell2010enhancing}.
In contrast, our setting is conservative and does not include extra side information: we know only user and item ids, and item implicit feedback.
We address interaction sparsity through novel data-driven regularization and training strategies~\cite{cikm18adv}.
Our goal is to enable a wide spectrum of neural group recommenders to seamlessly integrate suitable preference encoders and aggregators.

\textbf{Mutual Information:}
Recent neural MI estimation methods~\cite{mine} leverage the InfoMax~\cite{infomax} principle for self-supervised representation learning.
They exploit the intrinsic \textit{structure} of the input data (\textit{e.g.}, spatial locality in images, community structure in graphs) via MI maximization objectives to improve representational quality.
Recent advances employ auto-regressive models~\cite{cpc} and aggregation functions~\cite{dim, dgi, yeh2019qainfomax} with noise-contrastive loss functions to preserve MI between structurally related inputs.

We leverage the InfoMax principle to exploit the preference covariance \textit{structure} shared amongst group members.
A key novelty of our approach is MI-guided weighting to regularize group embeddings with the personal preferences of highly relevant members.

\section{Preliminaries}
\label{sec:groupim_prelim}

In this section, we first formally define the \textit{ephemeral group recommendation} problem. 
Then, we present a base neural group recommender $\mR$ that unifies existing neural methods into a general framework.
Finally, we analyze the key shortcomings of $\mR$ to 
discuss motivations for propose new model learning strategies for ephemeral group recommendation.

\subsection{{Problem Definition}}
We consider the implicit feedback setting (only visits or clicks, no explicit ratings) with a user set $\gU$, an item set $\gI$, a group set $\gG$, a binary $|\gU| \times |\gI|$ user-item interaction matrix $\mX_U$,
and a binary $|\gG| \times |\gI|$ group-item interaction matrix $\mX_G$. We denote $\vx_u$, $\vx_g$ as the corresponding rows for user $u$ and group $g$ in user-item $\mX_U$ and group-item $\mX_G$ interaction matrices, with $|\vx_u|$, $|\vx_g|$ indicating their respective number of interacted items.
\begin{definition}[\textbf{Ephemeral Group}] An \textit{ephemeral group} $g \in \gG$ comprises a set of $|g|$ users $u^g = \{u^g_1, \dots, u^g_{|g|} \} \subset \gU$ with sparse historical interactions $\vx_g$.
\end{definition}

\begin{definition}[\textbf{Ephemeral Group Recommendation}]
We evaluate group recommendation on \textit{strict ephemeral groups}, which have never interacted together during training.
Given a strict ephemeral group $g$ during testing, 
our goal is to generate a ranked list over the item set $\gI$ relevant to users in $u^g$, \textit{i.e.}, learn a 
function $f_G : P(\gU) \times \gI \mapsto \sR$
that maps an ephemeral group and an item to a relevance score, where $P(\gU)$ is the power set of $\gU$.
\end{definition}

\begin{figure}[t]
    \centering
    \includegraphics[width=0.9\linewidth]{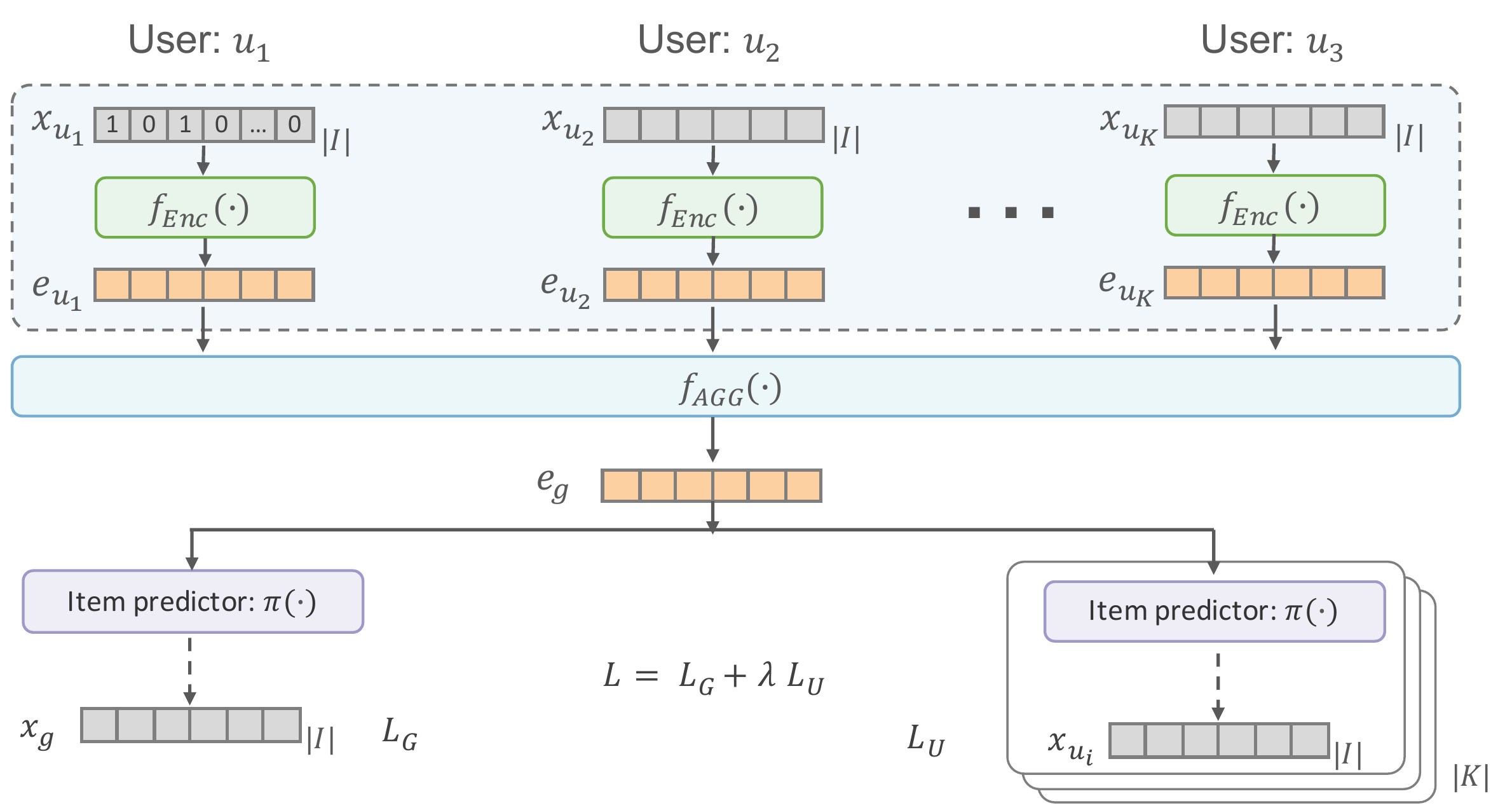}
    \caption{Neural architecture diagram of a base neural group recommender with user preference encoder, group preference aggregator and joint user and group losses.}
    \label{fig:base_arch}
\end{figure}

\subsection{{Base Neural Group Recommender}}
\label{sec:groupim_base_recommender}
Several neural group recommendation models have achieved impressive results~\cite{agree,agr}.
Despite their diversity in modeling group interactions, we remark that state-of-the-art neural methods share a clear model structure: we present a base group recommender $\mR$ that includes three modules:
a preference encoder; a preference aggregator; and a joint user and group interaction loss. Unifying these neural group recommenders within a single framework facilitates deeper analysis into their shortcomings in addressing ephemeral groups.

The base neural group recommender $\mR$ (Figure~\ref{fig:base_arch}) first computes user representations $\mE \in \sR^{|\gU| \times D}$ from individual user-item interactions $\mX_U$ using a preference encoder $f_{\textsc{enc}} (\cdot)$, followed by applying a neural preference aggregator $f_{\textsc{agg}} (\cdot)$ to compute the group representation $\ve_g \in \sR^D$ for group $g$. 
Finally, the latent group representation $\ve_g$ is jointly trained over the group-item $\mX_G$ and user-item $\mX_U$ interactions,
to make item recommendations for group $g$.

\subsubsection{\textbf{User Preference Representations.}}
User embeddings $\mE$ constitute a latent representation of their personal preferences, indicated in the interaction matrix $\mX_U$.
Since latent-factor collaborative filtering methods adopt a variety of strategies (such as matrix factorization, autoencoders, etc.) to learn user embeddings $\mE$, we define the preference encoder $f_{\textsc{enc}} : |\gU| \times \sZ_2^{|\gI|} \mapsto \sR^{D} $ with two inputs: user $u$ and associated binary personal preference vector $\vx_u$, defined by:

\begin{equation}
\ve_u =  f_{\textsc{enc}} (u, \vx_u) %
\;  \forall u \in \gU
\label{eqn:user_rep}
\end{equation}

We can augment the user representation $\ve_u$ with additional inputs, including user/item contextual attributes, item relationships, etc. via customized encoders~\cite{survey}.

\subsubsection{\textbf{Group Preference Aggregation.}}
A preference aggregator models the interactions among group members to compute an aggregate representation $\ve_g \in \sR^D$ for ephemeral group $g \in \gG$.
Since groups are sets of users with no inherent ordering, we consider the class of permutation-invariant functions (such as summation or pooling operations) on sets~\cite{deep_sets}.
Specifically, $f_{\textsc{agg}} (\cdot)$ is permutation-invariant to the order of group member embeddings $\{ e_{u_1}, \dots, e_{u_{|g|}} \}$.
We compute the group representation $\ve_g$ for group $g$ using an arbitrary preference aggregator $f_{\textsc{agg}} (\cdot)$ as:
\begin{equation}
    \ve_g = f_{\textsc{agg}} (\{ \ve_u : u \in u^g \}) \; \forall g \in \gG 
    \label{eqn:group_rep}
\end{equation}

\subsubsection{\textbf{Joint User and Group Loss.}}
The group representation $\ve_g$ is trained over the group-item interactions $\mX_G$ with group-loss $L_G$. 
The framework supports different recommendation objectives, including pairwise~\cite{bpr} and pointwise~\cite{ncf} ranking losses.
Here, we use a multinomial likelihood formulation owing to its impressive results in user-based neural collaborative filtering~\cite{vae-cf}. The group representation $\ve_g$ is transformed by a fully connected layer and normalized by a softmax function to produce a probability vector $\pi (\ve_g)$ over $\gI$. 
The group-interaction loss measures the KL-divergence between the normalized purchase history $\vx_{g}/|x_g|$ ($\vx_g$ indicates items interacted by group $g$) and predicted item probabilities $\pi (\ve_g)$, given by:

\begin{equation}
 \hspace{-5pt} L_G = - \sum\limits_{g \in \gG} \frac{1}{|\vx_g|} \sum\limits_{i \in \gI} x_{gi} \log \pi_i(\ve_g) ; \hspace{5pt} \pi(\ve_g)= \text{softmax}(\mW_I \ve_g) 
 \label{eqn:group_loss}
\end{equation}

Next, we define the user-loss $L_U$ that regularizes the user representations $\mE$ with user-item interactions $\mX_U$, thus facilitating joint training 
with shared encoder $f_{\textsc{enc}} (\cdot)$ and predictor ($\mW_I$) layers~\cite{agree}.
We use a similar multinomial likelihood-based formation, given by:
\begin{equation}
    L_U = - \sum\limits_{u \in \gU} \frac{1}{|\vx_u|} \sum\limits_{i \in \gI} x_{ui} \log \pi_i(\ve_u) ; \hspace{5pt} L_R = L_G + \lambda L_U
    \label{eqn:user_loss}
\end{equation}
where $L_R$ denotes the overall loss of the base recommender $\mR$ with balancing hyper-parameter $\lambda$.
Prior work AGREE~\cite{agree} trains an attentional group preference aggregator with a pairwise regression loss over both $\mX_U$ and $\mX_G$, while MoSAN~\cite{agr} trains a collection of sub-attentional aggregators with bayesian personalized ranking~\cite{bpr} loss on just $\mX_G$.
Thus, state-of-the-art neural methods AGREE~\cite{agree} and MoSAN~\cite{agr} are specific instances of the framework described by the base neural group recommender $\mR$.

\subsection{{Motivation}}
\label{sec:groupim_motivations}

To address ephemeral groups, we focus on \textit{learning} strategies that are \textit{independent} of the architectural choices in the base recommender $\mR$.
With the rapid advances in neural methods, we envision future enhancements in neural architectures for user representations and group preference aggregation.
Since ephemeral groups by definition purchase very few items together, base group recommenders suffer from inadequate training data in group interactions.
Here, the group embedding $\ve_g$ receives back-propagation signals from sparse interacted items in $\vx_g$, thus lacking evidence to reliably estimate the role of each member.
To address group interaction sparsity towards robust ephemeral group recommendation, we propose two 
data-driven model learning strategies that are independent of the base recommendation mechanisms to generate individual and group representations.

\subsubsection{\textbf{Self-supervised  Representation Learning}}

We note that users' preferences are \textit{group-dependent}; and users occurring together in groups typically exhibit \textit{covarying} preferences (\textit{e.g.}, shared cuisine tastes).
Thus, group interactions reveal \textit{distinctions} across groups (\textit{e.g.}, close friends versus colleagues) and latent \textit{user associations} (\textit{e.g.}, co-occurrence of users in similar groups), that are not directly evident when the base group recommender $\mR$ only predicts sparse group interactions.

We \textit{contrast} the preference representations of group members against those of non-member users with similar item histories, to effectively regularize the latent space of user and group representations.
This promotes the representations to encode 
latent discriminative characteristics shared by group members, that are not discernible from their sparse interactions. %

\begin{table}[t]
    \centering
    \begin{tabular}{@{}rl@{}}
        \toprule
        Symbol                & Description                                                               \\
        \midrule        
$\mX_U$ & Binary $ |\gU| \times |\gI|$ user-item interaction matrix \\
$\mX_G$ & Binary $ |\gG| \times |\gI|$ group-item interaction matrix \\
$\vx_u$ & $|\gI|$-dimensional row for user $u$ in matrix $\mX_U$ \\
$\vx_g$ & $|\gI|$-dimensional row for group $g$ in matrix $\mX_G$ \\
$\mR$ & Base neural group recommender \\
$f_{\textsc{enc}} $  & Individual user preference encoder \\
$f_{\textsc{agg}} $ & Group preference aggregator \\
$\rve_u$ & Latent user preference representation for user $u \in \gU$ \\
$\rve_g$ & Latent group preference representation for group $g \in \gG$\\
        \bottomrule
    \end{tabular}
    \caption{Notation}
    \label{tab:groupim_notations}
\end{table}

\subsubsection{\textbf{Group-adaptive Preference Prioritization.}}
To overcome group interaction sparsity, we critically remark that while groups are ephemeral with sparse interactions, the group members have comparatively richer individual interaction histories. Thus, we propose to \textit{selectively} exploit the personal preferences of group members to enhance the quality of group representations via contextual regularization.

The user-loss $L_U$ (equation~\ref{eqn:user_loss}) in base recommender $\mR$ attempts to regularize user embeddings $\mE$ based on their individual activities $\mX_U$.
A key weakness is that
$L_U$ forces $\ve_u$ to \textit{uniformly} predict preferences $\vx_u$ across all groups containing user $u$.
Since groups interact with items differently than individual members,
inaccurately utilizing $\mX_U$ can become counter-productive.
Fixed regularization results in degenerate models that either over-fit or are over-regularized, due to lack
of flexibility in adapting preferences per group.

To overcome group interaction sparsity, 
we \textit{contextually} identify members that are highly \textit{relevant} to the group and regularize the group representation to reflect their personal preferences.
To measure contextual relevance, we introduce group-specific relevance weights $w (u, g)$ for each user $u$ where $w(\cdot)$ is a learned weighting function of both the user and group representations.
This enhances the expressive power of the group recommender, thus effectively alleviating the challenges imposed by group interaction sparsity.

In this section, we defined ephemeral group recommendation, and 
presented a base group recommender architecture with three modules: user representations, group preference aggregation, and joint loss functions.
Finally, 
we motivated the need to:
contrastively regularize the user-group space to capture member associations and group distinctions; and learn group-specific weights $w(u,g)$ to regularize group representations with individual preferences.

\section{\textsc{GroupIM} Framework}

\label{sec:groupim_groupim}
We first motivate mutual information towards achieving our two proposed model learning strategies, followed by a detailed description of our proposed framework~\textsc{\groupim}.

\subsection{{Mutual Information Maximization.}}

We introduce our user-group mutual information maximization approach through a stylized example.
We extend the introductory example to illustrate how to regularize Alice's latent representation based on her interactions in two different groups.
Consider Alice who first goes out for lunch to an Italian restaurant with a visitor Bob, and later dines at a Mexican restaurant with her friend Charlie.

First, Alice plays different roles across the two groups (\textit{i.e.}, stronger influence among friends than with Bob) due to the differences in group context (visitors versus friends). Thus, we require a measure to quantify the contextual \textit{informativeness} of user $u$ in group $g$.

Second, we require the latent representation of Alice to capture associations with both visitor Bob and friend Charlie, yet express variations in her group activities.
Thus, it is necessary to not only \textit{differentiate} the role of Alice across groups, but also compute appropriate representations that make her presence in each group more \textit{coherent}.

\begin{figure}[t]
    \centering
    \includegraphics[width=0.9\linewidth]{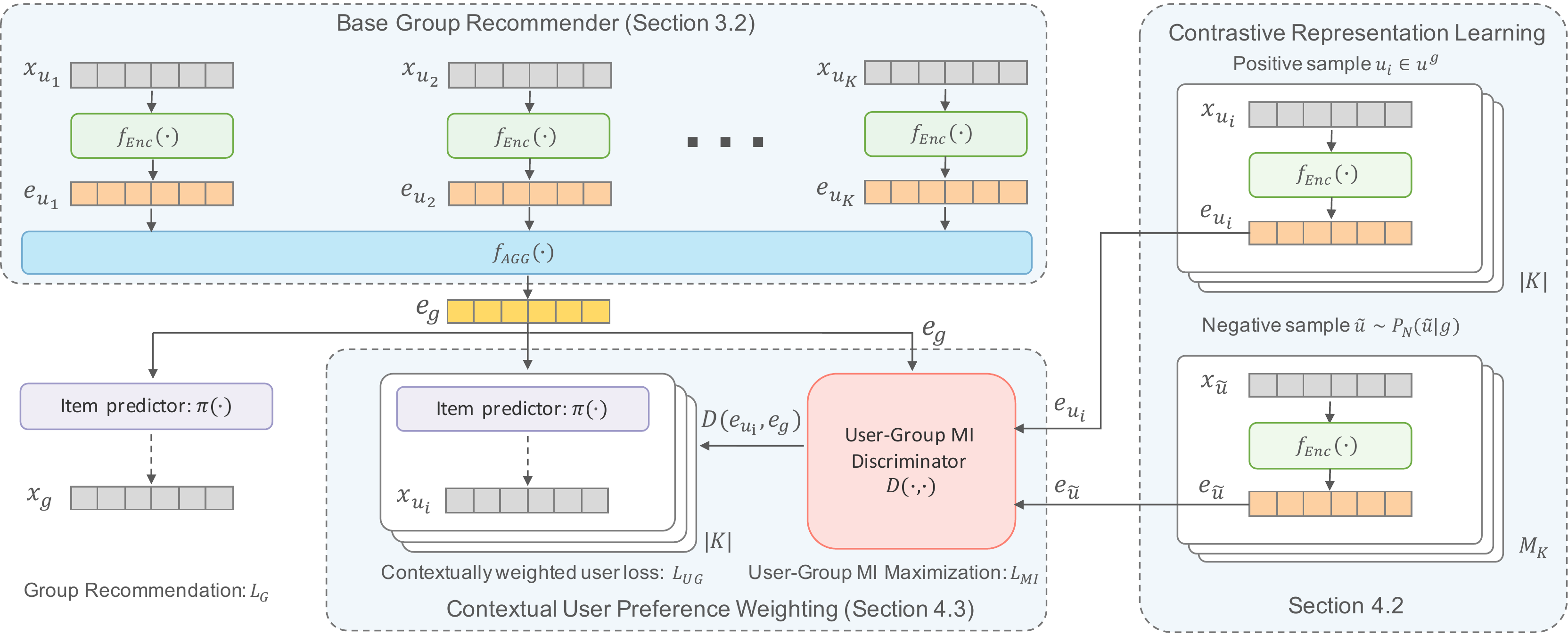}
    \caption{Neural architecture diagram of~\groupim~depicting the different model components and loss terms appearing in Equation~  \ref{eqn:model_objective}.}
    \label{fig:base_grec}
\end{figure}
To achieve these two goals at once, we maximize \textit{user-group mutual information} (MI) to regularize the latent space of user and group representations, and set group-specific relevance weights $w(u,g)$ in proportion to their estimated MI scores.
User-group MI measures the contextual informativeness of a member $u$ towards the group decision through the reduction in group decision uncertainty when user $u$ is included in group $g$.
Unlike correlation measures that quantify monotonic linear associations, mutual information captures complex non-linear statistical relationships between covarying random variables. Our proposed user-group MI maximization strategy enables us to achieve our two-fold motivation (Section~\ref{sec:groupim_motivations}): 

\begin{itemize}
    \item \textbf{Group Preference Co-variance Learning}: %
    Maximizing user-group MI 
    encourages the group embedding
    $\ve_g$ to encode preference covariance across group members, and regularizes the user embeddings $\mE$ to capture social associations in group interactions.
    \item 
    \textbf{Group-specific User Relevance}: 
    By quantifying $w(u,g)$ through
    user-group mutual information, 
    we accurately capture the extent of \textit{informativeness} for user $u$ in group $g$, thus guiding group-adaptive personal preference prioritization for regularization.
\end{itemize}

\subsection{{User-Group MI Maximization.}}
\label{sec:groupim_user-group-im}
 Neural MI estimation~\cite{mine} has demonstrated feasibility to maximize MI by training a classifier $\mD$ (\textit{a.k.a}, \textit{discriminator} network) to accurately separate \textit{positive} samples drawn from their joint distribution from \textit{negative} samples drawn from the product of marginals.

We maximize user-group MI between group member representations $\{ \ve_u : u \in u^g \}$ and the group representation $\ve_g$ (computed in equations~\ref{eqn:user_rep} and~\ref{eqn:group_rep} respectively).
We train
a contrastive \textit{discriminator} network 
$\mD: \sR^D \times \sR^D \mapsto \sR^{+}$, where $\mD(\ve_u, \ve_g)$ represents the probability score assigned to this user-group pair (higher scores for users who are members of group $g$).
The positive samples $(\ve_u, \ve_g)$ for $\mD$ are the preference representations of $(u,g)$ pairs such that $u \in u^g$, and negative samples are derived by pairing $\ve_g$ with the representations of non-member users sampled from a negative sampling distribution $P_{\gN} (u | g)$. 
The discriminator $\mD$ is trained on a noise-contrastive type objective with a binary cross-entropy (BCE) loss between samples from the joint (positive pairs), and the product of marginals (negative pairs), resulting in the following training objective:

\begin{align}
    \hspace{-0.9em} L_{MI} = - \frac{1}{|\gG|} %
    \sum\limits_{g \in \gG} \frac{1}{\alpha_g} \Big[ & \sum\limits_{u \in u^g}  \log  \mD_{ug} + \sum\limits_{j=1}^{M_g} \E_{\tilde{u} \sim P_{\gN}} \log (1 - \mD_{\tilde{u}g} ) \Big]
    \label{eqn:mi_loss}
\end{align}

where $\alpha_g = |g| + M_g$, $M_g$ is the number of negative users sampled for group $g$ and $\mD_{ug}$ is a shorthand for $\mD(\ve_u, \ve_g)$.
This objective maximizes MI between $\ve_u$ and $\ve_g$ based on the Jensen-Shannon divergence between the joint and the product of marginals~\cite{dgi}.

We employ a \textit{preference-biased} negative sampling distribution $P_{\gN} (\tilde{u} | g)$, which assigns higher likelihoods to \textit{non-member} users who have purchased the group items $\vx_{g}$.
These \textit{hard negative examples} encourage the discriminator to learn latent aspects shared by group members by contrasting against other users with similar individual item histories.
We define the negative sampling distribution  $P_{\gN} (\tilde{u} | g)$ as:
\begin{equation}
P_{\gN} (\tilde{u} | g) \propto  \eta \gI (\vx_{\tilde{u}}^T \cdot \vx_g > 0 \}) + (1 - \eta) \frac{1}{|\gU|}
\label{eqn:neg_sampler}
\end{equation}
where $\gI (\cdot)$ is an indicator function and $\eta$ controls the sampling bias.
We set $\eta=0.5$ across all our experiments.
In comparison to random negative sampling, our experiments indicate that preference-biased negative user sampling exhibits better discriminative abilities.

When $L_{MI}$ is trained jointly with the base recommender loss $L_R$ (equation~\ref{eqn:user_loss}), 
maximizing user-group MI enhances the quality of user and group representations computed by the encoder $f_{\textsc{enc}} (\cdot)$ and aggregator $f_{\textsc{agg}} (\cdot)$.
We now present our proposed approach to overcome the limitations due to the fixed regularizer $L_U$ (Section~\ref{sec:groupim_motivations}).

\subsection{{Contextual User Preference Weighting}}
\label{sec:groupim_context_weight}
In this section, we describe a contextual weighting strategy to identify and prioritize personal preferences of relevant group members, to overcome group interaction sparsity. 
We avoid degenerate solutions 
by varying the extent of regularization induced by each $\vx_u$ (for user $u$) across groups through group-specific relevance weights $w(u,g)$.
Contextual weighting accounts for user participation in diverse groups with different levels of shared interests.

By maximizing user-group mutual information, the trained discriminator $\mD$ outputs scores $\mD(\ve_u, \ve_g)$ that quantify the contextual informativeness of each $(u,g)$ pair (higher scores for informative users).
Thus, we set the relevance weight $w(u,g)$ for group member $u \in u^g$ to be proportional to the discriminator score $\mD(\ve_u,\ve_g)$. 
Instead of regularizing the user representations $\mE$ with $\vx_u$ in each group ($L_U$ in eqn~\ref{eqn:user_loss}), we directly regularize the group representation $\ve_g$ with $\vx_u$ in proportion to $\mD(\ve_u, \ve_g)$ for each group member $u$. 
Direct optimization of the group representation $\ve_g$ (instead of $\ve_u$) results 
in more effective 
model regularization with faster flow of gradients, especially with sparse group interactions in ephemeral groups.
Thus, we now define the contextually weighted user-loss $L_{UG}$ as:

\begin{equation}
L_{UG} = - \sum\limits_{g \in \gG} \frac{1}{|\vx_g|} \sum\limits_{i \in \gI} \sum\limits_{u \in u^g} \mD(\ve_u, \ve_g) \; x_{ui} \log \pi_i (\ve_g)
\label{eqn:user_context_loss}
\end{equation}

where $L_{UG}$ effectively regularizes the latent group representation $\ve_g$ with the individual activities of group member $u$ with contextual relevance weight $\mD(\ve_u,\ve_g)$.

The overall model objective of our framework~\groupim~includes three terms: $L_G$, $L_{UG}$, and $L_{MI}$, which is described in Section~\ref{sec:groupim_model_opt} in detail.
~\groupim~regularizes the latent representations computed by $f_{\textsc{enc}} (\cdot)$ and $f_{\textsc{agg}} (\cdot)$ through 
user-group MI maximization ($L_{MI}$) to contrastively capture group member associations; and contextual MI-guided weighting ($L_{UG}$) to prioritize individual preferences.

\subsection{{Model Details}}
\label{sec:groupim_model_details}
We now describe the architectural details of preference encoder $f_{\textsc{enc}}(\cdot)$, aggregator $f_{\textsc{agg}} (\cdot)$, discriminator $\mD$, and an alternative optimization approach to train our framework~\groupim.

\subsubsection{\textbf{User Preference Encoder}}
To encode individual user preferences $\mX_U$ into preference embeddings $\mE$, we use a Multi-Layer Perceptron with two fully connected dense layers, defined by:
\begin{equation}
 \ve_u = f_{\textsc{enc}} (\vx_u) = \sigma (\mW_2^T (  \sigma  (\mW_1^T \vx_u + b_1 ) + b_2  ) 
\end{equation}

where $\ve_u$ is the latent preference representation for user $u$, with learnable weight matrices $\mW_1 \in \sR^{|\gI| \times D}$ and $\mW_2 \in \sR^{D \times D}$, biases $b_1, b_2 \in \sR^D$, and $tanh(\cdot)$ activations for non-linearity $\sigma$.

\textbf{Pre-training:} We pre-train the weights and biases of the first encoder layer ($\mW_1, \vb_1$) on the user-item interaction matrix $\mX_U$ with user-loss $L_U$ (equation~\ref{eqn:user_loss}).
We use these pre-trained parameters to initialize the first layer of $f_{\textsc{enc}} (\cdot)$ before optimizing the overall objective of~\groupim.
Our model ablation studies in Section~\ref{sec:groupim_ablation} indicate significant improvements owing to this pre-trained initialization strategy.

\subsubsection{\textbf{Group Preference Aggregators}}
\label{sec:groupim_group_agg}
We consider three preference aggregators \textsc{Maxpool}, \textsc{Meanpool} and, \textsc{Attention}, which are widely used 
in graph neural networks~\cite{gnn_powerful, graphsage, dysat} and have close ties to aggregators examined in prior work, \textit{i.e.}, \textsc{Maxpool} and \textsc{Meanpool} mirror the heuristics of maximum satisfaction~\cite{max_satisfaction} and averaging~\cite{average}, while attentions learn varying member contributions~\cite{agree, agr}.
We define the three preference aggregators below:

\begin{itemize}[leftmargin=*]
    \item \textbf{Maxpool}: The preference embedding of each member is passed through \textsc{MLP} layers, followed by element-wise max-pooling to aggregate group member embeddings, given by:
    \begin{equation}
    \ve_g = \textsc{max}(\{ \sigma(\mW_{\textsc{agg}} \ve_{u} + b), \forall u \in u_g \})
    \end{equation}
    where \textsc{max} denotes the element-wise max operator and $\sigma(\cdot)$ is a nonlinear activation.
    Intuitively, the MLP layers compute features for each member, and max-pooling over each of the computed features effectively captures different aspects of group members.
    \item \textbf{Meanpool}: We similarly apply an element-wise mean-pooling operation after the \textsc{MLP} layers, to compute the latent group representation $\ve_g$ as:
    \begin{equation}
        \ve_g = \textsc{mean}(\{ \sigma(\mW_{\textsc{agg}} \ve_{u} + b), \forall u \in u_g \})
    \end{equation}
    \item \textbf{Attention}: To explicitly differentiate group members' roles, we employ neural attentions~\cite{attention} to compute a weighted sum of members' preference representations, where the weights are learned by an attention network, parameterized by a single \textsc{MLP} layer.
    \begin{equation}
    \ve_g = \sum\limits_{u \in u_g} \alpha_u \mW_{\textsc{agg}} \ve_u \hspace{10 pt} \alpha_u = \frac{\exp (\vh^T \mW_{\textsc{agg}} \ve_u) }{\sum\limits_{u^{'} \in u_g} \exp(\vh^T \mW_{\textsc{agg}} \ve_{u^{'}}) }
    \end{equation}
    where $\alpha_u$ indicates the contribution of a group member $u$ towards the group decision. We exclude bias terms for the sake of brevity.
    This formulation can be trivially extended to item-conditioned weighting~\cite{agree}, self-attention~\cite{wang2019group} and sub-attention networks~\cite{agr}.
\end{itemize}

\subsubsection{\textbf{Discriminator Architecture}}
The contrastive discriminator architecture learns a scoring function to assign higher scores to observed $(u,g)$ pairs relative to negative examples, thus parameterizing group-specific relevance weight $w(u,g)$.
Similar to existing work on neural mutual information estimation~\cite{dgi}, we use a simple bilinear function to score user-group representation pairs. The discriminator function $\mD$ is defined by the following equation:
\begin{equation}
    \mD (\ve_u, \ve_g) = \sigma (\ve_u^T \mW \ve_g)
\end{equation}
where $\mW$ is a learnable scoring matrix and $\sigma$ is the logistic sigmoid non-linearity function to convert raw scores into probabilities of $(\ve_u, \ve_g)$ being a positive example.
We leave investigation of further architectural variants for the discriminator $\mD$ to future work.

\subsubsection{\textbf{Model Optimization}}
\label{sec:groupim_model_opt}
The overall objective of~\groupim~is composed of three terms, the group-loss $L_G$ (Equation~\ref{eqn:group_loss}), contextually weighted user-loss $L_{UG}$ (Equation~\ref{eqn:user_context_loss}), and MI maximization loss $L_{MI}$ (Equation~\ref{eqn:mi_loss}).
The combined model training objective is given by:

\begin{equation}
   L =  \hspace{-5pt} \underbrace{L_G}_{\text{Group Recommendation Loss}} \hspace{-25pt} + \hspace{-25pt} \overbrace{\lambda L_{UG}}^{\text{Contextually Weighted User Loss}} \hspace{-25pt}+ \hspace{-25pt} \underbrace{L_{MI}}_{\text{User-Group MI Maximization Loss}}
   \label{eqn:model_objective}
\end{equation}

We train~\groupim~using an alternating optimization schedule. In the first step, the discriminator $\mD$ is held constant, while optimizing the group recommender on $L_G + \lambda L_{UG}$.
The second step trains $\mD$ on $L_{MI}$, resulting in gradient updates for both parameters of $\mD$ as well as those of the user encoder $f_{\textsc{enc}}(\cdot)$ and preference aggregator $f_{\textsc{agg}}(\cdot)$.

Thus, the discriminator $\mD$ only seeks to \textit{regularize} the model (\textit{i.e.,} encoder and aggregator) during training through loss terms $L_{MI}$ and $L_{UG}$.
During inference, we directly use the regularized encoder $f_{\textsc{enc}}(\cdot)$ and aggregator $f_{\textsc{agg}}(\cdot)$ to make group recommendations.

\subsubsection{\textbf{Model Objective Interpretation}}
To understand our model better, we examine a special case with fixed relevance weights $w(u,g) =1$ without MI maximization. Here, the overall model objective reduces to:

 \begin{align}
 L^{'} = - \sum\limits_{g \in \gG} \Big[ \frac{1}{|\vx_g|} \sum\limits_{i \in \gI} x_{gi} \log \pi_i(\ve_g) + \lambda  \sum\limits_{i \in \gI} \bar{x}_g \log \pi_i(\ve_g) \Big]
 \end{align}

 where $\bar{x}_{gi} =\frac{1}{|\vx_g|} \sum\limits_{u \in u^g} x_{ui}$ is the mean preference vector of users in group $g$ for item $i \in \gI$.
 The first term is a supervised loss to predict label $x_{gi}$ from group embedding $\ve_g$.
 The second term can be viewed as an unsupervised \textit{group reconstruction} loss that encodes member preferences $\{ \vx_u : u \in u^g \}$ into the group embedding $\ve_g$ to reconstruct their \textit{mean} preference vector $\bar{x}_{g}$.
 In the special case of $w(u,g) =1$,~\groupim~reduces to a classical semi-supervised learning objective (similar to Equation~\ref{eqn:user_loss}) for group recommendation.

 Thus, our framework~\groupim~can be overall regarded as jointly performing \textit{self-supervised} learning with \textit{user-group mutual information maximization} (via loss term $L_{MI}$) and \textit{semi-supervised} learning with \textit{group-adaptive} regularization (via loss term $L_{UG}$).

\section{Experiments}
\label{sec:groupim_experiments}
\begin{table}[t]
\centering

\begin{tabular}{@{}p{0.31\linewidth}K{0.12\linewidth}K{0.15\linewidth}K{0.14\linewidth}K{0.14\linewidth}@{}}

\toprule
\textbf{Dataset} &  \textbf{Yelp} & \textbf{Weeplaces} & \textbf{Gowalla}  & \textbf{Douban}\\
\midrule
\textbf{\# Users} &  7,037 & 8,643 & 25,405 & 23,032 \\
\textbf{\# Items} &  7,671 & 25,081 & 38,306 & 10,039\\
\textbf{\# Groups} &  10,214 & 22,733 & 44,565 & 58,983 \\
\textbf{\# U-I interactions} &  220,091 & 1,358,458 & 1,025,500 & 1,731,429 \\
\textbf{\# G-I Interactions} &  11,160 & 180,229 & 154,526 & 93,635 \\
\textbf{Average \# items/user} & 31.3 & 58.83 & 40.37 &  75.17\\
\textbf{Average \# items/group} & 1.09 & 2.95 & 3.47 &  1.59 \\
\textbf{Average group size} & 6.8 & 2.9 & 2.8 & 4.2 \\
\bottomrule
\end{tabular}
\caption{Statistics of four real-world datasets with ephemeral groups. Group-Item interactions are sparse: average number of interacted items per group $<$ 3.5.}
\label{tab:groupim_dataset_stats}
\end{table}
In this section, we present an extensive quantitative and qualitative analysis of our model. We first introduce datasets, baselines, and experimental setup (Section~\ref{sec:groupim_datasets},~\ref{sec:groupim_baselines}, and~\ref{sec:groupim_setup}), followed by our main group recommendation results (Section~\ref{sec:groupim_main_results}).
In Section~\ref{sec:groupim_ablation}, we conduct an ablation study 
to understand our gains over the base recommender.
In Section~\ref{sec:groupim_group_char}, we study how key group characteristics (group \textit{size, coherence, and aggregate diversity}) impact recommendation results.
In Section~\ref{sec:groupim_mi_analysis}, we qualitatively visualize the variation in discriminator scores assigned to group members, for different kinds of groups. Finally, we discuss the limitations of our approach in Section~\ref{sec:groupim_discussion}.

\subsection{{Datasets}}
\label{sec:groupim_datasets}
First, we conduct experiments on large-scale POI (point-of-interest) recommendation datasets extracted from three location-based social networks.
Since the POI datasets do not contain explicit group interactions, we 
construct group interactions by jointly using the check-ins and social network information:
check-ins at the same POI within 15 minutes by an individual and her subset of friends in the social network together constitutes a single group interaction, while the remaining check-ins at the POI correspond to individual interactions.
We define the group recommendation task as recommending POIs to ephemeral groups of users.
The datasets were pre-processed to retain users and items with five or more check-ins each. We present dataset descriptions below:
\begin{itemize}[leftmargin=*]
    \item \textbf{Weeplaces}~\footnote{\url{https://www.yongliu.org/datasets/}}: we extract check-ins on POIs over all major cities in the United States, across various categories including Food, Nightlife, Outdoors, Entertainment and Travel.
    \item \textbf{Yelp}~\footnote{\url{https://www.yelp.com/dataset/challenge}}: user check-ins on local restaurants located in the city of Los Angeles, California.
    \item \textbf{Gowalla}~\cite{gowalla}: we use restaurant check-ins across all cities in the United States, in the time period upto June 2011.
\end{itemize}
    
Second, we evaluate event venue recommendation on Douban, which is the largest online event-based social networking platform in China.
\begin{itemize}[leftmargin=*]
    \item \textbf{Douban}~\cite{socialgroup}: users organize and participate in social events, where users attend events together in groups and items correspond to event venues.
    During pre-processing, 
    we filter out users and venues with less than 10 interactions each.
\end{itemize}
Groups across all datasets are ephemeral since group interactions are sparse (average number of items per group $< 3.5$ in Table~\ref{tab:groupim_dataset_stats})

\subsection{{Baselines}}
\label{sec:groupim_baselines}
We present comparisons against state-of-the-art baselines in two categories: score aggregation methods with predefined aggregators, and data-driven preference aggregators.

\begin{itemize}[leftmargin=*]
    \item \textbf{Popularity}~\cite{popularity}: recommends items to groups based on item popularity, which is measured by its interaction count in the training set. 
    \item \textbf{User-based CF + Score Aggregation}: We utilize a 
    state-of-the-art neural recommendation model VAE-CF~\cite{vae-cf}, followed by score aggregation via:
averaging (AVG), least-misery (LM), maximum satisfaction (MAX), and relevance-disagreement (RD). %
    \item \textbf{COM}~\cite{com}: a probabilistic generative model that determines group decisions based on 
    group members' individual preferences and topic-dependent influence.
    \item \textbf{CrowdRec}~\cite{crowdrec}: a probabilistic generative model that extends COM through item-specific latent variables capturing their global popularity for group recommendation.
    \item \textbf{MoSAN}~\cite{agr}: a neural group recommender that employs a collection of sub-attentional networks to model group member interactions. Since MoSAN originally ignores individual activities $\mX_U$, we include $\mX_U$ into $\mX_G$ as pseudo-groups with single users.
    \item \textbf{AGREE}~\cite{agree}: a neural group recommender that utilizes attentional group preference aggregation to compute item-specific group member weights, for joint model training over individual and group interactions.
\end{itemize}

We tested~\textbf{\groupim}~by substituting three preference aggregators, \textsc{Maxpool}, \textsc{Meanpool}, and \textsc{Attention} (Section~\ref{sec:groupim_group_agg}). 
All experiments were conducted on a single Nvidia Tesla V100 GPU with \textit{PyTorch}~\cite{pytorch} implementations on the Linux platform. Our implementation of~\textbf{\groupim}~and datasets are publicly available\footnote{\url{https://github.com/CrowdDynamicsLab/GroupIM}}.

\renewcommand*{\factor}{0.049}
\begin{table}[ht]
\centering
\small
\begin{tabular}{@{}p{0.325\linewidth}@{\hspace{13pt}}
K{\factor\linewidth}K{\factor\linewidth}@{\hspace{13pt}}
K{\factor\linewidth}K{\factor\linewidth}@{\hspace{13pt}}
K{\factor\linewidth}K{\factor\linewidth}@{\hspace{13pt}}
K{\factor\linewidth}K{\factor\linewidth}K{\factor\linewidth}@{}} \\
\toprule
\multirow{1}{*}{\textbf{Dataset}} &  \multicolumn{2}{c}{\textbf{Yelp (LA)}} & \multicolumn{2}{c}{\textbf{Weeplaces}}  & \multicolumn{2}{c}{\textbf{Gowalla}} & \multicolumn{2}{c}{\textbf{Douban}} \\
\multirow{1}{*}{\textbf{Metric}} & \textbf{N@20} & \textbf{N@50} &  \textbf{N@20} & \textbf{N@50}  & \textbf{N@20}& \textbf{N@50} &
 \textbf{N@20}& \textbf{N@50} \\
\midrule
\multicolumn{9}{c}{\textbf{Predefined Score Aggregators}} \\
\midrule
\textbf{Popularity~\cite{popularity}}  & 0.000  & 0.000   & 0.063  & 0.074  &  0.075  & 0.088  & 0.003  & 0.005  \\
\textbf{VAE-CF + AVG~\cite{vae-cf,average}}  & 0.142  & 0.179  & 0.273  & 0.313   & 0.318  & 0.362 & 0.179  & 0.217   \\
\textbf{VAE-CF + LM~\cite{vae-cf,least_misery}}  & 0.097  & 0.120  & 0.277  & 0.311  & 0.375  & 0.409   & 0.221  & 0.252   \\
\textbf{VAE-CF + MAX~\cite{vae-cf,max_satisfaction}}  & 0.099  & 0.133   & 0.229  & 0.270  & 0.267  & 0.316 & 0.156  & 0.194  \\
\textbf{VAE-CF + RD~\cite{vae-cf,rd}}  & 0.143  & 0.181  & 0.239  & 0.279  & 0.294  & 0.339  & 0.178  & 0.216   \\
\midrule
\multicolumn{9}{c}{\textbf{Data-driven Preference Aggregators}}\\
\midrule
\textbf{COM~\cite{com}}  & 0.143  & 0.154   & 0.329  & 0.348  & 0.223  & 0.234  &  0.283  & 0.288   \\
\textbf{Crowdrec~\cite{crowdrec}}  & 0.082  & 0.101  &  0.353  & 0.370   & 0.325  & 0.338    & 0.121  & 0.188  \\
\textbf{AGREE~\cite{agree}}  & 0.123  & 0.168   & 0.242  & 0.292   & 0.160  & 0.223  & 0.126  & 0.173   \\
\textbf{MoSAN~\cite{agr}}  & 0.470  & 0.494   & 0.287  & 0.334  &  0.323  & 0.372  & 0.193  & 0.239 \\
\midrule
\multicolumn{9}{c}{\textbf{Group Information Maximization Recommenders (GroupIM)}}\\
\midrule
\textbf{{GroupIM-Maxpool} } & 0.488  & 0.501 & 0.479  & 0.505  &  0.433  & 0.463  & 0.291 & 0.313    \\
\textbf{{GroupIM-Meanpool}}  & 0.629  & 0.637   & 0.518  & 0.543  & 0.476  & 0.504  & 0.323  & 0.351  \\
\textbf{{GroupIM-Attention}}  & \textbf{0.633}  & \textbf{0.647}  & \textbf{0.521}  & \textbf{0.546}   & \textbf{0.477}  & \textbf{0.505}  &  \textbf{0.325}  & \textbf{0.356}  \\
\bottomrule
\end{tabular}
\caption{Group recommendation results on four datasets, N@K denotes \textsc{NDCG}@K metric at $K = 20$ and $50$.
The~\groupim~variants indicate \textsc{maxpool}, \textsc{meanpool}, and \textsc{attention} as preference aggregators in our MI maximization framework.
~\groupim~achieves significant performance gains of 31 to 62\% NDCG@20 over competing group recommenders. 
Notice that \textsc{meanpool} and \textsc{attention} variants of our model~\groupim~achieve comparable group recommendation performance across all datasets.
}

\label{tab:groupim_ndcg_results}
\end{table}

\subsection{{Experimental Setup}}
\label{sec:groupim_setup}
We randomly split the set of all groups into training (70\%), validation (10\%), and test (20\%) sets, while utilizing the individual interactions of all users for training.
Note that each group appears only in one of the three sets.
The test set contains \textit{strict ephemeral groups} (\textit{i.e.}, a specific combination of users) that do not occur in the training set. Thus, we train on ephemeral groups and test on strict ephemeral groups.
We use \textsc{NDCG@K} and \textsc{Recall@K} as evaluation metrics. %
\textsc{Recall@K} measures the percentage of relevant items in the top-K rank list for each group, while \textsc{NDCG@K} is position-sensitive and assigns higher scores for relevant items at the top.
We report model performance at $K = 20, 50$ in Tables~\ref{tab:groupim_ndcg_results} and~\ref{tab:groupim_recall_results}.

We tune the latent dimension (number of topics for COM and Crowdrec) in the range $\{ 32, 64, 128\}$ for all models 
and other baseline hyper-parameters in ranges centered at author-provided valuess on the validation set.
In GroupIM, we use two fully connected layers of size 64 each
in the MLP preference encoder $f_{\textsc{enc}} (\cdot)$
and tune $\lambda$ in the range $\{2^{-4}, 2^{-3}, \dots, 2^6\}$.
We use 5 negative samples for each true user-group pair to train the discriminator.
Our experimental results are averaged over 10 random runs with random initializations.

\renewcommand*{\factor}{0.049}
\begin{table}[ht]
\centering
\small
\begin{tabular}{@{}p{0.325\linewidth}@{\hspace{13pt}}
K{\factor\linewidth}K{\factor\linewidth}@{\hspace{13pt}}
K{\factor\linewidth}K{\factor\linewidth}@{\hspace{13pt}}
K{\factor\linewidth}K{\factor\linewidth}@{\hspace{13pt}}
K{\factor\linewidth}K{\factor\linewidth}K{\factor\linewidth}@{}} \\
\toprule
\multirow{1}{*}{\textbf{Dataset}} &  \multicolumn{2}{c}{\textbf{Yelp (LA)}} & \multicolumn{2}{c}{\textbf{Weeplaces}}  & \multicolumn{2}{c}{\textbf{Gowalla}} & \multicolumn{2}{c}{\textbf{Douban}} \\
\multirow{1}{*}{\textbf{Metric}} & \textbf{R@20} & \textbf{R@50} & \textbf{R@20} & \textbf{R@50}  & 
\textbf{R@20} & \textbf{R@50} & \textbf{R@20} & \textbf{R@50}\\
\midrule
\multicolumn{9}{c}{\textbf{Predefined Score Aggregators}} \\
\midrule
\textbf{Popularity~\cite{popularity}}  & 0.001  & 0.001  & 0.126  & 0.176  &  0.143  & 0.203   & 0.009  & 0.018  \\
\textbf{VAE-CF + AVG~\cite{vae-cf,average}}   & 0.322  & 0.513  & 0.502  & 0.666  &0.580  & 0.758  & 0.381  & 0.558  \\
\textbf{VAE-CF + LM~\cite{vae-cf,least_misery}}  & 0.198  & 0.316  & 0.498  & 0.640  & 0.610  & 0.750  & 0.414  & 0.555  \\
\textbf{VAE-CF + MAX~\cite{vae-cf,max_satisfaction}}   & 0.231  & 0.401  &  0.431  & 0.604  & 0.498  & 0.702  & .339  & 0.517  \\
\textbf{VAE-CF + RD~\cite{vae-cf,rd}}  & 0.321  & 0.513    & 0.466  & 0.634  &0.543  & 0.723  & 0.379  & 0.557  \\
\midrule
\multicolumn{9}{c}{\textbf{Data-driven Preference Aggregators}}\\
\midrule
\textbf{COM~\cite{com}}  & 0.232  & 0.286  &  0.472  & 0.557  &  0.326  & 0.365  &  0.417  & 0.436  \\
\textbf{Crowdrec~\cite{crowdrec}}  &  0.217  & 0.315  & 0.534  & 0.609  & 0.489  & 0.548  & 0.375  & 0.681  \\
\textbf{AGREE~\cite{agree}}  & 0.332  & 0.545  &  0.484  & 0.711  & 0.351  & 0.605  &  0.310  & 0.536  \\
\textbf{MoSAN~\cite{agr}}  &  0.757  & \textbf{0.875}  & 0.548  & 0.738  &  0.584  & 0.779  & 0.424  & 0.639  \\
\midrule
\multicolumn{9}{c}{\textbf{Group Information Maximization Recommenders (GroupIM)}}\\
\midrule
\textbf{{GroupIM-Maxpool} } & 0.676  & 0.769   & 0.676  & 0.776  & 0.628  & 0.747   & 0.524  & 0.637   \\
\textbf{{GroupIM-Meanpool}}  & 0.778  & 0.846  &  0.706  & 0.804  &0.682  & 0.788  & 0.569  & 0.709  \\
\textbf{{GroupIM-Attention}}   &\textbf{0.782}  & 0.851    & \textbf{0.716}  & \textbf{0.813}  &  \textbf{0.686}  & \textbf{0.796}   & \textbf{0.575}  & \textbf{0.714}   \\
\bottomrule
\end{tabular}
\caption{Group recommendation results on four datasets, R@K denotes the \textsc{Recall}@K metric at $K = 20$ and $50$.
The~\groupim~variants indicate \textsc{maxpool}, \textsc{meanpool}, and \textsc{attention} as preference aggregators.~\groupim~achieves significant gains of 3 to 28\% \textsc{Recall}@20 over competing group recommenders.}

\label{tab:groupim_recall_results}
\end{table}

\subsection{{Experimental Results}}
\label{sec:groupim_main_results}
We note the following key observations from our experimental results comparing our framework~\textbf{\groupim}~with its three aggregator variants, against competing baselines on ephemeral group recommendation (Tables~\ref{tab:groupim_ndcg_results} and~\ref{tab:groupim_recall_results}).

First, heuristic score aggregation with neural user-level recommenders (\textit{i.e.},~\textbf{VAE-CF}) performs comparable to (and often beats) 
probabilistic models 
(\textbf{COM, Crowdrec}).
Neural methods with multiple non-linear transformations, are expressive enough to identify latent groups of similar users just from their individual item interactions.

Second, there is no clear winner among the different pre-defined score aggregation strategies, \textit{e.g.}, \textbf{VAE-CF + LM} (least misery) outperforms the rest on Gowalla and Douban, while \textbf{VAE-CF + LM} (averaging) is superior on Yelp and Weeplaces. This empirically validates the
non-existence of a single optimal strategy for all datasets.

Third,~\textbf{MoSAN}~\cite{agr} outperforms both probabilistic models and fixed score aggregators on most datasets.
~\textbf{MoSAN}~achieves better results owing to the expressive power of neural preference aggregators (such as sub-attention networks) to capture group member interactions, albeit not explicitly differentiating personal and group activities.
Notice that naive joint training over personal and group activities via static regularization (as in \textbf{AGREE}~\cite{agree}) results in poor performance due to sparsity in group interactions.
Static regularizers on $\mX_U$ cannot distinguish the role of users across groups, resulting in models that lack generalization to ephemeral groups with sparse interactions.

~\textbf{\groupim}~variants outperform baselines significantly, with \textsc{attention} achieving overall best results.
In contrast to neural methods (\textit{i.e.}, \textbf{MoSAN} and \textbf{AGREE}), 
~\textbf{\groupim}~regularizes the latent representations by contextually weighting the personal preferences of informative members, thus effectively tackling group interaction sparsity.
The \textsc{maxpool} variant is noticeably inferior, due to the higher sensitivity of \textit{max} operation to outlier members.

Note that \textsc{Meanpool} performs comparably to \textsc{attention}. This is because in~\textbf{\groupim}, the discriminator $\mD$ does the heavy-lifting of contextually differentiating the role of users across groups to effectively regularize the encoder $f_{\textsc{enc}} (\cdot)$ and aggregator $f_{\textsc{agg}} (\cdot)$ modules.
If $f_{\textsc{enc}} (\cdot)$ and $\mD$ are expressive enough, efficient \textsc{meanpool} aggregation can achieve near state-of-the-art results (Tables~\ref{tab:groupim_ndcg_results} and ~\ref{tab:groupim_recall_results}).

An important implication is the \textit{reduced inference complexity} of our model, \textit{i.e.}, once trained using our MI maximizing framework, simple aggregators (such as \text{meanpool}) suffice to achieve state-of-the-art performance.
This is especially significant, considering that our closest baseline \textbf{MoSAN}~\cite{agr} utilizes sub-attentional preference aggregation networks that scale quadratically with group size.

\begin{figure}[t]
    \centering
    \includegraphics[width=0.9\linewidth]{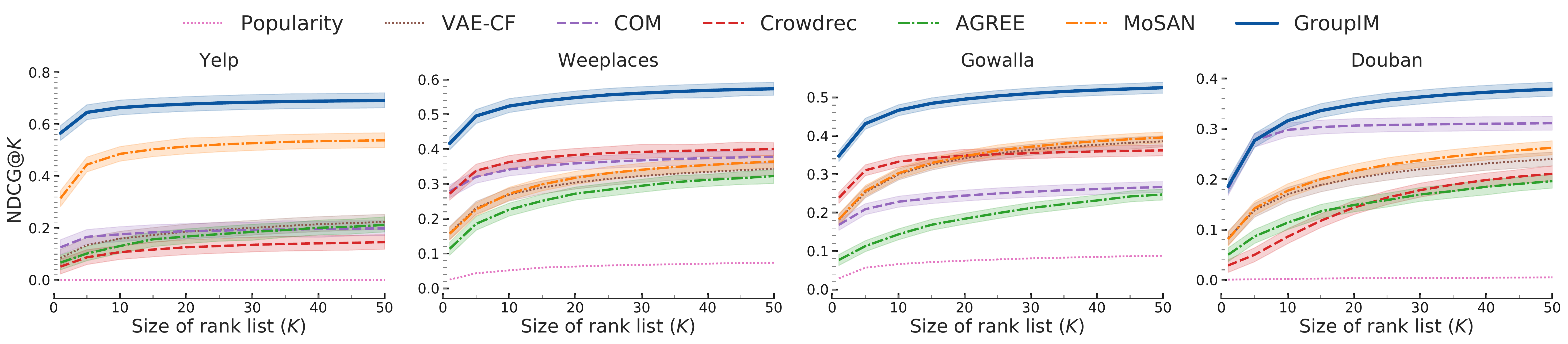}
    \caption{\textsc{NDCG}@K across size of rank list $K$. Variance bands indicate 95\% confidence intervals over 10 random runs.
    Existing methods underperform since they either disregard member roles (VAE-CF variants) or overfit to the sparse group activities.
    ~\groupim~contextually identifies informative members and regularizes their representations, to show strong gains.}
    \label{fig:groupim_ndcg}
\end{figure}

 \begin{figure}[t]
     \centering
     \includegraphics[width=0.9\linewidth]{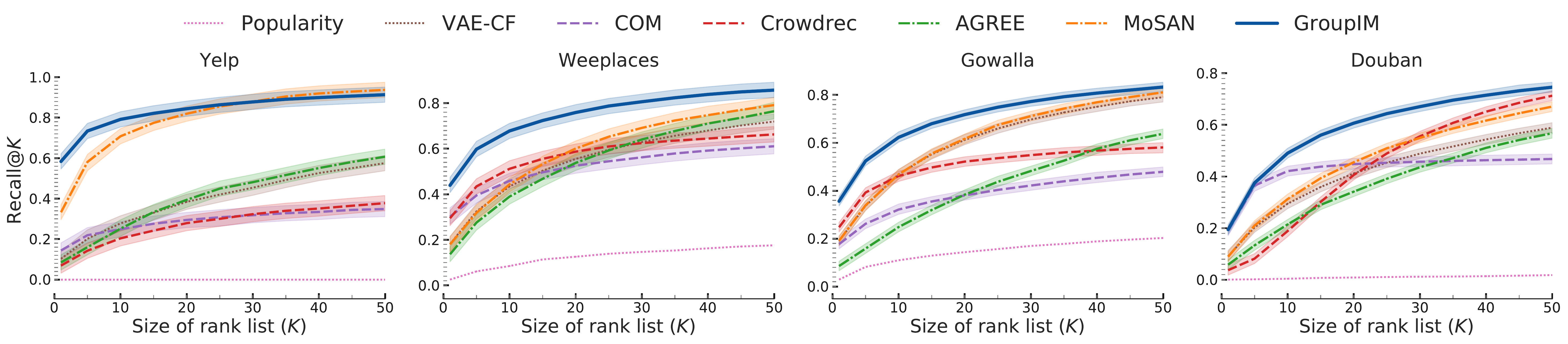}
     \caption{\textsc{Recall}@K across size of rank list $K$ (1 to 50). Variance bands indicate 95\% confidence intervals over 10 independent runs.
     ~\groupim~has larger recall gains for smaller $K (<20)$, indicating more accurate recommendations within top ranks.}
     \label{fig:groupim_recall}
 \end{figure}

We compare the variation in \textsc{NDCG} and \textsc{Recall} scores of all models with size of rank list $K$ (1 to 50) in figures~\ref{fig:groupim_ndcg} and~\ref{fig:groupim_recall}.
We only depict the best aggregator for \textbf{VAE-CF}.
~\textbf{\groupim} consistently generates more \textit{precise} recommendations across all datasets. We observe smaller gains in Douban, where the user-item interactions exhibit substantial correlation with corresponding group activities.~\textbf{\groupim} achieves significant performance gains in characterizing diverse groups, evidenced by our results in section~\ref{sec:groupim_group_char}.

\subsection{{Model Analysis}}
\label{sec:groupim_ablation}
In this section, we present an ablation study to analyze several variants of~\textbf{\groupim}, guided by our motivations (Section~\ref{sec:groupim_motivations}).
In our experiments, we choose \textsc{attention} as the aggregator due to its consistently high performance.
We conduct empirical studies on Weeplaces and Gowalla datasets to report \textsc{NDCG}@50 and \textsc{Recall}@50 in Table~\ref{tab:groupim_ablation_results}.

First, we examine the model performance of the base group recommender $\mR$ (Section~\ref{sec:groupim_base_recommender}) which does not utilize user-group MI estimation or maximization for model training. The ablation study results are shown in Table~\ref{tab:groupim_ablation_results}.

\renewcommand*{\factor}{0.088}
\begin{table}[hbtp]
\centering
\small
\begin{tabular}{@{}p{0.47\linewidth}K{\factor\linewidth}K{\factor\linewidth}
K{\factor\linewidth}@{\hspace{13pt}}K{\factor\linewidth}K{\factor\linewidth}K{\factor\linewidth}@{}} \\
\toprule
{\textbf{Dataset}} &  \multicolumn{2}{c}{\textbf{Weeplaces} } & \multicolumn{2}{c}{\textbf{Gowalla}} \\
\textbf{Metric} & \textbf{N@50} & \textbf{R@50} & \textbf{N@50} &  \textbf{R@50} \\
\midrule
\multicolumn{5}{c}{\textbf{Base Group Recommender Variants}} \\
\midrule
(1) \textbf{Base} ($L_G$) & 0.420 &  0.621  &  0.369 & 0.572  \\
(2) \textbf{Base} ($L_G  + \lambda L_U$) & 0.427 & 0.653 & 0.401  & 0.647 \\
\midrule
\multicolumn{5}{c}{\textbf{\groupim~Variants}} \\
\midrule
(3) \textbf{\groupim} ($L_G + L_{MI}$) & 0.431 & 0.646 & 0.391 & 0.625 \\
(4) \textbf{\groupim} (Uniform weights)  &  0.441  &  0.723 & 0.418 &  0.721 \\
(5) \textbf{\groupim} (Cosine similarity) & 0.488 & 0.757 &  0.445 & 0.739 \\
(6) \textbf{\groupim} (No pre-training) & 0.524 & 0.773 &  0.472 & 0.753 \\
(7) \textbf{\groupim} ($L_G+ \lambda L_{UG} + L_{MI}$)  & \textbf{0.543} & \textbf{0.804} & \textbf{0.505} & \textbf{0.796}  \\
\bottomrule
\end{tabular}
\caption{~\groupim~ablation study ($\textsc{NDCG}$ and $\textsc{Recall}$ at $K = 50$).
Contrastive representation learning (row 3) improves the base recommender (row 1), but is substantially more effective with group-adaptive preference weighting (row 7).}
\label{tab:groupim_ablation_results}
\end{table}

\subsubsection*{\textbf{Base Group Recommender.}} We examine two variants below:\\
(1) We train the base recommender $\mR$ on group interactions $\mX_G$ with loss $L_G$ (equation~\ref{eqn:group_loss}).\\
(2) We train the base recommender $\mR$ jointly on individual $\mX_U$ and group $\mX_G$ interactions with static regularization on $\mX_U$ using joint loss $L_R$ (Equation~\ref{eqn:user_loss}).

In comparison to similar neural group preference aggregator \textbf{MoSAN}, our base group recommender $\mR$ is stronger on \textsc{NDCG} but inferior on \textsc{Recall}.
The difference is likely due to the multinomial likelihood used to train $\mR$, in contrast to the ranking loss in \textbf{MoSAN}.
Static regularization via $\mX_U$ (row 1) results in higher gains for Gowalla (richer user-item interactions) with relatively larger margins for \textsc{Recall} than \textsc{NDCG}.
We now empirically examine the different model variants of~\textbf{\groupim}~in two parts:

\subsubsection*{\textbf{\groupim: Self-supervised Representation Learning.}} We analyze the benefits derived by just training the contrastive discriminator $\mD$ to capture group member associations through user-group MI maximization, \textit{i.e.}, we define a model variant (row 3) to optimize just $L_G + L_{MI}$, without the $L_{UG}$ term.
Direct user-group MI maximization (row 3)
improves over the base group recommender $\mR$ (row 1), thus validating the benefits of contrastive self-supervised learning via mutual information maximization, however still suffers from lack of user preference prioritization.

\subsubsection*{\textbf{\groupim: Group-adaptive Preference Prioritization}}
We analyze the utility of data-driven contextual weighting (via user-group MI), 
by examining two alternate \textit{fixed} strategies to define $w(u,g)$ in the loss term $L_{UG}$ of~\groupim:\\
(4) \textbf{\textit{Uniform weights}}: We assign the same relevance weight $w(u,g) = 1$ for each group member $u$ in group $g$, when optimizing the loss term $L_{UG}$. \\
(5) \textbf{\textit{Cosine similarity}}: To model user-group correlation, we set the relevance weight $w(u,g)$ as the cosine similarity between $\vx_u$ and $\vx_g$.

From table~\ref{tab:groupim_ablation_results} (rows 4 and 5), the uniform weights variant of loss $L_{UG}$ (row 4) surpasses the statically regularized model (row 2), due to more direct feedback from $\mX_U$ to the group embedding $\ve_g$ during model training.
Cosine similarity (row 5) achieves stronger gains owing to more accurate correlation-guided user weighting across groups.
Our model~\textbf{\groupim}~(row 7) has strong gains over the fixed weighting strategies as a result of its regularization strategy to contextually identify informative members across different groups. 
\subsubsection*{\textbf{\groupim: Pre-training $f_{\textsc{ENC}} (\cdot)$ on $\mX_U$}}
We depict model performance without pre-training (random initializations) in row 6.
Our model (row 7) achieves noticeable gains; pre-training identifies good model initialization points for better convergence.

\subsection{{Impact of Group Characteristics}}
\label{sec:groupim_group_char}

In this section, we examine our results closely to understand the reason for~\textbf{\groupim}'s gains over baselines in two datasets, Weeplaces and Gowalla. We study ephemeral groups along three facets: \textit{group size}; \textit{group coherence}; and \textit{group aggregate diversity}.

\begin{figure}[t]
    \centering
    \includegraphics[width=0.9\linewidth]{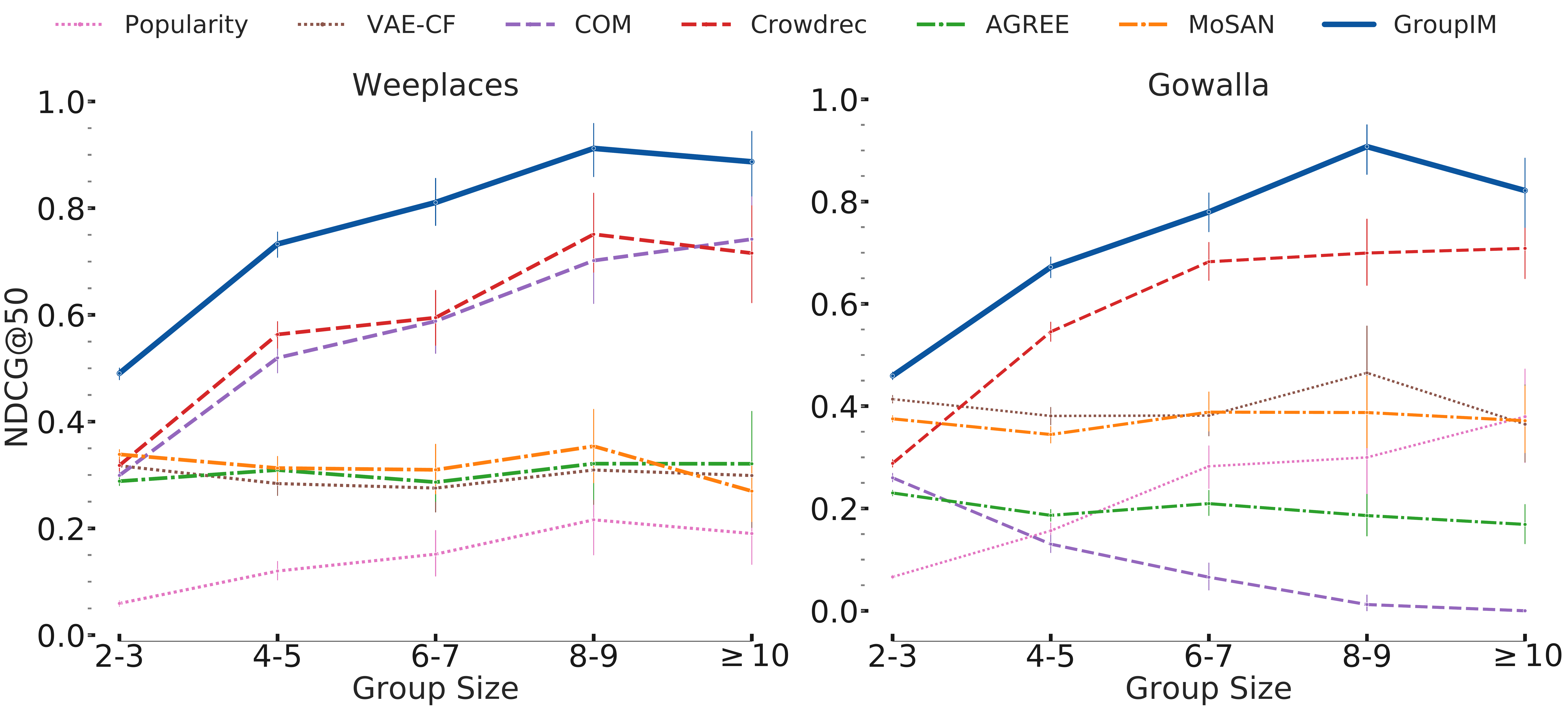}
    \caption{Performance (NDCG@50), across group sizes.~\groupim~has larger gains for larger groups due to accurate user associations learnt via MI maximization.}
    \label{fig:groupim_group_size}
\end{figure}

\subsubsection{\textbf{Group Size}}
We partition test groups into bins based on five levels of group size (2-3, 4-5, 6-7, 8-9, and  $\geq$10).
Figure~\ref{fig:groupim_group_size} depicts the variation in \textsc{NDCG}@50 scores on Weeplaces and Gowalla.

We make three key observations:  methods that explicitly distinguish individual and group activities (such as \textbf{COM, CrowdRec, GroupIM}), exhibit distinctive trends \textit{wrt} group size. In contrast, \textbf{MoSAN}~\cite{agr} and \textbf{AGREE}~\cite{agree}, which either uniformly mix both behaviors or apply static regularizers, show no noticeable variation; %
Performance generally increases with group size. 
Although test groups are previously unseen, for larger groups, 
subsets of inter-user interactions are more likely to be seen during training,
thus resulting in better performance; 
~\textbf{\groupim}~achieves higher (or steady) gains for groups of larger sizes owing to its more accurate prioritization of personal preferences for each member, \textit{e.g.}, ~\textbf{\groupim}~clearly has stronger gains for groups of sizes 8-9 and $\geq10$ in Gowalla.

\subsubsection{\textbf{Group Coherence}}
\begin{figure}[t]
    \centering
    \includegraphics[width=0.9\linewidth]{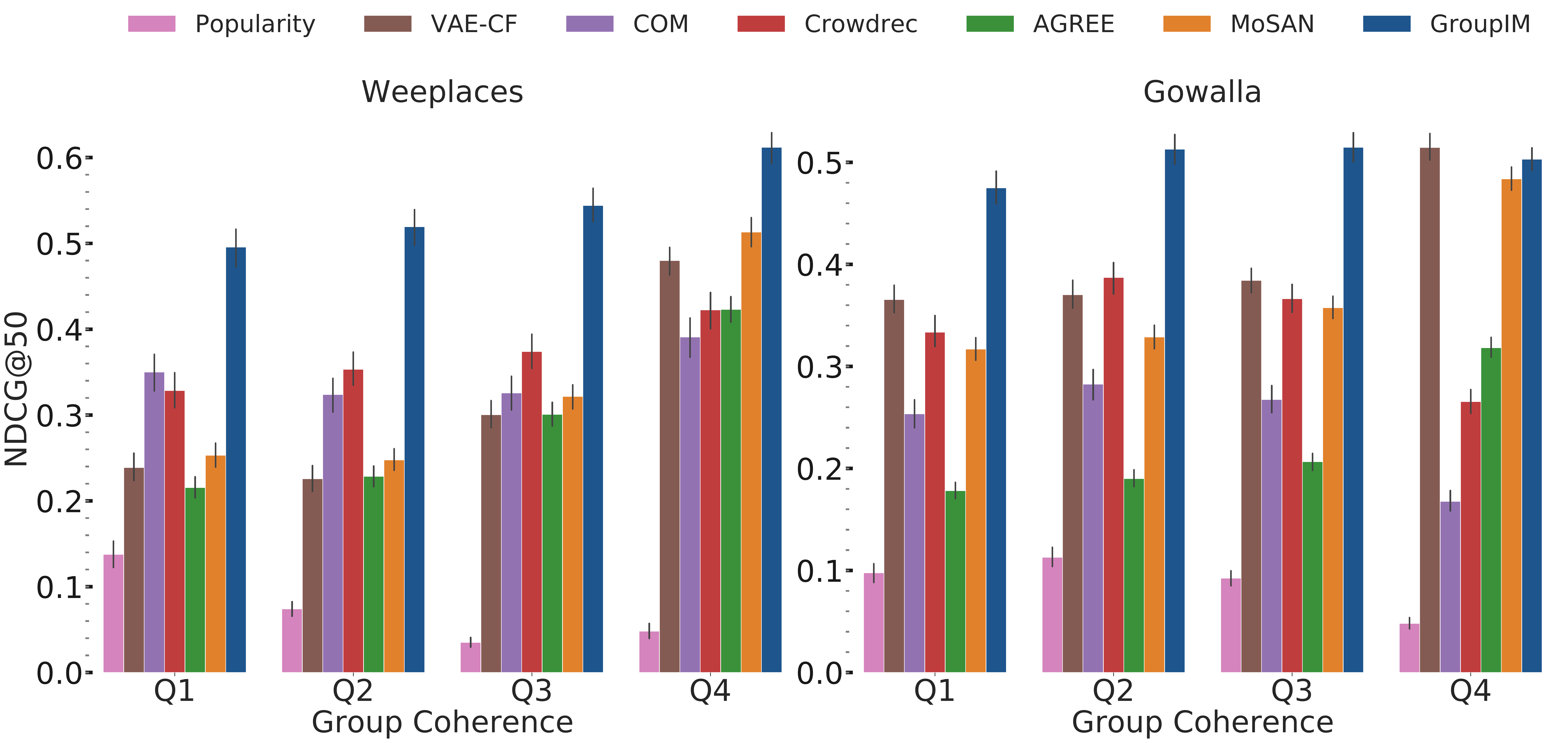}
    \caption{Performance (NDCG@50), across group coherence quartiles (Q1: lowest, Q4: highest).~\groupim~has larger gains in Q1 \& Q2 (low group coherence).}
    \label{fig:groupim_group_coherence}
\end{figure}

We define \textit{group coherence} as the mean pair-wise correlation of personal activities ($\vx_u$) of group members, \textit{i.e.}, if a group has users who frequently co-purchase items, it receives greater coherence.
We separate test groups into four quartiles by their coherence scores.
Figure~\ref{fig:groupim_group_coherence} depicts \textsc{NDCG}@50
for groups under each quartile (Q1 - Lower values). %

~\textbf{\groupim}~has stronger gains for groups with low coherence (quartiles Q1 and Q2), which empirically validates the efficacy of contextual user preference weighting in regularizing the encoder and aggregator, for groups with dissimilar member preferences.

\begin{figure}[t]
    \centering
    \includegraphics[width=0.9\linewidth]{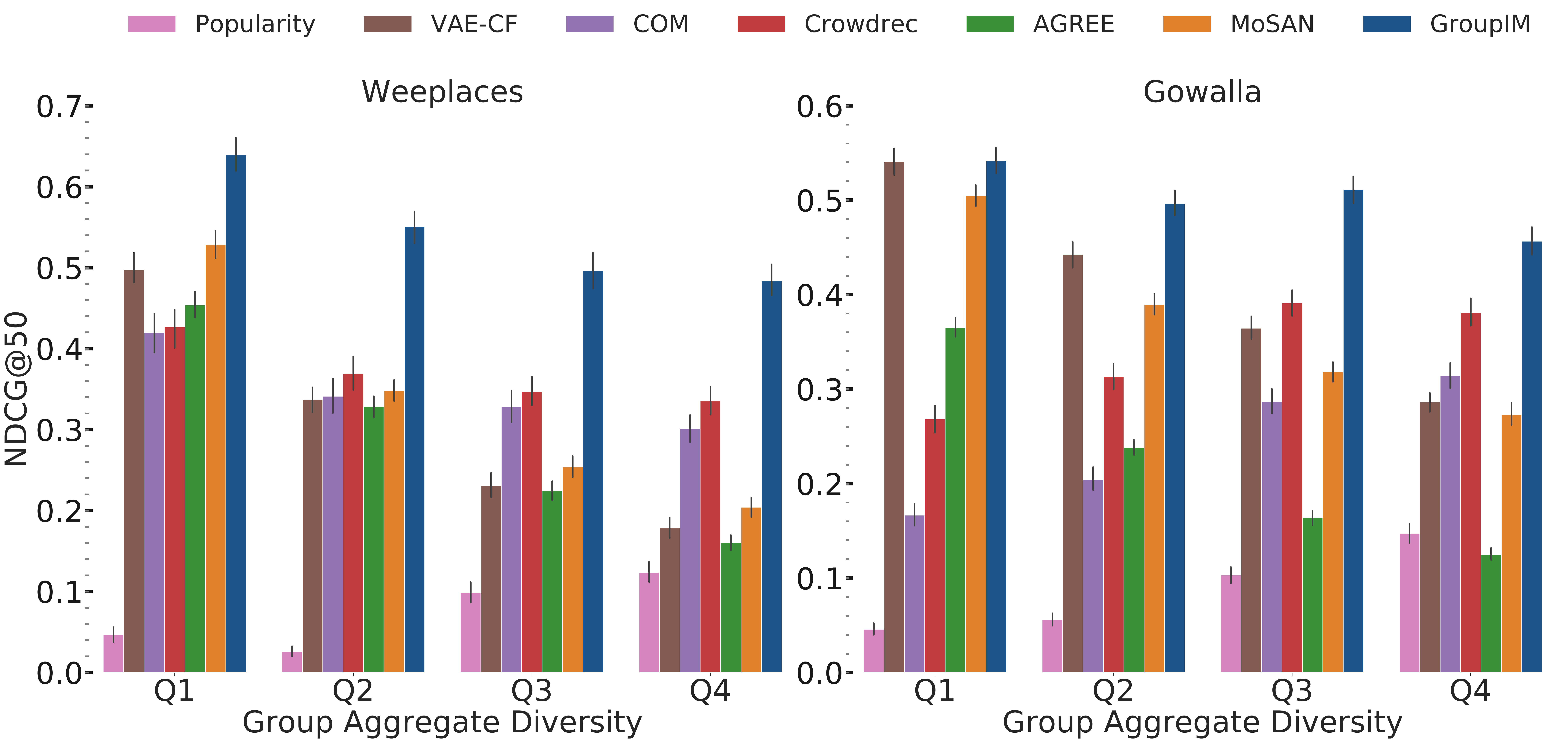}
    \caption{Performance (NDCG@50), across group aggregate diversity quartiles (Q1: lowest, Q4: highest).~\groupim~has larger gains in Q3 \& Q4 (high diversity).}
    \label{fig:groupim_group_coverage}
\end{figure}
\subsubsection{\textbf{Group Aggregate Diversity}}
We adapt the classical \textit{aggregate diversity} metric~\cite{diversity_metrics} to define \textit{group aggregate diversity} as the total number of distinct items interacted across all group members, 
\textit{i.e.}, if the set of all purchases of group members covers a wider range of items, then the group has higher aggregate diversity. 
We report NDCG@50 across aggregate diversity quartiles in figure~\ref{fig:groupim_group_coverage}.

Model performance typically decays (and stabilizes), with increase in group aggregate diversity.
Diverse groups with large candidate item sets pose an information overload for group recommenders, leading to worse results. 
Contextual prioritization with contrastive self-supervised learning benefits diverse groups, as evidenced by the higher relative gains of~\textbf{\groupim}~highly for diverse groups (quartiles Q3 and Q4).

\subsection{{Qualitative MI Discriminator Analysis}}
\label{sec:groupim_mi_analysis}
We examine the contextual relevance weights $w(u,g)$ estimated by \textbf{\groupim} over test ephemeral groups, across group size and coherence.

\begin{figure}[t]
    \centering
    \includegraphics[width=0.9\linewidth]{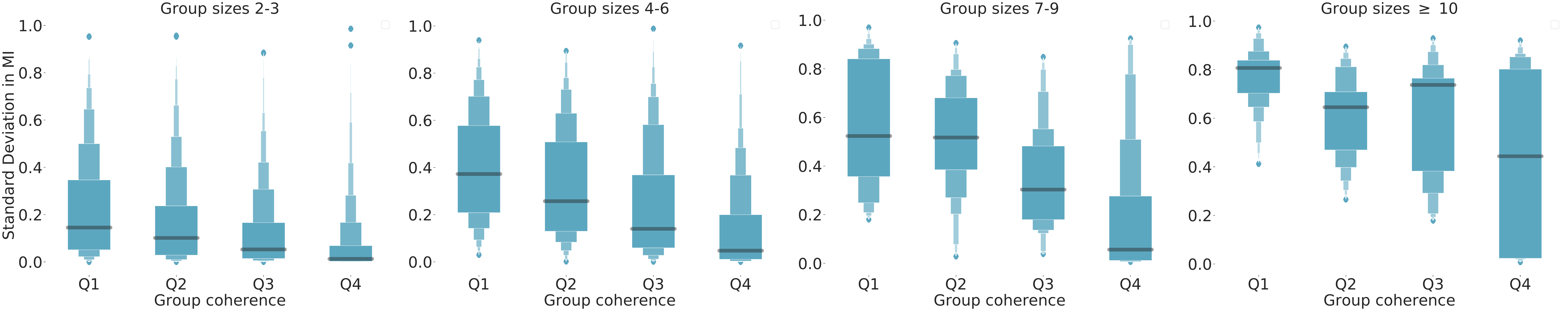}
    \caption{\textit{MI variation} (std. deviation in discriminator scores over members) per group coherence quartile across group sizes.
    For groups of a given size, as coherence increases, \textit{MI variation} decreases.
    As groups increase in size, \textit{MI variation} increases.}
    \label{fig:groupim_discriminator}
\end{figure}

We divide groups into four bins based on group sizes (2-3, 4-6, 7-9, and $\geq10$), and partition them into quartiles based on group coherence per bin.
To analyze the variation in \textit{contextual informativeness} across group members, 
we compute \textit{MI variation} as the standard deviation of scores given by $\mD$ over group members.
Figure~\ref{fig:groupim_discriminator} depicts letter-value plots of \textit{MI variation} for groups in corresponding coherence quartiles across group sizes on Weeplaces.

\textit{MI variation} increases with group size, since larger groups often comprise users with divergent roles and interests. 
Thus, the learned discriminator generalizes to unseen groups, to discern and estimate markedly different relevance scores for each group member.
To further examine the intuition conveyed by the scores, we compare \textit{MI variation} across group coherence quartiles within each group size-range.

\textit{MI variation} is negatively correlated with group coherence for groups of similar sizes, \textit{e.g.}, \textit{MI variation} is consistently higher for groups with low coherence (quartiles Q1 and Q2).
For highly coherent groups (quartile Q4), the discriminator $\mD$ assigns comparable scores across all members, which is consistent with our intuitions and earlier empirical results on the efficacy of simple averaging strategies for coherent groups.

\subsection{{Sensitivity Analysis}}
\label{sec:groupim_sensitivity}
\begin{figure}[t]
     \centering
     \includegraphics[width=0.9\linewidth]{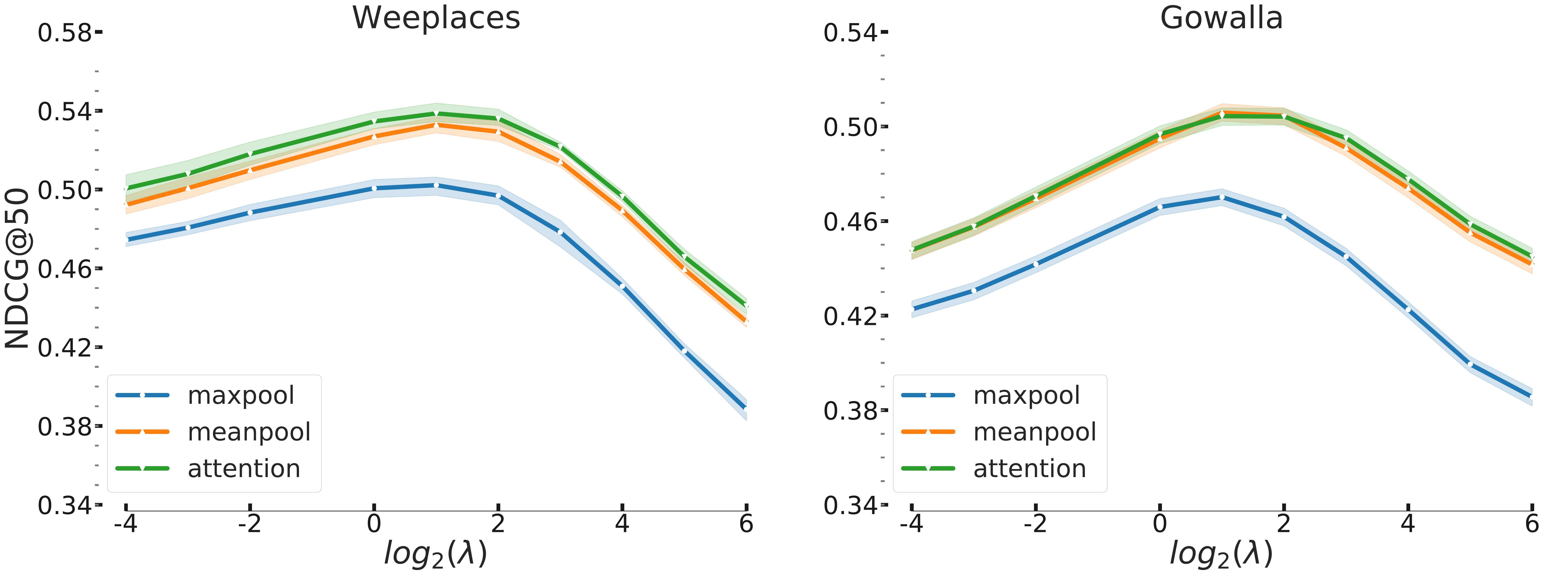}
 \caption{ 
 $\lambda$ is varied in $\{2^{-4}, 2^{-3}, \dots, 2^6\}$.
 Larger $\lambda$ values result in performance drops due to overfitting on $\mX_U$.~\groupim~is robust in a wide range of values.
 }
     \label{fig:groupim_lambda}
 \end{figure}

In this section, we analyze parameter sensitivity of our framework~\groupim~with respect to the user-preference weight $\lambda$. We show a plot depicting model performance (\textsc{NDCG}@50) versus $\lambda$ in figure~\ref{fig:groupim_lambda}.
Varying the hyper-parameter $\lambda$ in different ranges of values results in performance drops on either side, but for differing reasons.
Low $\lambda$ values result in overfitting to the group activities $\mX_G$, while substantially larger values result in degenerate representations that lack group distinctions.
 Although we tune $\lambda$ for each dataset, we find recommendation performance of our model to be stable in a broad range of values that transfer across different datasets.

\subsection{{Limitations}}
\label{sec:groupim_discussion}
We identify two limitations of our work. Despite \textit{learning} to contextually prioritize users' preferences across groups, $\lambda$ controls the overall strength of preference regularization.
Since optimal $\lambda$ varies across datasets and applications, we plan to explore meta-learning approaches to eliminate such hyper-parameters~\cite{meta-reweight}.

\textbf{\groupim}~relies on user-group MI estimation to contextually identify informative members, which might become challenging when users have sparse individual interaction histories.
In such a scenario, side information (\textit{e.g.}, social network of users), or contextual factors (\textit{e.g.}, location, interaction time)~\cite{cross-domain} can prove effective.

\section{Conclusion}
\label{sec:groupim_conclusion}
This chapter introduces a recommender architecture-agnostic framework~\groupim~that integrates arbitrary neural preference encoders and aggregators for ephemeral group recommendation.
To overcome group interaction sparsity,~\groupim~regularizes the user-group representation space by maximizing user-group MI to contrastively capture preference covariance among group members. %
Unlike prior work that incorporate individual preferences through static regularizers, we
dynamically prioritize the preferences of informative members through MI-guided contextual preference weighting.
Our extensive experiments on four real-world datasets show significant gains for~\groupim~over state-of-the-art methods.

In this chapter, we explored our second inductive learning application of generating item recommendations to ephemeral groups of users in a multipartite interaction setting involving users, groups, and items. We designed a self-supervised learning framework to specifically overcome group interaction sparsity challenges.
Finally, in the next chapter, we examine a practical industrial pplication of generating friend suggestions in large-scale social platforms, where we design inductive learning models with the capability to seamlessly incorporate new users and dynamic user interactions.

\chapter{\textsc{GraFRank}: Multi-Faceted Friend Ranking in Social Platforms}
\label{chap:grafrank}

\section{Introduction}
\label{sec:grafrank_introduction}

Learning latent user representations has become increasingly important in advancing  user understanding, and has seen widespread adoption in various industrial settings, \textit{e.g.}, video recommendations on YouTube~\cite{youtube}, related pin recommendations on Pinterest~\cite{pinsage} etc.
The user representations learned using deep models are \textit{effective} at complementing, or even replacing conventional collaborative filtering methods~\cite{CF}, and are \textit{versatile}, \textit{e.g.}, the learned user embeddings can be used to suggest new friendships and also to infer profile attributes (\textit{e.g.}, age, gender) in social networks.

Learning latent representations of nodes in graphs has prominent applications in  multiple \textit{academic} settings, such as link prediction~\cite{gnn_linkpred}, community detection~\cite{gnn_community}, and \textit{industrial} recommender systems, including e-commerce~\cite{alibaba_ecommerce, m2grl}, content discovery~\cite{pinsage, multisage}, and food delivery~\cite{uber}.
Graph Neural Networks (GNNs)~\cite{gnn_review} have emerged as a popular paradigm for graph representation learning due to their ability to learn representations combining graph structure and node/link attributes, without relying on expensive feature engineering.
GNNs can be formulated as a \textit{message passing} framework where node representations are learned by propagating features from local graph neighborhoods via trainable aggregators.
Recently, GNNs have demonstrated promising results in a few industrial systems designed for item recommendations in bipartite~\cite{pinsage} or multipartite~\cite{multisage} user-to-item interaction graphs.

Despite their rich representational ability, GNNs have been relatively unexplored in large-scale \textit{user-user social interaction} modeling applications, like friend suggestion. Recommending new potential friends to encourage users to expand their networks, is a cornerstone of social networking, and plays an important role towards user retention, and promoting engagement within the platform.

Prior efforts typically formulate friend suggestion as link prediction (or matrix completion) with a rich literature of graph-based heuristics~\cite{linkpred} to quantify user-user affinity, \textit{e.g.}, two users are more likely to connect if they have many common friends.
A few GNN models target link prediction in academic settings, learn aggregators over enclosing subgraphs around each candidate link~\cite{wlnm, gnn_linkpred, igmc};
such models do not scale to industry-scale social graphs with over millions of nodes and billions of edges.  Still, GNNs have enormous potential for learning expressive user representations in social networks, due to their intuitive message-passing paradigm that enables attention to social influence
from friends in their ego-network.

Yet, designing GNNs for friend recommendations in large-scale social media poses unique challenges.
First, social networks are characterized by \textit{heavy-tailed} degree distributions, \textit{e.g.}, many networks approximately follow power-law distributions~\cite{tail}.
This poses a key challenge of \textit{limited structural information} for a significant proportion of users with very few friends.
A related challenge is \textit{activity sparsity} where a very small fraction of users actively form new friendships at any given time.
Secondly, contemporary social networking platforms offer a multitude of avenues for users to interact, \textit{e.g.}, users can communicate with friends either by \textit{directly} exchanging messages, pictures, and videos, or \textit{indirectly} through a variety of social actions, including liking, sharing, and commenting on posts.
Extracting knowledge from such \textit{heterogeneous} in-platform user actions is challenging, yet extremely \textit{valuable} to address sparsity challenges for a vast majority of inactive users. %

\begin{figure}[t]
    \centering
    \includegraphics[width=0.9\linewidth]{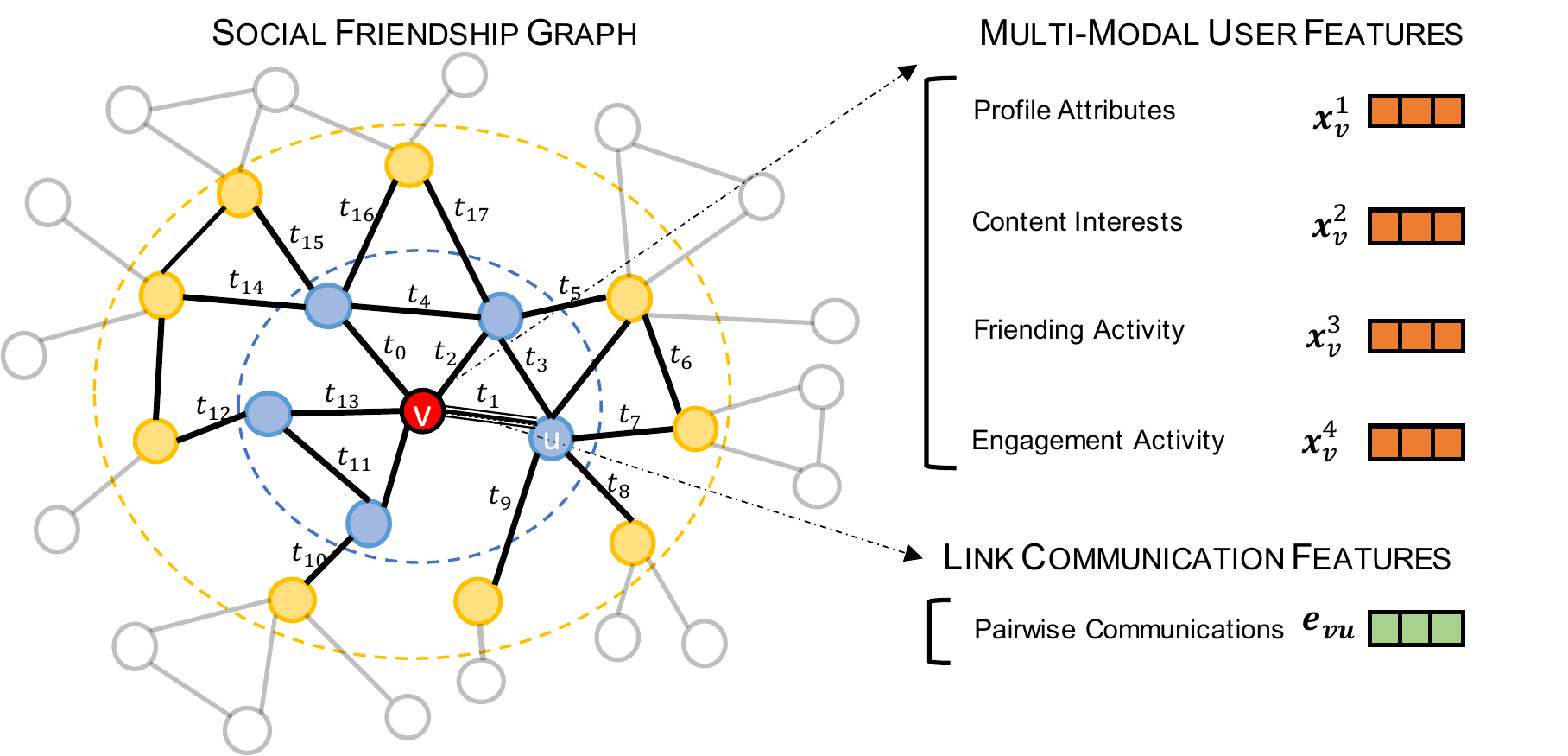}
    \caption{Desiderata for Multi-faceted Friend Ranking: temporally evolving friendship graph with multi-modal user features and pairwise link communication features.}
    \label{fig:grafrank_desiderata}
\end{figure}

\textbf{Present Work:} In this work, we overcome structural and interactional sparsity by exploiting the rich knowledge of heterogeneous in-platform actions. We formulate friend recommendation on social networks as \textit{multi-faceted friend ranking} on an \textit{evolving} friendship graph, with multi-modal user features and link communication features (Figure~\ref{fig:grafrank_desiderata}).
We represent users with heterogeneous feature sets spanning multiple \textit{modalities}, that include a collection of \textit{static} profile attributes (\textit{e.g.}, demographic information) and \textit{time-sensitive} in-platform activities (\textit{e.g.}, content interests and interactions).
We also leverage pairwise link features on existing friendships, which capture recent communication activities across multiple direct (\textit{e.g.}, messages) and indirect (\textit{e.g.}, stories) channels within the platform.

To understand the complexity of user interactions and gain insights into various factors impacting friendship formation,
we conduct an empirical analysis to investigate \textit{attribute homophily} with respect to different user feature modalities. %
Our analysis reveals \textit{diverse homophily distributions} across modalities and users, and indicates non-trivial \textit{cross-modality} correlations.
Motivated by these observations, we design an end-to-end GNN architecture, \grafrank (Graph Attentional Friend Ranker) for multi-faceted friend ranking.

\grafrank generates user representations by \textit{modality-specific} neighbor aggregation and \textit{cross-modality} attention.
Here, we handle heterogeneity in modality homophily by learning modality-specific message-passing aggregators to compute a set of latent representations for each user. Specifically, the neighbor aggregator is driven by a friendship attention mechanism that captures the influence of individual features and pairwise communications. %
We introduce a \textit{cross-modality} attention module to compute the effective user representation by attending over the different modality-specific representations for each user, thereby learning non-linear correlations across modalities. We summarize our key contributions below:

\begin{description}
    \item \textbf{Graph-Neural Friend Ranking}:
    To our knowledge, ours is the first work to investigate \textit{graph neural network} usage and design for social user-user interaction modeling applications.
    Unlike prior work that typically view friend recommendation as structural link prediction, we present a novel formulation with multi-modal user features and link features, to leverage knowledge of rich heterogeneous user activities in social networking platforms.

    \item \textbf{\grafrank Model:}
    Motivated by our empirical study that reveals heterogeneity in modality homophily and cross-modality correlations, we design a novel GNN architecture, \grafrank, to learn \textit{multi-faceted} user representations.
    Distinct from conventional GNNs that operate on a single homogeneous feature space, \grafrank encapsulates information from multiple correlated feature modalities and user-user interactions.

    \item \textbf{Robust Experimental Results:} Our experiments on two large-scale datasets from a popular social networking platform Snapchat, indicate significant gains for \grafrank over state-of-the-art baselines on friend candidate retrieval (relative MRR gains of 30\%) and ranking  (relative MRR gains of 20\%) tasks. Our qualitative analysis indicates stronger gains for the crucial population of \textit{less-active} and \textit{low-degree} users. %
\end{description}

\section{Related Work}
\label{sec:grafrank_related_work}
We briefly review a few related lines of work on friend recommendations,  graph neural networks, and multi-modal learning.

\textbf{Friend Recommendation:}
The earliest methods were carefully designed \textit{graph-based heuristics} of user-user proximity in social networks~\cite{linkpred}, \textit{e.g.}, path-based Katz centrality~\cite{katz} or common neighbor-based Adamic/Adar \cite{adamic_adar}.
Supervised learning techniques exploited a collection of such extracted pairwise features to train classifiers and ranking models~\cite{heuristic_prediction, baydnn}.
However, computing pairwise heuristics \textit{on-the-fly} for every potential link, is \textit{infeasible} in large-scale evolving social networks.

More recently, \textit{graph embedding} methods to learn latent node representations in graphs that capture the structural properties of a node and its neighborhoods, gained popularity owing to the scalability and efficacy of skip-gram models~\cite{embedding_survey}.
In particular, graph embedding models like node2vec~\cite{node2vec} and Deepwalk~\cite{deepwalk} learn unsupervised node embeddings to maximize the likelihood of co-occurrence in fixed-length random walks, with applications in downstream tasks such as node classification and link prediction.

Graph embedding methods cannot directly incorporate node/link features, and more importantly learn latent embeddings for each node; this implies that number of \textit{model parameters} scales linearly with the \textit{size of the graph}, which is \textit{prohibitive} for large-scale social networks with over multiple millions of users.

\textbf{Graph Neural Networks:}
GNNs learn node representations by recursively propagating features (\textit{i.e.}, message passing) from local neighborhoods through the use of aggregation and activation functions~\cite{gnn_review}. A key feature of GNNs is their ability to learn node representations combining graph topology and node/link attributes.
Graph Convolutional Networks (GCNs)~\cite{gcn} learn degree-weighted neighborhood aggregators by operating on the graph Laplacian.
Many models generalize GCN with a wide range of learnable aggregators, \textit{e.g.}, self-attentions~\cite{gat}, mean and max pooling functions~\cite{graphsage, encoder_decoder}; these approaches have consistently outperformed embedding techniques based upon random walks~\cite{deepwalk, node2vec}.
In contrast to most GNN models that store the entire graph in GPU memory, GraphSAGE~\cite{graphsage} is an inductive variant that reduces memory footprint by sampling a fixed-size set of neighbors in each GNN layer. A few scalable extensions include minibatch training with variance reduction~\cite{variance_reduction, sto_gcn}, subgraph sampling~\cite{graphsaint}, and graph clustering~\cite{clustergcn}.

Despite the successes of GNNs in diverse applications, \textit{very few} industrial systems have developed large-scale GNN implementations. One recent system, PinSage~\cite{pinsage} extends GraphSAGE for Pinterest recommendation on a user-item bipartite graph.
MultiSage~\cite{multisage} extends PinSage to multipartite user-item graphs.

However, GNNs remain unexplored for large-scale \textit{user-user} social modeling applications where users exhibit multifaceted behaviors by interacting with different functionalities on social platforms.
In our work, we design GNNs for the important application of \textit{friend suggestion}, through a \textit{novel multi-faceted friend ranking} formulation with \textit{multi-modal user features} and \textit{link communication features}.

\textbf{Multi-Modal Learning:}
Deep learning techniques for multi-modal feature fusion over diverse modalities such as text, images, video, and graphs, have diverse applications~\cite{user_profiling, rumour}.
Specifically, multi-modal extensions of GNNs have been explored in micro-video recommendation~\cite{micro_video} and urban computing~\cite{urban_computing} applications.
In contrast to prior work that regard modalities as largely \textit{independent} data sources, user feature modalities in social networks tend to be correlated.
Exploiting knowledge from multiple channels have been shown to benefit recommendations in retail and e-commerce~\cite{multi_channel}.
In this work, we propose a novel \textit{cross-modality attention} layer that captures non-linear modality correlations, to learn \textit{multi-faceted} user representations.

\section{Preliminaries}
\label{sec:grafrank_prob_defn}

We first formulate the problem of \textit{multi-faceted friend ranking} in large-scale social networking platforms (Section~\ref{sec:grafrank_prob_defn}) and then briefly introduce relevant background and notations on \textit{graph neural networks} (Section~\ref{sec:grafrank_base_gnn}).

\subsection{Problem Formulation}
In this section, we introduce the different information sources in a social networking platform, that are relevant to friend recommendation.
Each registered individual in the platform is denoted by a \textit{user} $u$ or $v$ and a pair of users $(u,v)$ may be connected by a \textit{friendship}, which is an undirected relationship, \textit{i.e.}, if $u$ is a friend of $v$, $v$ is also a friend of $u$.
We assume that the social network has a set of $N$ users $\gV$
introduced until our latest observation time of the platform.
The friendship graph $\gG$ evolves when new friendships form and when new users join the platform.
In this work, we only consider the emerging of new users and friendships while leaving the removal of existing users and edges as future work.

Prior work typically represent a dynamic network as a sequence of static snapshots, primarily due to scaling concerns.
However, graph snapshots are coarse approximations of the actual continuous-time network and rely on a user-specified discrete time interval for snapshot creation~\cite{graph_snapshot, shah2015timecrunch}.
We also assume multiple time-aware user-level features (across modalities) and link(edge)-level features capturing pairwise user-user communications. In industrial settings, such features are commonly extracted by routine batch jobs and populated in an upstream database at regular time intervals (\textit{e.g.}, daily batch inference jobs), to facilitate efficient model training.%

Thus, we adopt a hybrid data model that achieves the best of both worlds.
We formulate the \textit{friendship graph} as a continuous-time dynamic graph (CTDG) with the expressivity to record friendships at the finest possible temporal granularity; and represent \textit{features} as a sequence of daily snapshots where the time-sensitive features (\textit{e.g.}, engagement activity) are recorded at different time scales.

\textbf{Friendship Graph}:
Let us consider an observation time window $(t_s, t_e)$
such that friendships created in this window specify the training data for the friend ranking model.
We divide this window $(t_s, t_e)$ into a sequence of $S$ daily snapshots, denoted by $1, 2, \dots, S$. %
Formally, we model the friendship graph $\gG$ as a timed sequence of friend creation events over the entire time range $(0, t_e)$, defined as:

\begin{definition}[Friendship Graph]
Given a graph $\gG = (\gV, \gE, \gT)$, let $\gV$ be the set of users, $\gE \subseteq \gV \times \gV \times \sR$ be the set of friendship links between users in $\gG$. At the finest granularity, each link $e = (u,v,t) \in \gE$ is assigned a unique timestamp $t \in \sR^{+} ; 0 < t < t_e$ that denotes the link creation time, and $\gT: \sR^{+} \mapsto [0, S] $ is a function that maps each timestamp $t$ to a corresponding snapshot.
\end{definition}

Note that the time window $(t_s, t_e)$ corresponds to snapshots $[1, S]$, and snapshot 0 is a placeholder for any $t < t_s$.
However, friendship graph $\gG$ includes the entire set of friendships created with all time-stamped links in $(0, t_e)$.
The set of temporal neighbors of user $v$ at time $t$ includes all friends created before $t$, defined as $N_t(v) = \{ w: e = (v, w, t^{'}) \in \gE \wedge t^{'} < t \}$.

\textbf{Multi-Modal Evolving User Features:}
In a social network, users typically use multiple features, such as posting videos, exchanging messages or pictures with friends, or liking and sharing posts and articles, which can be indicative of both their stable traits and mutable interests.
Since users with shared interests tend to form social connections due to homophily~\cite{homophily},
we extract user attributes spanning a total of $K=4$ modalities, which include profile attributes, in-app interests, friend creation activities, and user engagement activities, described in detail below:
\begin{description}
    \item \textit{Profile Attributes}: a set of (mostly) static demographic features describing the user, including age, gender, languages, etc., that are listed or inferred from their profile.
    \item \textit{Content Interests}: a real-valued feature vector describing the textual content (\textit{e.g.}, posts, news) interacted by the user within the platform, \textit{e.g.}, topics of posts liked by the user on Facebook.
    \item \textit{Friending Activity}: this feature modality records the aggregated number of friend requests sent/received, reciprocated friendships, and viewed suggestions made by the user in different time ranges (\textit{e.g.}, daily, weekly, and monthly).
    \item \textit{Engagement Activity}: this modality records the aggregated number of in-app direct and indirect engagements made by the user (\textit{e.g.}, direct messages, pictures, and comments on posts) with other friends in different time ranges.
\end{description}

The user feature modalities include a combination of \textit{static} and \textit{time-sensitive} features, \textit{i.e.}, the profile attributes are static while the rest of the modalities are time-sensitive and often evolve at different scales across users, \textit{e.g.}, a long-time active user may frequently communicate with a stable set of friends, while a new user is more likely to quickly add new friends before communicating.

The feature vector of a user $u \in \gV$ in snapshot $s \in [1, S]$ is defined by $\rvx_v^s = [ \rvx_v^{s, 1}, \dots, \rvx_v^{s, K}]$ where $\rvx_v^{s, k} \in \sR^D_k$ is the $k$-th user feature modality and $[\cdot]$ denotes row-wise concatenation.

\textbf{Pairwise Link Communication Features}:
Social networks evolve centering around communication artifacts, with two predominant types of online communication channels: \textit{conversations} and \textit{social actions}.
Conversations can include exchanges of text messages, pictures, and video content with friends, which indicate direct user-user interactions via communication.
In contrast, social actions facilitate indirect user interactions through a variety of actions, \textit{e.g.}, posting a Snapchat Story (image or video content) or liking a Facebook post, results in a passive broadcast to friends.

For conversational channels, we extract bidirectional edge features between users and their friends reflecting the number of communications sent and received by each pair of users. We capture indirect social actions by recording the number of user actions associated with each friend.
Similar to user features, we extract edge features per snapshot by aggregating communications at different time intervals.
The link feature vector for a pair of users $(u, v)$ at time $t$ (who became friends before $t$), is denoted by
$\rve^{s}_{uv} \in \sR^F$ where $s = \gT(t)$ is the snapshot corresponding to timestamp $t$, and $E$ is the number of extracted link communication features.

We now formally define the problem of \textit{multi-faceted friend ranking} in large-scale social platforms, over friendship graph $\gG$ with multi-modal user features and pairwise link features.

\begin{problem}[\textbf{Multi-Faceted Friend Ranking}]
Our objective is to leverage the multi-modal user features $\{ \rvx^s_v : v \in \gV, 1 \leq s \leq S \}$,
link communication features $\{ \rve^s_{uv} : s = \gT(t) \;, \; (u,v,t) \in \gE \} $ and the structure of the friendship graph $\gG$, to generate temporal user representations $\{\rvh_v (t) \in \sR^D : v \in \gV \}$ at time $t$, that can be used for friend ranking, \textit{e.g.}, via nearest-neighbor lookup for candidate retrieval, or friend candidate re-ranking.
\end{problem}

\subsection{{Background on GNNs}}
\label{sec:grafrank_base_gnn}
We briefly introduce a generic formulation of a graph neural network layer with message-passing for neighborhood aggregation.

GNNs use multiple layers to learn node representations. At each layer $l > 0$ ($l = 0$ is the input layer), GNNs compute a representation for node $v$ by aggregating features from its neighborhood, through a learnable aggregator $F_{\theta, l}$ per layer. Stacking $k$ layers allows the $k$-hop neighborhood of a node to influence its representation.

\begin{equation}
\mathbf{h}_{v,l} = F_{\theta,l} \Big( \mathbf{h}_{v, l-1}, \{  \mathbf{h}_{u, l-1} \} \Big), \quad u \in N(v)
\label{eq:grafrank_basic_agg}
\end{equation}

\Cref{eq:grafrank_basic_agg} indicates that the node embedding $\mathbf{h}_{v,l} \in \mathbb{R}^{D}$ for node $v$ at the $l$-th layer is a non-linear aggregation $F_{\theta, l}$ of the embeddings from layer $l-1$ of node $v$ and the embeddings of immediate network neighbors $u \in \mathcal{N}(v)$ of node $v$. The function $F_{\theta, l}$ defines the message passing function at layer $l$ and can be instantiated using a variety of aggregator architectures, including graph convolution~\cite{gcn}, graph attention~\cite{gat}, and pooling~\cite{graphsage}. The node representation for $v$ at the input layer is $\mathbf{h}_{v, 0}$, where $\mathbf{h}_{v, 0} = \rvx_v$ and $\rvx_v \in \mathbb{R}^{D}$.
The representation of node $v$ at the final GNN layer is typically trained using a supervised objective.

The above formulation (Equation~\ref{eq:grafrank_basic_agg}) operates under the setting of a static graph and assumes a single static input feature vector for each node. In contrast, our social network has a time-evolving friendship graph $\gG$ with pairwise link features and multi-modal node (user features), which poses unique modeling challenges.
In the next section, we present a graph neural network model for multi-faceted friend ranking. 
\section{Graph Neural Friend Ranking}
\label{sec:grafrank_methods}

\begin{figure}
    \centering
    \includegraphics[width=0.9\linewidth]{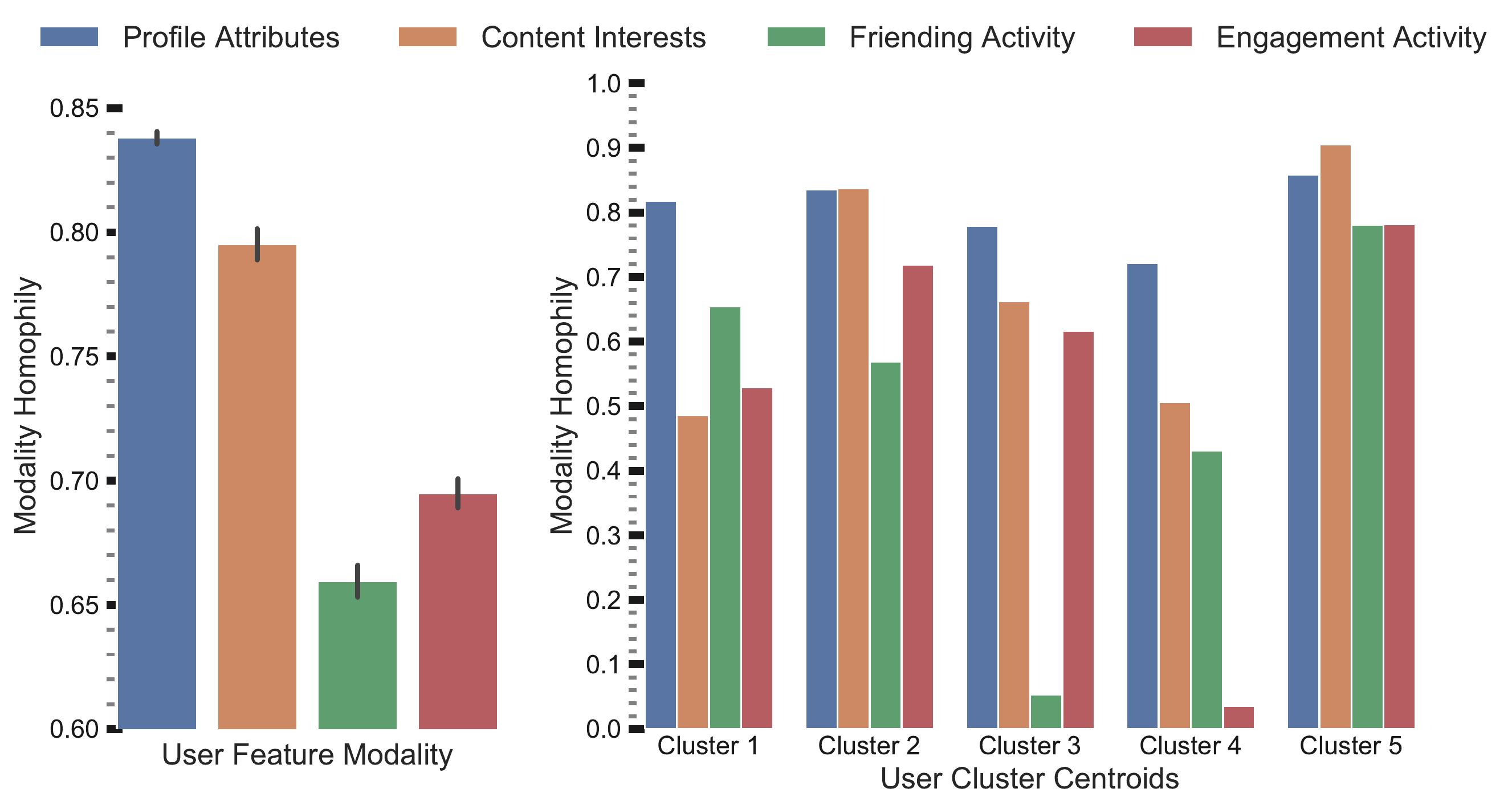}
    \caption{Users exhibit different extents of homophily across feature modalities. (a) Overall modality homophily scores, with 95\% confidence interval bands (b) five representative cluster centroids identified by clustering users based on their homophily distributions over the $K$ modalities. }
    \label{fig:grafrank_modality_kmeans}
\end{figure}

In this section, we present our approach to inductively learn user representations in a dynamic friendship graph with pairwise link features and multi-modal node features.

We first conduct an empirical analysis on user feature modalities, to gain insights into various factors impacting friendship formation (Section~\ref{sec:grafrank_motivation}).
We then formulate the design choices of our model \grafrank for friend ranking grounded on our acquired insights (Section~\ref{sec:grafrank_grafrank}), followed by model training details (Section~\ref{sec:grafrank_training})

\subsection{{Motivating Insight: Modality Analysis}}
\label{sec:grafrank_motivation}
We conduct an empirical study that helps us formulate the design choices in our model.
We aim to validate the existence and understand the extent and variance of attribute homophily with respect to the different user feature modalities.
We begin by analyzing users' ego-networks to characterize \textit{modality homophily}, both overall and broken-down across different user segments.
The definition of \textit{modality homophily} echoes the standard definition of attribute homophily~\cite{homophily}, but generalized to include a modality of attributes, \textit{i.e.,} the tendency of users in a social graph to associate with others who are \textit{similar} to them along attributes of a certain modality.

We define a homophily measure $\rvm_{vu}^{k}$ between a user $v$ and her friend $u$ on modality $k$ by the standard cosine similarity~\cite{homophily_cosine},
which is a normalized metric that accounts for heterogeneous activity across users.
We compute a modality homophily score $\rvm_u^{k}$ for user $v$ on modality $k$ by the mean over all her neighbors, defined by:

\begin{equation}
  \rvm_v^{k} = \frac{1}{|N_v|} \sum\limits_{u \in N_v} \rvm_{vu}^{k}  \hspace{15pt} \rvm_{uv}^{k} =  cos(\rvx^{k}_u, \rvx^{k}_v)
\end{equation}

Note that, we omit the snapshot $s$ above since the discussion is restricted to a single feature snapshot.
Figure~\ref{fig:grafrank_modality_kmeans} (a) shows the overall modality homophily scores (averaged across all users), for each of the $K$ modalities.
We observe differing extents of attribute homophily across modalities, with higher variance for the time-sensitive modalities (\textit{e.g.}, friending and engagement activities).

We further extend our analysis to examine the \textit{homophily distribution} over modalities at the granularity of individual users, to understand if modality homophily varies across different users.
To study this, we first represent each user $u$ by a $K$-dimensional \textit{modality vector} $\rvm_{u} =  [\rvm_u^{1}, \dots, \rvm_u^{K}]$ that describes the homophily distribution over the $K$ feature modalities.
We then use $k$-means~\cite{kmeans} to cluster the set of all users based on their computed modality vectors; Figure~\ref{fig:grafrank_modality_kmeans} (b) shows the centroid vectors of five representative clusters identified by $k$-means.
We observe stark differences in the modality vector centroids across the five clusters, indicating the existence of user segments with diverse homophily distributions over the $K$ modalities.
This motivates our first key observation:
\begin{hypothesis}[\textbf{Heterogeneity in Modality Homophily}]
Users exhibit different extents of homophily across feature modalities, and the homophily distribution over modalities varies significantly across diverse user segments.
\label{hyp:hyp_1}
\end{hypothesis}

Each modality enables identification of a subset of friends that exhibit modality homophily. However, this poses a question:
do the $K$ modalities induce the same (or disparate) subsets of homophilous friends, or are the friends that exhibit homophily in each modality, correlated? We investigate this relationship in this section.

For every feature modality $k$, we cluster the ego-network (set of direct friends) $N(v)$ of each user $v \in \gV$ by representing it as a set of modality-specific feature vectors $\{ \rvx^k_u \in \sR^{D_k} : u \in N(v)\}$; this results in ego-clustering assignments for each modality $k$.
To quantify \textit{cross-modality correlations}, we compute a \textit{correlation} score for each pair of modalities, measured by the \textit{consensus} between their corresponding ego-clustering assignments. %

We use two standard measures: Normalized Mutual Information (NMI) and Adjusted Rand Index (ARI) to evaluate consensus between clusterings~\cite{clustering_evaluation}.
NMI measures the statistical correlation between two clustering assignments; however, NMI increases with the number of distinct clusters. ARI measures the percentage of correct pairwise assignments, and is chance-corrected with an expected value of zero. Note that NMI and ARI are symmetric metrics.

Figure~\ref{fig:grafrank_cross_modality} depicts average NMI and ARI scores for each pair of feature modalities. We observe substantial correlation in cluster assignments across a few modalities (\textit{e.g.}, time-sensitive modalities $M_1$ and $M_3$) while some (\textit{e.g.,} static modality $M_0$) are quite distinct from the rest. Our key takeaway regarding modality correlation is:

\begin{figure}[t!]
    \centering
    \includegraphics[width=0.9\linewidth]{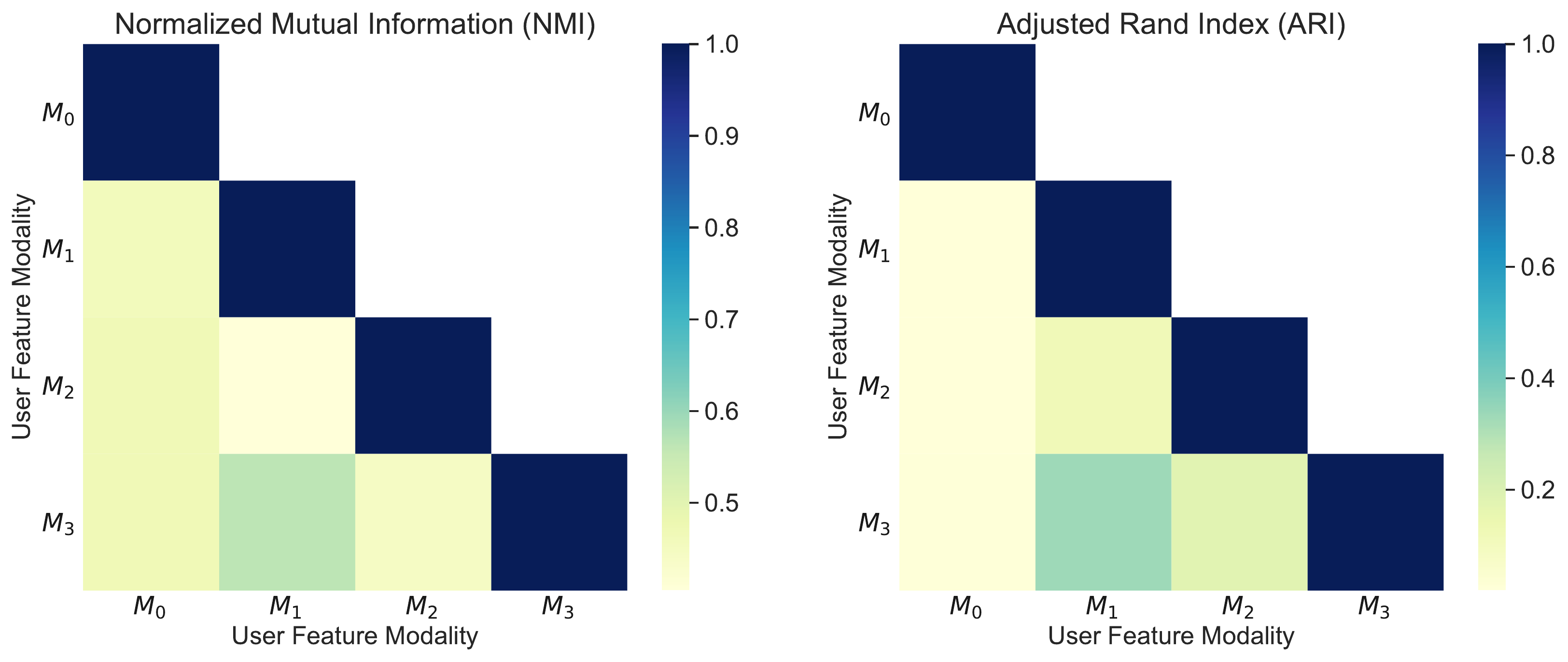}
    \caption{Cross-modality Correlation Study: NMI (a) and ARI (b) metrics for each pair of modalities, quantifying pairwise correlation by consensus in ego-clustering assignments (obtained independently with respect to each modality). We observe substantial correlations across pairs of modalities.}
    \label{fig:grafrank_cross_modality}
\end{figure}

\begin{hypothesis}[\textbf{Cross-Modality Correlation}]
There exist non-trivial correlations between pairs of feature modalities, indicated
by the consensus in their induced ego-network clustering of users.
\label{hyp:hyp_2}
\end{hypothesis}

\subsection{{\grafrank: Multi-Faceted Friend Ranking}}
\label{sec:grafrank_grafrank}
\begin{figure*}[t]
    \centering
    \includegraphics[width=0.9\linewidth]{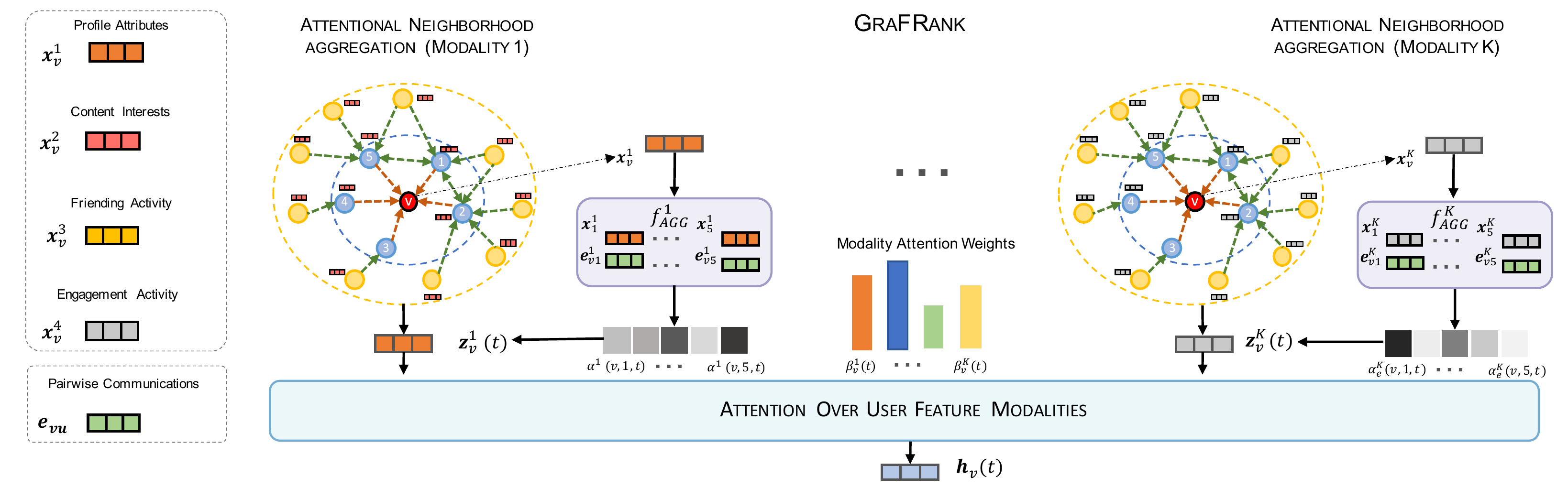}
    \caption{Overall framework of \grafrank: a set of $K$ modality-specific neighbor aggregators (parameterized by individual modality-specific user features and link communication features) to compute $K$ intermediate user representations; cross-modality attention layer to compute final user representations by capturing discriminative facets of each modality.}
    \label{fig:grafrank_grafrank}
\end{figure*}

In this section, we first introduce the key components of our model \grafrank~(Graph Attention Friend Ranker) for friend ranking. Our modeling choices in designing a multi-modal GNN, follow from our observations.~\grafrank~has three modules (Figure~\ref{fig:grafrank_grafrank}):
\begin{itemize}
    \item Modality-specific neighbor aggregation.
    \item Cross-modality attention layer.
    \item Pairwise ranking objective for model training.
\end{itemize}
Below, we present a detailed description of each module.

\subsubsection{\textbf{Modality-specific Neighbor Aggregation}}
Since the feature modalities vary in the extent of induced homophily (Observation~\ref{hyp:hyp_1}), we treat each modality individually as opposed to the popular choice of directly combining multi-modal user features by concatenation.
Thus, our objective is to learn a modality-specific representation $\rvz^{k}_v (t) \in \sR^{D}$ for each user $v \in \gV$ at time $t \in \sR^{+}$, that encapsulates information exclusively from modality $k$.
We endow each user $v$ with the expressivity to flexibly prioritize different friends in her temporal neighborhood $N_t(v)$, hence accounting for the variance in homophily distribution across user segments.

To achieve this, we design a \textit{modality-specific neighbor aggregation} module to compute $K$ representations $\{\rvz_v^{1} (t), \dots, \rvz_v^{K} (t) \}, \rvz_v^{k} (t) \in \sR^D$ for each user $v$ at time $t \in \sR^{+}$, where each $\rvz_v^{k} (t) $ is obtained using an independent and unique message-passing function per modality. %
We begin by describing a single layer, which consists of two major operations:
\textit{message propagation} and \textit{message aggregation}.
We then discuss generalization to multiple successive layers.
\begin{figure}[t!]
    \centering
    \includegraphics[width=0.9\linewidth]{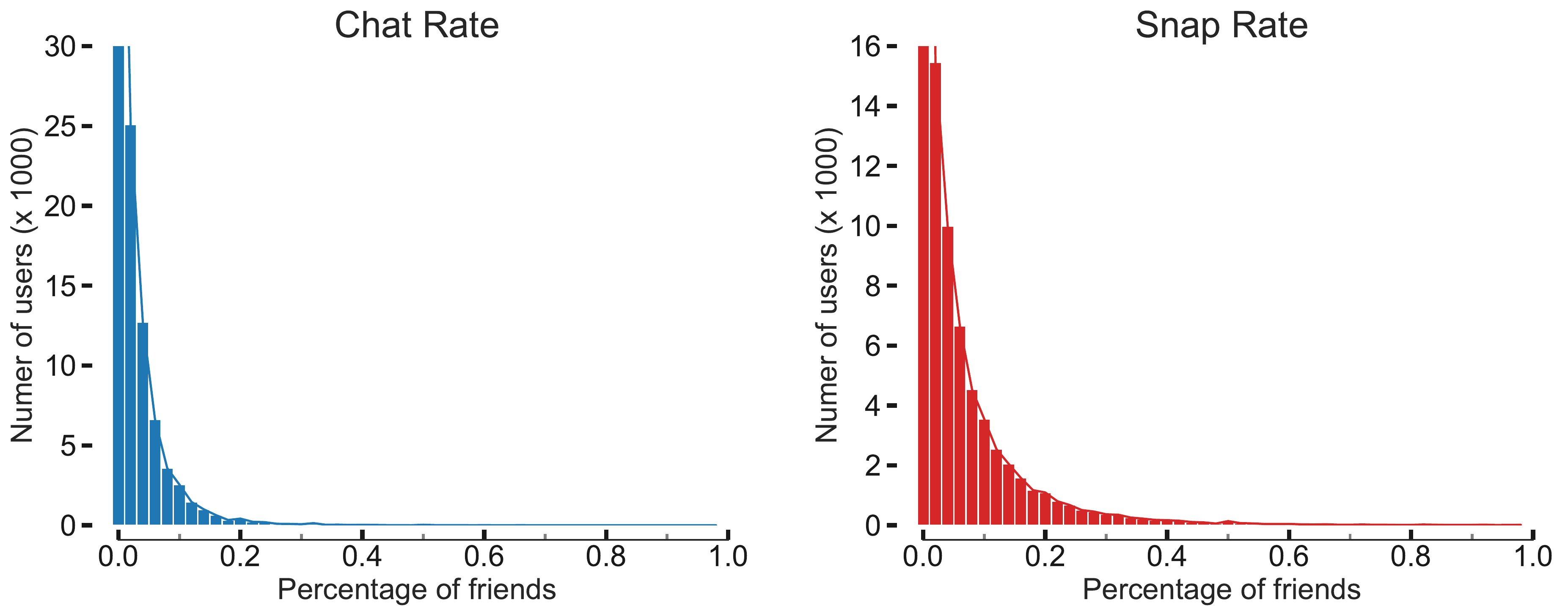}
    \caption{Friend communication rate in two direct channels (Chat, Snap) on Snapchat. Most users communicate frequently only with a subset ($\leq20$\%) of their friends, making influence modeling  critical during neighbor aggregation.}
    \label{fig:grafrank_communication_rate}
\end{figure}

\textbf{Message Propagation:}
We define the message-passing mechanism to aggregate information from the ego-network $N_t(v)$ of user $v$ at time $t$. Specifically, the propagation step for modality $k$ aggregates the $k$-th modality features $\{ \rvx^{s, k}_u : u \in N_t(v), s = \gT(t) \}$ from the corresponding snapshot $s = \gT(t)$ of temporal neighbors $N_t(v)$.
To quantify the importance of each friend $u$ in the ego-network, we propose a friendship attention~\cite{gat} which takes embeddings $\rvx^{s, k}_u$ and $\rvx^{s, k}_v$ as input, and computes an attentional coefficient $\alpha^{k} (v, u, t)$ to control the influence of friend $u$ on $v$ at time $t$.

\begin{equation}
    \alpha^{k} (v, u, t) = \text{LeakyRELU} \Big( \rva_k^T  \big( \mW^{k}_1 \rvx^{s, k}_{v} \; || \; \mW^{k}_1 \rvx^{s, k}_{u} \big) \Big)  \hspace{10pt} s = \gT (t)
    \label{eqn:grafrank_node_attn}
\end{equation}

where $\mW^{k}_1 \in \sR^{D_k \times D}$ is a shared linear transformation applied to each user, $||$ is the concatenation operation, and the friendship attention is modeled as a  single feed-forward layer parameterized by weight vector $\rva_k \in \sR^{2D}$ followed by the LeakyReLU nonlinearity.
We then normalize the attentional coefficients across all friends connected with $v$ at time $t$ by adopting the softmax function:

\begin{equation}
\alpha^{k} (v, u, t) = \frac{\exp \big(\alpha^{k}(v, u, t) \big) }{\sum_{w \in N_t (v)} \exp \big(\alpha^{k} (v, w, t) \big) }
        \label{eqn:grafrank_node_attn_softmax}
\end{equation}

We now define an ego-network representation $\rvz^{k}_v \big(t, N_v(t)\big) \in \sR^{D_k}$ for user $v$ in modality $k$ that captures messages propagated from first-order neighbors in the ego-network $N_v(t)$.
The message $\rvm^{k}_{v \leftarrow u} \in \sR^D$ propagated from friend $u$ to user $v$ at time $t$ is defined as the transformed friend embedding, \textit{i.e.},  $\rvm^{k}_{v \leftarrow u} = \mW^k_1 \rvx^{s, k}_v$.
We then compute $\rvz^{k}_v \big(t, N_v(t) \big)$ through a weighted average of message embeddings from each friend $u \in N_v(t)$ and guided by normalized friendship weights $\alpha^{k} (u, v, t)$, which is defined as:

\begin{equation}
   \rvz^{k}_v \big(t, N_v(t)\big) = \sum_{u \in N_v(t)} \alpha^{k} (v, u, t) \rvm^{k}_{v \leftarrow u}
\label{eqn:grafrank_node_emb}
\end{equation}

While the above equation specifies different contributions to friends, these weights are learnt merely based on the connectivity structure of the ego-network.
In reality, most users only have a few close friends, and users with many friends only frequently communicate with a few of them.
To empirically validate this hypothesis, we compute the \textit{friend communication rate} of all users, defined by the percentage of friends that a user has directly communicated with at least once (directly sent a Chat/Snap with on Snapchat) in a one-month window.
From Figure~\ref{fig:grafrank_communication_rate}, we find that a vast majority of users communicate primarily with a small percentage (10-20\%) of their friends; thus, we posit that \textit{friendship activeness} is critical to precisely model user affinity. %

Towards this goal, we incorporate \textit{pairwise link communication} features to parameterize both the attentional coefficients and the message aggregated from friends in the ego-network.
Specifically, we formulate the message $\rvm^{k}_{v \leftarrow u} \in \sR^D$ from friend $u$ to user $v$ at time $t$ as a function of both friend feature $\rvx^{s, k}_{u}$ and link feature $\rve^{s}_{vu}$. %

\begin{equation}
\rvm^{k}_{v \leftarrow u} = \mW^{k}_2  \rvx^{s, k}_{u} + \mW^{k}_e \rve^{s}_{vu} + \rvb  \qquad s = \gT (t)
\label{eqn:grafrank_edge_message}
\end{equation}

where $\mW^{k}_2 \in \sR^{D_k \times D}, \mW^{k}_e \in \sR^{E \times D} $ are trainable weight matrices operating on the user and link features respectively, and $\rvb \in \sR^D$ is a learnable bias vector.
The attentional co-efficient $\alpha^{k} (v, u, t)$ is then computed as a function of user feature $\rvx_v^{s, k}$ and message embedding $\rvm^{k}_{v \leftarrow u} \in \sR^D$ from friend $u$ to user $v$, defined by:

\begin{equation}
 \alpha^{k} (v, u, t) = \sigma \bigg( \rva_k^T  \Big( \mW^{k}_1 \rvx^{s, k}_{v} \; || \; \rvm^{k}_{v \leftarrow u} \Big) \bigg)  \qquad s = \gT (t)
\label{eqn:grafrank_edge_attn}
\end{equation}

where $\sigma$ is a non-linearity such as LeakyRELU.
We similarly normalize the attentional coefficients $\alpha^{k} (v, u, t)$ using Equation~\ref{eqn:grafrank_node_attn_softmax} and compute the ego-network representation $ \rvz^{k}_v (t, N_v(t))$ for user $v$ on modality $k$ using Equation~\ref{eqn:grafrank_node_emb} with the message embedding $\rvm^{k}_{v \leftarrow u}$ from Equation~\ref{eqn:grafrank_edge_message} conditioned on link features.

Distinct from conventional graph convolution and attention networks~\cite{gcn, graphsage, pinsage, gat} that only consider user $\rvx^{s, k}_v$ and friend $\rvx^{s, k}_v$ features to parameterize the neighbor aggregation, we additionally encode the link communication features $\rve^{s}_{vu}$ into the attentional co-efficient $\alpha^{k} (v, u, t)$ (Equation~\ref{eqn:grafrank_edge_attn}) and the message $\rvm^{k}_{v \leftarrow u}$ (Equation~\ref{eqn:grafrank_edge_message}) passed from friend $u$; this encourages the aggregation to be cognizant of the pairwise communications with friends, \textit{e.g.}, passing more messages from the active friendships.
In addition, we observe a boost in friend ranking performance (Section~\ref{sec:grafrank_ablation}) owing to our novel communication-aware message passing strategy.

\textbf{Message Aggregation:} We refine the latent representation for user $v$ by aggregating the messages propagated from friends in $N_v(t)$.
In addition to the messages from $N_v(t)$, we additionally consider the self-connection of $v$, \textit{i.e.}, $\rvm^k_{v \leftarrow v} = \mW^k_1 \rvx_v^{s, k}$ to preserve knowledge of the original features ($\mW_1$ is the same transformation used in Equation~\ref{eqn:grafrank_node_attn}).
Specifically, we concatenate the ego-network and self-representations of user $v$, and further transform the concatenated embedding through a dense layer $F_{\theta}^{k}$, defined by:

\begin{align}
\rvz^{k}_v (t) \quad &=  \quad F^{k}_{\theta} \Big( \rvm^k_{v \leftarrow v} \; ,  \rvz^{k}_v \big(t, N_v(t)\big)  \Big) \\
\quad &=  \quad \sigma \bigg( \mW_a^{k} \Big(  \rvz^{k}_v (t, N_v(t)) \; || \; \rvm^k_{v \leftarrow v} \Big) + \rvb_a \bigg) \;\; %
\label{eqn:grafrank_message_concat}
\end{align}

where $\mW_a^k \in \sR^{D \times D}$, $\rvb_a \in \sR^D$ are trainable weight parameters of the message aggregator, and $\sigma$ denotes the ELU activation function, which allows messages to encode both positive and small negative signals.
Empirically we observe significant gains owing to retaining knowledge of self-representation via concatenation, instead of directly using the propagated ego-network representation from the neighborhood, as in GCN~\cite{gcn}, and GAT~\cite{gat}.

\textbf{Higher-order Propagation:}
We stack multiple aggregation layers to model high-order connectivity information, \textit{i.e.}, propagate features from $l$-hop neighbors.
The inputs to layer $l$ depend on the user representations output from layer $(l-1)$ where the initial (\textit{i.e.}, “layer 0”) representations are set to the input user features in modality $k$.
By stacking $l$ layers, we recursively formulate the representation $\rvz^{k}_{v, l}$ of user $v$ at the end of layer $l$ by:

\begin{equation}
\rvz^{k}_{v, l} = F^{k}_{\theta, l} \Big(\rvm^k_{v \leftarrow v, l-1} ,  z^{k}_{v, l-1} \big(t, N_v(t) \big)  \Big) \quad  \rvm^k_{v \leftarrow v, l-1} = z^{k}_{v, l-1}
\label{eq:grafrank_multi_hop}
\end{equation}

where $z^{k}_{v, l-1}$ is the representation of user $v$ in modality $k$ after $(l-1)$ layers and
$z^{k}_{v, l-1} (t, N_v(t))$ denotes the $(l-1)$-ego-network representation of user $v$.
We apply $L$ neighbor aggregation layers to generate the layer-$L$ representation $\rvz^{k}_{v, L}$ of user $v$ in modality $k$.

\subsubsection{\textbf{Cross Modality Attention}}
To learn complex non-linear correlations across feature modalities, we design a \textit{cross-modality attention} mechanism. Specifically, we learn modality attention weights $\beta^{k}_{v} (t)$ to distinguish the influence of each modality $k$ using a two-layer Multi-Layer Perceptron, by:

\begin{equation}
\beta^{k}_v (t) = \frac{\exp \big( \rva^T_m  \; \mW_m \rvz^{k}_{v, L} + b_m  \big) } { \sum_{k^{\prime}=1}^K  \exp \big( \rva^T_m \; \mW_m \rvz^{k^{\prime}}_{v, L} + b_m \big) }
\label{eq:grafrank_cross_attention}
\end{equation}
with weights $\mW_m \in \sR^{D \times D}, \rva_m \in \sR^D$ and scalar bias $b_m$.
The final representation $\rvh_v (t) \in \sR^D$ of user $v$ is computed by a weighted aggregation of the layer-$L$ modality-specific user representations $\{\rvz^{1}_{v, L}, \dots, \rvz^{K}_{v, L}\}$, guided by modality weights $\beta^{k}_{v} (t)$, defined by:

\begin{equation}
 \rvh_v (t) = \sum\limits_{k=1}^K \beta^{k}_v (t) \; \mW_m \rvz^{k}_{v, l}
\label{eq:grafrank_final_emb}
\end{equation}

\subsection{{Model Training}}
\label{sec:grafrank_training}
We train \grafrank using a pairwise ranking objective to differentiate \textit{positive} and \textit{negative} neighborhoods.
We assume access to training data described by a set of timestamped links $\gL$ created in a time span $(t_s, t_e)$, where $(v, u, t) \in \gL \; $ indicates a bi-directional link between \textit{source} user $u$ and \textit{target} friend $v$ formed at time $t \in (t_s, t_e)$.
To learn the parameters of \grafrank, we define a triplet like training objective based on max-margin ranking.

\subsubsection{\textbf{Pairwise Ranking Objective}}
We define a time-sensitive ranking loss over the user embeddings ($\rvh_v (t)$ for user $v$ at time $t$) to rank the inner product of positive links $(v, u, t) \in \gL$, higher than sampled negatives $(v, n, t)$ by a margin $\Delta$, as:

\begin{equation}
    L = \sum\limits_{(u, v, t) \in \gL} \E_{n \sim P_n (v)} \max \{0, \rvh_v (t) \cdot \rvh_n (t) -  \rvh_v (t) \cdot \rvh_u (t) + \Delta \}
    \label{eq:grafrank_loss}
\end{equation}

where $\Delta$ is a margin hyper-parameter and $P_n (v)$ is the negative sampling distribution for user $v$. This generic contrastive learning formulation enables usage of the same framework to learn user embeddings for different recommendation tasks such as candidate retrieval and ranking, with diverse negative sampling distributions.

\subsubsection{\textbf{Friend Candidate Retrieval and Ranking Tasks}}
We learn user embeddings towards two key use-cases in friend ranking: \textit{candidate retrieval} and \textit{candidate ranking}.
Candidate retrieval aims to generate a list of top-$N$ (\textit{e.g.}, $N=100$) potential friend suggestions out of a very large candidate pool (over millions of users), while candidate ranking involves fine-grained re-ranking within a much smaller pool of the generated candidates, to determine the top-$n$ (\textit{e.g.}, $n=10$) suggestions shown to end users in the platform. We define different negative sampling distributions $P_n (v)$ for each task owing to their different ranking granularties, as follows:

\begin{description}
    \item \textit{Candidate Retrieval}: For the \textit{coarse-grained} task of candidate retrieval, we uniformly sample five \textit{random negative} users for each positive link, from the entire user set $\gV$. Generating random negatives is efficient and effective at quickly training the model to identify potential friend candidates for each user. However, random negatives are often too \textit{easy} to distinguish and may not provide the requisite resolution for the model to learn \textit{fine-grained distinctions} necessary for candidate friend ranking.
\item \textit{Candidate Ranking}: To enhance model resolution for candidate ranking, we also consider \textit{hard negative} examples for each positive pair $(u, v)$ that are related to the source $u$, but not as relevant as the target friend $v$.
For each user $u$, we generate five hard negatives based on graph proximity. Specifically, users in $k$-hop ($3 \leq k \leq 4$) graph neighborhoods of $u$ (that are not within the 2-hop neighborhood), are randomly sampled as hard negatives. In practice, we pre-compute a set of hard negatives for each user, to facilitate efficient negative sampling during model training.

We adopt a \textit{two-phase} learning approach for candidate ranking. We pre-train the model on random negatives (as in candidate retrieval), to identify good model initialization points, followed by fine-tuning on hard negatives. Ranking hard negatives is more challenging, hence encouraging the model to progressively learn friend distinctions at a finer granularity. We empirically show notable gains on candidate ranking due to our two-phase strategy, compared to training individually on random or hard negatives.

\end{description}

\subsubsection{\textbf{Temporal Neighborhood Sampling}}
We learn a temporal user representation $\rvh_v (t)$ for user $v$ at time $t$ by selecting a fixed number of friends from $N_v(t)$ for neighbor aggregation at each layer; this controls the memory footprint during training~\cite{graphsage}.

To efficiently identify and sample neighbors of $v$ at any time $t$, we represent the time-evolving friendship graph $\gG$ as a \textit{temporal adjacency list} at its latest time $t_s$ where each user $v$ has a list of (friend, time) tuples sorted by link creation times.
This data representation enables $O (\log d)$ neighbor lookup at an arbitrary timestamp $t$ via binary search where $d$ is the average user degree in the graph.

\subsubsection{\textbf{Multi-GPU Minibatch Training}}
We train \grafrank with minibatches of links from $\gL$ using multiple GPUs on a single shared memory machine.
The temporal adjacency list of $\gG$ and feature matrices $\mX, \mE$ are placed in shared CPU memory to enable fast parallel neighborhood sampling and feature lookup.
We adopt a \textit{producer-consumer} architecture~\cite{pinsage} that alternates between CPUs and GPUs for model training.
A CPU-bound producer constructs friend neighborhoods, looks up user and link features, and generates negative samples for the links of a minibatch.
We then partition each minibatch across multiple GPUs, to compute forward passes and gradients with a PyTorch model over dynamically constructed computation graphs. The gradients from different GPUs are synchronized using PyTorch's Distributed Data Parallel~\cite{pyt_distributed}.

\section{Experiments}
\label{sec:grafrank_experiments}

To analyze the quality of user representations learned by \grafrank, we propose five research questions to guide our experiments:

\begin{enumerate}[label=(\subscript{\textbf{RQ}}{{\textbf{\arabic*}}}), leftmargin=*]
\item Can \grafrank outperform feature-based models and state-of-the-art GNNs on candidate \textit{retrieval} and \textit{ranking} tasks?
\item How does \grafrank compare to prior work under alternative metrics of \textit{reciprocated} and \textit{communicated} friendships?
\item How do  different \textit{architectural} design choices and \textit{training} strategies in \grafrank impact performance?
\item How do training strategies and hyper-parameters impact convergence and performance of \grafrank?
\item How do the learned user embeddings in \grafrank perform across diverse user cohorts?
\end{enumerate}

\renewcommand*{\factor}{0.2}
\begin{table}[t]
\centering
\noindent\setlength\tabcolsep{2.3pt}
\begin{tabular}{@{}p{0.4\linewidth}@{\hspace{10pt}}
K{\factor\linewidth}K{\factor\linewidth}@{\hspace{10pt}}}
\toprule
\multirow{1}{*}{\textbf{Dataset}} &  \textbf{Region 1} & \textbf{Region 2}  \\
\midrule
\textbf{\# users} & 3.1 M & 17.1 M   \\
\textbf{\# links} & 286 M & 2.36 B \\
\textbf{\# user features} &  79 & 79 \\
\textbf{\# link features} &  6 & 6  \\
\textbf{\# test set friend requests} &  46K & 340K \\
\bottomrule
\end{tabular}
\caption{Dataset statistics}
\label{tab:grafrank_stats}
\end{table}

\subsection{{Experimental Setup}}
We now present our experimental setup with a brief description of datasets,  evaluation metrics, and model training details.
\subsubsection{\textbf{Datasets}}
We evaluate the user representations learned by \grafrank for friend recommendations, on two large-scale datasets from Snapchat. Each dataset is constructed from the interactions among Snapchat users belonging to a specific country (details obscured for privacy reasons).
In total, we collect 79 user features spanning four feature modalities and six pairwise link communication features, as described in Section~\ref{sec:grafrank_prob_defn}.
All features are standardized using zero-mean and unit-variance normalization before model training.
We define the set $\gL$ of positive training examples (Equation~\ref{eq:grafrank_loss}) using all timestamped friendships created over a span of one week.
Since our objective is to evaluate the quality of friendship recommendations, the test set comprises all \textit{friend add requests} generated by users over the next four days. We use 10\% of the labeled examples as a validation set for hyper-parameter tuning.
Table~\ref{tab:grafrank_stats} shows the detailed statistics of the two datasets from Snapchat.

\subsubsection{\textbf{Evaluation Metrics}}
We experiment on two friend recommendation tasks: candidate retrieval and candidate ranking, as outlined in Section~\ref{sec:grafrank_training}.
To evaluate friend recommendation, we use standard ranking metrics Hit-Rate (HR@K), Normalized Discounted Cumulative Gain (NDCG@K) and Mean Reciprocal Rank (MRR).
We adopt the standard negative-sample evaluation~\cite{ncf} to generate $N$ negative samples per positive pair of users $(u, v)$ in the test set (user $u$ has sent a friend request to user $v$).
We then compute ranking metrics for each test pair $(u, v)$ by ranking $v$ among the $N$ negative samples via inner products in the latent space.

To evaluate candidate retrieval, we use $N = 10000$ \textit{randomly sampled negative users} for each positive test pair, to emulate retrieval from a large candidate pool.
Ideally, candidate ranking should operate over a shortlisted list of potential friends identified by the retrieval system. However, we aim to provide a fair benchmark comparison of different models for candidate ranking that is agnostic to the biases of the upstream retrieval system.
Thus, we instead generate $N=500$ \textit{hard negatives samples} per test pair based on $K$-hop neighborhoods (Section~\ref{sec:grafrank_training}), to ensure an unbiased comparison.

\subsubsection{\textbf{Training Details}}
We train \grafrank using $L = 2$ message passing layers per feature modality with a hidden dimension size of 64 and output embedding dimension $D = 32$.
In each layer, we sample 15 first-order neighbors and 15 second-order neighbors for each sampled first-order neighbor; each user receives messages propagated from upto 225 friends.
During model training, we apply dropout with rate of 0.3 in the two aggregation layers.
The model is trained for a maximum of 30 epochs with a batch size of 256 positive pairs (apart from 5 random/hard negatives per pair) and learning rate of 0.001 using Adam optimizer.
We benchmark our experiments using a machine with 32 cores, 200 GB shared CPU memory, and a single Nvidia Tesla P100 GPU with PyTorch~\cite{pyt_distributed} implementations on the Linux platform.

\subsection{{Baselines}}
We compare the performance of \grafrank on friend ranking against strong feature-based machine learning models, and state-of-the-art graph neural networks that learn user representations.

\begin{itemize}
    \item \textbf{LogReg}~\cite{ml}: Logistic regression classifier for link prediction. The input feature for each pair of users, is a concatenation of source and target features across the $K$ user feature modalities.
    \item \textbf{XGBoost}~\cite{xgboost}: Tree boosting model for pairwise learning to rank, trained using the same input features as LogReg. It is currently deployed at Snapchat for quick-add friend suggestions.
    \item \textbf{MLP}~\cite{ml}: Two-layer Multi-Layer Perceptron with fully-connected layers and ReLU activations to learn user representations. We train MLP using the same ranking loss as our model (Equation~\ref{eq:grafrank_loss}).
    \item \textbf{GCN}~\cite{gcn}: Graph convolutional networks with degree-weighted neighbor aggregation and neighborhood sampling for scalable training. For each user, we concatenate features across the $K$ user feature modalities into a new nodal feature vector.
    \item \textbf{GAT}~\cite{gat}: Graph attention networks with self-attentional aggregation and neighborhood sampling for scalable training.
    \item \textbf{SAGE + Max}~\cite{graphsage}: Element-wise max pooling for neighbor aggregation and self-embedding concatenation at each layer.
    \item \textbf{SAGE + Mean}~\cite{graphsage}: Same as SAGE + Max with element-wise mean pooling function for neighbor aggregation.
\end{itemize}

Note that graph autoencoders~\cite{gae} and graph convolutional matrix completion models~\cite{gcmc} are not empirically comparable because they cannot scale to our large-scale social network datasets.

For each baseline, we train separate models for candidate retrieval and ranking tasks. We use random negatives for retrieval model training while resorting to hard negatives for ranking; we empirically find training separate models to be vastly superior to a training a single model using a mixture of random and hard negatives, or even curriculum training~\cite{pinsage}.
\renewcommand*{\factor}{0.072}
\begin{table}[t]
\centering
\scriptsize
\noindent\setlength\tabcolsep{1.3pt}
\begin{tabular}{@{}p{0.15\linewidth}@{\hspace{5pt}}
K{\factor\linewidth}K{\factor\linewidth}K{\factor\linewidth}K{\factor\linewidth}K{0.054\linewidth}@{\hspace{10pt}}
K{\factor\linewidth}K{\factor\linewidth}K{\factor\linewidth}K{\factor\linewidth}K{0.054\linewidth}@{\hspace{10pt}}@{}} \\
\toprule
\multirow{1}{*}{\textbf{Dataset}} &  \multicolumn{5}{c}{\textbf{Region 1}} & \multicolumn{5}{c}{\textbf{Region 2}}  \\
\cmidrule(lr){2-6} \cmidrule(lr){7-11}  %
\multirow{1}{*}{\textbf{Metric}} &
\textbf{N@5}  &  \textbf{N@50} &
\textbf{HR@5} & \textbf{HR@50} & \textbf{MRR} &
\textbf{N@5}  &  \textbf{N@50} &
\textbf{HR@5} & \textbf{HR@50} & \textbf{MRR}
\\
\midrule
\textbf{LogReg} & 0.1752 & 0.2460 & 0.2452 & 0.5262 & 0.1751 & 0.0761 & 0.1367 & 0.1134 & 0.3654 & 0.0831 \\
\textbf{MLP}  & 0.1923 & 0.2679 & 0.2721 & 0.5726 & 0.1903 & 0.0973 & 0.1720 & 0.1466 & 0.4541 & 0.1046 \\
\textbf{XGBoost} & 0.2099 & 0.2865 & 0.2932 & 0.5957 & 0.2071 & 0.1366 & 0.2097 & 0.1936 & 0.4921 & 0.1409 \\
\textbf{GCN} & 0.0934 & 0.1836 & 0.1490 & 0.5154 & 0.1034 & 0.1651 & 0.2634 & 0.2503 & 0.6427 & 0.1678 \\
\textbf{GAT} & 0.0851 & 0.1813 & 0.1424 & 0.5352 & 0.0960 & 0.1797 & 0.2794 & 0.2698 & 0.6663 & 0.1812 \\
\textbf{SAGE + Max} & 0.1790 & 0.2736 & 0.2695 & 0.6409 & 0.1797 & 0.1520 & 0.2505 & 0.2315 & 0.6269 & 0.1566 \\
\textbf{SAGE + Mean} & 0.2378 & 0.3240 & 0.3338 & 0.6757 & 0.2333 & 0.2870 & 0.3805 & 0.4005 & 0.7655 & 0.2790 \\
\midrule
\textbf{\grafrank} & \textbf{0.3152} & \textbf{0.3983} & \textbf{0.4318} & \textbf{0.7533} & \textbf{0.3035} & \textbf{0.4166} & \textbf{0.4950} & \textbf{0.5386} & \textbf{0.8395} & \textbf{0.4012} \\
\textbf{Percent Gains} & 32.55\% & 22.93\% & 29.36\% & 11.48\% & 30.09\% & 45.16\% & 30.09\% & 34.48\% & 9.67\% & 43.8\% \\
\bottomrule
\end{tabular}
\caption{\grafrank outperforms feature-based models and GNNs (relative gains of 30-43 \% MRR  with respect to the best baseline) on \textit{candidate retrieval} in Regions 1 and 2. HR@K and N@K denote Hit-Rate@K and NDCG@K metrics for $K=5, 50$.}
\label{tab:grafrank_main_retrieval}
\end{table}

\subsection{{Experimental Results}}
We first present our main empirical results comparing our proposed model \grafrank against competing baselines on candidate retrieval and ranking tasks, followed by additional comparisons using alternative measures of friendship quality.

\renewcommand*{\factor}{0.072}
\begin{table}[t]
\centering
\scriptsize
\noindent\setlength\tabcolsep{1.3pt}
\begin{tabular}{@{}p{0.15\linewidth}@{\hspace{5pt}}
K{\factor\linewidth}K{\factor\linewidth}K{\factor\linewidth}K{\factor\linewidth}K{0.054\linewidth}@{\hspace{10pt}}
K{\factor\linewidth}K{\factor\linewidth}K{\factor\linewidth}K{\factor\linewidth}K{0.054\linewidth}@{\hspace{10pt}}@{}} \\
\toprule
\multirow{1}{*}{\textbf{Dataset}} &  \multicolumn{5}{c}{\textbf{Region 1}} & \multicolumn{5}{c}{\textbf{Region 2}}  \\
\cmidrule(lr){2-6} \cmidrule(lr){7-11}  %
\multirow{1}{*}{\textbf{Metric}} &
\textbf{N@5}  &  \textbf{N@10} &
\textbf{HR@5} & \textbf{HR@10} & \textbf{MRR} &
\textbf{N@5}  &  \textbf{N@10} &
\textbf{HR@5} & \textbf{HR@10} & \textbf{MRR}
\\
\midrule
\textbf{LogReg} & 0.1521 & 0.1795 & 0.2268 & 0.3116 & 0.1523 & 0.1398 & 0.1711 & 0.2136 & 0.3106 & 0.1449 \\
\textbf{MLP}  & 0.1873 & 0.2190 & 0.2663 & 0.3644 & 0.1915 & 0.1927 & 0.2241 & 0.2721 & 0.3695 & 0.1967 \\
\textbf{XGBoost} & 0.1714 & 0.2002 & 0.2394 & 0.3287 & 0.1779 & 0.1844 & 0.2174 & 0.2605 & 0.363 & 0.1911 \\
\textbf{GCN} & 0.1345 & 0.1698 & 0.2039 & 0.3136 & 0.1462 & 0.1758 & 0.2147 & 0.2619 & 0.3826 & 0.1831 \\
\textbf{GAT} & 0.1416 & 0.1776 & 0.2197 & 0.3313 & 0.1503 & 0.2028 & 0.2445 & 0.2984 & 0.4276 & 0.2077 \\
\textbf{SAGE + Max} & 0.2063 & 0.2441 & 0.2980 & 0.4151 & 0.2094 & 0.2426 & 0.2818 & 0.3443 & 0.4654 & 0.2426 \\
\textbf{SAGE + Mean} & 0.2232 & 0.2607 & 0.3165 & 0.4330 & 0.2255 & 0.2766 & 0.3164 & 0.3835 & 0.5064 & 0.2744 \\
\midrule
\textbf{\grafrank} & \textbf{0.2684} & \textbf{0.3098} & \textbf{0.3772} & \textbf{0.5051} & \textbf{0.2669} & \textbf{0.3342} & \textbf{0.3767} & \textbf{0.4529} & \textbf{0.5841} & \textbf{0.3282} \\
\textbf{Percent Gains} & 20.25\% & 18.83\% & 19.18\% & 16.65\% & 18.36\% & 20.82\% & 19.06\% & 18.1\% & 15.34\% & 19.61\% \\
\bottomrule
\end{tabular}
\caption{\grafrank achieves significant improvements (relative gains of 18-20\% MRR with respect to the best baseline) over both feature-based models and prior GNNs in all ranking metrics on friend \textit{candidate ranking} in both Region 1 and Region 2.}
\label{tab:grafrank_main_ranking}
\end{table}

\subsubsection{\textbf{Friend Candidate Retrieval and Ranking ($\text{RQ}_1$)}}
\label{sec:grafrank_main_results}
We compare friend recommendation performance (based on add requests) of various approaches on retrieval and ranking in Tables~\ref{tab:grafrank_main_retrieval} and~\ref{tab:grafrank_main_ranking} respectively.
Interestingly, we find that SAGE based models variants often outperform popular GNN models GCN and GAT.
A possible explanation is the impact of feature space heterogeneity in social networks and stochastic neighbor sampling; this results in noisy user representations for GNN models (GCN, GAT) that recursively aggregate neighbor features without emphasizing self-connections.
Preserving knowledge of the original user features by explicitly concatenating the self-embedding in each layer results in noticeable gains (SAGE variants).

\grafrank significantly outperforms state-of-the-art approaches with over 20-30\% relative MRR gains.
In contrast to singular aggregation over the entire feature space by prior GNNs, \grafrank handles variance in homophily across different modalities through
modality-specific communication-aware neighbor aggregation.
Further, the final user representations are learnt by a correlation-aware attention layer to capture discriminative facets of each modality.

\subsubsection{\textbf{Alternative Friendship Quality Indicators ($\text{RQ}_2$)}}
In addition to evaluating friend suggestion based on the generated friend \textit{addition} requests, we consider other metrics to quantify friendship quality, \textit{e.g.}, social platforms often want to incentivize friendships that result in greater downstream engagement.
We therefore define friendship \textit{reciprocation} and future bi-directional \textit{communication} as two alternative measures of friendship quality.
We present comparisons to evaluate \textit{reciprocated} and \textit{communicated} friend recommendation performance on the retrieval task in Table~\ref{tab:grafrank_alternative_metrics}.

We observe consistently high gains for \grafrank on the reciprocated and communicated friend retrieval tasks; this also demonstrates the generality of our pairwise friend ranking objective (Equation~\ref{eq:grafrank_loss}) in learning user representations that promote downstream engagement.  Designing multi-criteria ranking objectives to balance different quality measures is worth exploring in the future.

\renewcommand*{\factor}{0.13}
\begin{table}[t]
\centering
\small
\noindent\setlength\tabcolsep{1.5pt}
\begin{tabular}{@{}p{0.22\linewidth}@{\hspace{8pt}}
K{0.11\linewidth}K{0.1\linewidth}@{\hspace{8pt}}
K{0.11\linewidth}K{0.1\linewidth}@{\hspace{8pt}}
K{0.11\linewidth}K{0.1\linewidth}@{\hspace{8pt}}}
\toprule
\multirow{1}{*}{\textbf{Dataset}} &  \multicolumn{2}{c}{\textbf{Add}} & \multicolumn{2}{c}{\textbf{Reciprocate}} & \multicolumn{2}{c}{\textbf{Communicate}}  \\
\cmidrule(lr){2-3} \cmidrule(lr){4-5} \cmidrule(lr){6-7}
\multirow{1}{*}{\textbf{Metric}} &
\textbf{HR@50} & \textbf{MRR} & \textbf{HR@50} & \textbf{MRR}& \textbf{HR@50} & \textbf{MRR} \\
\midrule
\textbf{LogReg} & 0.5262 & 0.1751 & 0.5582 & 0.2029 & 0.5495 & 0.1811 \\
\textbf{MLP}  & 0.5726 & 0.1903 & 0.6006 & 0.2165 & 0.6001 & 0.1979 \\
\textbf{XGBoost} & 0.5957 & 0.2071 & 0.6286 & 0.2322 & 0.6407 & 0.2274 \\
\textbf{GCN} & 0.5154 & 0.1034 & 0.5329 & 0.1113 & 0.5273 & 0.1038 \\
\textbf{GAT} & 0.5352 & 0.0960 & 0.5596 & 0.1045 & 0.5654 & 0.0971 \\
\textbf{SAGE + Max} & 0.6409 & 0.1797 & 0.6653 & 0.2043 & 0.6670 & 0.1834 \\
\textbf{SAGE + Mean} & 0.6757 & 0.2333 & 0.6984 & 0.2609 & 0.7056 & 0.2446 \\
\midrule
\textbf{GraFrank} & \textbf{0.7533} & \textbf{0.3035} & \textbf{0.7756} & \textbf{0.3367} & \textbf{0.7942} & \textbf{0.3152} \\
\textbf{Percent Gains} & 11.48 \% & 30.09\% & 11.05\% & 29.05\% & 12.56\% & 28.86 \% \\
\bottomrule
\end{tabular}
\caption{Comparison on \textit{add}, \textit{reciprocate} and \textit{communicate} friendship retrieval tasks (reported on Region 1). \grafrank has consistent gains across all tasks.}
\label{tab:grafrank_alternative_metrics}
\end{table}

\subsection{{Ablation Study ($\text{RQ}_3$)}}
\label{sec:grafrank_ablation}
In this section, we present two model ablation studies to analyze the \textit{architectural modeling} choices and \textit{training strategies} in \grafrank respectively. We briefly describe the two studies and our empirical findings below. 

\subsubsection{\textbf{Model Architecture}}
We design three variants of \grafrank to study the utilities of \textit{communication-aware} and \textit{modality-specific aggregation} towards friend suggestion.

\begin{itemize}
    \item \textbf{$\text{GraFRank}_{UM}$ (User-Modality)}: We analyze the contribution of link features by parameterizing the $k$-th modality aggregator with \textit{just} the $k$-th modality user features (Equation~\ref{eqn:grafrank_node_attn}). Note that link features are excluded during neighbor aggregation.
    \item  \textbf{$\text{GraFRank}_{UL}$ (User-Link)}: To study the effectiveness of learning \textit{modality-specific} aggregators, we define a \textit{single modality-agnostic} aggregator over user feature vectors obtained by concatenation across the $K$ modalities and link features.
    \item \textbf{$\text{GraFRank}_{U}$ (User)}: We remove link features from the aggregator in \textbf{$\text{GraFRank}_{UL}$} to further test the standalone benefits of link features in parameterizing a single neighbor aggregator.
\end{itemize}

\renewcommand*{\factor}{0.13}
\begin{table}[t]
\centering
\noindent\setlength\tabcolsep{2.5pt}
\begin{tabular}{@{}p{0.3\linewidth}@{\hspace{8pt}}
K{\factor\linewidth}K{\factor\linewidth}@{\hspace{8pt}}
K{\factor\linewidth}K{\factor\linewidth}@{\hspace{8pt}}}
\toprule
\multirow{1}{*}{\textbf{Dataset}} &  \multicolumn{2}{c}{\textbf{Retrieval}} & \multicolumn{2}{c}{\textbf{Ranking}}   \\
\cmidrule(lr){2-3} \cmidrule(lr){4-5}
\multirow{1}{*}{\textbf{Metric}} &
\textbf{HR@50} & \textbf{MRR} &  \textbf{HR@10} & \textbf{MRR}    \\
\midrule
(a) \textbf{$\grafrank_U$} & 0.6968 & 0.2346 & 0.4255 & 0.2164 \\
(b) \textbf{$\grafrank_{UL}$} & 0.7069 & 0.2423 & 0.4450 & 0.2301 \\
(c) \textbf{$\grafrank_{UM}$} & 0.7239 & 0.2823 & 0.4887 & 0.2480 \\
\textbf{\grafrank} & \textbf{0.7533} & \textbf{0.3035} & \textbf{0.5051} & \textbf{0.2669} \\
\bottomrule
\end{tabular}
\caption{Model architecture ablation study of \grafrank. Removing (c) link communication features, (b) modality-specific aggregation, or (a) both, hurt model performance.}
\label{tab:grafrank_architecture_ablation}
\end{table}

The model performance of all architectural variations are reported in Table~\ref{tab:grafrank_architecture_ablation}.
${\grafrank}_{UL}$ performs much worse than \grafrank, thus highlighting the benefits of learning multiple modality-specific aggregators to account for varying extents of modality homophily.

Communication-aware neighbor aggregation is effective at identifying actively engaged friends during neighbor aggregation; this is evidenced by the performance gains of \grafrank over ${\grafrank}_{UM}$ (modality-aware user feature aggregation).
We find noticeable gains from parameterizing the neighbor aggregator with link features even in the absence of modality-specific aggregation;  the comparison between ${\grafrank}_{UL}$ and ${\grafrank}_{U}$ (single user feature aggregator without link features) further validates this benefits of utilizing link features during message passing.

\subsubsection{\textbf{Training Strategy}}
We examine different training strategies to learn GNN models for friend recommendation.
We train \grafrank with random negatives for candidate retrieval, but adopt two-phase hard negative fine-tuning (with random negative pretraining) for candidate ranking.
To validate our choices, we examine three model training settings: (a) random negative training, (b) hard negative training, and (c) fine-tuning (after pretraining on random negatives), for two GNN models: SAGE + Mean, and \grafrank, across both candidate retrieval and ranking tasks. Note that training with combination of random and hard negatives, as proposed in~\cite{pinsage}, is excluded since it consistently performs worse than the above three strategies on both retrieval and ranking tasks.

\renewcommand*{\factor}{0.12}
\begin{table}[h]
\centering
\noindent\setlength\tabcolsep{1.8pt}
\begin{tabular}{@{}p{0.41\linewidth}@{\hspace{8pt}}
K{\factor\linewidth}K{\factor\linewidth}@{\hspace{8pt}}
K{\factor\linewidth}K{\factor\linewidth}@{\hspace{8pt}}}
\toprule
\multirow{1}{*}{\textbf{Dataset}} &  \multicolumn{2}{c}{\textbf{Retrieval}} & \multicolumn{2}{c}{\textbf{Ranking}}   \\
\cmidrule(lr){2-3} \cmidrule(lr){4-5}
\multirow{1}{*}{\textbf{Metric}} &
\textbf{HR@50} & \textbf{MRR} &  \textbf{HR@10} & \textbf{MRR}    \\
\midrule
\textbf{SAGE + Mean (random)} & 0.6757 & 0.2333 & 0.3943 & 0.1923 \\
\textbf{SAGE + Mean (hard)} & 0.3275 & 0.0766 & 0.4330 & 0.2255 \\
\textbf{SAGE + Mean (fine-tune)} & 0.3978 & 0.0965 & 0.4561 & 0.2372 \\
\midrule
\textbf{\grafrank (random)} & \textbf{0.7533} & \textbf{0.3035} & 0.4655 & 0.2254 \\
\textbf{\grafrank (hard)} & 0.4542 & 0.1461 & 0.4823 & 0.2594 \\
\textbf{\grafrank (fine-tune)} & 0.5283 & 0.1871 & \textbf{0.5051} & \textbf{0.2669} \\
\bottomrule
\end{tabular}
\caption{Training strategy comparison of two GNNs across retrieval and ranking tasks. Random negative training achieves best results for retrieval. Random negative pretraining with hard negative fine-tuning benefits ranking.}
\label{tab:grafrank_training_ablation}
\end{table}

We make three consistent observations from the performance comparison (Table~\ref{tab:grafrank_training_ablation}) across all of the compared GNN models:
\begin{itemize}
    \item Random negative training achieves best results for retrieval, but performs poorly on ranking; such models lack the \textit{resolution} to  discriminate amongst potential candidates for re-ranking.
    \item Training on hard negatives improves candidate ranking as expected, yet results in poor retrieval performance. Learning fine profile-oriented distinctions among graph-based neighbors, is actually detrimental to the coarse-grained task of retrieval.
    \item Random negative pretraining yields good parameter initialization points that are more conducive for effective fine-tuning on hard negatives.  Fine-tuning improves results for all GNNs over direct hard negative training on ranking, but is ineffective for retrieval.
\end{itemize}

\subsection{{Training and Sensitivity Analysis ($\text{RQ}_4$)}}
In this section, we quantitatively analyze model \textit{convergence} and model \textit{sensitivity} to sampled \textit{neighborhood sizes} in GNN models.

\subsubsection{\textbf{Model Training Analysis}}
\begin{figure}[t]
    \centering
    \includegraphics[width=0.9\linewidth]{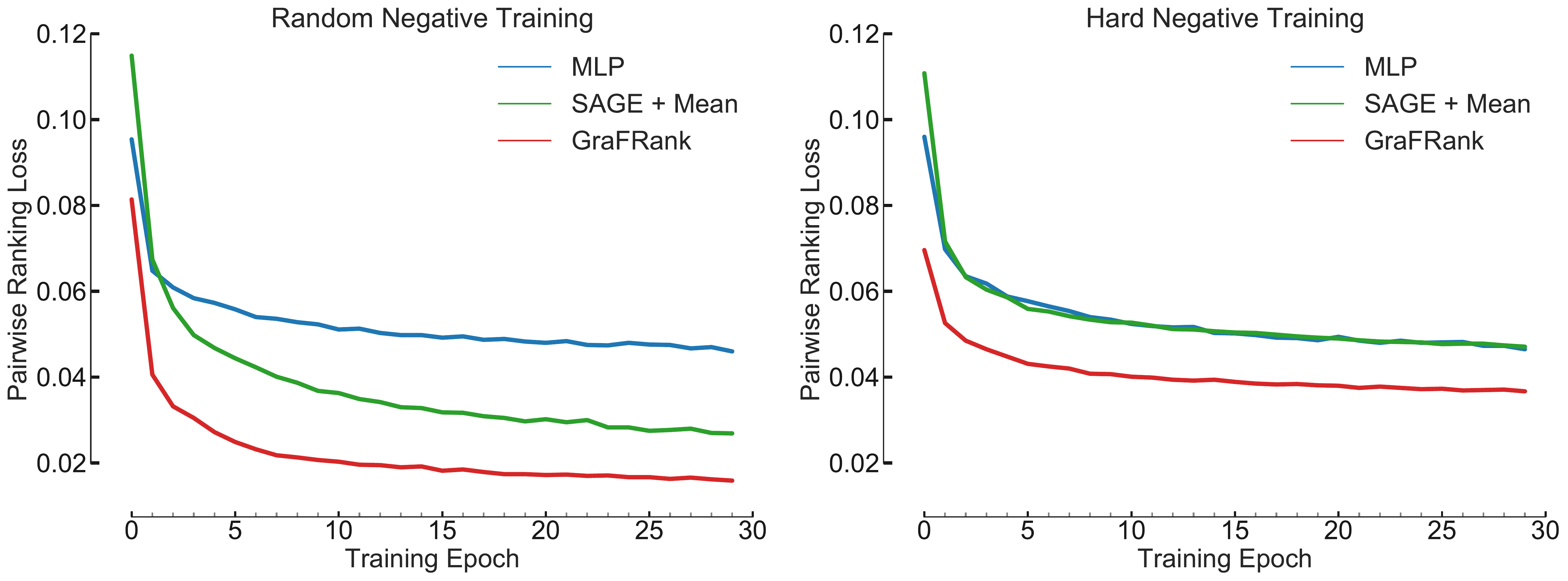}
    \caption{\grafrank converges faster to better optimization minima in \textit{random} and \textit{hard} negative settings, which translates to notable gains on both retrieval and ranking tasks.}
    \label{fig:grafrank_loss_curve}
\end{figure}

We investigate the relative abilities of different models to optimize the pairwise friend ranking objective (Equation~\ref{eq:grafrank_loss}).
We compare the convergence rates of baselines MLP, SAGE + Mean, and \grafrank under both random and hard negative training settings, by examining the average training loss per epoch in Figures~\ref{fig:grafrank_loss_curve} (a) and (b) respectively.

As expected, all models converge to a lower training loss against random negatives (Figure~\ref{fig:grafrank_loss_curve} (a)) when compared to hard negatives (Figure~\ref{fig:grafrank_loss_curve} (b)).
Interestingly, SAGE + Mean shows similar training convergence as MLP in Figure~\ref{fig:grafrank_loss_curve} (b), but achieves better test results; this indicates better generalization for GNNs over feature-based models.
Compared to baselines, \grafrank converges to a better optimization minimum, under both random and hard negative settings, that also generalizes to better test results (Tables~\ref{tab:grafrank_main_retrieval} and~\ref{tab:grafrank_main_ranking}).

\subsubsection{\textbf{Runtime and Sensitivity Analysis}}
\begin{figure}[t]
    \centering
    \includegraphics[width=0.9\linewidth]{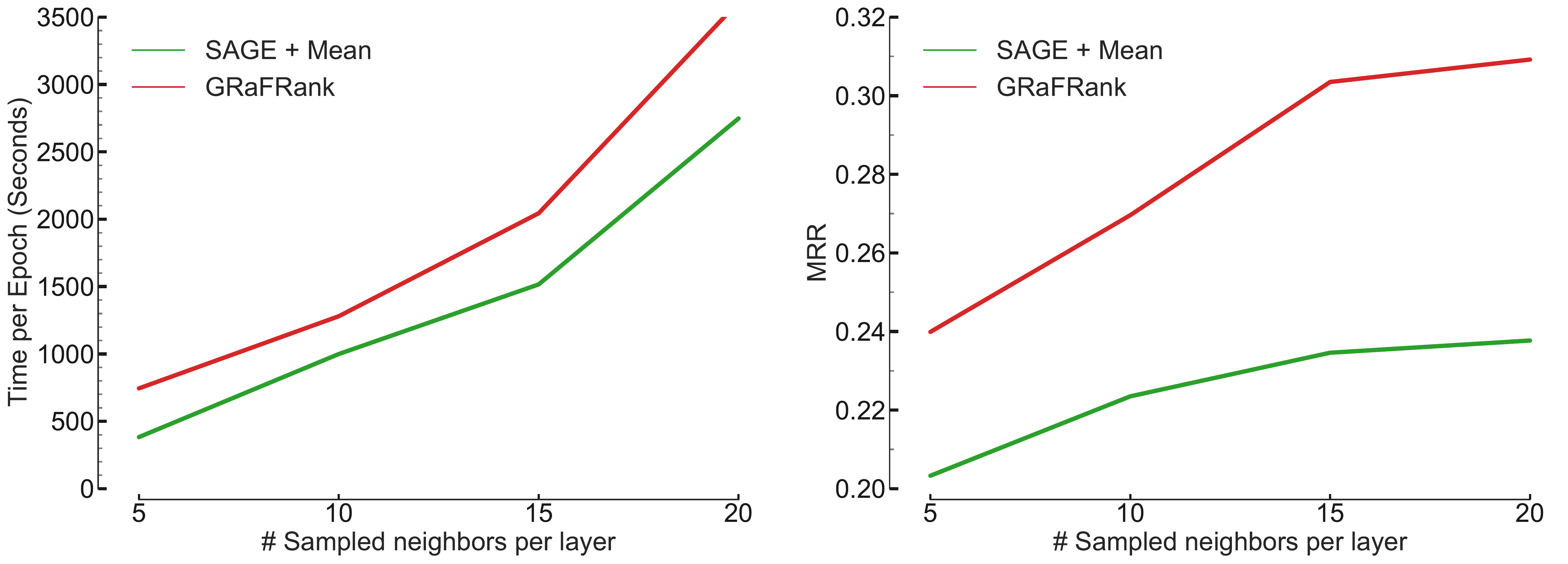}
    \caption{We observe diminishing returns in MRR after neighborhood size $T=15$; \grafrank has significant gains over SAGE + Mean, with marginally higher training times.}
    \label{fig:grafrank_runtime_sensitivity}
\end{figure}

A key trade-off in training scalable GNN models lies in choosing the size of sampled neighborhoods $T$ in each message-passing layer.
In our experiments, we train two-layer GNN models for friend ranking.
Figure~\ref{fig:grafrank_runtime_sensitivity} shows the runtime and performance of SAGE + Mean and \grafrank for different sizes of sampled neighborhoods $T$ from 5 to 20.

Model training time generally increases linearly with $T$, but a greater slope after $T=15$.
We also observe diminishing returns in model performance (MRR) with increase in the size of sampled neighborhood $T$ after $T=15$.
Thus, we select a two-layer GNN model with layer-wise neighborhood size of 15, to provide an effective trade-off between computational cost and performance.

Compared to SAGE + Mean, \grafrank has marginally higher training times, yet achieves significant performance gains (20\% MRR), justifying the added cost of modality-specific aggregation.

\subsection{User Cohort Analysis ($\text{RQ}_5$)}
In this section, we present multiple \textit{qualitative} analyses to examine model performance across user segments with varied node \textit{degree} and friending \textit{activity} levels, and compare \textit{t-SNE visualizations} of user representations learned by different neural models.

\subsubsection{\textbf{Impact of degree and activity}}
\begin{figure}[t]
    \centering
    \includegraphics[width=0.9\linewidth]{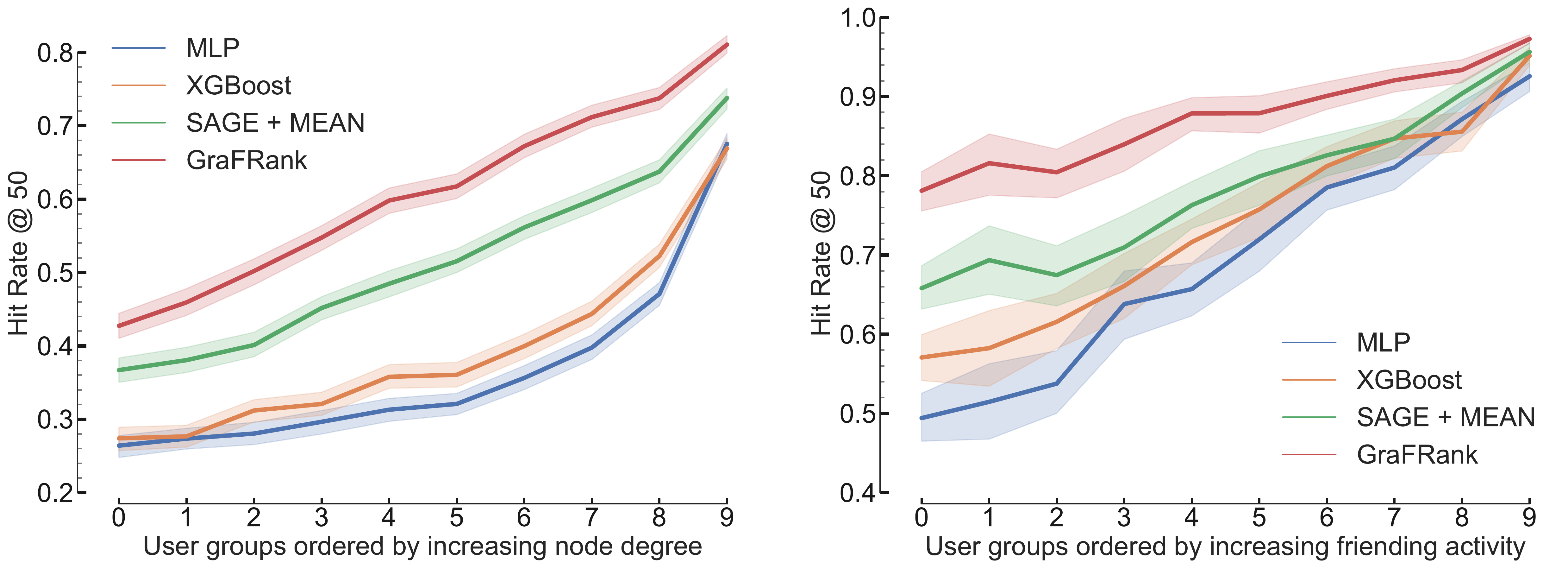}
    \caption{\grafrank has significant gains across all user segments, with notably larger gains for \textit{low-to-medium degree} users (a), and \textit{less-active} users (b).} %
    \label{fig:grafrank_user_cohort}
\end{figure}

We examine friend recommendation performance, across users with different node degrees and friending activities.
Specifically, for each test user, \textit{degree} is the number of friends, and \textit{activity} is the number of friend requests sent/received in the past 30 days.
We divide the test users into groups, independently based on their degree and activity levels.
We compare \grafrank with feature-based models MLP, XGBoost and the best GNN baseline SAGE + Mean.
Figures~\ref{fig:grafrank_user_cohort}(a) and (b) depict friend candidate retrieval performance HR@50 across user segments with different degrees and activities respectively.

From Figure~\ref{fig:grafrank_user_cohort}(a), overall model performance generally increases with node degree, due to the availability of more structural information.
\grafrank has significant improvements across all user segments, with notably higher gains for \textit{low-to-medium} degree users (relative gains of 20\%).
\grafrank prioritizes active friendships by communication-aware message-passing, which compensates for the lack of sufficient local connectivities in the ego-network.

The performance variation across users with different \textit{activity} levels in Figure~\ref{fig:grafrank_user_cohort}(b), exhibits more distinctive trends with clear gains for GNN models over feature-based MLP and XGBoost for less-active users.
Significantly, \grafrank has much stronger gains over SAGE + Mean, in \textit{less-active user segments}, owing to its \textit{multi-faceted modeling} of heterogeneous in-platform user actions.
\grafrank effectively overcomes sparsity concerns for less-active users, through
modality-specific neighbor aggregation over multi-modal user features to learn expressive user representations for friend ranking.

\subsubsection{\textbf{Visualization}}
\begin{figure}[t]
    \centering
    \includegraphics[width=0.9\linewidth]{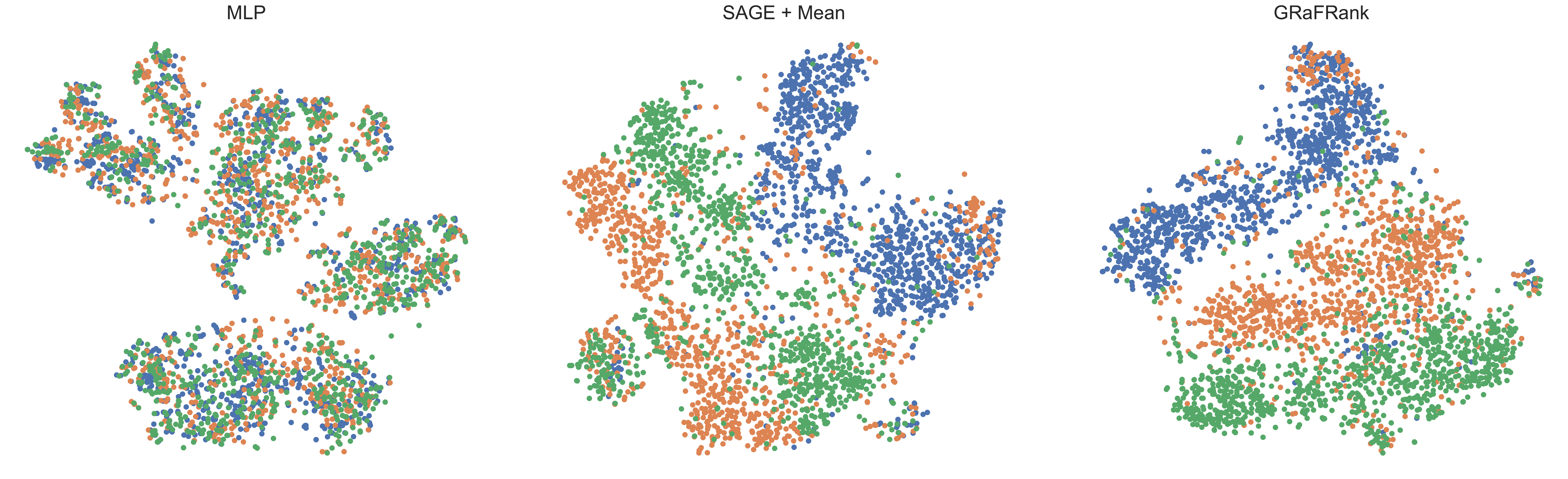}
    \caption{Visualization of two-dimensional t-SNE transformed user representations from feature-based MLP, and GNN models: SAGE + Mean, and \grafrank. Users with the same color belong to the same city. Compared to MLP and SAGE + Mean, the friendship relationships learnt by \grafrank result in well-separated user clusters capturing geographical proximity.}
    \label{fig:grafrank_emb_viz}
\end{figure}

To analyze the \textit{versatility} of learned user embeddings, we present a qualitative visualization to compare different models on their expressivity to capture geographical user proximity.
We randomly select users from three different cities within Region 1 and use t-SNE~\cite{tsne} to transform their learned embeddings into two-dimensional vectors. Figure~\ref{fig:grafrank_emb_viz} compares the visualization results from different neural models.  Evidently, the visualization learned by MLP does not capture geographical proximity, while the GNN models are capable of grouping users located within the same city.
Compared to SAGE + Mean, \grafrank forms even more well-segmented groups with minimal inter-cluster overlap.

\section{Conclusion}
\label{sec:grafrank_conclusion}
This chapter investigates graph neural network usage and design for friend suggestion in large-scale social platforms.
We formulate friend suggestion as a multi-faceted friend ranking with multi-modal user features and link communication features.
Motivated by our insights from an empirical analysis on user feature modalities, we design a neural architecture \grafrank that handles heterogeneity in modality homophily via modality-specific neighbor aggregators, and learns non-linear modality correlations through cross-modality attention.
Our experiments on two multi-million social network datasets from Snapchat reveal significant performance gains in friend candidate retrieval (30\% MRR gains) and ranking (20\% MRR gains).
Further, our qualitative analysis indicates stronger gains for a crucial population of less-active and low-degree users.

In this chapter, we concluded our exploration of inductive user modeling scenarios with an industrial application of generating friend suggestions in social platforms. In summary, this dissertation covers inductive user behavior modeling tasks involving new users, user-generated content, and ephemeral groups, across diverse online interaction platforms.

\chapter{Conclusions and Future Directions}
\label{chap:conclusion}
In this chapter, we first briefly summarize the key research contributions of this dissertation, and then discuss immediate extensions and broader research avenues for future work.

\section{Research Contributions}
We designed neural frameworks to model user behavior in online platforms (\textit{e.g.}, social networking and
e-commerce platforms), involving activities and interactions with other users and functionalities.
We examined behavior modeling applications across diverse scenarios, involving graph-structured interactions in social networks, bipartite user-item interactions in e-commerce platforms, and multipartite user-group-item interactions in group activities.

\subsection{{Technical Challenges}}
Designing neural user modeling frameworks poses several technical challenges in handling the massive scale of behavioral data and the diversity of interaction types involving over millions of entities.
Despite the potential to learn neural models that harness massive interaction logs, a central theme of this dissertation is addressing \textit{data sparsity} challenges that manifest in different ways across applications, which are briefly summarized below:
\begin{description}
\item \textbf{Ground-truth Label Sparsity}: In large-scale online platforms, we typically have access to very limited ground-truth labels for entity attributes, \textit{e.g.}, age, gender of users in social networks or aspect ratings of products in e-commerce platforms.

In chapters~\ref{chap:infomotif} and~\ref{chap:groupim}, to compensate for the lack of sufficient training label information, we designed self-supervised learning strategies that derive auxiliary supervision signals from the intrinsic structure (\textit{e.g.}, higher-order connectivity patterns or group membership structures) of the underlying data.

\item \textbf{Entity-level Interaction Sparsity}:
Heavy-tailed distributions in user interests and interaction patterns are commonly observed in user behavioral data~\cite{barabasi_tail} across several online platforms.
Despite the availability of behavioral data at massive scales, we have very limited historical records at the granularity of individual entities due to the highly skewed interaction distribution. 
Thus, learning meaningful trends or insights for a vast majority of entities is challenging due to \textit{entity-level interaction sparsity}.

We explore deep generative modeling (Chapter~\ref{chap:infvae}), self-supervised learning (Chapter~\ref{chap:groupim}) and meta-learning (Chapter~\ref{chap:protocf}) paradigms to enable accurate personalized recommendation and inference for data-poor tail entities (such as long-tailed users, items and ephemeral groups) with severe interaction sparsity.

\item \textbf{Feature Diversity and Skew}: In inductive learning applications, entities are often represented using a combination of diverse features, \textit{e.g.}, user features in social networking platforms may include static profile attributes, dynamic communication and engagement activities, while item features in e-commerce platforms may include textual descriptions, product reviews, and other attributes (such as brand or price).
Here, the key modeling challenges are feature heterogeneity and skew: the features often belong to different modalities and exhibit non-trivial correlations; we also observe skewed occurrences of features across different entities.

To handle attribute diversity in local graph neighborhoods, we introduced graph neural network training objectives that learn local structure and attribute co-variation in chapter~\ref{chap:infomotif}. 
We also designed multi-faceted graph neural architectures that can effectively learn from multiple correlated feature modalities in chapter~\ref{chap:grafrank}.
\end{description}

In this dissertation, we primarily developed architecture-agnostic frameworks for user behavior modeling, with generalizable learning strategies that are targeted towards addressing the aforementioned sparsity challenges.
Our learning frameworks enable personalized inference at scale for broad application scenarios across diverse platforms.
Furthermore, we also present an example of a specific deep learning architecture design that effectively handles sparsity concerns in a large-scale industrial application.

Our problem settings span transductive and inductive learning scenarios, where transductive learning models behavior of entities seen during training and inductive learning targets unseen entities that are only observed during inference.
Thus, our key research contributions are also organized below into \textit{transductive} and \textit{inductive} user behavior modeling applications.

\subsection{{Transductive User Behavior Modeling}}
We examined two transductive learning settings: inference and recommendation in \textit{graph-structured} interactions (Chapter~\ref{chap:infomotif}) and \textit{bipartite user-item} interactions (Chapter~\ref{chap:protocf}).

First, in Chapter~\ref{chap:infomotif}, we formulate user profiling in online platforms as semi-supervised (transductive) learning over graphs connecting different entities.
To address ground-truth \textit{label sparsity} in node attribute inference, we utilize higher-order connectivity structures (network motifs) to effectively regularize arbitrary graph neural networks through self-supervised learning objectives.
Our framework learns attributed structural roles for nodes to identify structurally similar nodes with co-varying local attributes independent of graph proximity, thus effectively handling \textit{feature sparsity and diversity} in local neighborhoods.

Thereafter, in Chapter~\ref{chap:protocf}, we focus on learning to recommend \textit{few-shot} items with neural collaborative filtering models for personalized recommendation.
To eliminate the distribution disparity between head items (with \textit{abundant} interactions) and tail items (with \textit{sparse} interactions), our novel meta-learning framework learns-to-compose robust prototype representations for few-shot items via knowledge transfer from arbitrary base recommenders.

\subsection{{Inductive User Behavior Modeling}}
We examined three inductive learning settings with a focus on prediction and recommendation tasks for entities that are only observed during inference-time. Specifically, we modeled the spread or diffusion of \textit{user-generated content} in social networks (Chapter~\ref{chap:infvae}), and enabled item recommendations for \textit{ephemeral (unseen) groups} (Chapter~\ref{chap:groupim}).
Finally, we designed inductive ranking models for the application of \textit{friend suggestion} in large-scale social platforms (Chapter~\ref{chap:grafrank}).

First, in Chapter~\ref{chap:infvae}, we model the spread (or diffusion) of user-generated content in social networks. Here, we encounter \textit{interaction sparsity} challenges since a vast population of users seldom post content (sparse diffusion actions).
To address interaction sparsity, we introduced a deep generative modeling framework that models users as probability distributions in the latent space with variational priors parameterized by graph neural networks.
Our social regularization framework incorporates the co-variance of temporal context (recent posting activities) with structural graph connectivity for diffusion prediction.

Then, in Chapter~\ref{chap:groupim}, we focus on personalized item recommendations for ephemeral groups with limited or no historical interactions together.
To address \textit{group interaction sparsity}, we introduced self-supervised learning objectives that exploit the preference
co-variance among group members for group recommender training.
To enable inductive generalization to ephemeral groups, our framework relies on mutual information estimation and maximization over group membership structures, to effectively regularize base group recommenders.

Finally, in Chapter~\ref{chap:grafrank}, we proposed multi-modal inference with graph neural networks for the industrially ubiquitous application of generating friend suggestion in social networking platforms.
To address \textit{interaction sparsity} challenges for \textit{less-active} and \textit{low-degree users}, we presented an inductive graph neural network model that captures knowledge from multiple correlated user feature modalities and user-user interactions over a time-evolving friendship graph to achieve multi-faceted friend ranking.

\subsection{{Summary of Learning Paradigms}}
In this dissertation, we have explored several architectural modeling choices and training strategies to address the sparsity-oriented technical challenges across the different chapters. Here, we provide a brief summary of the central learning paradigms:

\subsubsection{\textbf{Graph Neural Networks (design and training)}}
Graph Neural Networks (GNNs) are a powerful paradigm to learn node representations combining graph topology and node/edge features for both transuctive and inductive learning applications.
Their localized message passing framework makes them to extremely effective to directly alleviate entity-level label and interaction sparsity: GNNs can directly consider structural connectivity and interactions with other related entities through local message-passing propagations; GNNs can easily incorporate side information for different entities through node and edge features belonging to diverse modalities.

In this dissertation, we have advanced graph neural network design, training, and applications in several directions, as described below:

\begin{itemize}
\item Model training strategies to capture node structural roles via higher-order connectivity structures (Chapter~\ref{chap:infomotif}).

\item Social regularization of sequential models via variational graph autoencoders for diffusion prediction. (Chapter~\ref{chap:infvae}).

\item Model design to learn from multifaceted interactions in large-scale social networking platforms (Chapter~\ref{chap:grafrank}).

\end{itemize}

\subsubsection{\textbf{Deep Generative Modeling}}
Deep Generative Models are powerful neural networks that aim to approximate any kind of data distribution~\cite{dgm}. We leverage their expressive power to model data-driven priors for entities (\textit{e.g.}, users, items) that are represented as probability distributions in the latent space.
By parameterizing latent space priors through deep generative models that can capture different modeling hypotheses, we effectively address entity-level sparsity concerns.

Variational Autoencoders (VAEs)~\cite{vae} and Generative Adversarial Networks (GANs)~\cite{gan} are two of the most commonly used frameworks for deep generative modeling. VAEs maximize a variational lower bound of the data log-likelihood, while GANs aim to achieve an equilibrium between a generator and discriminator through alternating optimization.

In chapter~\ref{chap:infvae}, we design a variational autoencoder framework for diffusion modeling, where a graph VAE regularizes the user latent space to capture structural graph connectivity information.
We have also explored adversarial learning to model item space priors based on co-occurence statistics to address item-level sparsity~\cite{cikm18adv} (not included in this dissertation).

\subsubsection{\textbf{Self-supervised Learning}}
The paradigm of self-supervised learning~\cite{self_ssl_survey} alleviates label and interaction sparsity challenges by setting up new learning objectives based on additional supervision signals extracted from the intrinsic structure of the data.
For instance, auxiliary supervision signals for images may be created by rotating, cropping, or colorizing images, each associated with appropriate supervised training objectives.
Despite promising recent advances owing to self-supervised learning in computer vision~\cite{dim} and natural language~\cite{bert} applications, this paradigm is relatively unexplored in the graph mining and recommendation domains.

In chapter~\ref{chap:infomotif}, we design self-supervised learning objectives to regularize graph neural networks for transductive node classification by learning attribute correlations in higher-order connectivity structures (network motifs).
In chapter~\ref{chap:groupim}, we presented a self-supervised learning framework that relies on mutual information estimation and maximization over group memberships, to overcome group interaction sparsity for ephemeral group recommendation.

\subsubsection{\textbf{Meta-Learning for Few-shot Inference}}
The meta-learning (learning to learn) paradigm intends to rapidly learn new concepts or tasks given a limited number of few training examples.
There are three common approaches: metric-based~\cite{protonet}, gradient-based~\cite{maml} and optimization-based models for meta-learning.
Notably, meta-learning has achieved considerable success in few-shot supervised learning such as image classification applications in computer vision.
Yet, designing meta-learning frameworks for user behavior modeling in online platforms with severe interaction sparsity, poses unique scaling challenges.

In chapter~\ref{chap:protocf}, we formulate long-tail recommendations  in online platforms as a few-shot learning problem of learning-to-recommend entities (users/items) with very few interactions.
We introduced a novel metric-based few-shot item recommendation framework that outperforms state-of-art neural recommendation approaches on overall item recommendation (by 5\% Recall@50) while achieving significant performance gains (of 60-80\% Recall@50) for tail items (with less than 20 interactions).

\section{Future Work}
There are multiple avenues of future work that can improve upon our proposed frameworks for sparsity-aware user behavior modeling in online platforms.
We first discuss immediate extensions that can directly extend the learning paradigms and applications discussed in this dissertation. 
Later, we conclude with a discussion on broader research directions (beyond sparsity-oriented challenges) that are critical to online user behavior models.
\subsection{{Extensions to our Frameworks}}
We first discuss a few extensions to our approaches introduced in the preceding chapters. 

\subsubsection{\textbf{Self-supervised Pre-training and Learning over Graphs}}
In chapter~\ref{chap:infomotif}, we designed self-supervised learning objectives over graphs to learn attribute correlations in higher-order connectivity structures (network motifs).
We can further generalize this self-supervised learning framework beyond motifs to preserve other structural properties and social hypotheses, including pairwise distances, global community structures, attribute homophily, etc.
An important direction is leveraging self-supervised learning objectives to pre-train graph neural networks over pretext tasks that explore a variety of different learning perspectives.
With the rapid proliferation of e-commerce and social platforms, pre-training graph neural networks over large-scale rich, attributed, and heterogeneous graphs will be become critical, similar to BERT~\cite{bert} models in natural language applications.

\subsubsection{\textbf{Modeling Group Formation and Evolution}}
We have briefly examined group interaction modeling in Chapter~\ref{chap:groupim} through personalized item recommendations for ephemeral groups with limited or no historical interactions together.
In reality, group interactions occur in a variety of different contexts in modern social platforms (\textit{e.g.}, Facebook, Snapchat), where new groups frequently form and existing groups evolve over time with respect to both memberships and interests.
Compared to prior dynamic graph evolution literature that capture node-level structural and behavior evolution, group interactions often involve complex dynamics among different members.
Thus, an important future research direction is modeling group formation and evolution in social networking platforms, by formulating appropriate temporal link prediction and user recommendation problems.

\subsubsection{\textbf{Deep Generative Modeling for Cold-start Recommendation}}
In chapter~\ref{chap:infvae}, we explored deep generative models~\cite{dgm} for robust behavior modeling in social networks with sparsity in content-posting activities of users.
Modeling entity representations as probability distributions in the latent space, has advantages: it allows for flexible distribution-aware and data-driven prior regularization of the latent space with potential gains for long-tail and cold-start entities; flexible probabilistic models better capture uncertainty in the latent space, including data-point variances.
Most of the prior neural collaborative filtering approaches are discriminative recommendation models that are effective for previously observed entities.
To enable cold-start recommendations, one effective approach is to share probabilistic priors within entities of the same category (\textit{e.g.}, restaurants of the same cuisine), which can be learned from the auxiliary side information (\textit{e.g.}, content, knowledge graphs, etc.) associated with entities.
We can build on advances in variational autoencoders (VAEs)~\cite{vae} and generative adversarial networks (GANs)~\cite{gan} to design effective deep generative recommendation models in variety of different settings that entail cold-start entities.

\subsubsection{\textbf{Meta-Learning for Cross-Domain Recommendation}}
In chapter~\ref{chap:protocf}, we proposed a metric-based meta-learning framework for few-shot recommendation over user-item interactions in a single domain.
Our meta-learning approach enables knowledge transfer by relying on the assumption of similar training and inference interaction distributions, which is expected to be satisfied across items within a single domain.
Cross-domain recommendation settings are commonly observed in large-scale e-commerce platforms where domains may describe products belonging to different categories or interactions in different devices (voice versus web shopping) over a shared user space.
In such a scenario, we can explore meta-learning frameworks supplemented with appropriate domain adaptation~\cite{darec} strategies to account for distributional shifts across domains.
Apart from metric-based approaches that enable knowledge via shared latent embeddings, we may also explore gradient-based and optimization-based meta-learning models, which can potentially provide more flexibility in handling cross-domain distribution disparities.

\subsection{{Broader Research Challenges beyond Sparsity}}
We finally discuss a few research directions that go beyond the sparsity-oriented challenges discussed in the other chapters of this dissertation.

\subsubsection{\textbf{Detecting Abuse and Malicious Activity}}

In this dissertation, we build neural user modeling frameworks that are robust to a variety of data sparsity challenges that manifest in behavioral data. An underlying assumption is that the data utilized for model training and inference is clean, reliable, and of high-quality, \textit{i.e.}, the behavioral data accurately represents the information about the interacting entities.
In online platforms, we often observe malicious activities that deny, disrupt, degrade and deceive other participants with a wide range of intents, \textit{e.g.}, fake accounts (which pose potential threats to the safety and security of online communities), and threatening or disparaging content (such as hate-speech, misinformation, bullying and harassment).

One approach to handle malicious activities is to pro-actively detect instances of abusive behavior and remove the corresponding malicious accounts in the data processing stage itself before input to our neural behavior models~\cite{predator}.
Abusive behavior detection can be formulated as a binary classification problem~\cite{malicious_detect}, which is severely imbalanced due to label sparsity associated with labeling malicious activities.	
Our few-shot learning approaches to handle long-tailed distributions can potentially be generalized to learn effective abuse detection models. It is also worthwhile to leverage the power of crowd-sourcing~\cite{abuse_crowd} for annotating posts with abuse-related labels in social platforms.

\subsubsection{\textbf{Model Robustness to Adversarial Attacks}} 
Despite the efficacy of abuse detection strategies, malicious interaction data may invariably become a part of training sets that are used for model learning.
A few recent studies have demonstrated the vulnerability of deep learning models to adversarial examples~\cite{adv_attack_defense}, which are subtle (but non-random) perturbations that induce erroneous outputs from the model (\textit{e.g.}, misclassification), \textit{e.g.}, adversarial examples in e-commerce applications may include fake user profiles that aim to promote (or demote) specific item(s) in the top-$K$ recommendation list of users, or to recommend irrelevant items (to create a mistrust on a system)~\cite{adv_oblivious_rec, adv_rec_survey}.

Thus, it is important to design models that are robust to adversarial attacks, including both poisoning (training-time) and evasion (inference-time) attacks.
Crafting adversarial examples (\textit{e.g.}, fake user profiles) for graph-structured data or recommender systems is significantly more challenging due to the mostly discrete nature of online interactions.
We can explore new learning frameworks for generating adversarial examples, grounded on model compression strategies such as knowledge distillation~\cite{distillation} and deep generative modeling paradigms of GANs~\cite{gan} and VAEs~\cite{vae}.
Since the details of most deployed machine learning models are not publicly available, it is also critical to develop black-box adversarial attack strategies that do not have access to the architectural details or model parameters.
The final step is the design of specialized training and optimization strategies (such as adversarial training or denoising) to enhance model robustness against adversarial examples.

\subsubsection{\textbf{Privacy-preserving User Modeling}}

In this dissertation, we have assumed complete access to online interaction data across all users for model training. However, this data often includes personal information about users (\textit{e.g.}, age, gender, location, who likes what etc.), which may reveal sensitive information with potential for misuse~\cite{weinsberg2012blurme} and can further result in de-anonymization~\cite{deanon} with the use of sophisticated techniques.
With recent government regulations and laws on privacy protection (\textit{e.g.}, General Data Protection Regulation (GDPR) in EU) and increased privacy awareness, it is important to design user modeling techniques that ensure user privacy while generating personalized recommendations.

In contrast to the traditional paradigm of collecting, storing, and processing sensitive user data on a centralized external backend server, we can explore distributed federated learning paradigms~\cite{federated_mf} which can ensure that sensitive user data never leaves their accounts (or devices).
In this paradigm, a master model is distributed to the end clients (users), who in turn can use their locally stored private data and personalized model for both inference and updates. The updates from different users are aggregated on the server to update the master model, which can then be redistributed to the users.
Though federated learning can enable a higher level of privacy than conventional machine learning, it is also critical to derive theoretical estimates of information leakage in such privacy-preserving solutions.

\subsubsection{\textbf{Bias and Fairness Issues in Behavior Models}}
In recent times, several studies have shown the potential for deployed machine learning systems to amplify social inequities and unfairness;  a few examples include an automated hiring system that had a higher likelihood of recommending from certain racial, gender, or age groups~\cite{hiring_bias}, and an insurance company that involuntarily discriminated against elderly patients when using machine learning to workout insurance premiums.
In this context, two central concepts of \textit{bias} and \textit{fairness} are important. 
In general, a machine learning system exhibits bias if it systematically and unfairly discriminates against certain individuals or groups of individuals (more generally entities) in favor of others.
The goal of fairness is to design machine learning algorithms that make fair predictions across various entities or groups of entities, \textit{e.g.,} groups may be defined based on different protected attributes (\textit{e.g.}, race, gender) of users~\cite{bias_fairness_survey}.

In several chapters of this dissertation, we have examined \textit{popularity bias} issues arising due to interaction skew in social networks and recommender systems (popular entities interact frequently, while the majority of other entities do not get much attention).
Beyond popularity bias, it is important to quantify and measure other forms of societal biases in online social media and recommendation applications, to ensure appropriate exposure of the protected attributes such as user demographic properties (age, gender, race, ethnicity, location, etc.) and diversity (item category or genres, supplier, etc.).
The overall objective is the design fairness-aware recommendation and inference algorithms for mitigating different biases without severely impacting key business metrics; we note that fairness-aware approaches can be applied at several parts of the pipeline: data collection and generation (pre-processing), at the point of modeling (in-processing), or after modeling (post-processing).

\bibliographystyle{IEEE_ECE}

\bibliography{thesisrefs, infomotif/infomotif,protocf/protocf,infvae/infvae,groupim/groupim, grafrank/grafrank}

\appendix

\backmatter

\end{document}